\documentclass{article}

\usepackage[final,nonatbib]{neurips_2022}

\usepackage[utf8]{inputenc} 
\usepackage[T1]{fontenc}    
\usepackage{hyperref}       
\usepackage{url}            
\usepackage{booktabs}       
\usepackage{amsfonts}       
\usepackage{nicefrac}       
\usepackage{microtype}      
\usepackage{xcolor}         
\usepackage{microtype}
\usepackage{graphicx}
\usepackage{booktabs} 

\usepackage{amssymb,bm,amsthm}
\usepackage{enumerate}
\usepackage{color, colortbl}
\usepackage{booktabs, multirow}
\definecolor{darkgreen}{rgb}{0, 0.5, 0}
\definecolor{red}{rgb}{1, 0, 0}
\definecolor{purple}{rgb}{0.5, 0, 0.5}
\usepackage{chngpage}
\usepackage{comment}
\usepackage{mfirstuc}
\usepackage{mathtools}
\usepackage{lettrine}

\newcommand\ie{\textit{i.e.,}}
\newcommand\eg{\textit{e.g.,}}

\newcommand\etc{\textit{etc.}}
\newcommand\iid{\textit{i.i.d.}}

\newcommand{\red}{\textcolor{red}}

\newcommand{\beq}{\begin{equation}}
\newcommand{\eeq}{\end{equation}}
\newcommand{\beqnn}{\begin{equation*}}
\newcommand{\eeqnn}{\end{equation*}}
\newcommand{\beqy}{\begin{eqnarray}}
\newcommand{\eeqy}{\end{eqnarray}}
\newcommand{\beqynn}{\begin{eqnarray*}}
\newcommand{\eeqynn}{\end{eqnarray*}}
\newcommand{\bit}{\begin{itemize}}
\newcommand{\eit}{\end{itemize}}
\newcommand{\ben}{\begin{enumerate}}
\newcommand{\een}{\end{enumerate}}
\newcommand{\bex}{\begin{example}}
\newcommand{\eex}{\end{example}}
\newcommand{\trace}{\mathrm{trace}}

\newcommand{\balg}[1]{\begin{algorithm} \caption{#1}}
\newcommand{\ealg}{\end{algorithm}}

\newcommand{\balgc}{\begin{algorithmic}[1]}
\newcommand{\ealgc}{\end{algorithmic}}

\newcommand{\bary}{\begin{array}}
\newcommand{\eary}{\end{array}}
\newcommand{\bmx}{\begin{bmatrix}}
\newcommand{\emx}{\end{bmatrix}}
\newcommand{\bsmx}{\left[\begin{smallmatrix}}
\newcommand{\esmx}{\end{smallmatrix}\right]}
\newcommand{\bmxc}[1]{\left[\begin{array}{@{}#1@{}}}
\newcommand{\emxc}{\end{array}\right]}
\newcommand{\bcn}{\begin{center}}
\newcommand{\ecn}{\end{center}}

\newcommand{\diag}{\mathrm{diag}}

\newcommand{\Rbb}{{\mathbb{R}}}

\newcommand{\e}{\boldsymbol{e}}

\renewcommand{\u}{\boldsymbol{u}}

\newcommand{\x}{{\boldsymbol{x}}}

\providecommand{\abs}[1]{\left| #1 \right|}

\newenvironment{theorem}[2][Theorem]{\begin{trivlist}
		\item[\hskip \labelsep {\bfseries #1}\hskip \labelsep {\bfseries #2.}]}{\end{trivlist}}

\newenvironment{proposition}[2][Proposition]{\begin{trivlist}
		\item[\hskip \labelsep {\bfseries #1}\hskip \labelsep {\bfseries #2.}]}{\end{trivlist}}

\usepackage{amsmath,amsfonts,bm}

\def\eqref#1{equation~\ref{#1}}

\def\1{\bm{1}}

\DeclareMathAlphabet{\mathsfit}{\encodingdefault}{\sfdefault}{m}{sl}
\SetMathAlphabet{\mathsfit}{bold}{\encodingdefault}{\sfdefault}{bx}{n}

\usepackage{subfig}
\usepackage{floatrow}
\usepackage{wrapfig}
\usepackage{nccmath}

\newtheorem{definition}{Definition}

\usepackage{amsmath}
\usepackage{amssymb}
\usepackage{mathtools}
\usepackage{amsthm}

\usepackage[capitalize,noabbrev]{cleveref}

\title{Revisiting Heterophily For Graph Neural Networks}

\author{
Sitao Luan$^{1,2}$, Chenqing Hua$^{1,2}$, Qincheng Lu$^{1}$, Jiaqi Zhu$^{1}$, Mingde Zhao$^{1,2}$, Shuyuan Zhang$^{1,2}$,\\\textbf{Xiao-Wen Chang$^{1}$, Doina Precup$^{1,2,3}$}\\
\{sitao.luan@mail, chenqing.hua@mail, qincheng.lu@mail, jiaqi.zhu@mail, mingde.zhao@mail,\\
shuyuan.zhang@mail, chang@cs, dprecup@cs\}.mcgill.ca\\
$^1$McGill University; $^2$Mila; $^3$DeepMind\\

}

\begin{document}

\maketitle

\vspace{-0.3cm}
\begin{abstract}
Graph Neural Networks (GNNs) extend basic Neural Networks (NNs) by using graph structures based on the relational inductive bias (homophily assumption). While GNNs have been commonly believed to outperform NNs in real-world tasks, recent work has identified a non-trivial set of datasets where their performance compared to NNs is not satisfactory. Heterophily has been considered the main cause of this empirical observation and numerous works have been put forward to address it. In this paper, we first revisit the widely used homophily metrics and point out that their consideration of only graph-label consistency is a shortcoming. Then, we study heterophily from the  perspective of post-aggregation node similarity and define new homophily metrics, which are potentially advantageous compared to existing ones. Based on this investigation, we prove that some harmful cases of heterophily can be effectively addressed by local diversification operation. Then, we propose the Adaptive Channel Mixing (ACM), a framework to adaptively exploit aggregation, diversification and identity channels node-wisely to extract richer localized information for diverse node heterophily situations. ACM is more powerful than the commonly used uni-channel framework for node classification tasks on heterophilic graphs and is easy to be implemented in baseline GNN layers. When evaluated on $10$ benchmark node classification tasks, ACM-augmented baselines consistently achieve significant performance gain, exceeding state-of-the-art GNNs on most  tasks without incurring significant computational burden.
\end{abstract}

\vspace{-0.3cm}
\section{Introduction}
\vspace{-0.3cm}
\label{sec:introduction}

Deep Neural Networks (NNs) \cite{lecun2015deep} have revolutionized many machine learning areas, including image recognition \cite{krizhevsky2012imagenet}, speech recognition \cite{graves2013speech} and natural language processing \cite{bahdanau2014neural}, due to their effectiveness in learning latent representations from Euclidean data. Recent research has shifted focus on non-Euclidean data~\cite{bronstein2016geometric}, \eg{} relational data or graphs. Combining graph signal processing and convolutional neural networks~\cite{lecun1998gradient}, numerous Graph Neural Network (GNN) architectures have been proposed ~\cite{scarselli2008graph,defferrard2016fast,hamilton2017inductive,velivckovic2017attention,kipf2016classification,luan2019break}, which empirically outperform traditional NNs on graph-based machine learning tasks such as node classification, graph classification, link prediction and graph generation, \etc 
GNNs are built on the homophily assumption~\cite{mcpherson2001birds}: connected nodes tend to share similar attributes with each other \cite{hamilton2020graph}, which offers additional information besides node features. This relational inductive bias~\cite{battaglia2018relational} is believed to be a key factor leading to GNNs' superior performance over NNs' in many tasks. 

However, growing empirical evidence suggests that GNNs are not always advantageous compared to traditional NNs. In some cases, even simple Multi-Layer Perceptrons (MLPs) can outperform GNNs by a large margin  on relational data~\cite{zhu2020beyond,liu2021non, luan2020complete,chien2021adaptive}. An important reason for this is believed to be the heterophily problem: the homophily assumption does not always hold, so connected nodes may in fact have different attributes. Heterophily has received lots of attention recently and an increasing number of models have been put forward to address this problem  \cite{zhu2020beyond,liu2021non,luan2020complete,chien2021adaptive,zhu2020graph,yan2021two,ma2021homophily, he2021bernnet,li2022finding}.  
In this paper, we first show that by only considering graph-label consistency, existing homophily metrics are not able to describe the effect of some cases of heterophily on aggregation-based GNNs. 
We propose a post-aggregation node similarity matrix,
and based on it, we derive new homophily metrics, whose advantages are illustrated on synthetic graphs (Sec.~\ref{sec:heterophily_analysis}). Then, we prove that diversification operation can help to address some harmful cases of heterophily (Sec.~\ref{sec:acm_framework}). Based on this, we 
propose the Adaptive Channel Mixing (ACM) GNN framework which augments uni-channel baseline GNNs, allowing them to exploit aggregation, diversification and identity channels adaptively, node-wisely and locally in each layer. ACM significantly boosts the performance of 3 uni-channel baseline GNNs by 2.04\% $\sim$ 27.5\% for node classification tasks on 7 widely used benchmark heterophilic graphs, exceeding SOTA models (Sec.~\ref{sec:experiments}) on all of them. For 3 homophilic graphs, ACM-augmented GNNs can perform at least as well as the uni-channel baselines and are competitive compared with SOTA.
\vspace{-0.3cm}
\paragraph{Contributions} 1. To our knowledge, we are the first to analyze heterophily from post-aggregation node similarity perspective. 2. The proposed ACM framework is highly different from adaptive filterbank with multiple channels and existing GNNs for heterophily: 1) the traditional adaptive filterbank channels \cite{vary2006adaptive} uses a scalar weight for each filter and this weight is shared by all nodes. In contrast, ACM provides a mechanism so that different nodes can learn different weights to utilize information from different channels to account for diverse local heterophily; 2) Unlike existing methods that leverage the high-order filters and global property of high-frequency signals \cite{zhu2020beyond,liu2021non,chien2021adaptive, he2021bernnet} which require more computational resources, ACM successfully addresses heterophily by considering only the \textbf{nodewise local information adaptively}.
3. Unlike existing methods that try to facilitate learning filters with high expressive power \cite{zhu2020beyond,zhu2020graph,chien2021adaptive, he2021bernnet}, ACM aims that, when given a filter with certain expressive power, we can extract richer information from additional channels in a certain way to address heterophily. This makes ACM more flexible and easier to be implemented.
\vspace{-0.3cm}
\section{Preliminaries}
\vspace{-0.3cm}
\label{sec:prelimiary_notation}
In this section, we introduce notation and background knowledge. We use \textbf{bold} font for vectors (\eg{} $\bm{v}$). Suppose we have an undirected connected graph $\mathcal{G}=(\mathcal{V},\mathcal{E}, A)$, where $\mathcal{V}$ is the node set with $\abs{\mathcal{V}}=N$; $\mathcal{E}$ is the edge set without self-loops; $A \in \mathbb{R}^{N\times N}$ is the symmetric adjacency matrix with $A_{i,j}=1$ \textit{if} $e_{ij} \in \mathcal{E}$,
otherwise $A_{i,j}=0$.
Let $D$ denote the diagonal degree matrix of ${\cal G}$, \ie{} $D_{i,i} = d_i = \sum_j A_{i,j}$. Let $\mathcal{N}_i$ denote the neighborhood set of node $i$, \ie{}  $\mathcal{N}_i=\{j: e_{ij} \in \mathcal{E}\}$. A graph signal is a vector $\bm{x} \in \mathbb{R}^N$ defined on $\mathcal{V}$, where $\bm{x}_i$ is associated with node $i$. We also have a feature matrix ${X} \in \mathbb{R}^{N\times F}$, whose columns are graph signals and whose $i$-th row  ${X_{i,:}}$ is a feature vector of node $i$. We use $Z\in \mathbb{R}^{N\times C}$ to denote the label encoding matrix, whose $i$-th row  $Z_{i,:}$ is the one-hot encoding of the label of node $i$. 

\subsection{Graph Laplacian, Affinity Matrix and Variants} 
\vspace{-0.1cm}
\label{sec:laplacian_affinity_matrix}
The (combinatorial) graph Laplacian is defined as $L = D - A$, which is Symmetric Positive Semi-Definite (SPSD)  \cite{chung1997spectral}. Its eigendecomposition is $L=U\Lambda U^T$, where the columns $\u_i$ of $U\in \Rbb^{N\times N}$ are orthonormal eigenvectors, namely the \textit{graph Fourier basis}, $\Lambda = \diag(\lambda_1, \ldots, \lambda_N)$ with $\lambda_1 \leq \cdots \leq \lambda_N$. These eigenvalues are also called \textit{frequencies}. 

In additional to $L$, some variants are also commonly used, \eg{} the symmetric normalized Laplacian $L_{\text{sym}} = D^{-1/2} L D^{-1/2} = I-D^{-1/2} A D^{-1/2}$ and the random walk normalized Laplacian $L_{\text{rw}} = D^{-1} L = I - D^{-1} A$. The graph Laplacian and its variants can be considered as high-pass filters for graph signals. 
The affinity (transition) matrices can be derived from the Laplacians, \eg{} $A_\text{rw} = I - L_\text{rw} = D^{-1} A$, $A_\text{sym} = I-L_\text{sym} = D^{-1/2} A D^{-1/2}$ and are considered to be low-pass filters \cite{maehara2019revisiting}.
Their eigenvalues satisfy $\lambda_i(A_\text{rw}) = \lambda_i(A_\text{sym}) = 1- \lambda_i(L_\text{sym}) = 1- \lambda_i(L_\text{rw}) \in (-1,1]$. 
Applying the renormalization trick \cite{kipf2016classification} to affinity and Laplacian matrices respectively leads to $\hat{A}_\text{sym} = \tilde{D}^{-1/2} \tilde{A} \tilde{D}^{-1/2}$ and $\hat{L}_{\text{sym}} = I - \hat{A}_\text{sym}$, 
where $\tilde{A} \equiv A+I$ and $\tilde{D} \equiv D+I$. The renormalized affinity matrix essentially adds a self-loop to each node in the graph, and is widely used in Graph Convolutional Network (GCN) \cite{kipf2016classification} as follows:
\begin{equation}
    \label{eq:gcn_original}
   Y = \text{softmax} (\hat{A}_\text{sym} \; \text{ReLU} (\hat{A}_\text{sym} {X} W_0 ) \; W_1 )
\end{equation}
where $W_0 \in \Rbb^{F\times F_1}$ and $W_1 \in \Rbb^{F_1\times O}$ are learnable parameter matrices. GCNs can be trained by minimizing the following cross entropy loss

\begin{equation}
\label{eq:cross_entropy_loss}
     \mathcal{L}  = -\trace(Z^T \log Y)
\end{equation}
where $\log(\cdot)$ is a component-wise logarithm operation. The random walk renormalized matrix $\hat{A}_{\text{rw}} = \tilde{D}^{-1} \tilde{A}$,
which shares the same eigenvalues as $\hat{A}_{\text{sym}}$, can also be applied in GCN. The corresponding Laplacian is defined as $\hat{L}_{\text{rw}} = I - \hat{A}_\text{rw}$. The matrix $\hat{A}_{\text{rw}}$ is essentially a random walk matrix and
behaves as a mean aggregator that is applied in spatial-based GNNs \cite{hamilton2017inductive,hamilton2020graph}. To bridge spectral and spatial methods, we use $\hat{A}_{rw}$ in this paper.

\subsection{Metrics of Homophily}
\vspace{-0.1cm}

\label{sec:homophily_metrics}
The homophily metrics are defined by considering different relations between node labels and graph structures. There are three commonly used homophily metrics: edge homophily \cite{abu2019mixhop,zhu2020beyond}, node homophily \cite{pei2020geom} and class homophily \cite{lim2021new}~\footnote{\cite{lim2021new} did not name this homophily metric. We named it \textit{class homophily} based on its definition.}, defined as follows:
\begin{equation}
\begin{aligned}
\label{eq:definition_homophily_metrics}
    H_\text{edge}(\mathcal{G}) &= \frac{\big|\{e_{uv} \mid e_{uv}\in \mathcal{E}, Z_{u,:}=Z_{v,:}\}\big|}{|\mathcal{E}|}, 
    H_\text{node}(\mathcal{G}) = \frac{1}{|\mathcal{V}|} \sum_{v \in \mathcal{V}}  H_\text{node}^v= \frac{1}{|\mathcal{V}|} \sum_{v \in \mathcal{V}} 
    \frac{\big|\{u \mid u \in \mathcal{N}_v, Z_{u,:}=Z_{v,:}\}\big|}{d_v}, \\
    H_\text{class}(\mathcal{G}) &\!=\! \frac{1}{C\!-\!1} \sum_{k=1}^{C}\bigg[h_{k}
    \!-\! \frac{\big|\{v \!\mid\! Z_{v,k} \!=\! 1 \}\big|}{N}\bigg]_{+}, \ \
h_{k}\! =\! \frac{\sum_{v \in \mathcal{V}} \big|\{u \!\mid\! Z_{v,k}\! =\! 1, u \in \mathcal{N}_v, Z_{u,:}\!=\!Z_{v,:}\}\big| }{\sum_{v \in \{v|Z_{v,k}=1\}} d_{v}}
\end{aligned}
\end{equation}
where $H_\text{node}^v$ is the local homophily value for node $v$; $[a]_{+}=\max (a, 0)$; $h_{k}$ is the class-wise homophily metric \cite{lim2021new}. All metrics are in the range of $[0,1]$; a value close to $1$ corresponds to strong homophily, while a value close to $0$ indicates strong heterophily. $H_\text{edge}(\mathcal{G})$ measures the proportion of edges that connect two nodes in the same class; $H_\text{node}(\mathcal{G})$ evaluates the average proportion of edge-label consistency of all nodes; $H_\text{class}(\mathcal{G})$ tries to avoid sensitivity to imbalanced classes, which can make $H_\text{edge}(\mathcal{G})$ misleadingly large. The above definitions are all based on the \textbf{linear feature-independent graph-label consistency}. The inconsistency relation is implied to have a negative effect to the performance of GNNs. With this in mind, in the following section, we give an example to illustrate the shortcomings of the above metrics and propose new feature-independent metrics that are defined from post-aggregation node similarity perspective, which is novel.

\begin{wrapfigure}{R}{0.5\textwidth}
  \begin{center}
    \includegraphics[width=1.0\textwidth]{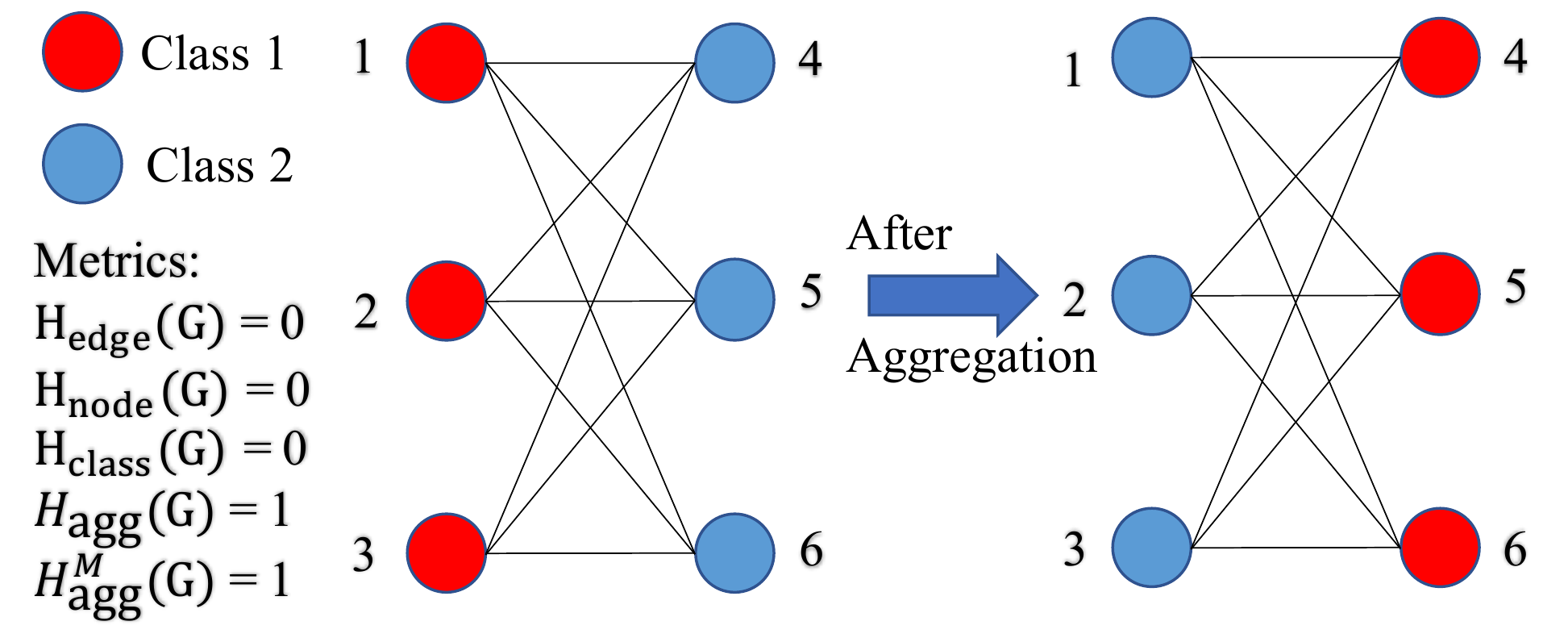}
  \end{center}
  \caption{Example of harmless heterophily}
  \label{fig:example_harmless_heterophily}
\end{wrapfigure}

\vspace{-0.3cm}
\section{Analysis of Heterophily}
\vspace{-0.2cm}
\label{sec:heterophily_analysis}
\subsection{Motivation and Aggregation Homophily}
\vspace{-0.1cm}
Heterophily is widely believed to be harmful for message-passing based GNNs \cite{zhu2020beyond,pei2020geom,chien2021adaptive} because, intuitively, features of nodes in different classes will be falsely mixed,
leading nodes to be indistinguishable \cite{zhu2020beyond}.
Nevertheless, it is not always the case, \eg{} the bipartite graph\footnote{\cite{ma2021homophily} use the same example but not to demonstrate the deficiency of homophily metrics.}  shown in Figure \ref{fig:example_harmless_heterophily} is highly heterophilic according to the existing homophily metrics in \eqref{eq:definition_homophily_metrics}, but after mean aggregation, the nodes in classes 1 and 2 just exchange colors and are still distinguishable\footnote{\cite{chien2021adaptive} also point out the insufficiency of $H_\text{node}$ by examples to show that different graph typologies with the same $H_\text{node}(\mathcal{G})$ can carry different label information.}. This example tells us that, besides graph-label consistency, we need to study the relation between nodes after aggregation step. To this end, we first define the post-aggregation node similarity matrix as follows:
\begin{equation}
\label{eq:post_agg_similarity_matrix}
    S(\hat{A},X) \equiv \hat{A}X (\hat{A}X)^T \in \mathbb{R}^{N\times N}
\end{equation}
where $\hat{A} \in \mathbb{R}^{N\times N}$ denotes a general aggregation operator. $S(\hat{A},X)$ is essentially the gram matrix that measures the similarity between each pair of aggregated node features.

\vspace{-0.2cm}
\paragraph{Relationship Between $S(\hat{A},X)$ and Gradient of SGC}
SGC \cite{wu2019simplifying} is one of the most simple but representative GNN models and its output can be written as:
\begin{equation}
    \begin{aligned}
    Y  = \text{softmax} (\hat{A}  X W ) = \text{softmax} (Y')
    \end{aligned}
\end{equation}
With the loss function in \eqref{eq:cross_entropy_loss}, after each gradient descent step, we have $\Delta W = \gamma \frac{d \mathcal{L}}{d W}$, where $\gamma$ is the learning rate. The update of $Y'$ is (see Appendix \ref{appendix:details_of_nll_loss_explanation} for  derivation):
\begin{equation}
    \begin{aligned}
    \label{eq:gradient_descent_update}
    \Delta Y' & = \hat{A} X \Delta W = \gamma \hat{A}X \frac{d \mathcal{L}}{d W} \propto \hat{A}X \frac{d \mathcal{L}}{d W} = \hat{A}X X^T\hat{A}^T (Z-Y) = S(\hat{A},X) (Z-Y)
    \end{aligned}
\end{equation}
where $Z-Y$ is the prediction error matrix. The update direction of the prediction for node $i$ is essentially a weighted sum of the prediction error, \ie{} $\Delta (Y')_{i,:} = \sum_{j\in \mathcal{V}} \big[S(\hat{A},X)\big]_{i,j} (Z-Y)_{j,:}$ and $\big[S(\hat{A},X)\big]_{i,j}$ can be considered as the weights. Intuitively, a high similarity value $\big[S(\hat{A},X)\big]_{i,j}$ means node $i$ tends to be updated to the same class as node $j$. This indicates that $S(\hat{A},X)$ is closely related to a single layer GNN model.

Based on the above definition and observation, we define the aggregation similarity score as follows.


\begin{definition} The aggregation similarity score is: 
\begin{equation}
\begin{aligned}
\label{eq:aggregation_similarity}
     & S_\text{agg}\big(S(\hat{A},X)\big)  \\
    &  = \frac{1}{\left| \mathcal{V} \right|} \left| \left\{v   \,\big| \,
    \mathrm{Mean}_u\big( \{S(\hat{A},X)_{v,u} | Z_{u,:}=Z_{v,:} \}\big) 
    \geq  \mathrm{Mean}_u\big(\{S(\hat{A},X)_{v,u} | Z_{u,:} \neq Z_{v,:} \} \big) \right\} \right|
\end{aligned}
\end{equation}
where $\mathrm{Mean}_u\left(\{\cdot\}\right)$ takes the average over $u$ of a given multiset of values or variables.
\end{definition}
$S_\text{agg}(S(\hat{A},X))$ measures the proportion of nodes $v\in\mathcal{V}$ as which the average weights on the set of nodes in the same class (including $v$) is larger than that 
in other classes. In practice, we observe that in most datasets, we will have $S_\text{agg}(S(\hat{A},X)) \geq 0.5$ \footnote{See Appendix \ref{appendix:extension_of_proposition1} for an intuitive explanation under certain conditions.}. To make the metric range in [0,1], like existing metrics, we rescale \eqref{eq:aggregation_similarity} to the following modified aggregation similarity, 
\begin{equation}
\label{eq:modified_aggregation_homophily}
  S^M_\text{agg}\big(S(\hat{A},X)\big) = \big[2 S_\text{agg}\big(S(\hat{A},X)\big)-1\big]_{+}
\end{equation}

In order to measure the consistency between labels and graph structures without considering node features and to make a fair comparison with the existing homophily metrics in \eqref{eq:definition_homophily_metrics}, we define the graph ($\mathcal{G}$) aggregation ($\hat{A}$) homophily and its modified version \footnote{In practice, we will only check $H_{\text{agg}}(\mathcal{G})$ when $H_{\text{agg}}^M(\mathcal{G})=0$.} as:
\begin{equation}
    \label{eq:aggregation_homophily_metrics}
    H_{\text{agg}}(\mathcal{G}) = S_\text{agg}\big(S(\hat{A},Z)\big)\!, \; H_{\text{agg}}^M(\mathcal{G}) = S_\text{agg}^M\big(S(\hat{A},Z)\big)
\end{equation}
As the example shown in Figure \ref{fig:example_harmless_heterophily}, when $\hat{A} = \hat{A}_\text{rw}$, it is easy to see that $H_{\text{agg}}(\mathcal{G}) = H_{\text{agg}}^M(\mathcal{G}) = 1$ and other metrics are 0. Thus, this new metric reflects the fact that nodes in classes 1 and 2 are still highly distinguishable after aggregation, while other metrics mentioned before fail to capture such information and misleadingly give value 0. This shows the advantage of $H_{\text{agg}}(\mathcal{G})$ and $H_{\text{agg}}^M(\mathcal{G})$, which additionally exploit information from aggregation operator $\hat{A}$ and the similarity matrix.

To comprehensively compare $H_{\text{agg}}^M(\mathcal{G})$ with the existing metrics on their ability to elucidate the influence of graph structure on GNN performance, we generate synthetic graphs with different homophily levels and evaluate SGC \cite{wu2019simplifying} and GCN \cite{kipf2016classification} on them in the next subsection.
\begin{figure*}[htbp]
\centering
    {
     \subfloat[$H_\text{edge}(\mathcal{G})$]{
     \captionsetup{justification = centering}
     \includegraphics[height=0.11\textheight]{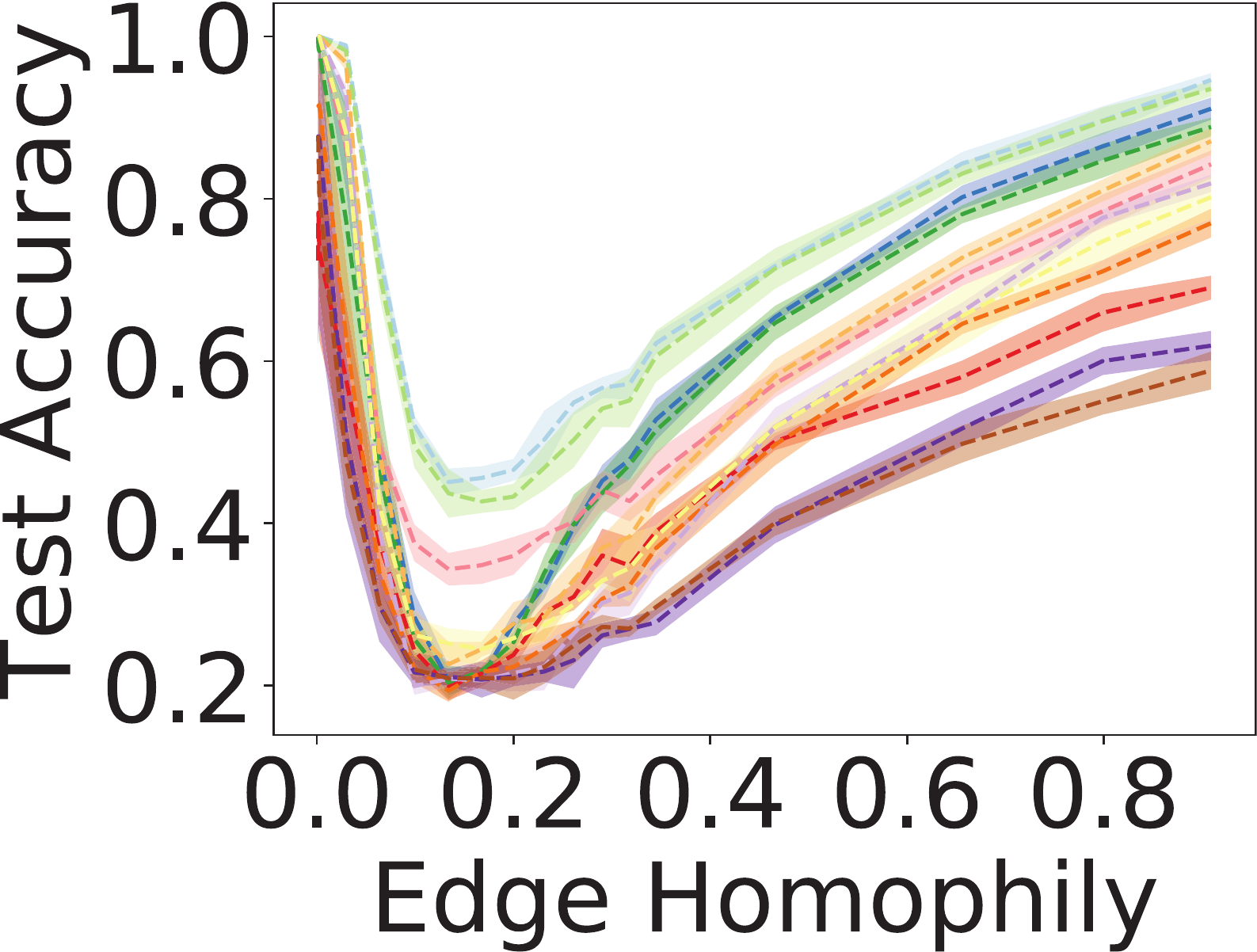}
     } 
     \subfloat[$H_\text{node}(\mathcal{G})$]{
     \captionsetup{justification = centering}
     \includegraphics[height=0.11\textheight]{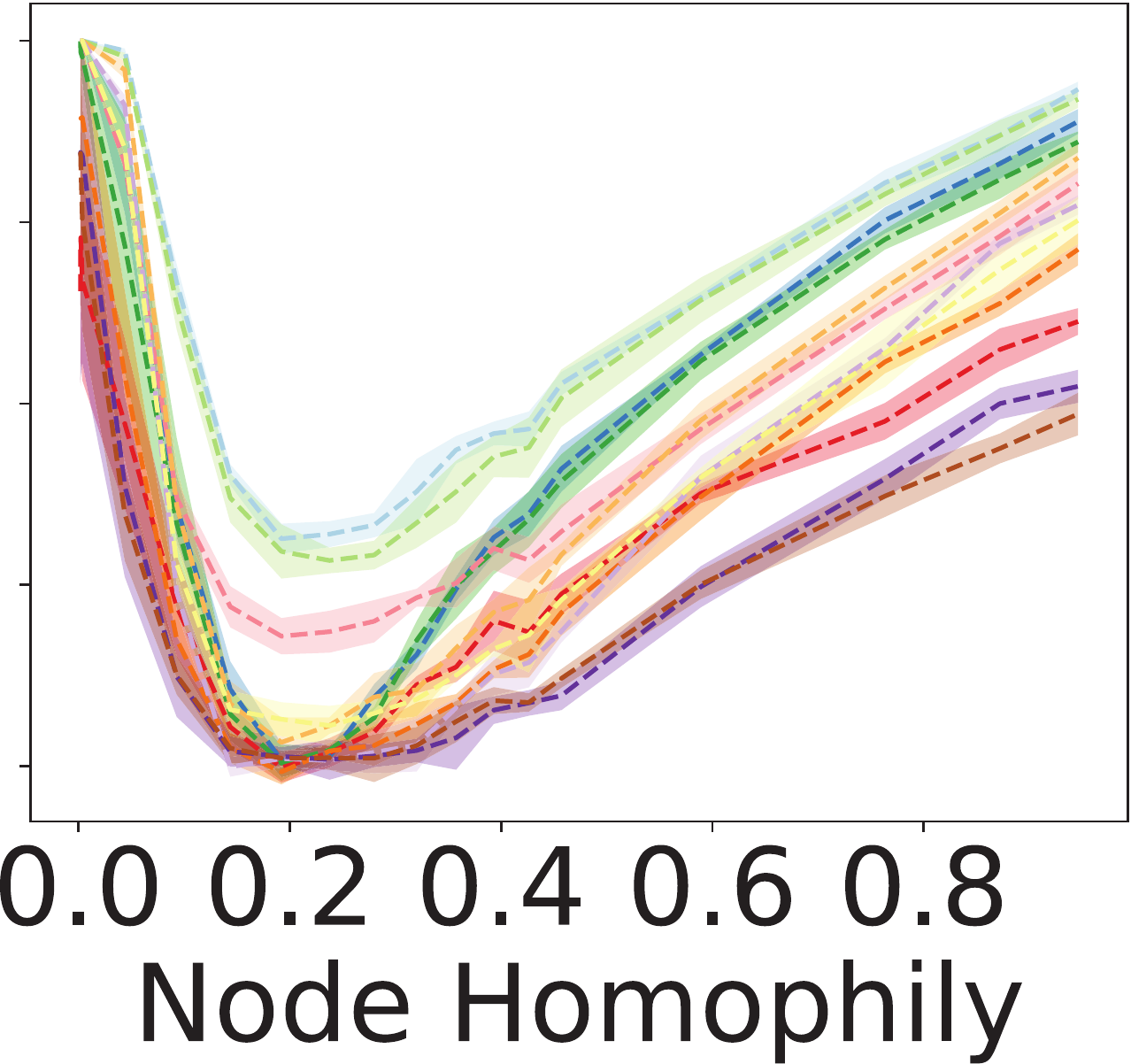}
     }
     \subfloat[$H_\text{class}(\mathcal{G})$]{
     \captionsetup{justification=centering}
     \includegraphics[height=0.11\textheight]{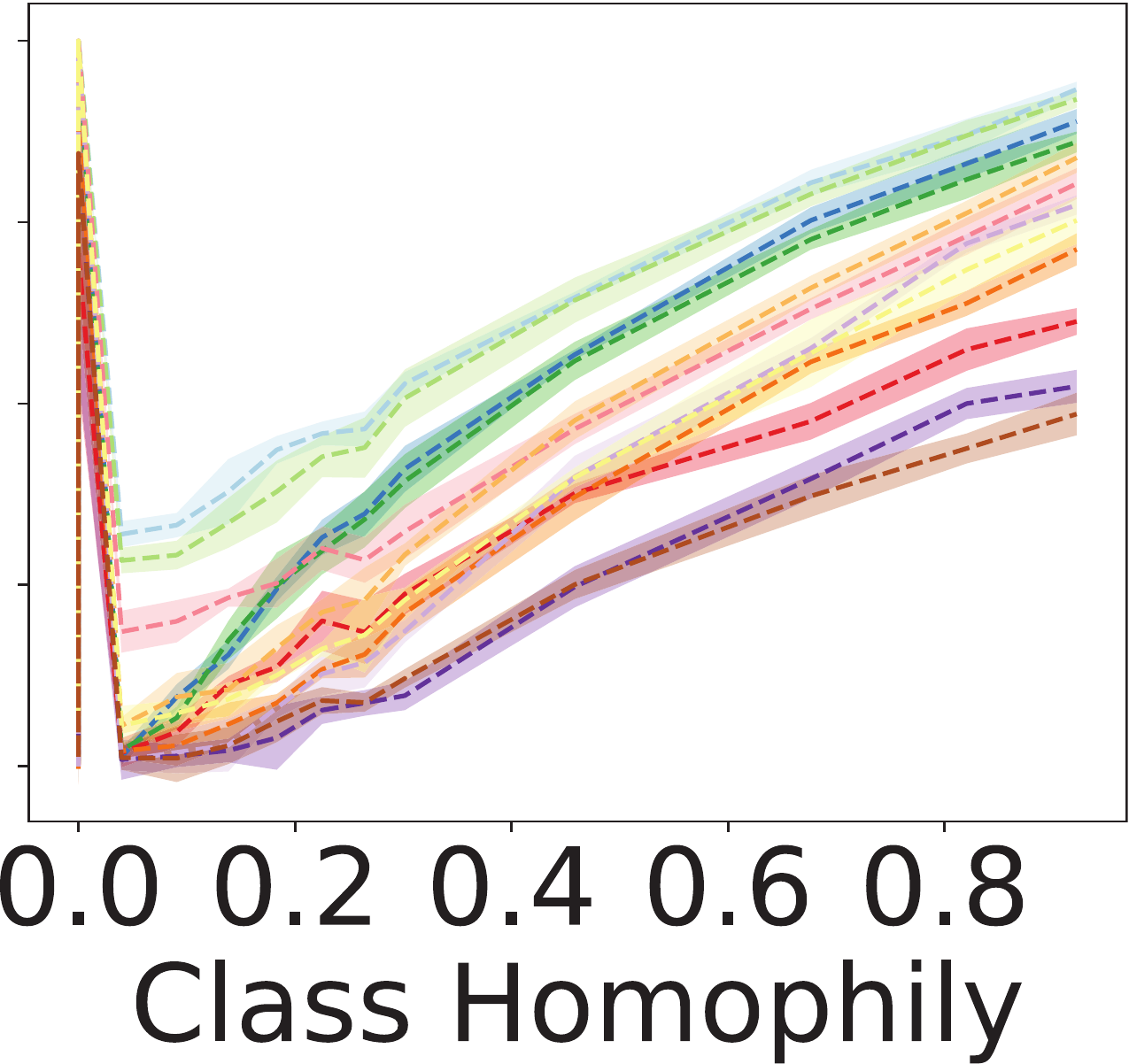}
     } 
     \subfloat[$H_{\text{agg}}^M(\mathcal{G})$]{
     \captionsetup{labelsep=newline,format=plain,indention=-100pt}
     \includegraphics[height=0.1115\textheight]{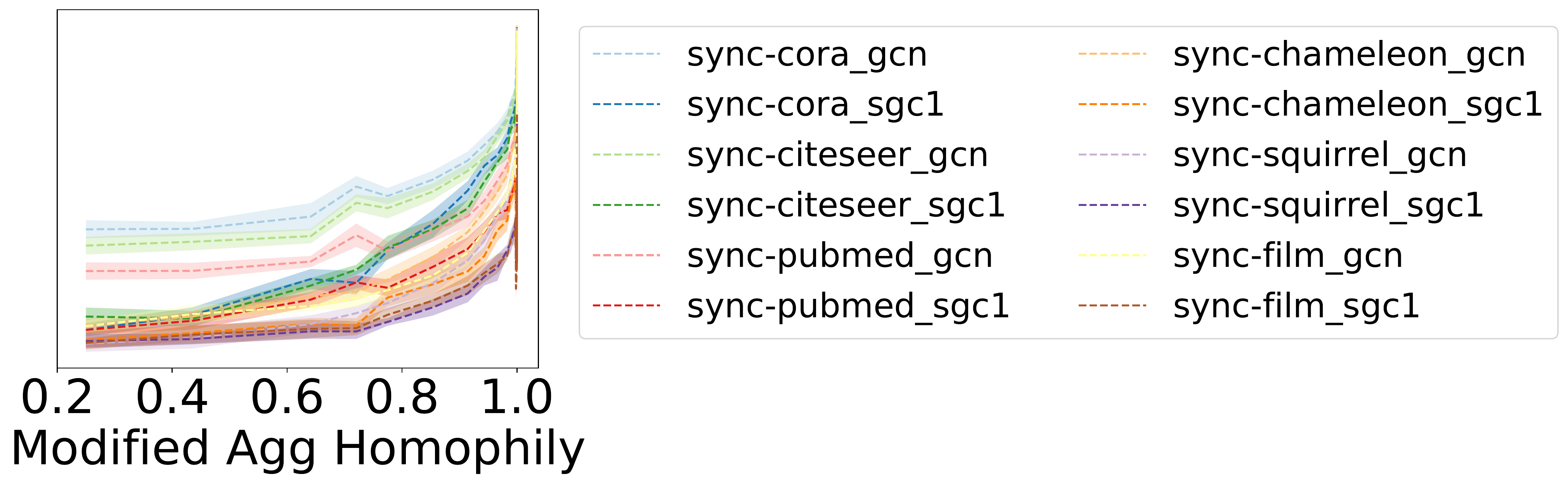}
     }
     }
     \caption{Comparison of baseline performance under different homophily metrics.}
     \label{fig:comparison_homophily_metrics}
\end{figure*}

\subsection{Empirical Evaluation and Comparison on Synthetic Graphs}
\vspace{-0.1cm}
In this subsection, we conduct experiments on synthetic graphs generated with different levels of $H_{\text{edge}}^M(\mathcal{G})$ to assess the output of $H_{\text{agg}}^M(\mathcal{G})$ in comparison with existing metrics.
\vspace{-0.2cm}
\paragraph{Data Generation \& Experimental Setup}
We first generated 10 graphs for each of 28 edge homophily levels, from 0.005 to 0.95, for a total of $280$ graphs. In every generated graph, we had 5 classes, with 400 nodes in each class. For nodes in each class, we randomly generated 800 intra-class edges and [$\frac{800}{H_\text{edge}(\mathcal{G})} -800$] inter-class edges. The features of nodes in each class are sampled from node features in the corresponding class of $6$ base datasets (\textit{Cora, CiteSeer, PubMed, Chameleon, Squirrel, Film}). Nodes were randomly split into  train/validation/test sets, in proportion of 60\%/20\%/20\%. We trained 1-hop SGC (\textit{sgc-1}) \cite{wu2019simplifying} and GCN \cite{kipf2016classification} on the synthetic graphs \footnote{See Appendix \ref{appendix:hyperparameter_space_synthetic_graphs} for a description of the hyperparameter searching range and Appendix \ref{appendix:synthetic_setup_discussion} for more a detailed description of the data generation process}. For each value of $H_\text{edge}(\mathcal{G})$, we take the average test accuracy and standard deviation of runs over the 10 generated graphs with that value. For each generated graph, we also calculate $H_{\text{node}}(\mathcal{G}), H_{\text{class}}(\mathcal{G})$ and $H_{\text{agg}}^M(\mathcal{G})$. Model performance with respect to different homophily values is shown in Figure \ref{fig:comparison_homophily_metrics}.
\vspace{-0.2cm}
\paragraph{Comparison of Homophily Metrics}
The performance of SGC-1 and GCN is expected to be monotonically increasing if the homophily metric is informative. However, Figure \ref{fig:comparison_homophily_metrics}(a)(b)(c) show that the performance curves under $H_{\text{edge}}(\mathcal{G}), H_{\text{node}}(\mathcal{G})$ and $H_{\text{class}}(\mathcal{G})$ are $U$-shaped \footnote{A similar J-shaped curve for $H_\text{edge}(\mathcal{G})$ is found in \cite{zhu2020beyond}, though using different data generation processes. The authors do not mention the insufficiency of edge homophily.}, while Figure \ref{fig:comparison_homophily_metrics}(d) 
reveals a nearly monotonic curve with a little numerical perturbation around 1. This indicates that $H_{\text{agg}}^M(\mathcal{G})$ provides a better indication of the way in which the graph structure affects the performance of SGC-1 and GCN than existing metrics. (See more discussion on aggregation homophily and theoretical results for regular graphs in Appendix \ref{appendix:synthetic_setup_discussion}.)

\vspace{-0.3cm}
\section{Adaptive Channel Mixing (ACM)}
\vspace{-0.2cm}
\label{sec:acm_framework}
In prior work~\cite{luan2020complete,chien2021adaptive,bo2021beyond}, it has been shown that
high-frequency graph signals, which can be extracted by a high-pass filter (HP), is empirically useful for addressing heterophily. In this section, based on the similarity matrix in \eqref{eq:gradient_descent_update}, we theoretically prove that a diversification operation, \ie{} HP filter, can address some cases of harmful heterophily locally. Besides, a node-wise analysis shows that different nodes may need different filters to process their neighborhood information. Based on the above analysis, in Sec.~\ref{sec:filterbank_acm_gnn_architecture} we propose Adaptive Channel Mixing (ACM), a 3-channel architecture which can adaptively exploit local and node-wise information from aggregation, diversification and identity channels. 
\begin{figure*}[t!]
  \begin{center}
    \includegraphics[width=1\textwidth]{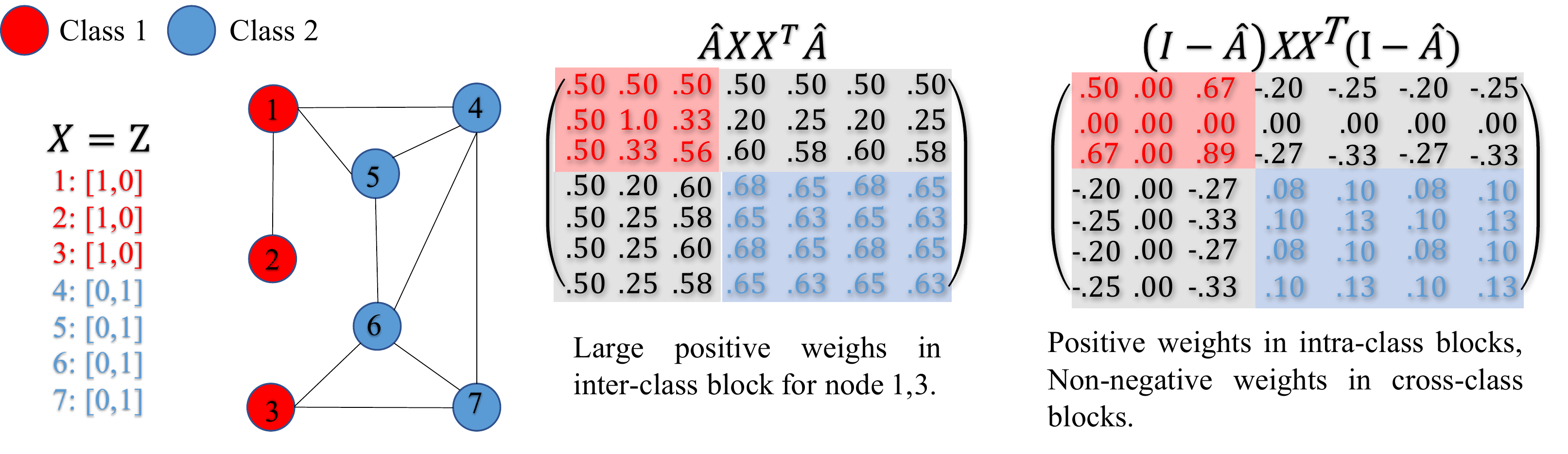}
  \end{center}
  \caption{Example of how diversification can address harmful heterophily}
  \label{fig:successful_example_hp_filter}
\end{figure*}
\vspace{-0.2cm}
\subsection{Diversification Helps with Harmful Heterophily}
\vspace{-0.3cm}
\label{sec:how_diversification_operation_helps}
We first consider the example shown in Figure \ref{fig:successful_example_hp_filter}. From $S(\hat{A},X)$, we can see that nodes $\{1,3\}$ assign relatively large positive weights to nodes in class 2 after aggregation, which will 
make nodes $\{1,3\}$ hard to be distinguished from nodes in class 2. However, we can still 
distinguish nodes $\{1,3\}$ and $\{4,5,6,7\}$ by considering 
their neighborhood differences: nodes $\{1,3\}$ are different from most of their neighbors while nodes $\{4,5,6,7\}$ are similar to most of their neighbors. This indicates that although some nodes become similar after aggregation, they are still distinguishable through their local surrounding dissimilarities. 

This observation leads us to introduce the \textit{diversification operation}, \ie{} HP filter $I-\hat{A}$ \cite{ekambaram2014graph} to extract information regarding neighborhood differences, thereby addressing harmful heterophily. As $S(I-\hat{A},X)$ in Fig.~\ref{fig:successful_example_hp_filter} shows, nodes $\{1,3\}$ will assign negative weights to nodes $\{4,5,6,7\}$ after the diversification operation, 
\ie{} nodes 1,3 treat nodes 4,5,6,7 as negative samples and will move away from them during backpropagation. This example reveals that there are cases in which the diversification operation is helpful to handle heterophily, while the aggregation operation is not. Based on this observation, we first define the diversification distinguishability of a node and the graph diversification distinguishability value,  which measures the proportion of nodes for which the diversification operation is potentially helpful.

\begin{definition} Diversification Distinguishability (DD) based on $S(I-\hat{A},X)$.

Given $S(I-\hat{A},X)$, a node $v$ is diversification distinguishable if the following two conditions are satisfied at the same time,
\vspace{-0.2cm}
\begin{equation}
\label{eq:diversification_distinguishability}
\begin{split}
    \textbf{1.}\ \mathrm{Mean}_u \left(\{S(I-\hat{A},X)_{v,u}|u \in \mathcal{V} \land Z_{u,:}=Z_{v,:}\} \right) \geq 0; \\
 \textbf{2.}\ \mathrm{Mean}_u \left(\{S(I-\hat{A},X)_{v,u}|u \in \mathcal{V} \land Z_{u,:} \neq Z_{v,:}\} \right) \leq 0
    \end{split}
\end{equation}
\vspace{-0.3cm}
Then, graph diversification distinguishability value is defined as

\begin{equation}
\begin{aligned}
    &\mathrm{DD}_{\hat{A},X}(\mathcal{G}) = \frac{1}{\left|\mathcal{V}\right|} \Big|\{v|v \in \mathcal{V} \land v \mbox{ is diversification distinguishable}\}\Big|
\end{aligned}
\end{equation}
\end{definition}
\vspace{-0.2cm}
We can see that $\mathrm{DD}_{\hat{A},X}(\mathcal{G}) \in [0,1]$. Based on Def.~$2$, the effectiveness of diversification in addressing heterophily can be theoretically proved under certain conditions:
\vspace{-0.2cm}
\begin{theorem} 1
(See Appendix \ref{appendix:proof_theorem2} for proof). 
For $C=2$, suppose $X=Z, \hat{A}=\hat{A}_{\text{rw}}$. Then for any $I-\hat{A}_{\text{rw}}$, all nodes are diversification distinguishable and $\mathrm{DD}_{\hat{A},Z}(\mathcal{G})=1$.
\end{theorem}
\vspace{-0.2cm}
With the above results for HP filters, we will now introduce the concept of filterbank which combines both LP (aggregation) and HP (diversification) filters and can potentially handle various local heterophily cases. We then develop ACM framework in the following subsection.
\vspace{-0.3cm}
\subsection{Filterbank and Adaptive Channel Mixing (ACM) Framework}
\vspace{-0.1cm}
\label{sec:filterbank_acm_gnn_architecture}
\begin{wrapfigure}{R}{0.3\textwidth}
  \begin{center}
    \includegraphics[width=1\textwidth]{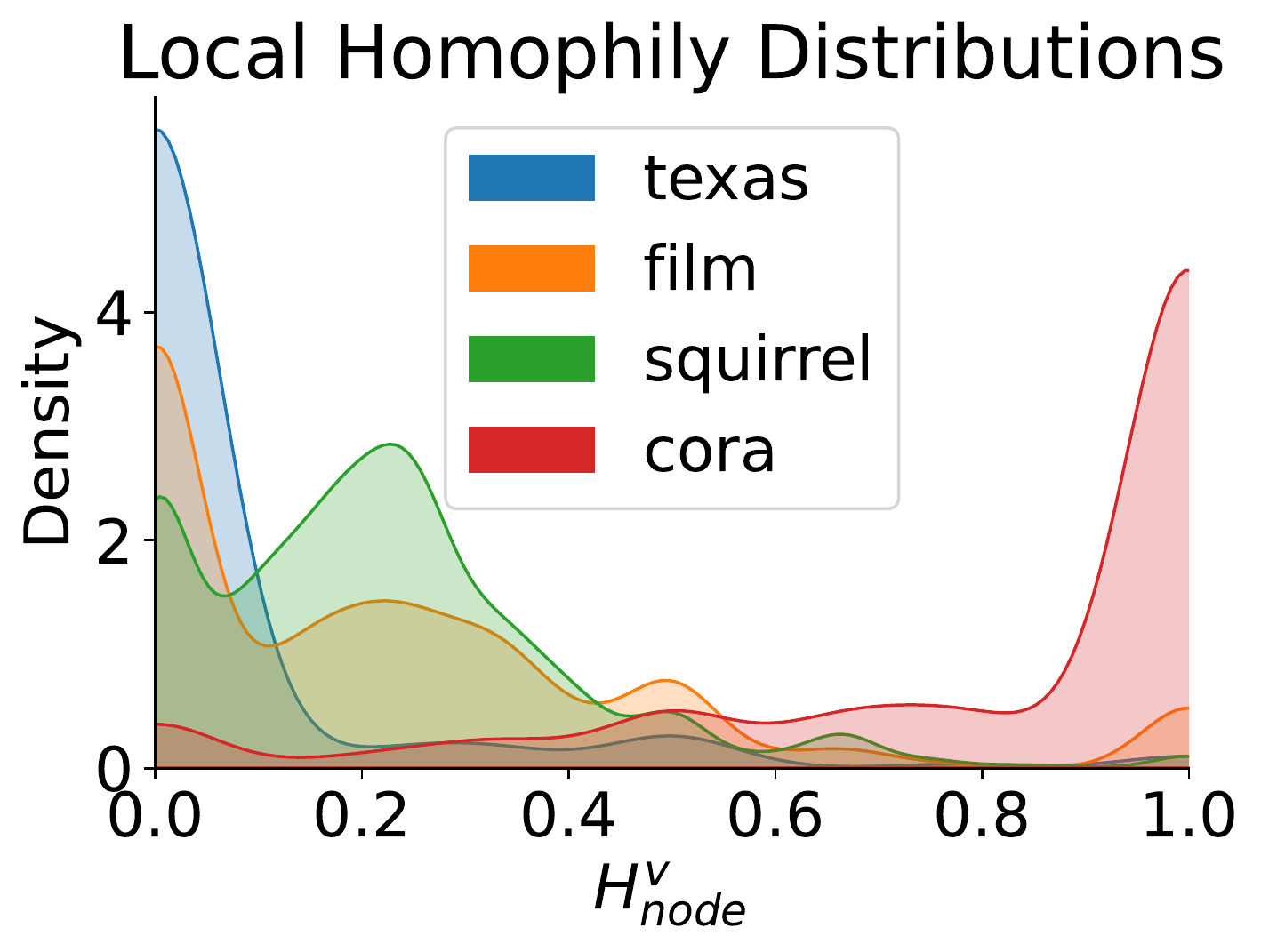}
  \end{center}
  \caption{$H_\text{node}^v$ distributions}
  \label{fig:local_homophily_4}
\end{wrapfigure}
\paragraph{Filterbank} For the graph signal $\bm{x}$ defined on $\mathcal{G}$, a 2-channel linear (analysis) filterbank \cite{ekambaram2014graph} \footnote{In graph signal processing, an additional synthesis filter \cite{ekambaram2014graph} is required to form the 2-channel filterbank. But a synthesis filter is not needed in our framework.} includes a pair of  filters $H_\text{LP}, H_\text{HP}$, which retain the low-frequency and high-frequency content of $\bm{x}$, respectively.
Most existing GNNs use a uni-channel filtering architecture \cite{kipf2016classification,velivckovic2017attention,hamilton2017inductive} with either LP or HP channel, which only partially preserves the input information. Unlike the uni-channel architecture, filterbanks with $H_\text{LP} + H_\text{HP} = I$ do not lose any information from the input signal, which is called the perfect reconstruction property \cite{ekambaram2014graph}. Generally, the Laplacian matrices ($L_\text{sym}$, $L_\text{rw}$, $\hat{L}_\text{sym}$, $\hat{L}_\text{rw}$) can be regarded as HP filters \cite{ekambaram2014graph} and affinity matrices ($A_\text{sym}$, $A_\text{rw}$, $\hat{A}_\text{sym}$, $\hat{A}_\text{rw}$) can be treated as LP filters \cite{maehara2019revisiting, hamilton2020graph}. Moreover,  we extend the concept of filterbank and view MLPs as using the identity (full-pass) filterbank with $H_\text{LP}=I$ and $H_\text{HP}=0$, which also satisfies $H_\text{LP} + H_\text{HP} = I+0 = I$. 
\vspace{-0.3cm}
\paragraph{Node-wise Channel Mixing for Diverse Local Homophily} The example in Figure \ref{fig:successful_example_hp_filter} also shows that different nodes may need the local information extracted from different channels, \eg{} nodes $\{1,3\}$ demand information from the HP channel while node 2 only needs information from the LP channel. Figure \ref{fig:local_homophily_4} reveals that nodes have diverse distributions of node local homophily $H_\text{node}^v$ across different datasets. In order to adaptively leverage the LP, HP and identity channels in GNNs to deal with the diverse local heterophily situations, we will now describe our proposed Adaptive Channel Mixing (ACM) framework.
\vspace{-0.2cm}
\paragraph{Adaptive Channel Mixing (ACM)}
We will use GCN \footnote{See more variants in Appendix \ref{appendix:details_implementation_acm_acmII}.} as an example to introduce the ACM framework in matrix form, but the framework can be combined in a similar manner to many different GNNs.
The ACM framework includes the following steps:
\vspace{-0.1cm}
\begin{equation*}
\begin{aligned}
\label{eq:acm_gnn_spectral}
& \textbf{{Step 1. Feature Extraction for Each Channel:}} \\
& \text{Option 1: } {H}^{l}_L = \text{ReLU}\left(H_\text{LP} {H^{l-1}} W^{l-1}_L\right), {{H}^{l}_H} =  \text{ReLU}\left(H_\text{HP} {H^{l-1}} W^{l-1}_H\right), {H}^{l}_I  = \ \text{ReLU}\left(I {H^{l-1}} W^{l-1}_I\right); \\
& \text{Option 2: } {H}^{l}_L = H_\text{LP} \text{ReLU}\left({H^{l-1}} W^{l-1}_L\right), {{H}^{l}_H} = H_\text{HP} \text{ReLU}\left({H^{l-1}} W^{l-1}_H\right), {H}^{l}_I  = I\ \text{ReLU}\left({H^{l-1}} W^{l-1}_I\right); \\
& H^0=X \in \mathbb{R}^{N\times F_0}, \ W_L^{l-1},\ W_H^{l-1}, \ W_I^{l-1} \in \mathbb{R}^{F_{l-1} \times F_l},\ l = 1,\dots,L;\\
& \textbf{Step 2. Row-wise Feature-based Weight Learning} \\
&\tilde{\alpha}_L^l = \text{Sigmoid} \left({H}^{l}_L \tilde{W}^{l}_L\right),\ \tilde{\alpha}_H^l = \text{Sigmoid} \left({H}^{l}_H \tilde{W}^{l}_H\right), \tilde{\alpha}_I^l = \text{Sigmoid} \left({H}^{l}_I \tilde{W}^{l}_I\right),\ \tilde{W}_L^{l-1},\ \tilde{W}_H^{l-1},\ \tilde{W}_I^{l-1} \in \mathbb{R}^{F_l \times 1}\\ 
& \left[{\alpha}_L^l, {\alpha}_H^l, {\alpha}_I^l \right] = \text{Softmax}\left((\left[\tilde{\alpha}_L^l,\tilde{\alpha}_H^l,\tilde{\alpha}_I^l\right]/T) W_\text{Mix}^l \right) \in \mathbb{R}^{N\times 3}, T \in \mathbb{R} \text{ temperature},\ W_\text{Mix}^l \in \mathbb{R}^{3\times 3}; \\
&\textbf{Step 3. Node-wise Adaptive Channel Mixing:}&\\
&{H^{l}}  =  \text{ReLU}\left(\text{diag}(\alpha_L^l){H}^{l}_L + \text{diag}(\alpha_H^l){H}^{l}_H + \text{diag}(\alpha_I^l){H}^{l}_I \right)&
\end{aligned}
\end{equation*}
We will refer to the instantiation which uses option 1 in step 1 as ACM and to the one using option 2 as ACMII. In step 1, ACM(II)-GCN implement different feature extractions for $3$ channels using a set of filterbanks. Three filtered components, $H_L^l,H_H^l, H_I^l$, are obtained. To adaptively exploit information from each channel, ACM(II)-GCN first extract nonlinear information from the filtered signals, then use $W_\text{Mix}^l$ to learn which channel is important for each node, leading to the row-wise weight vectors $\alpha_L^l,\alpha_H^l,\alpha_I^l \in \mathbb{R}^{N\times 1}$ whose $i$-th elements are the weights for node $i$ \footnote{See Appendix \ref{appendix:ablation_w_mix} and \ref{appendix:raw_combined_feature_comparison} for more discussion of the components in ACM architecture.}. These three vectors are then used as weights in defining the updated $H^l$ in step 3.

\vspace{-0.2cm}
\paragraph{Complexity} The number of learnable parameters in layer $l$ of ACM(II)-GCN is $3F_{l-1}(F_l +1)+9$, compared to $F_{l-1}F_l$ in GCN. The computation of steps 1-3 takes $NF_l(8+6F_{l-1}) + 2F_l(\text{nnz}(H_\text{LP}) + \text{nnz}(H_\text{HP}))+ 18N$ flops, while the GCN layer takes $2NF_{l-1}F_l + 2F_l(\text{nnz}(H_\text{LP}))$ flops, where $\text{nnz}(\cdot)$ is the number of non-zero elements. An ablation study and a detailed comparison on running time are conducted in Sec.~\ref{sec:ablation_tests_running_time}.
\vspace{-0.2cm}
\paragraph{Limitations of Diversification} Like any other method, there exists some cases of harmful heterophily that diversification operation cannot work well. For example, suppose we have an imbalanced dataset where several small clusters with distinctive labels are densely connected to a large cluster. In this case, the surrounding differences of nodes in small clusters are similar, \ie{} the neighborhood differences  mainly come from their connections to the same large cluster, and this can lead to the diversification operation failing to discriminate them. See Appendix \ref{appendix:limitation_diversification} for a more detailed discussion.

\vspace{-0.3cm}
\section{Related Work}
\vspace{-0.3cm}
\label{sec:related_works}
We now discuss relevant work on addressing heterophily in GNNs. \cite{abu2019mixhop} acknowledges the difficulty of learning on graphs with weak homophily and propose MixHop to extract features from multi-hop neighborhoods to get more information.  \cite{hou2019measuring} propose measurements based on feature smoothness and label smoothness that are potentially helpful to guide GNNs when dealing with heterophilic graphs. Geom-GCN \cite{pei2020geom} precomputes unsupervised node embeddings and uses the graph structure defined by geometric relationships in the embedding space to define the bi-level aggregation process to handle heterophily.  H$_2$GCN \cite{zhu2020beyond} combines 3 key designs to address heterophily: (1) ego- and neighbor-embedding separation; (2) higher-order neighborhoods; (3) combination of intermediate representations. CPGNN \cite{zhu2020graph} models label correlations through a compatibility matrix, which is beneficial for heterophilic graphs, and propagates a prior belief estimation into the GNN by using the compatibility matrix.  FAGCN \cite{bo2021beyond} learns edge-level aggregation weights as GAT \cite{velivckovic2017attention} but allows the weights to be negative, which enables the network to capture high-frequency components in the graph signals. GPRGNN \cite{chien2021adaptive} uses learnable weights that can be both positive and negative for feature propagation. This allows GPRGNN to adapt to heterophilic  graphs and  to handle both high- and low-frequency parts of the graph signals (See Appendix \ref{appendix:difference_with_sota_methods} for a more comprehensive comparison between ACM-GNNs, ACMII-GNNs and FAGCN, GPRGNN). BernNet \cite{he2021bernnet} designs a scheme to learn arbitrary graph spectral filters with Bernstein polynomial to address heterophily. \cite{ma2021homophily} points out that homophily is not necessary for GNNs and characterizes conditions that GNNs can perform well on heterophilic graphs.

\vspace{-0.1cm}
\section{Empirical Evaluation}
\vspace{-0.2cm}
\label{sec:experiments}
In this section, we evaluate the proposed ACM and ACMII framework on real-world datasets (see Appendix \ref{appendix:model_comparison_synthetic_datasets} for a performance comparison with basline models on synthetic datasets). We first conduct ablation studies in Sec.~\ref{sec:ablation_tests_running_time} to validate the effectiveness and efficiency of different components of ACM and ACMII. Then, we compare with  state-of-the-art (SOTA) models in Sec.~\ref{sec:comparison_with_sota}.  The hyperparameter searching range and computing resources are described in Appendix \ref{appendix:hyperparameter}.
\begin{figure}[H]
    \centering
     {
    \subfloat[ Input Feature]{
     \captionsetup{justification = centering}
     \includegraphics[width=0.3\textwidth]{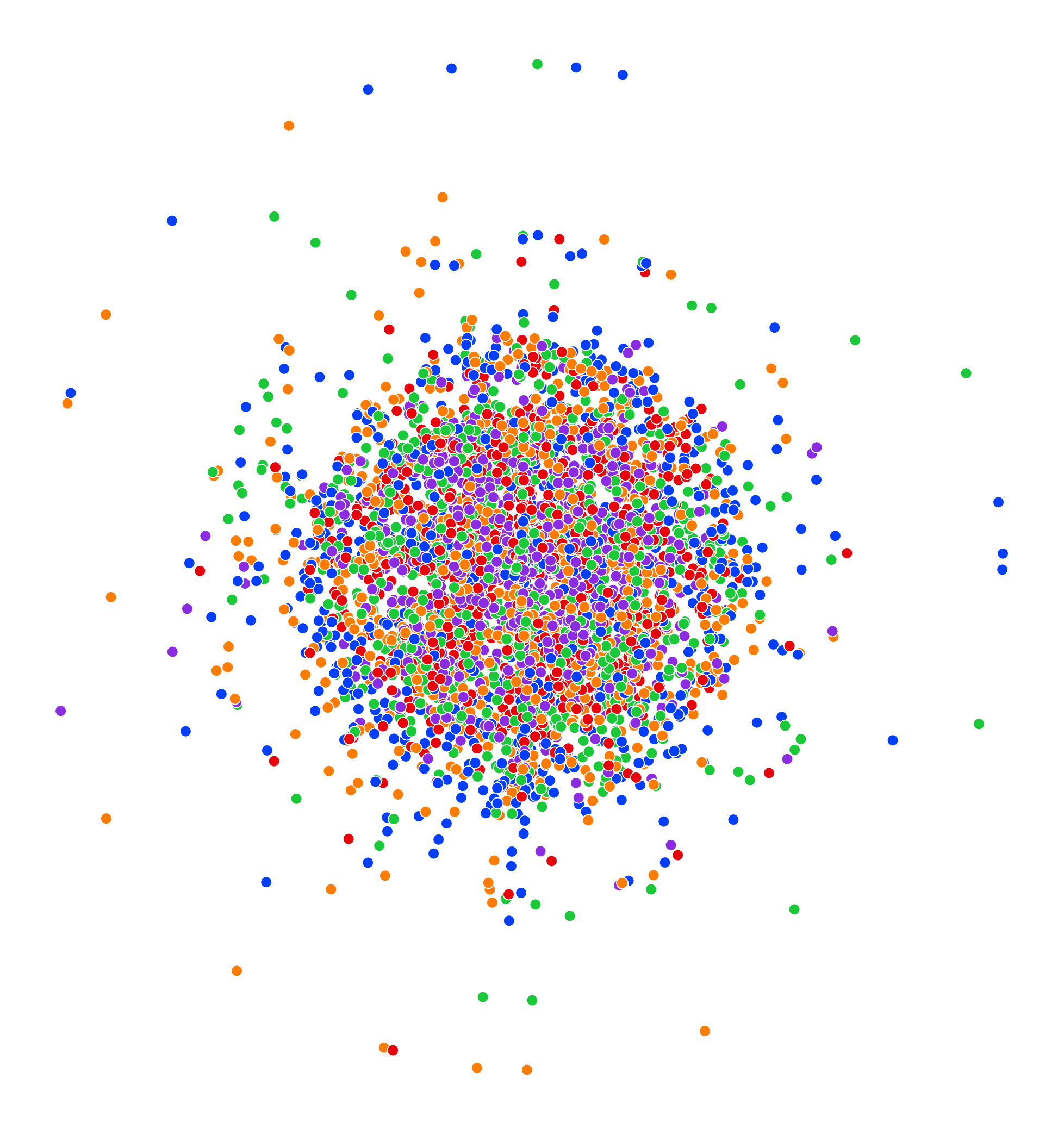}
     } 
     \subfloat[GCN Output]{
     \captionsetup{justification = centering}
     \includegraphics[width=0.3\textwidth]{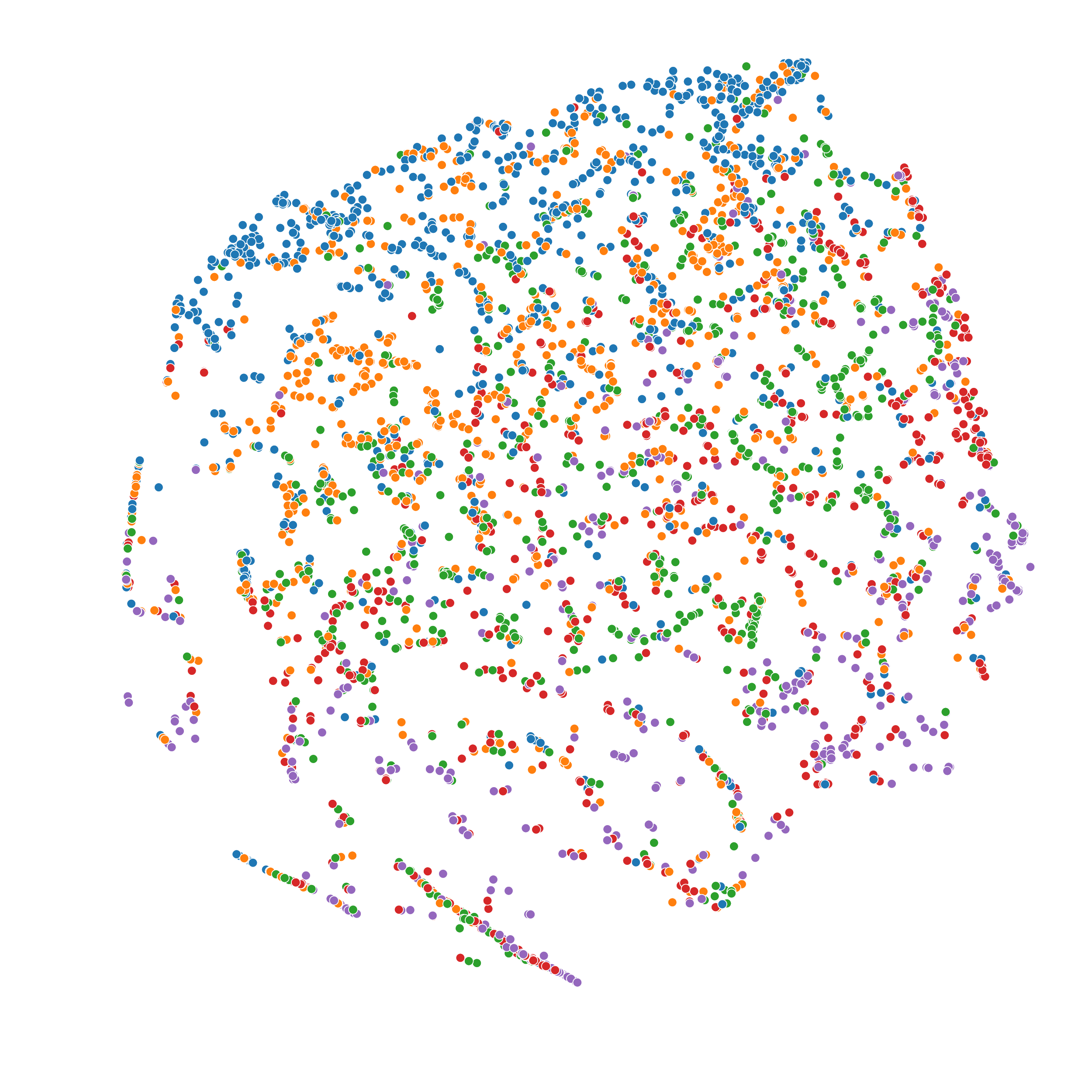}
     } 
     \subfloat[ACM-GCN Output]{
     \captionsetup{justification = centering}
     \includegraphics[width=0.3\textwidth]{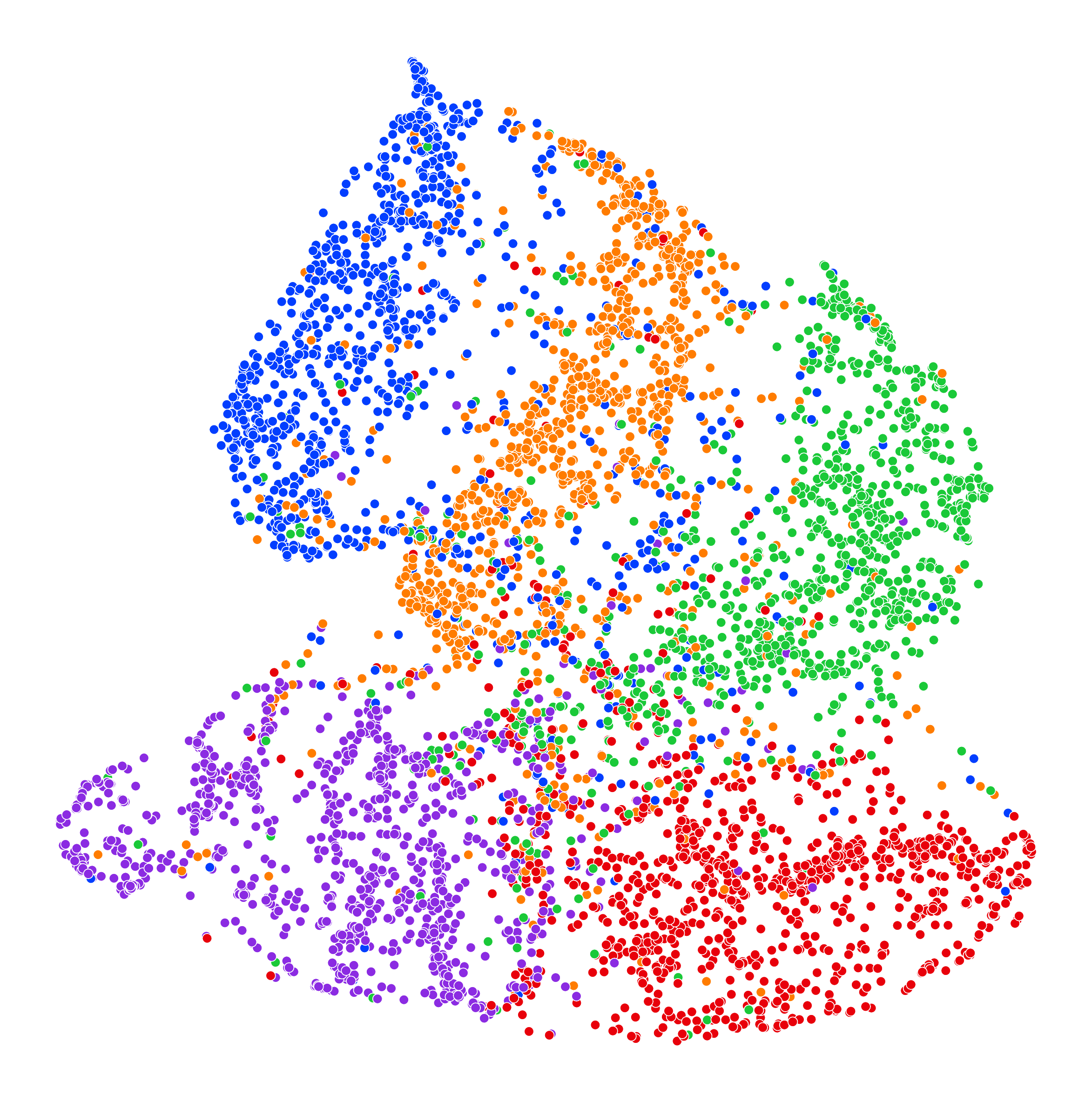}
     } \\
     
     \subfloat[Low-pass Channel]{
     \captionsetup{justification = centering}
     \includegraphics[width=0.22\textwidth]{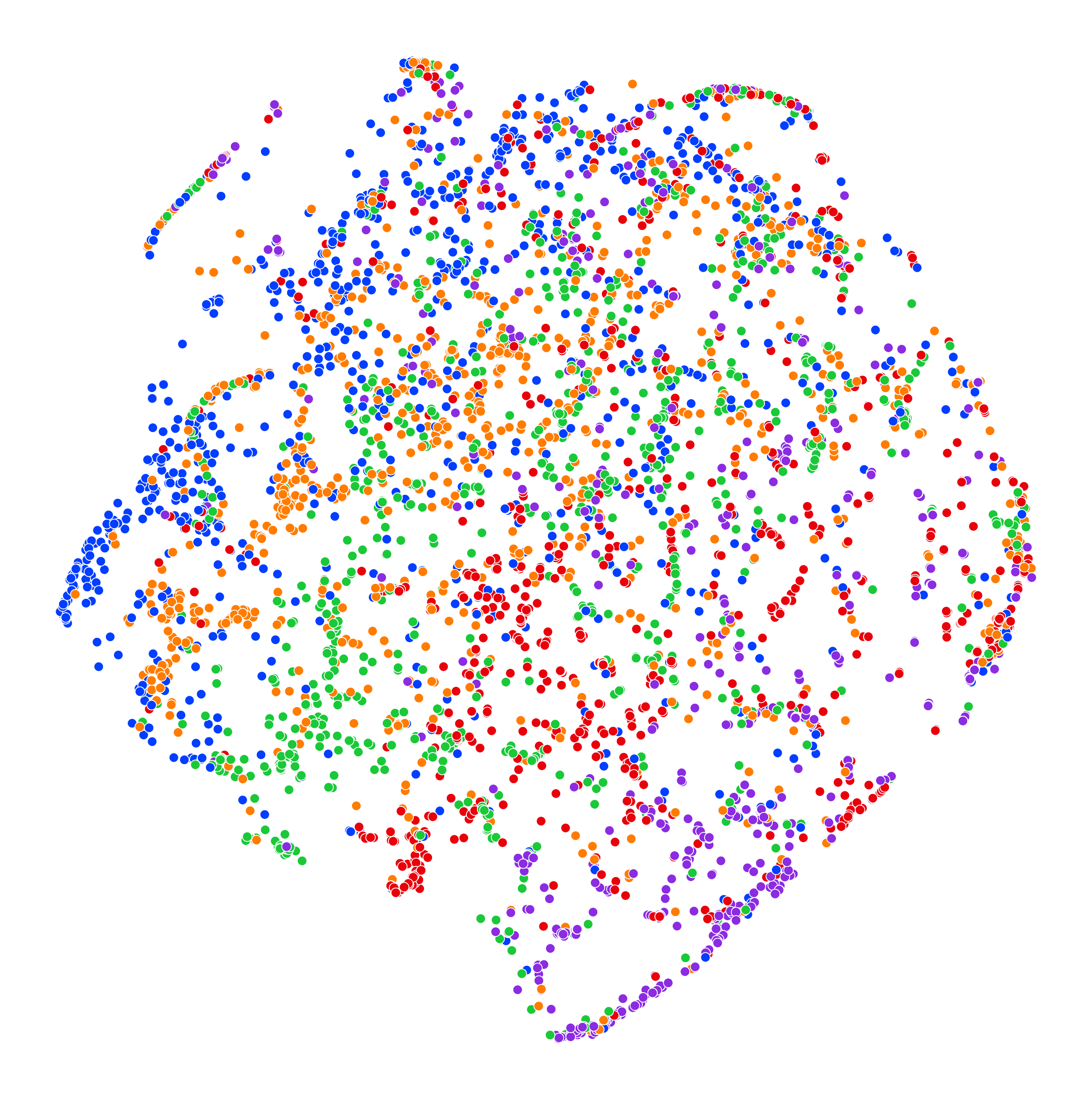}
     } 
     \subfloat[High-pass Channel]{
     \captionsetup{justification = centering}
     \includegraphics[width=0.22\textwidth]{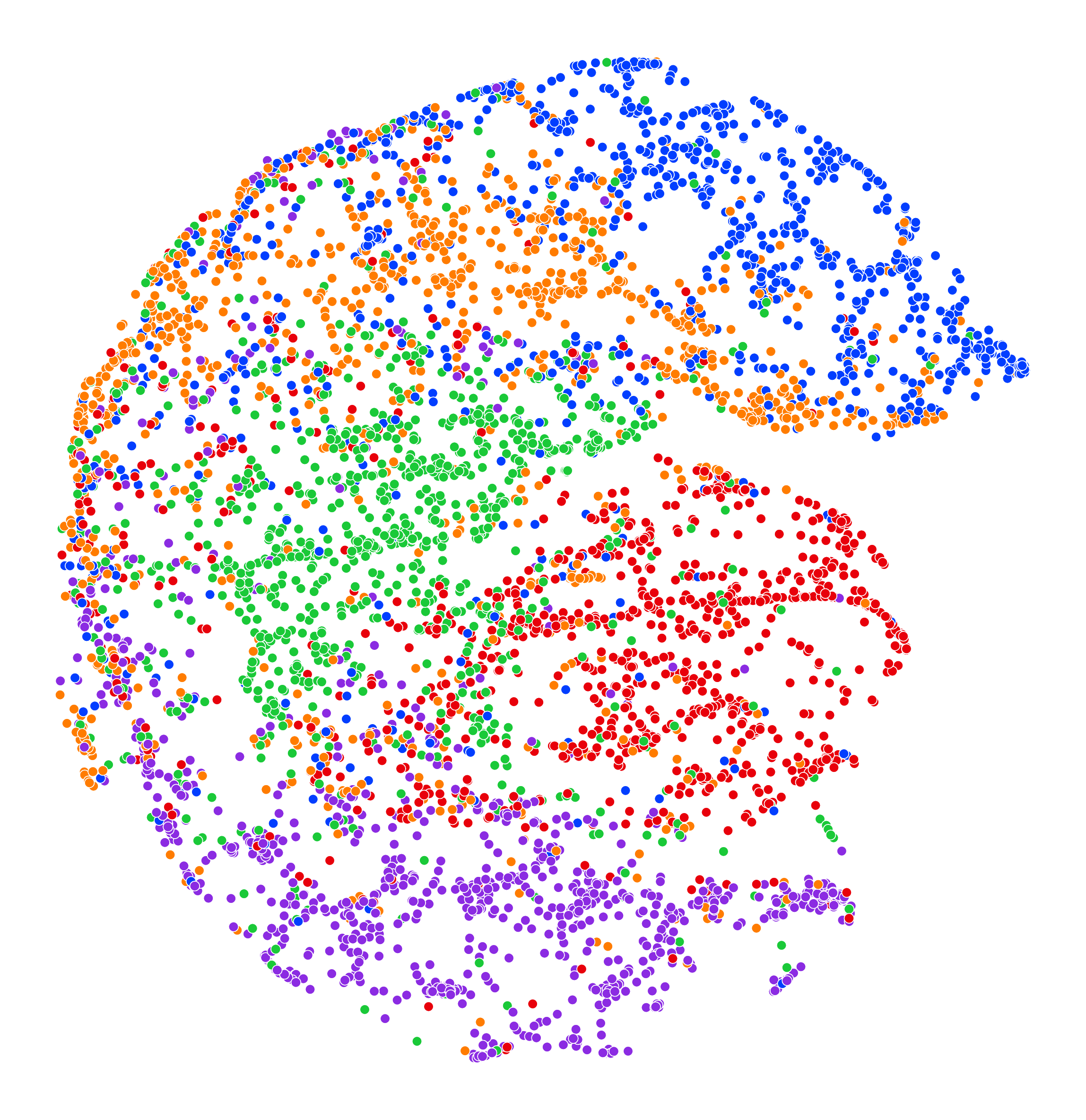}
     } 
     \subfloat[Identity Channel]{
     \captionsetup{justification = centering}
     \includegraphics[width=0.22\textwidth]{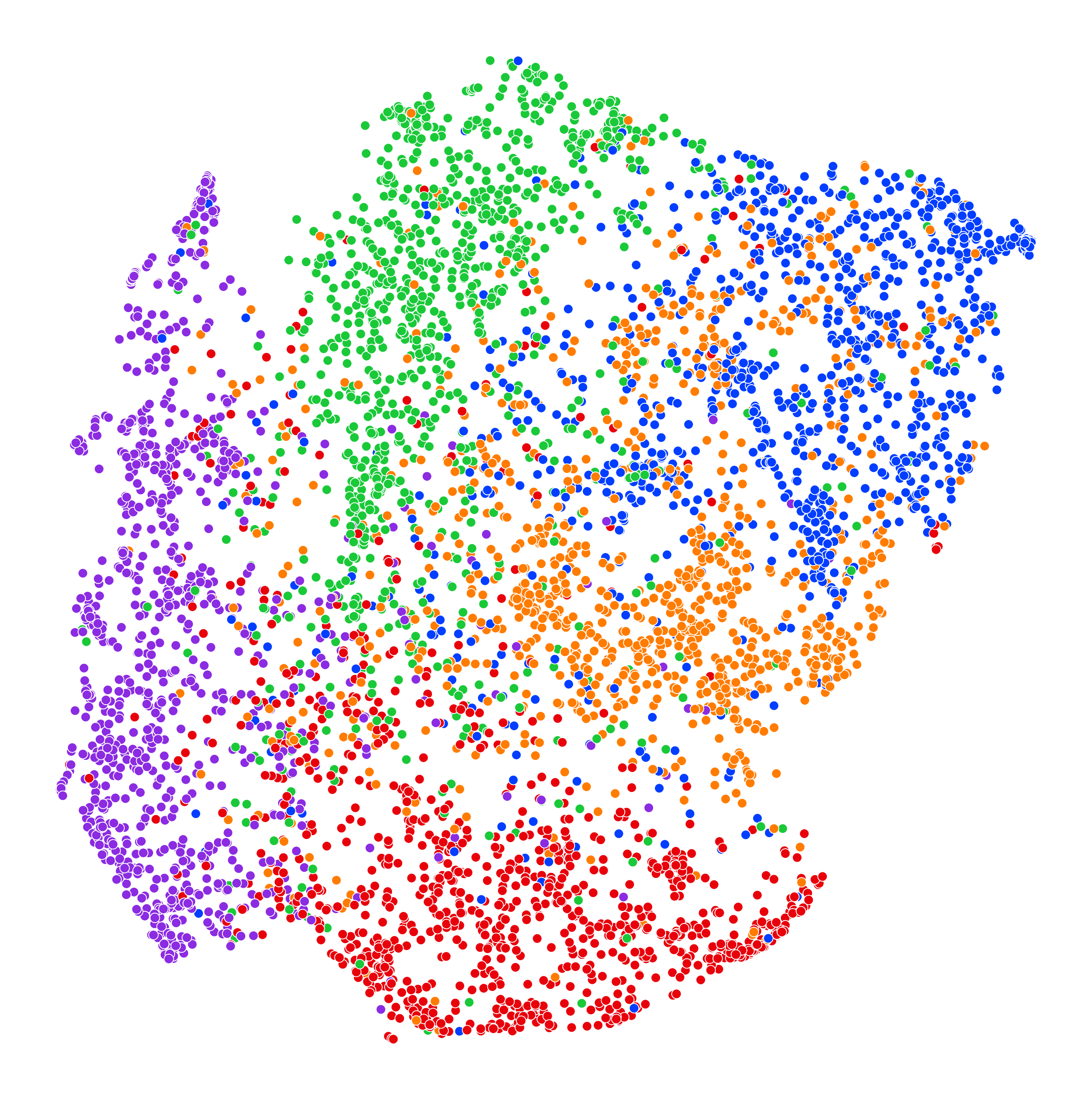}
     }
     \subfloat[Learned Weights ${\alpha}$]{
     \captionsetup{justification = centering}
     \includegraphics[width=0.22\textwidth]{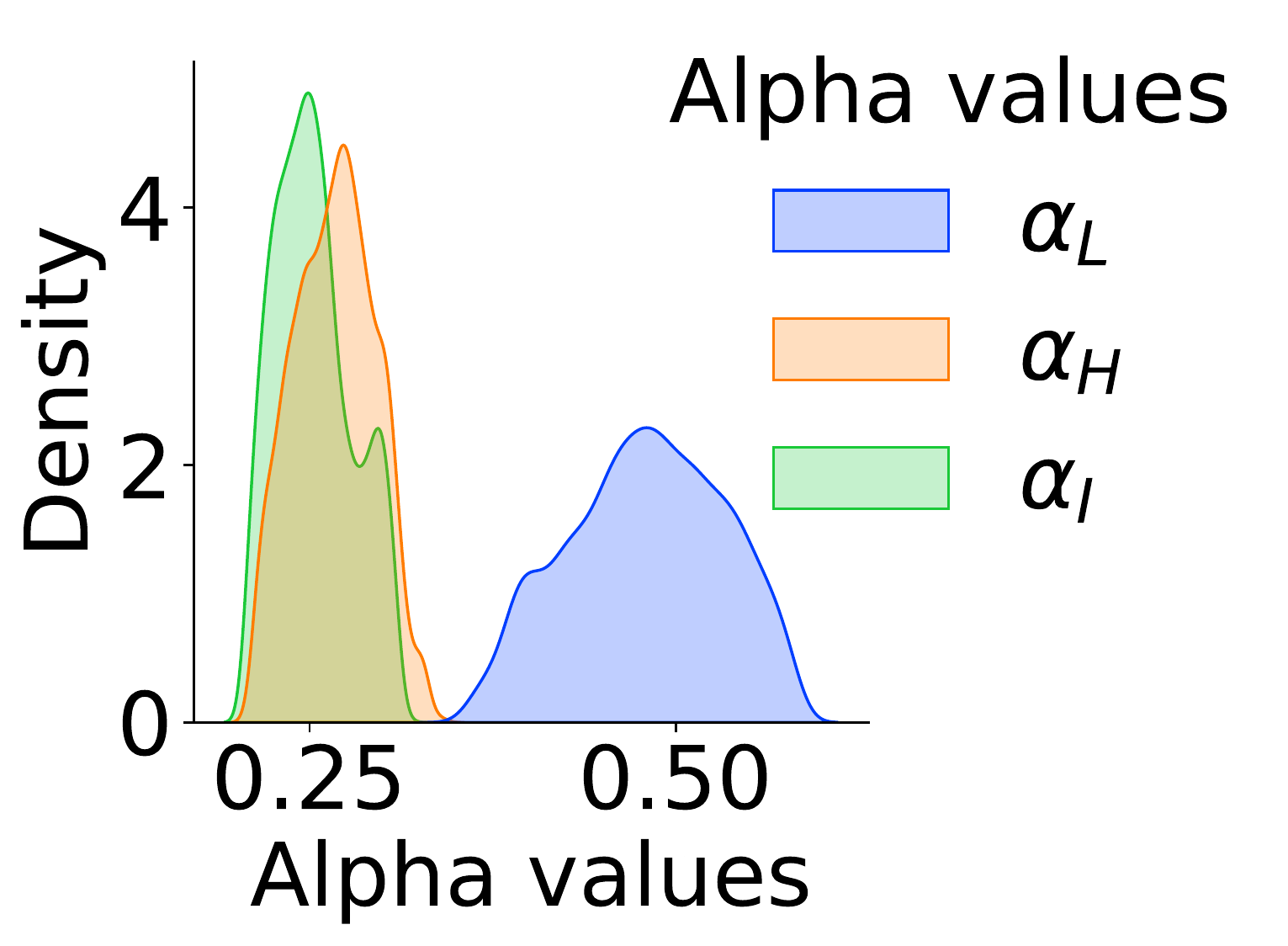}
     } 
     }
     \caption{t-SNE visualization of the output layer of ACM-GCN and GCN trained on Squirrel}
     \label{fig:visualization_acmgcn}
\end{figure}
\vspace{-0.4cm}
\subsection{Ablation Study \& Efficiency}
\vspace{-0.1cm}
\label{sec:ablation_tests_running_time}
\begin{table*}[htbp]
  \centering
  \tiny
  \caption{Ablation study on 9 real-world datasets \cite{pei2020geom}. Cell with \checkmark means the component is applied to the baseline model. The best test results are highlighted.}
  \label{tab:ablation_study}
  \setlength{\tabcolsep}{1pt}
    \begin{tabular}{c|cccc|ccccccccc|r}
    \toprule
    \toprule
    \multicolumn{14}{c}{Ablation Study on Different Components in ACM-SGC and ACM-GCN (\%)}                       &  \\
    \midrule
    \multicolumn{1}{p{4.125em}|}{Baseline    } & \multicolumn{4}{c|}{Model Components} & Cornell & Wisconsin & Texas & Film  & Chameleon & Squirrel & Cora  & CiteSeer & PubMed & \multicolumn{1}{c}{\multirow{1}[4]{*}{Rank}} \\
\cmidrule{2-14}    \multicolumn{1}{p{4.125em}|}{Models} & LP    & HP    & Identity & Mixing & Acc $\pm$ Std & Acc $\pm$ Std & Acc $\pm$ Std & Acc $\pm$ Std & Acc $\pm$ Std & Acc $\pm$ Std & Acc $\pm$ Std & Acc $\pm$ Std & Acc $\pm$ Std &  \\
    \midrule
    \multicolumn{1}{c|}{\multirow{5}[1]{*}{ACM-SGC-1 w/}} & $\checkmark$ &       &       &       & 70.98 $\pm$ 8.39 & 70.38 $\pm$ 2.85 & 83.28 $\pm$ 5.43 & 25.26 $\pm$ 1.18 & 64.86 $\pm$ 1.81 & 47.62 $\pm$ 1.27 & 85.12 $\pm$ 1.64 & 79.66 $\pm$ 0.75 & 85.5 $\pm$ 0.76 & \multicolumn{1}{c}{12.89} \\
          & $\checkmark$ & $\checkmark$ &       & $\checkmark$ & 83.28 $\pm$ 5.81 & 91.88 $\pm$ 1.61 & 90.98 $\pm$ 2.46 & 36.76 $\pm$ 1.01 & 65.27 $\pm$ 1.9 & 47.27 $\pm$ 1.37 & 86.8 $\pm$ 1.08 & 80.98 $\pm$ 1.68 & 87.21 $\pm$ 0.42 & \multicolumn{1}{c}{10.44} \\
          & $\checkmark$ &       & $\checkmark$ & $\checkmark$ & 93.93 $\pm$ 3.6 & 95.25 $\pm$ 1.84 & 93.93 $\pm$ 2.54 & 38.38 $\pm$ 1.13 & 63.83 $\pm$ 2.07 & 46.79 $\pm$ 0.75 & 86.73 $\pm$ 1.28 & 80.57 $\pm$ 0.99 & 87.8 $\pm$ 0.58 & \multicolumn{1}{c}{9.44} \\
          & $\checkmark$ & $\checkmark$ & $\checkmark$ &       & 88.2 $\pm$ 4.39 & 93.5 $\pm$ 2.95 & 92.95 $\pm$ 2.94 & 37.19 $\pm$ 0.87 & 62.82 $\pm$ 1.84 & 44.94 $\pm$ 0.93 & 85.22 $\pm$ 1.35 & 80.75 $\pm$ 1.68 & 88.11 $\pm$ 0.21 & \multicolumn{1}{c}{11.00} \\
          & $\checkmark$ & $\checkmark$ & $\checkmark$ & $\checkmark$ & 93.77 $\pm$ 1.91 & 93.25 $\pm$ 2.92 & 93.61 $\pm$ 1.55 & 39.33 $\pm$ 1.25 & 63.68 $\pm$ 1.62 & 46.4 $\pm$ 1.13 & 86.63 $\pm$ 1.13 & 80.96 $\pm$ 0.93 & 87.75 $\pm$ 0.88 & \multicolumn{1}{c}{10.00} \\
          \midrule
    \multicolumn{1}{c|}{\multirow{5}[0]{*}{ACM-GCN w/}} & $\checkmark$ &       &       &       & 82.46 $\pm$ 3.11 & 75.5 $\pm$ 2.92 & 83.11 $\pm$ 3.2 & 35.51 $\pm$ 0.99 & 64.18 $\pm$ 2.62 & 44.76 $\pm$ 1.39 & 87.78 $\pm$ 0.96 & 81.39 $\pm$ 1.23 & 88.9 $\pm$ 0.32 & \multicolumn{1}{c}{11.44} \\
          & $\checkmark$ & $\checkmark$ &       & $\checkmark$ & 82.13 $\pm$ 2.59 & 86.62 $\pm$ 4.61 & 89.19 $\pm$ 3.04 & 38.06 $\pm$ 1.35 & \cellcolor[rgb]{ .816,  .808,  .808}\textbf{69.21 $\pm$ 1.68} & 57.2 $\pm$ 1.01 & 88.93 $\pm$ 1.55 & \cellcolor[rgb]{ .816,  .808,  .808}\textbf{81.96 $\pm$ 0.91} & 90.01 $\pm$ 0.8  & \multicolumn{1}{c}{7.22} \\
          & $\checkmark$ &       & $\checkmark$ & $\checkmark$ & 94.26 $\pm$ 2.23 & 96.13 $\pm$ 2.2 & 94.1 $\pm$ 2.95 & 41.51 $\pm$ 0.99 & 67.44 $\pm$ 2.14 & 53.97 $\pm$ 1.39 & 88.95 $\pm$ 0.9 & 81.72 $\pm$ 1.22 & 90.88 $\pm$ 0.55 & \multicolumn{1}{c}{4.44} \\
          & $\checkmark$ & $\checkmark$ & $\checkmark$ &       & 91.64 $\pm$ 2 & 95.37 $\pm$ 3.31 & \cellcolor[rgb]{ .816,  .808,  .808}\textbf{95.25 $\pm$ 2.37} & 40.47 $\pm$ 1.49 & 68.93 $\pm$ 2.04 & 54.78 $\pm$ 1.27 & \cellcolor[rgb]{ .816,  .808,  .808}\textbf{89.13 $\pm$ 1.77} & \cellcolor[rgb]{ .816,  .808,  .808}\textbf{81.96 $\pm$ 2.03} & \cellcolor[rgb]{ .816,  .808,  .808}\textbf{91.01 $\pm$ 0.7} & \multicolumn{1}{c}{3.11} \\
          & $\checkmark$ & $\checkmark$ & $\checkmark$ & $\checkmark$ & 94.75 $\pm$ 2.62 & \cellcolor[rgb]{ .816,  .808,  .808}\textbf{96.75 $\pm$ 1.6} & 95.08 $\pm$ 3.2 & 41.62 $\pm$ 1.15 & 69.04 $\pm$ 1.74 & \cellcolor[rgb]{ .816,  .808,  .808}\textbf{58.02 $\pm$ 1.86} & 88.95 $\pm$ 1.3 & 81.80 $\pm$ 1.26 & 90.69 $\pm$ 0.53 & \multicolumn{1}{c}{\cellcolor[rgb]{ .816,  .808,  .808}\textbf{2.78}} \\
          \midrule
    \multicolumn{1}{c|}{\multirow{4}[1]{*}{ACMII-GCN w/}} & $\checkmark$ & $\checkmark$ &       & $\checkmark$ & 82.46 $\pm$ 3.03 & 91.00 $\pm$ 1.75 & 90.33 $\pm$ 2.69 & 38.39 $\pm$ 0.75 & 67.59 $\pm$ 2.14 & 53.67 $\pm$ 1.71 & \cellcolor[rgb]{ .816,  .808,  .808}\textbf{89.13 $\pm$ 1.14} & 81.75 $\pm$ 0.85 & 89.87 $\pm$ 0.39 & \multicolumn{1}{c}{7.44} \\
          & $\checkmark$ &       & $\checkmark$ & $\checkmark$ & 94.26 $\pm$ 2.57 & 96.00 $\pm$ 2.15 & 94.26 $\pm$ 2.96 & 40.96 $\pm$ 1.2 & 66.35 $\pm$ 1.76 & 50.78 $\pm$ 2.07 & 89.06 $\pm$ 1.07 & 81.86 $\pm$ 1.22 & 90.71 $\pm$ 0.67 & \multicolumn{1}{c}{4.67} \\
          & $\checkmark$ & $\checkmark$ & $\checkmark$ &       & 91.48 $\pm$ 1.43 & 96.25 $\pm$ 2.09 & 93.77 $\pm$ 2.91 & 40.27 $\pm$ 1.07 & 66.52 $\pm$ 2.65 & 52.9 $\pm$ 1.64 & 88.83 $\pm$ 1.16 & 81.54 $\pm$ 0.95 & 90.6 $\pm$ 0.47 & \multicolumn{1}{c}{6.67} \\
          & $\checkmark$ & $\checkmark$ & $\checkmark$ & $\checkmark$ & \cellcolor[rgb]{ .816,  .808,  .808}\textbf{95.9 $\pm$ 1.83} & 96.62 $\pm$ 2.44 & 95.25 $\pm$ 3.15 & \cellcolor[rgb]{ .816,  .808,  .808}\textbf{41.84 $\pm$ 1.15} & 68.38 $\pm$ 1.36 & 54.53 $\pm$ 2.09 & 89.00 $\pm$ 0.72 & 81.79 $\pm$ 0.95 & 90.74 $\pm$ 0.5 & \multicolumn{1}{c}{\cellcolor[rgb]{ .816,  .808,  .808}\textbf{2.78}} \\
\cmidrule{1-14}    \multicolumn{14}{c|}{Comparison of Average Running Time Per Epoch(ms)}                                        &  \\
\cmidrule{1-14}    \multicolumn{1}{c|}{\multirow{5}[2]{*}{ACM-SGC-1 w/}} & $\checkmark$ &       &       &       & 2.53  & 2.83  & 2.5   & 3.18  & 3.48  & 4.65  & 3.47  & 3.43  & 4.04  &  \\
          & $\checkmark$ & $\checkmark$ &       & $\checkmark$ & 4.01  & 4.57  & 4.24  & 4.55  & 4.76  & 5.09  & 5.39  & 4.69  & 4.75  &  \\
          & $\checkmark$ &       & $\checkmark$ & $\checkmark$ & 3.88  & 4.01  & 4.04  & 4.43  & 4.06  & 4.5   & 4.38  & 3.82  & 4.16  &  \\
          & $\checkmark$ & $\checkmark$ & $\checkmark$ &       & 3.31  & 3.49  & 3.18  & 3.7   & 3.53  & 4.83  & 3.92  & 3.87  & 4.24  &  \\
          & $\checkmark$ & $\checkmark$ & $\checkmark$ & $\checkmark$ & 5.53  & 5.96  & 5.43  & 5.21  & 5.41  & 6.96  & 6     & 5.9   & 6.04  &  \\
\cmidrule{1-14}    \multicolumn{1}{c|}{\multirow{5}[2]{*}{ACM-GCN w/}} & $\checkmark$ &       &       &       & 3.67  & 3.74  & 3.59  & 4.86  & 4.96  & 6.41  & 4.24  & 4.18  & 5.08  &  \\
          & $\checkmark$ & $\checkmark$ &       & $\checkmark$ & 6.63  & 8.06  & 7.89  & 8.11  & 7.8   & 9.39  & 7.82  & 7.38  & 8.74  &  \\
          & $\checkmark$ &       & $\checkmark$ & $\checkmark$ & 5.73  & 5.91  & 5.93  & 6.86  & 6.35  & 7.15  & 7.34  & 6.65  & 6.8   &  \\
          & $\checkmark$ & $\checkmark$ & $\checkmark$ &       & 5.16  & 5.25  & 5.2   & 5.93  & 5.64  & 8.02  & 5.73  & 5.65  & 6.16  &  \\
          & $\checkmark$ & $\checkmark$ & $\checkmark$ & $\checkmark$ & 8.25  & 8.11  & 7.89  & 7.97  & 8.41  & 11.9  & 8.84  & 8.38  & 8.63  &  \\
\cmidrule{1-14}    \multicolumn{1}{c|}{\multirow{4}[2]{*}{ACMII-GCN w/}} & $\checkmark$ & $\checkmark$ &       & $\checkmark$ & 6.62  & 7.35  & 7.39  & 7.62  & 7.33  & 9.69  & 7.49  & 7.58  & 7.97  &  \\
          & $\checkmark$ &       & $\checkmark$ & $\checkmark$ & 6.3   & 6.05  & 6.26  & 6.87  & 6.44  & 6.5   & 6.14  & 7.21  & 6.6   &  \\
          & $\checkmark$ & $\checkmark$ & $\checkmark$ &       & 5.24  & 5.27  & 5.46  & 5.72  & 5.65  & 7.87  & 5.48  & 5.65  & 6.33  &  \\
          & $\checkmark$ & $\checkmark$ & $\checkmark$ & $\checkmark$ & 7.59  & 8.28  & 8.06  & 8.85  & 8     & 10    & 8.27  & 8.5   & 8.68  &  \\
    \bottomrule
    \bottomrule
    \end{tabular}%
\end{table*}%

We will now investigate the effectiveness and efficiency of adding HP, identity channels and the adaptive mixing mechanism in the proposed framework by performing an ablation study. Specifically, we apply the components of ACM to SGC-1 \cite{wu2019simplifying} \footnote{We only test ACM-SGC-1 because SGC-1 does not contain any non-linearity which makes ACM-SGC-1 and ACMII-SGC-1 exactly the same.} and the components of ACM and ACMII to GCN \cite{kipf2016classification} separately. We run 10 times on each of the 9 benchmark datatsets, \textit{Cornell}, \textit{Wisconsin}, \textit{Texas}, \textit{Film}, \textit{Chameleon}, \textit{Squirrel}, \textit{Cora}, \textit{Citeseer} and \textit{Pubmed} used in \cite{musae,pei2020geom}, with the same 60\%/20\%/20\% random splits for train/validation/test used in \cite{chien2021adaptive} and report the average test accuracy as well as the standard deviation. We also record the average running time per epoch (in milliseconds) to compare the computational efficiency. We set the temperature $T$ in \eqref{eq:acm_gnn_spectral} to be $3$, which is the number of channels. 

The results in Table 1 show that on most datasets, the additional HP and identity channels are helpful, even for strong homophily datasets such as \textit{Cora, CiteSeer and PubMed}. The adaptive mixing mechanism also has an advantage over directly adding the three channels together. This illustrates the necessity of learning to customize the channel usage adaptively for different nodes. The t-SNE visualization in Figure \ref{fig:visualization_acmgcn} demonstrates that the high-pass channel(e) and identity channel(f) can extract meaningful patterns, which the low-pass channel(d) is not able to capture. The output of ACM-GCN(c) shows clearer boundaries among classes than GCN(b). The running time is approximately doubled in the ACM and ACMII framework compared to the original models.

\begin{figure*}[h]
     {
     \subfloat[$\uparrow$ 10.16 \% $\sim$ 13.44 \%]{
     \captionsetup{justification = centering}
     \includegraphics[height=0.15\textheight]{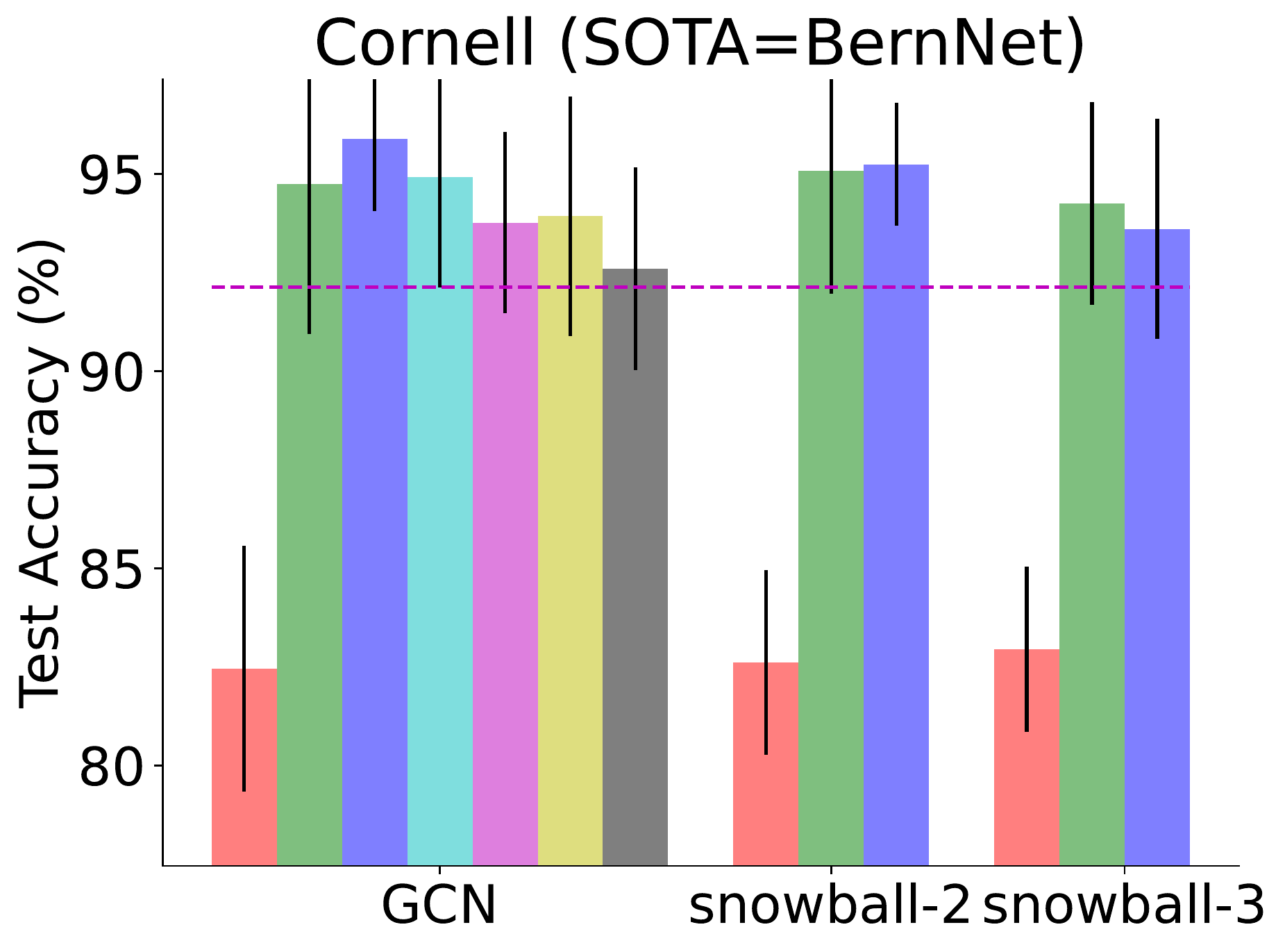}
     } 
     \subfloat[$\uparrow$ 20.25 \% $\sim$ 27.50 \%]{
     \captionsetup{justification = centering}
     \includegraphics[height=0.15\textheight]{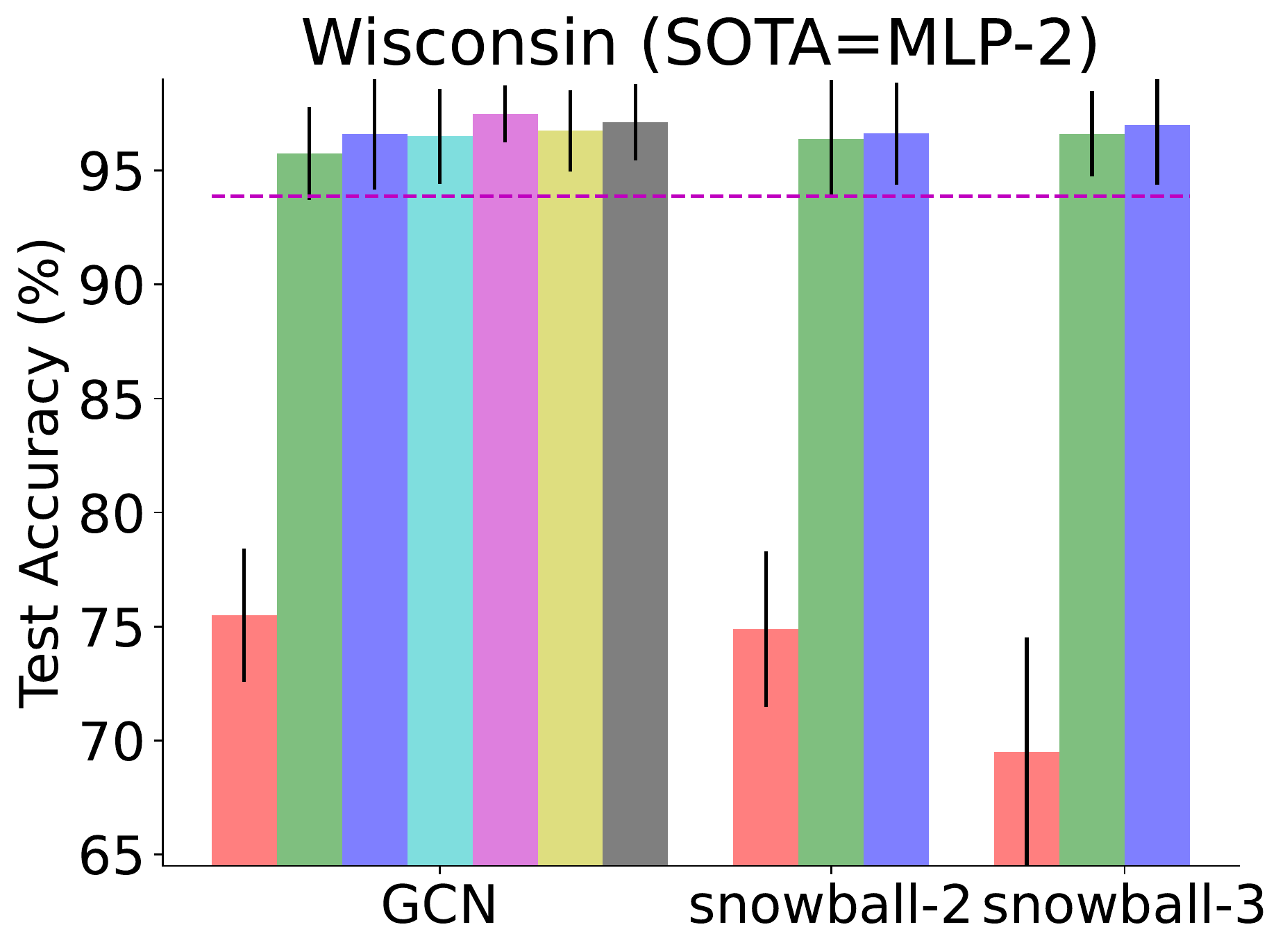}
     } 
     \subfloat[$\uparrow$ 11.64 \% $\sim$ 13.45 \%]{
     \captionsetup{justification=centering}
     \includegraphics[height=0.15\textheight]{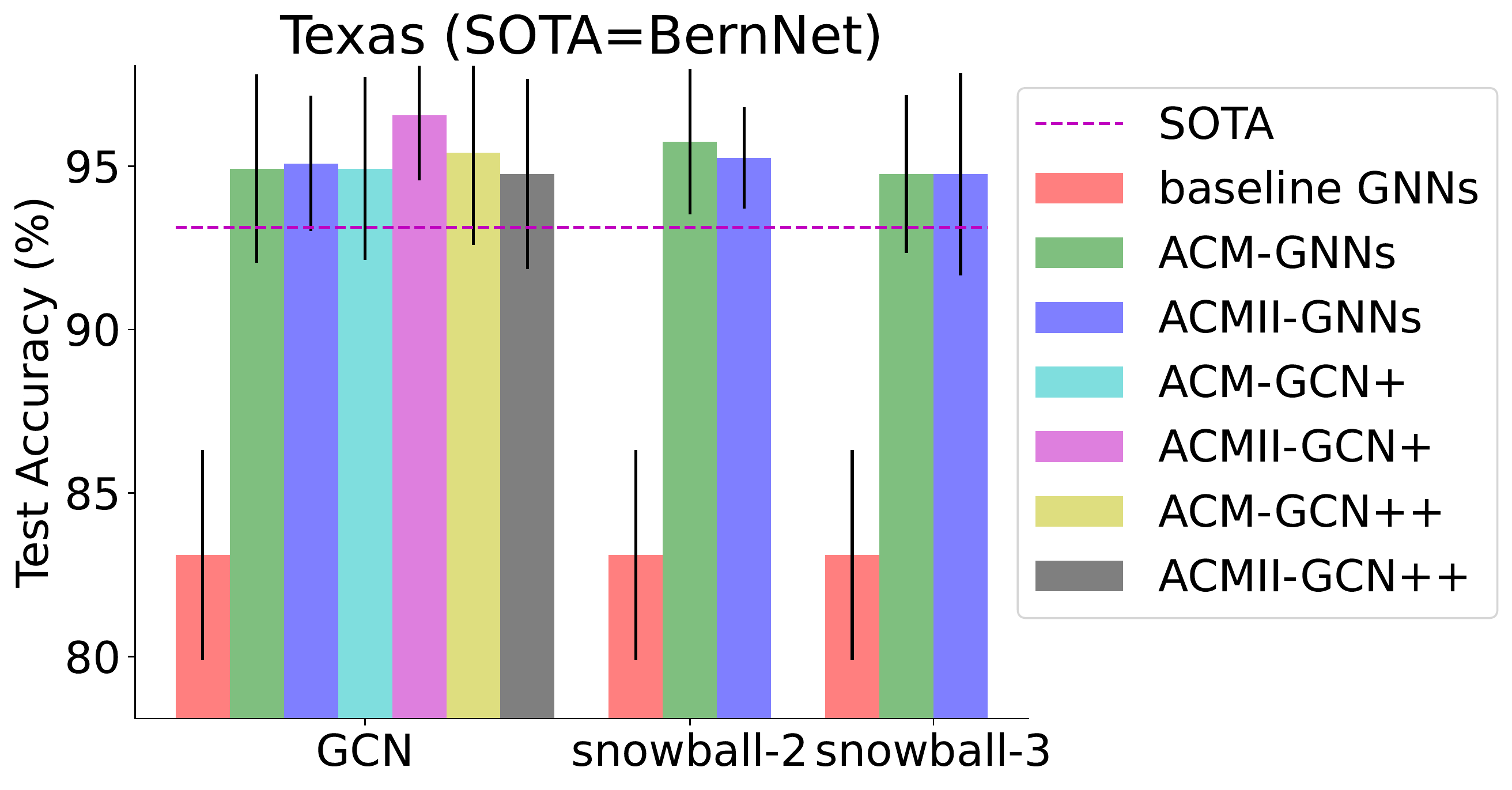}
     }\\
     \vspace{-0.3cm}
    \subfloat[$\uparrow$ 2.04 \% $\sim$ 11.90 \%]{
     \captionsetup{justification = centering}
     \includegraphics[height=0.15\textheight]{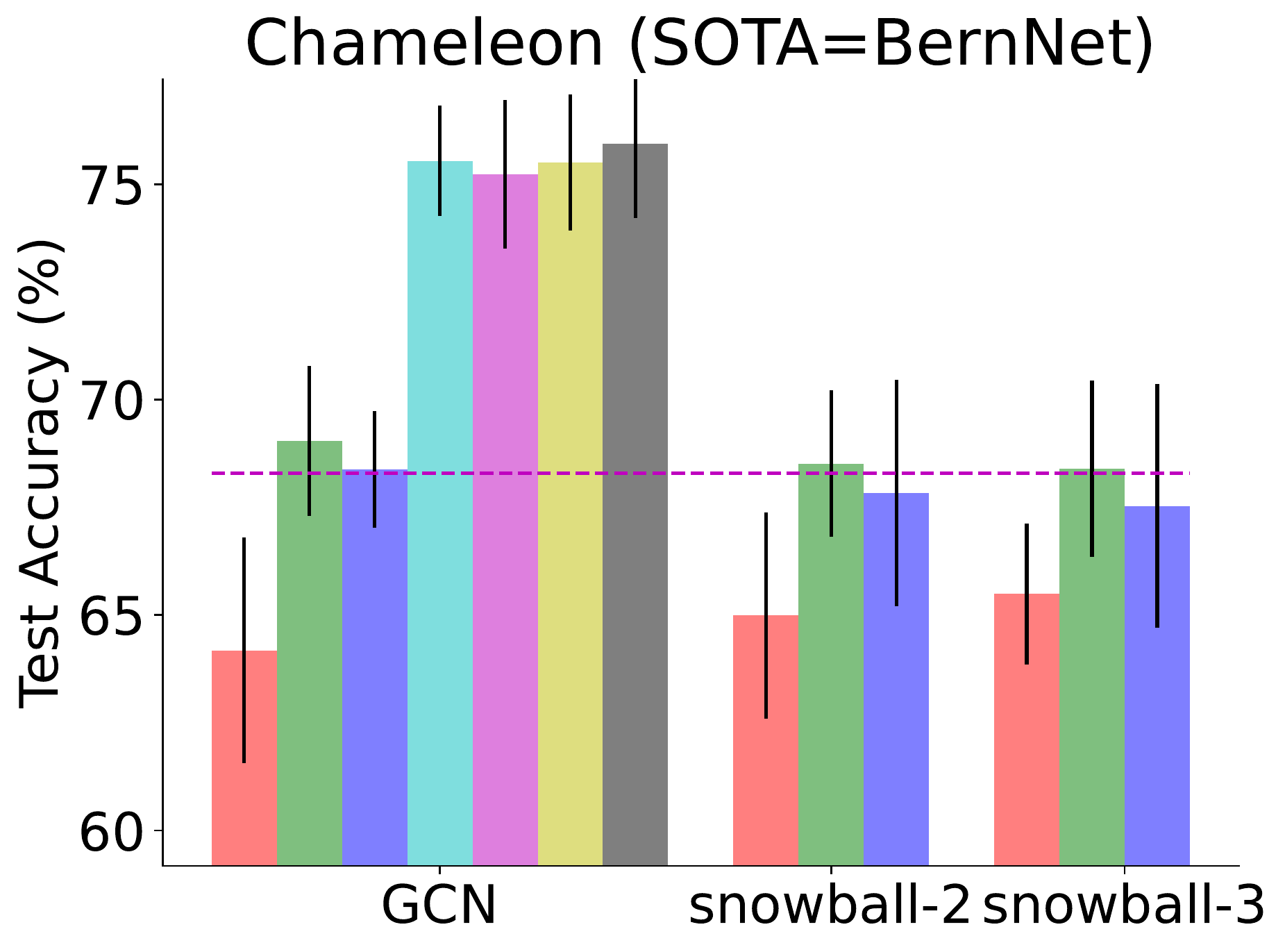}
     } 
     \subfloat[$\uparrow$ 4.31 \% $\sim$ 6.35 \%]{
     \captionsetup{justification = centering}
     \includegraphics[height=0.15\textheight]{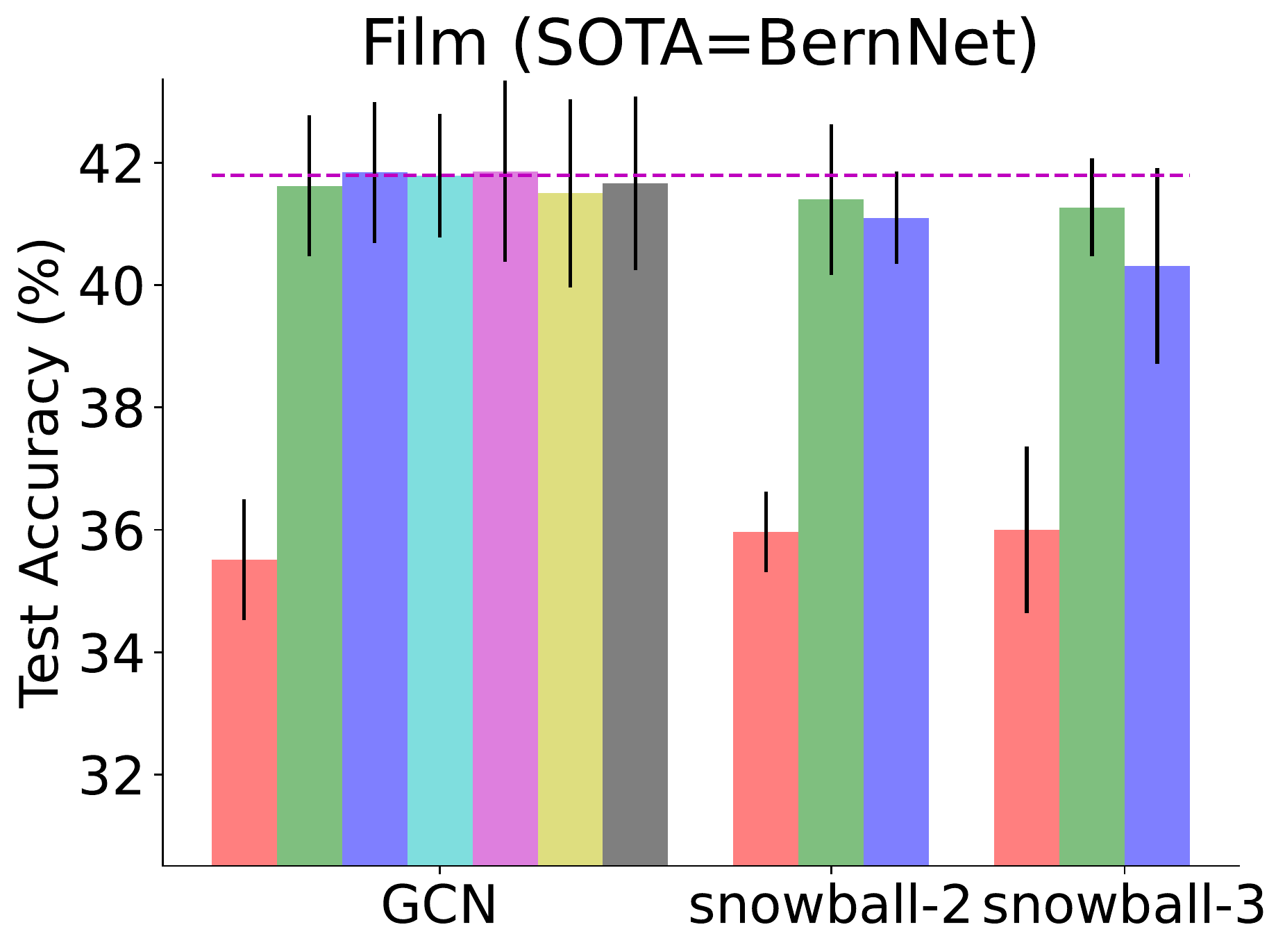}
     } 
     \subfloat[$\uparrow$ 4.06 \% $\sim$ 25.22 \%]{
     \captionsetup{justification=centering}
     \includegraphics[height=0.15\textheight]{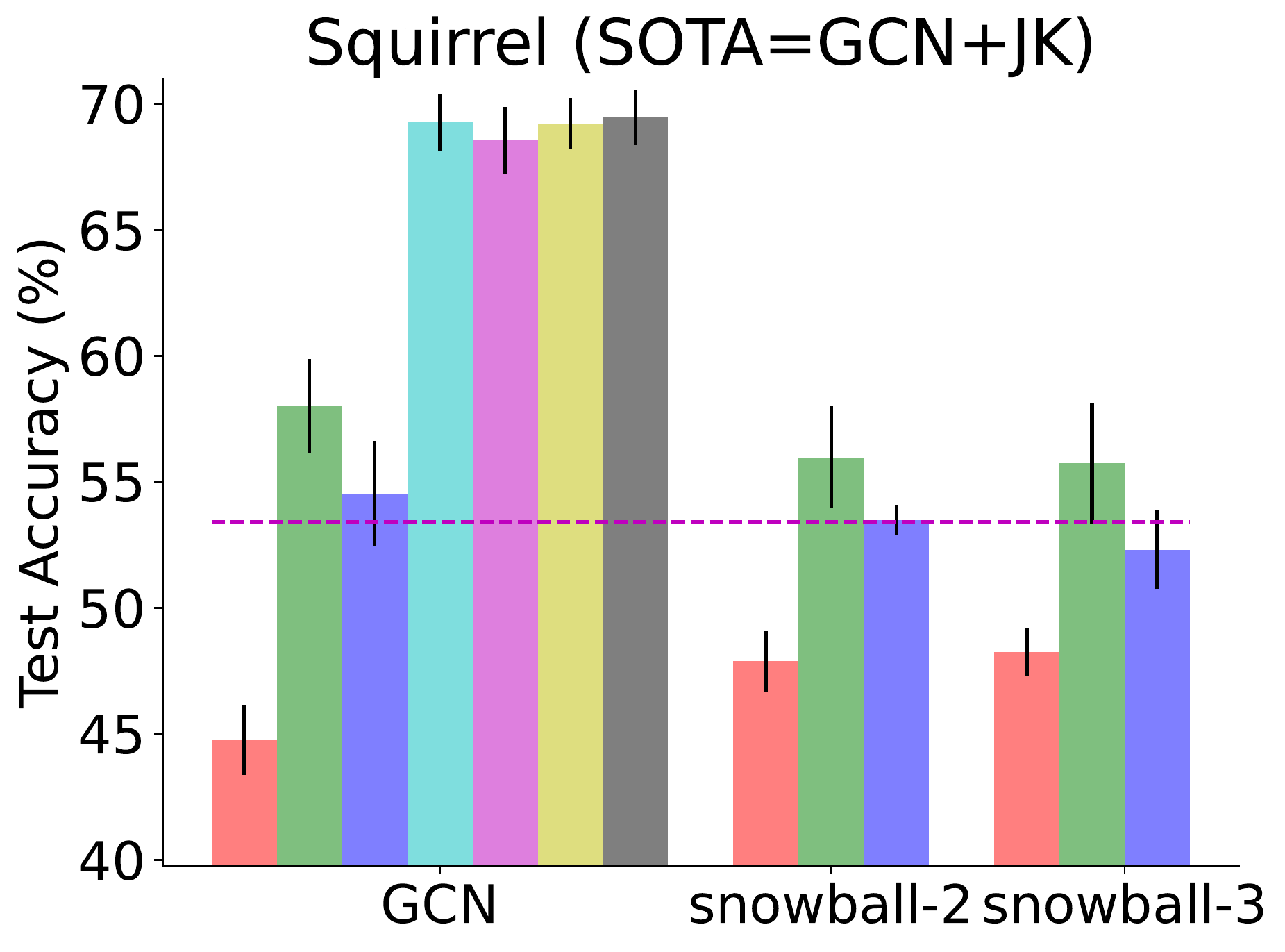}
     }
     }
     \vspace{-0.2cm}
     \caption{Comparison of baseline GNNs (red), ACM-GNNs (green), ACMII-GNNs (blue) with SOTA (magenta line) models on $6$ selected datasets. The black lines indicate the standard deviation. The symbol “$\uparrow$” shows the range of performance improvement (\%) of ACM-GNNs and ACMII-GNNs over baseline GNNs. See Appendix \ref{appendix:estimation_similarity_homophily_diversification_metrics} for a detailed discussion of the relation between $H_\text{agg}^M$ and GNN performance. }
     \label{fig:selected_comparison_with_sota}
\end{figure*}
\vspace{-0.3cm}
\subsection{Comparison with Baseline and SOTA Models}
\vspace{-0.2cm}
\label{sec:comparison_with_sota}
\paragraph{Datasets \& Experimental Setup}
In this section, we evaluate SGC \cite{wu2019simplifying} with 1 hop and 2 hops (SGC-1, SGC-2), GCNII \cite{chen2020simple}, GCNII$^*$ \cite{chen2020simple}, GCN \cite{kipf2016classification} and snowball networks \cite{luan2019break} with 2 and 3 layers (snowball-2, snowball-3) and combine them with the ACM or ACMII framework\footnote{GCNII and GCNII$^*$ are hard to implement with the ACMII framework. See Appendix \ref{appendix:details_implementation_acm_acmII} for explanation.}. We use $\hat{A}_\text{rw}$ as the LP filter and the corresponding HP filter is $I-\hat{A}_\text{rw}$ \footnote{See Appendix \ref{appendix:rw_symmetric_comparision} for the comparison of $\hat{A}_\text{rw}$ and $\hat{A}_\text{sym}$.}. Both filters are deterministic. We compare these approaches with several baselines and SOTA GNN models: MLP with 2 layers (MLP-2), GAT \cite{velivckovic2017attention},  APPNP \cite{klicpera2018predict}, GPRGNN \cite{chien2021adaptive}, H$_2$GCN \cite{zhu2020beyond}, MixHop \cite{abu2019mixhop}, GCN+JK \cite{kipf2016classification, pmlr-v80-xu18c, lim2021new}, GAT+JK \cite{velivckovic2017attention, pmlr-v80-xu18c, lim2021new}, FAGCN \cite{bo2021beyond}, GraphSAGE \cite{hamilton2017inductive}, Geom-GCN \cite{pei2020geom} and BernNet \cite{he2021bernnet}. In addition to the 9 benchmark datasets used in section \ref{sec:ablation_tests_running_time}, we further test the above models on a new benchmark dataset, \textit{Deezer-Europe} \cite{feather}\footnote{We choose \textit{Deezer-Europe} because  MLP outperforms GCN on it \cite{lim2021new}.}.

On each dataset used in \cite{musae,pei2020geom}, we test the models 10 times following the same early stopping strategy, the same 60\%/20\%/20\% random data split \footnote{See table \ref{tab:performance_comparison_fixed_splits} in Appendix \ref{appendix:results_on_fixed_splits_as_geomgcn} for the performance comparison with several SOTA models, \eg{} LINKX \cite{lim2021large} and GloGNN \cite{li2022finding}, on the fixed 48\%/32\%/20\% splits provided by \cite{pei2020geom}.} and Adam \cite{kingma2014adam} optimizer as used in GPRGNN \cite{chien2021adaptive}.  
For \textit{Deezer-Europe}, we test the above models 5 times with the same early stopping strategy, the same fixed splits and Adam used in \cite{lim2021new}. 
\vspace{-0.3cm}
\paragraph{Structure information channel and residual connection} Besides the filtered features, some recent SOTA models additionally use graph structure information, \ie{} $\text{MLP}_\theta(A)$, and residual connection to address heterophily problem, \eg{} LINKX \cite{lim2021large} and GloGNN \cite{li2022finding}. $\text{MLP}_\theta(A)$ and residual connection can be directly incorporated into ACM and ACMII framework, which leads us to ACM(II)-GCN+ and ACM(II)-GCN++. See the details of implementation in Appendix \ref{appendix:details_implementation_acm_acmII}.

To visualize the performance, in Fig.~\ref{fig:selected_comparison_with_sota}, we plot the bar charts of the test accuracy of SOTA models, three selected baselines (GCN, snowball-2, snowball-3), their ACM(II) augmented models, ACM(II)-GCN+ and ACM(II)-GCN++ on the $6$ most commonly used benchmark heterophily datasets (See Table \ref{tab:sota} in Appendix \ref{sec:full_result_comparison} for the full results, comparison and ranking). From Fig.~\ref{fig:selected_comparison_with_sota}, we can see that (1) after being combined with the ACM or ACMII framework, the performance of the three baseline models is \textbf{significantly boosted, by 2.04\%$\sim$27.50\%} on all the 6 tasks. The ACM and ACMII in fact achieve SOTA performance. (2) On \textit{Cornell, Wisconsin, Texas, Chameleon} and \textit{Squirrel}, the augmented baseline models \textbf{significantly outperform the current SOTA models}. Overall, these results suggest that the proposed approach can help GNNs to generalize better on node classification tasks on heterophilic graphs, without adding too much computational cost.
\vspace{-0.3cm}
\section{Conclusions and Limitations}
\vspace{-0.2cm}
We have presented an analysis of existing homophily metrics and proposed new metrics which are more informative in terms of correlating with GNN performance. 
 To our knowledge, this is the first work analyzing heterophily from the perspective of post-aggregation node similarity.
 The similarity matrix and the new metrics we defined  mainly capture linear feature-independent relationships of each node. This might be insufficient when nonlinearity and feature-dependent information is important for classification. In the future, it would be useful to investigate if a similarity matrix could be defined which is capable of capturing nonlinear and feature-dependent relations between aggregated node.


We have also proposed a multi-channel mixing mechanism which leverages the intuitions gained in the first part of the paper and can be combined with different GNN architectures, enabling adaptive filtering (high-pass, low-pass or identity) at different nodes. Empirically, this approach shows very promising results, improving the performance of the base GNNs with which it is combined and achieving SOTA results at the cost of a reasonable increase in computation time.
As discussed in Sec.~\ref{sec:filterbank_acm_gnn_architecture}, however, the filterbank method cannot properly handle all cases of harmful  heterophily, and alternative ideas should be explored as well in the future.

\vspace{-0.3cm}
\section{Acknowledge}
\vspace{-0.2cm}
The authors would like to give very special thanks to William L. Hamilton for valuable discussion and advice. The project was partially supported by DeepMind and NSERC.
\clearpage
\bibliography{references.bib}
\bibliographystyle{abbrv}

\clearpage

\section*{Checklist}

\begin{enumerate}

\item For all authors...
\begin{enumerate}
  \item Do the main claims made in the abstract and introduction accurately reflect the paper's contributions and scope?
    \answerYes{}
  \item Did you describe the limitations of your work?
    \answerYes{}
  \item Did you discuss any potential negative societal impacts of your work?
    \answerNA{}
  \item Have you read the ethics review guidelines and ensured that your paper conforms to them?
    \answerYes{}
\end{enumerate}

\item If you are including theoretical results...
\begin{enumerate}
  \item Did you state the full set of assumptions of all theoretical results?
    \answerYes{}
        \item Did you include complete proofs of all theoretical results?
    \answerYes{}
\end{enumerate}

\item If you ran experiments...
\begin{enumerate}
  \item Did you include the code, data, and instructions needed to reproduce the main experimental results (either in the supplemental material or as a URL)?
    \answerYes{}
  \item Did you specify all the training details (e.g., data splits, hyperparameters, how they were chosen)?
    \answerYes{}
        \item Did you report error bars (e.g., with respect to the random seed after running experiments multiple times)?
    \answerYes{}
        \item Did you include the total amount of compute and the type of resources used (e.g., type of GPUs, internal cluster, or cloud provider)?
    \answerYes{}
\end{enumerate}

\item If you are using existing assets (e.g., code, data, models) or curating/releasing new assets...
\begin{enumerate}
  \item If your work uses existing assets, did you cite the creators?
    \answerYes{}
  \item Did you mention the license of the assets?
    \answerNA{}
  \item Did you include any new assets either in the supplemental material or as a URL?
    \answerNo{}
  \item Did you discuss whether and how consent was obtained from people whose data you're using/curating?
    \answerNo{}
  \item Did you discuss whether the data you are using/curating contains personally identifiable information or offensive content?
    \answerNo{}
\end{enumerate}

\item If you used crowdsourcing or conducted research with human subjects...
\begin{enumerate}
  \item Did you include the full text of instructions given to participants and screenshots, if applicable?
    \answerNA{}
  \item Did you describe any potential participant risks, with links to Institutional Review Board (IRB) approvals, if applicable?
    \answerNA{}
  \item Did you include the estimated hourly wage paid to participants and the total amount spent on participant compensation?
    \answerNA{}
\end{enumerate}

\end{enumerate}

\clearpage

\appendix

\section{More Experimental Results}

\subsection{Comparison with SOTA Models on 60\%/20\%/20\% Random Splits}
\label{sec:full_result_comparison}
The main results of the full sets of experiments \footnote{The splits for all these experiments are random 60\%/20\%/20\% splits for train/valid/test. The open source code we use is from \url{https://github.com/jianhao2016/GPRGNN/blob/f4aaad6ca28c83d3121338a4c4fe5d162edfa9a2/src/utils.py\#L16}. See table \ref{tab:performance_comparison_fixed_splits} in Appendix \ref{appendix:results_on_fixed_splits_as_geomgcn} for the performance comparison with several SOTA models on the fixed 48\%/32\%/20\% splits provided by \cite{pei2020geom}.} with statistics of datasets are summarized in Table \ref{tab:sota}, where we report the mean accuracy (\%) and standard deviation. We can see that after applied in ACM or ACMII framework, the performance of baseline models are boosted on almost all tasks and achieve SOTA performance on $9$ out of $10$ datasets. Especially, ACMII-GCN+ performs the best in terms of average rank (4.40) across all datasets. Overall, It suggests that ACM or ACMII framework can significantly increase the performance of GNNs on node classification tasks on heterophilic graphs and maintain highly competitive performance on homophilic datasets.
\begin{table}[htbp]
  \centering
  \tiny
  \setlength{\tabcolsep}{0.5pt}
  \caption{Experimental results: average test accuracy $\pm$ standard deviation on $10$ real-world benchmark datasets. The best results are highlighted in grey and the best baseline results (SOTA in Figure \ref{fig:selected_comparison_with_sota}) are underlined. Results "*" are reported from \cite{chien2021adaptive,lim2021new} and results "$^\dagger$" are from \cite{pei2020geom}. NA means the reported results are not available and OOM means out of memory. }
\begin{tabular}{c|cccccccccc|c}
    \toprule
    \toprule
          & Cornell & Wisconsin & Texas & Film  & Chameleon & Squirrel & Deezer-Europe & Cora  & CiteSeer & PubMed &  \\
    \midrule
    \#nodes & 183   & 251   & 183   & 7,600 & 2,277 & 5,201 & 28,281 & 2,708 & 3,327 & 19,717 &  \\
    \#edges & 295   & 499   & 309   & 33,544 & 36,101 & 217,073 & 92,752 & 5,429 & 4,732 & 44,338 &  \\
    \#features & 1,703 & 1,703 & 1,703 & 931   & 2,325 & 2,089 & 31,241 & 1,433 & 3,703 & 500   &  \\
    \#classes & 5     & 5     & 5     & 5     & 5     & 5     & 2     & 7     & 6     & 3     &  \\
    $H_\text{edge}$ & 0.5669 & 0.4480 & 0.4106 & 0.3750 & 0.2795 & 0.2416 & 0.5251 & 0.8100 & 0.7362 & 0.8024 &  \\
    $H_\text{node}$ & 0.3855 & 0.1498 & 0.0968 & 0.2210 & 0.2470 & 0.2156 & 0.5299 & 0.8252 & 0.7175 & 0.7924 &  \\
    $H_\text{class}$ & 0.0468 & 0.0941 & 0.0013 & 0.0110 & 0.0620 & 0.0254 & 0.0304 & 0.7657 & 0.6270 & 0.6641 &  \\
    Data Splits & 60\%/20\%/20\% & 60\%/20\%/20\% & 60\%/20\%/20\% & 60\%/20\%/20\% & 60\%/20\%/20\% & 60\%/20\%/20\% & 50\%/25\%/25\% & 60\%/20\%/20\% & 60\%/20\%/20\% & 60\%/20\%/20\% &  \\
    $H_\text{agg}^{M}(G)$ & 0.8032 & 0.7768 & 0.694 & 0.6822 & 0.61  & 0.3566 & 0.5790 & 0.9904 & 0.9826 & 0.9432 &  \\
\midrule
    \midrule
          & \multicolumn{10}{c|}{Test Accuracy (\%) of State-of-the-art Models, Baseline GNN Models and ACM-GNN models} & Rank \\
    \midrule
    MLP-2 & 91.30 $\pm$ 0.70 & \underline{93.87 $\pm$ 3.33} & 92.26 $\pm$ 0.71 & 38.58 $\pm$ 0.25 & 46.72 $\pm$ 0.46 & 31.28 $\pm$ 0.27 & 66.55 $\pm$ 0.72 & 76.44 $\pm$ 0.30 & 76.25 $\pm$ 0.28 & 86.43 $\pm$ 0.13 & 23.40 \\
    \midrule
    GAT   & 76.00 $\pm$ 1.01 & 71.01 $\pm$ 4.66 & 78.87 $\pm$ 0.86 & 35.98 $\pm$ 0.23 & 63.9 $\pm$ 0.46 & 42.72 $\pm$ 0.33 & 61.09 $\pm$ 0.77 & 76.70 $\pm$ 0.42 & 67.20 $\pm$ 0.46 & 83.28 $\pm$ 0.12 & 26.20 \\
    APPNP & 91.80 $\pm$ 0.63 & 92.00 $\pm$ 3.59 & 91.18 $\pm$ 0.70 & 38.86 $\pm$ 0.24 & 51.91 $\pm$ 0.56 & 34.77 $\pm$ 0.34 & 67.21 $\pm$ 0.56 & 79.41 $\pm$ 0.38 & 68.59 $\pm$ 0.30 & 85.02 $\pm$ 0.09 & 22.80 \\
    GPRGNN & 91.36 $\pm$ 0.70 & 93.75 $\pm$ 2.37 & 92.92 $\pm$ 0.61 & 39.30 $\pm$ 0.27 & 67.48 $\pm$ 0.40 & 49.93 $\pm$ 0.53 & 66.90 $\pm$ 0.50 & 79.51$\pm$ 0.36 & 67.63 $\pm$ 0.38 & 85.07 $\pm$ 0.09 & 19.20 \\
    H2GCN & 86.23 $\pm$ 4.71 & 87.5 $\pm$ 1.77 & 85.90 $\pm$ 3.53 & 38.85 $\pm$ 1.17 & 52.30 $\pm$ 0.48 & 30.39 $\pm$ 1.22 & \underline{67.22 $\pm$ 0.90} & 87.52 $\pm$ 0.61 & 79.97 $\pm$ 0.69 & 87.78 $\pm$ 0.28 & 21.80 \\
    MixHop & 60.33 $\pm$ 28.53 & 77.25 $\pm$ 7.80 & 76.39 $\pm$ 7.66 & 33.13 $\pm$ 2.40 & 36.28 $\pm$ 10.22 & 24.55 $\pm$ 2.60 & 66.80 $\pm$ 0.58 & 65.65 $\pm$ 11.31 & 49.52 $\pm$ 13.35 & 87.04 $\pm$ 4.10 & 28.30 \\
    GCN+JK & 66.56 $\pm$ 13.82 & 62.50 $\pm$ 15.75 & 80.66 $\pm$ 1.91 & 32.72 $\pm$ 2.62 & 64.68 $\pm$ 2.85 & \underline{53.40 $\pm$ 1.90} & 60.99 $\pm$ 0.14 & 86.90 $\pm$ 1.51 & 73.77 $\pm$ 1.85 & \underline{90.09 $\pm$ 0.68} & 23.40 \\
     GAT+JK & 74.43 $\pm$ 10.24 & 69.50 $\pm$ 3.12 & 75.41 $\pm$ 7.18 & 35.41 $\pm$ 0.97 & 68.14 $\pm$ 1.18 & 52.28 $\pm$ 3.61 & 59.66 $\pm$ 0.92 & \underline{89.52 $\pm$ 0.43} & 74.49 $\pm$ 2.76 & 89.15 $\pm$ 0.87 & 20.90 \\
    FAGCN & 88.03 $\pm$ 5.6 & 89.75 $\pm$ 6.37 & 88.85 $\pm$ 4.39 & 31.59 $\pm$ 1.37 & 49.47 $\pm$ 2.84 & 42.24 $\pm$ 1.2 & 66.86 p, 0.53 & 88.85 $\pm$ 1.36 & \underline{\cellcolor[rgb]{ .816,  .808,  .808}\textbf{82.37 $\pm$ 1.46}} & 89.98 $\pm$ 0.54 & 18.20 \\
    BernNet & \underline{92.13 $\pm$ 1.64} & NA    & \underline{93.12 $\pm$ 0.65} & \underline{41.79 $\pm$ 1.01} & \underline{68.29 $\pm$ 1.58} & 51.35 $\pm$ 0.73 & NA    & 88.52 $\pm$ 0.95 & 80.09 $\pm$ 0.79 & 88.48 $\pm$ 0.41 & 14.75 \\
    GraphSAGE & 71.41 $\pm$ 1.24 & 64.85 $\pm$ 5.14 & 79.03 $\pm$ 1.20 & 36.37 $\pm$ 0.21 & 62.15 $\pm$ 0.42 & 41.26 $\pm$ 0.26 & OOM   & 86.58 $\pm$ 0.26 & 78.24 $\pm$ 0.30 & 86.85 $\pm$ 0.11 & 25.78 \\
    Geom-GCN* & 60.81 & 64.12 & 67.57 & 31.63 & 60.9  & 38.14 & NA    & 85.27 & 77.99 & 90.05 & 27.44 \\
    \midrule
    SGC-1 & 70.98 $\pm$ 8.39 & 70.38 $\pm$ 2.85 & 83.28 $\pm$ 5.43 & 25.26 $\pm$ 1.18 & 64.86 $\pm$ 1.81 & 47.62 $\pm$ 1.27 & 59.73 $\pm$ 0.12 & 85.12 $\pm$ 1.64 & 79.66 $\pm$ 0.75 & 85.5 $\pm$ 0.76 & 24.90 \\
    SGC-2 & 72.62 $\pm$ 9.92 & 74.75 $\pm$ 2.89 & 81.31 $\pm$ 3.3 & 28.81 $\pm$ 1.11 & 62.67 $\pm$ 2.41 & 41.25 $\pm$ 1.4 & 61.56 $\pm$ 0.51 & 85.48 $\pm$ 1.48 & 80.75 $\pm$ 1.15 & 85.36 $\pm$ 0.52 & 25.40 \\
    GCNII & 89.18 $\pm$ 3.96 & 83.25 $\pm$ 2.69 & 82.46 $\pm$ 4.58 & 40.82 $\pm$ 1.79 & 60.35 $\pm$ 2.7 & 38.81 $\pm$ 1.97 & 66.38 $\pm$ 0.45 & 88.98 $\pm$ 1.33 & 81.58 $\pm$ 1.3 & 89.8 $\pm$ 0.3 & 19.30 \\
    GCNII* & 90.49 $\pm$ 4.45 & 89.12 $\pm$ 3.06 & 88.52 $\pm$ 3.02 & 41.54 $\pm$ 0.99 & 62.8 $\pm$ 2.87 & 38.31 $\pm$ 1.3 & 66.42 $\pm$ 0.56 & 88.93 $\pm$ 1.37 & 81.83 $\pm$ 1.78 & 89.98 $\pm$ 0.52 & 16.40 \\
    GCN   & 82.46 $\pm$ 3.11 & 75.5 $\pm$ 2.92 & 83.11 $\pm$ 3.2 & 35.51 $\pm$ 0.99 & 64.18 $\pm$ 2.62 & 44.76 $\pm$ 1.39 & 62.23 $\pm$ 0.53 & 87.78 $\pm$ 0.96 & 81.39 $\pm$ 1.23 & 88.9 $\pm$ 0.32 & 20.90 \\
    Snowball-2 & 82.62 $\pm$ 2.34 & 74.88 $\pm$ 3.42 & 83.11 $\pm$ 3.2 & 35.97 $\pm$ 0.66 & 64.99 $\pm$ 2.39 & 47.88 $\pm$ 1.23 & OOM   & 88.64 $\pm$ 1.15 & 81.53 $\pm$ 1.71 & 89.04 $\pm$ 0.49 & 19.78 \\
    Snowball-3 & 82.95 $\pm$ 2.1 & 69.5 $\pm$ 5.01 & 83.11 $\pm$ 3.2 & 36.00 $\pm$ 1.36 & 65.49 $\pm$ 1.64 & 48.25 $\pm$ 0.94 & OOM   & 89.33 $\pm$ 1.3 & 80.93 $\pm$ 1.32 & 88.8 $\pm$ 0.82 & 19.11 \\
    \midrule
    ACM-SGC-1 & 93.77 $\pm$ 1.91 & 93.25 $\pm$ 2.92 & 93.61 $\pm$ 1.55 & 39.33 $\pm$ 1.25 & 63.68 $\pm$ 1.62 & 46.4 $\pm$ 1.13 & 66.67 $\pm$ 0.56 & 86.63 $\pm$ 1.13 & 80.96 $\pm$ 0.93 & 87.75 $\pm$ 0.88 & 17.00 \\
    ACM-SGC-2 & 93.77 $\pm$ 2.17 & 94.00 $\pm$ 2.61 & 93.44 $\pm$ 2.54 & 40.13 $\pm$ 1.21 & 60.48 $\pm$ 1.55 & 40.91 $\pm$ 1.39 & 66.53 $\pm$ 0.57 & 87.64 $\pm$ 0.99 & 80.93 $\pm$ 1.16 & 88.79 $\pm$ 0.5 & 17.70 \\
    ACM-GCNII & 92.62 $\pm$ 3.13 & 94.63 $\pm$ 2.96 & 92.46 $\pm$ 1.97 & 41.37 $\pm$ 1.37 & 58.73 $\pm$ 2.52 & 40.9 $\pm$ 1.58 & 66.39 $\pm$ 0.56 & 89.1 $\pm$ 1.61 & 82.28 $\pm$ 1.12 & 90.12 $\pm$ 0.4 & 14.30 \\
    ACM-GCNII* & 93.44 $\pm$ 2.74 & 94.37 $\pm$ 2.81 & 93.28 $\pm$ 2.79 & 41.27 $\pm$ 1.24 & 61.66 $\pm$ 2.29 & 38.32 $\pm$ 1.5 & 66.6 $\pm$ 0.57 & 89.00 $\pm$ 1.35 & 81.69 $\pm$ 1.25 & 90.18 $\pm$ 0.51 & 14.20 \\
    ACM-GCN & 94.75 $\pm$ 3.8 & 95.75 $\pm$ 2.03 & 94.92 $\pm$ 2.88 & 41.62 $\pm$ 1.15 & 69.04 $\pm$ 1.74 & 58.02 $\pm$ 1.86 & 67.01 $\pm$ 0.38 & 88.62 $\pm$ 1.22 & 81.68 $\pm$ 0.97 & 90.66 $\pm$ 0.47 & 7.90 \\
    ACM-GCN+ & 94.92 $\pm$ 2.79 & 96.5 $\pm$ 2.08 & 94.92 $\pm$ 2.79 & 41.79 $\pm$ 1.01 & \cellcolor[rgb]{ .816,  .808,  .808}\textbf{76.08 $\pm$ 2.13} & 69.26 $\pm$ 1.11 & 67.4 $\pm$ 0.44 & \cellcolor[rgb]{ .816,  .808,  .808}\textbf{89.75 $\pm$ 1.16} & 81.65 $\pm$ 1.48 & 90.46 $\pm$ 0.69 & 4.90 \\
    ACM-GCN++ & 93.93 $\pm$ 1.05 & \cellcolor[rgb]{ .816,  .808,  .808}\textbf{97.5 $\pm$ 1.25} & \cellcolor[rgb]{ .816,  .808,  .808}\textbf{96.56 $\pm$ 2} & \cellcolor[rgb]{ .816,  .808,  .808}\textbf{41.86 $\pm$ 1.48} & 75.23 $\pm$ 1.72 & 68.56 $\pm$ 1.33 & 67.3 $\pm$ 0.48 & 89.33 $\pm$ 0.81 & 81.83 $\pm$ 1.65 & 90.39 $\pm$ 0.33 & 4.30 \\
    ACM-Snowball-2 & 95.08 $\pm$ 3.11 & 96.38 $\pm$ 2.59 & 95.74 $\pm$ 2.22 & 41.4 $\pm$ 1.23 & 68.51 $\pm$ 1.7 & 55.97 $\pm$ 2.03 & OOM   & 88.83 $\pm$ 1.49 & 81.58 $\pm$ 1.23 & 90.81 $\pm$ 0.52 & 7.44 \\
    ACM-Snowball-3 & 94.26 $\pm$ 2.57 & 96.62 $\pm$ 1.86 & 94.75 $\pm$ 2.41 & 41.27 $\pm$ 0.8 & 68.4 $\pm$ 2.05 & 55.73 $\pm$ 2.39 & OOM   & 89.59 $\pm$ 1.58 & 81.32 $\pm$ 0.97 & \cellcolor[rgb]{ .816,  .808,  .808}\textbf{91.44 $\pm$ 0.59} & 7.22 \\
    \midrule
    ACMII-GCN & \cellcolor[rgb]{ .816,  .808,  .808}\textbf{95.9 $\pm$ 1.83} & 96.62 $\pm$ 2.44 & 95.08 $\pm$ 2.07 & 41.84 $\pm$ 1.15 & 68.38 $\pm$ 1.36 & 54.53 $\pm$ 2.09 & 67.15 $\pm$ 0.41 & 89.00 $\pm$ 0.72 & 81.79 $\pm$ 0.95 & 90.74 $\pm$ 0.5 & 5.90 \\
    ACMII-Snowball-2 & 95.25 $\pm$ 1.55 & 96.63 $\pm$ 2.24 & 95.25 $\pm$ 1.55 & 41.1 $\pm$ 0.75 & 67.83 $\pm$ 2.63 & 53.48 $\pm$ 0.6 & OOM   & 88.95 $\pm$ 1.04 & 82.07 $\pm$ 1.04 & 90.56 $\pm$ 0.39 & 7.56 \\
    ACMII-Snowball-3 & 93.61 $\pm$ 2.79 & 97.00 $\pm$ 2.63 & 94.75 $\pm$ 3.09 & 40.31 $\pm$ 1.6 & 67.53 $\pm$ 2.83 & 52.31 $\pm$ 1.57 & OOM   & 89.36 $\pm$ 1.26 & 81.56 $\pm$ 1.15 & 91.31 $\pm$ 0.6 & 9.00 \\
    ACMII-GCN+ & 93.93 $\pm$ 3.03 & 96.75 $\pm$ 1.79 & 95.41 $\pm$ 2.82 & 41.5 $\pm$ 1.54 & 75.51 $\pm$ 1.58 & 69.81 $\pm$ 1.11 & 67.44 $\pm$ 0.31 & 89.18 $\pm$ 1.11 & 81.87 $\pm$ 1.38 & 90.96 $\pm$ 0.62 & \cellcolor[rgb]{ .816,  .808,  .808}\textbf{4.4} \\
    ACMII-GCN++ & 92.62 $\pm$ 2.57 & 97.13 $\pm$ 1.68 & 94.75 $\pm$ 2.91 & 41.66 $\pm$ 1.42 & 75.93 $\pm$ 1.71 & \cellcolor[rgb]{ .816,  .808,  .808}\textbf{69.98 $\pm$ 1.53} & \cellcolor[rgb]{ .816,  .808,  .808}\textbf{67.5 $\pm$ 0.53} & 89.47 $\pm$ 1.08	 & 81.76 $\pm$ 1.25 & 90.63 $\pm$ 0.56 & 5.10 \\
    \bottomrule
    \bottomrule
    \end{tabular}%
  \label{tab:sota}%
\end{table}%

\subsection{Comparison with SOTA Models on Fixed 48\%/32\%/20\% Splits}
See table \ref{tab:performance_comparison_fixed_splits} for the results \label{appendix:results_on_fixed_splits_as_geomgcn}
and table \ref{tab:optimal_hyperparameters_fixed_splits} \ref{tab:hyperparameters_acmgcn+_acmgcn++} the optimal searched hyperparameters. The results and comparison give us the same conclusion as in Appendix \ref{sec:full_result_comparison}.
\begin{table}[htbp]
  \centering
  \caption{Experimental results on fixed splits provided by \cite{pei2020geom}: average test accuracy $\pm$ standard deviation on 9 real-world benchmark datasets. The best results are highlighted. Results of Geom-GCN, H$_2$GCN and GPRGNN, LINX, GloGNN, GloGNN++, Diag-NSD, O(d)-NSD, Gen-NSD, NLMLP, NLGCN and NLGAT are from \cite{pei2020geom,zhu2020beyond,lingam2021simple, lim2021new,li2022finding, bodnar2022neural, liu2021non}; results on the rest models are run by ourselves and the hyperparameter searching range is the same as table \ref{tab:real_world_datasets_hyperparameter_searching_range}.}
  \setlength{\tabcolsep}{3pt}
   \scalebox{.65}{
    \begin{tabular}{c|ccccccccc|c}
    \toprule
    \toprule
    Datasets/Models & Cornell & Wisconsin & Texas & Film  & Chameleon & Squirrel & Cora  & Citeseer & PubMed & Average Rank \\
    \midrule
    Geom-GCN & 60.54 $\pm$ 3.67 & 64.51 $\pm$ 3.66 & 66.76 $\pm$ 2.72 & 31.59 $\pm$ 1.15 & 60.00 $\pm$ 2.81 & 38.15 $\pm$ 0.92 & 85.35 $\pm$ 1.57 & \cellcolor[rgb]{ .816,  .808,  .808}\textbf{78.02 $\pm$ 1.15} & 89.95 $\pm$ 0.47 & 18.22 \\
    H2GCN & 82.70 $\pm$ 5.28 & 87.65 $\pm$ 4.98 & 84.86 $\pm$ 7.23 & 35.70 $\pm$ 1.00 & 60.11 $\pm$ 2.15 & 36.48 $\pm$ 1.86 & 87.87 $\pm$ 1.20 & 77.11 $\pm$ 1.57 & 89.49 $\pm$ 0.38 & 15.11 \\
    GPRGCN & 78.11 $\pm$ 6.55 & 82.55 $\pm$ 6.23 & 81.35 $\pm$ 5.32 & 35.16 $\pm$ 0.9 & 62.59 $\pm$ 2.04 & 46.31 $\pm$ 2.46 & 87.95 $\pm$ 1.18 & 77.13 $\pm$ 1.67 & 87.54 $\pm$ 0.38 & 17.67 \\
    FAGCN & 76.76 $\pm$ 5.87 & 79.61 $\pm$ 1.58 & 76.49 $\pm$ 2.87 & 34.82 $\pm$ 1.35 & 46.07 $\pm$ 2.11 & 30.83 $\pm$ 0.69 & 88.05 $\pm$ 1.57 & 77.07 $\pm$ 2.05 & 88.09 $\pm$ 1.38 & 20.00 \\
    GCNII & 77.86 $\pm$ 3.79  & 80.39 $\pm$ 3.40 & 77.57 $\pm$ 3.83 & 37.44 $\pm$ 1.30 & 63.86 $\pm$ 3.04  & 38.47 $\pm$ 1.58 & \cellcolor[rgb]{ .816,  .808,  .808}\textbf{88.37 $\pm$ 1.25 } & 77.33 $\pm$ 1.48 & \cellcolor[rgb]{ .816,  .808,  .808}\textbf{90.15 $\pm$ 0.43} & 12.44 \\
    MixHop & 73.51 $\pm$ 6.34  & 75.88 $\pm$ 4.90  & 77.84 $\pm$ 7.73  & 32.22 $\pm$ 2.34 & 60.50 $\pm$ 2.53  &  43.80 $\pm$ 1.48  & 87.61 $\pm$ 0.85  & 76.26 $\pm$1.33  & 85.31 $\pm$ 0.61 & 20.78 \\
    WRGAT & 81.62 $\pm$3.90  & 86.98 $\pm$ 3.78  & 83.62 $\pm$ 5.50  & 36.53 $\pm$ 0.77  & 65.24 $\pm$ 0.87  & 48.85 $\pm$ 0.78 & 88.20 $\pm$ 2.26  & 76.81 $\pm$ 1.89  & 88.52 $\pm$ 0.92 & 14.33 \\
    GGCN  & 85.68 $\pm$ 6.63  & 86.86 $\pm$ 3.29  & 84.86 $\pm$ 4.55 & 37.54 $\pm$ 1.56  & 71.14 $\pm$1.84  & 55.17 $\pm$ 1.58 & 87.95 $\pm$ 1.05  & 77.14 $\pm$ 1.45  & 89.15 $\pm$ 0.37 & 10.22 \\
    LINKX &  77.84 $\pm$ 5.81  & 75.49 $\pm$ 5.72 & 74.60 $\pm$ 8.37  & 36.10 $\pm$ 1.55  & 68.42 $\pm$ 1.38  & 61.81 $\pm$ 1.80 & 84.64 $\pm$ 1.13 & 73.19 $\pm$ 0.99 & 87.86 $\pm$ 0.77 & 18.78 \\
    GloGNN & 83.51 $\pm$ 4.26 & 87.06 $\pm$ 3.53  & 84.32 $\pm$ 4.15  & 37.35 $\pm$ 1.30 & 69.78 $\pm$ 2.42  & 57.54 $\pm$ 1.39  & 88.31 $\pm$ 1.13 & 77.41 $\pm$ 1.65  & 89.62 $\pm$ 0.35 & 8.78 \\
    GloGNN++ & 85.95 $\pm$ 5.10  &  88.04 $\pm$ 3.22  & 84.05 $\pm$ 4.90 & 37.70 $\pm$ 1.40  & 71.21 $\pm$ 1.84  & 57.88 $\pm$ 1.76  & 88.33 $\pm$ 1.09  & 77.22 $\pm$ 1.78  & 89.24 $\pm$ 0.39 & 7.33 \\
    ACM-SGC-1 & 82.43 $\pm$ 5.44 & 86.47 $\pm$ 3.77 & 81.89 $\pm$ 4.53 & 35.49 $\pm$ 1.06 & 63.99 $\pm$ 1.66 & 45.00 $\pm$ 1.4 & 86.9 $\pm$ 1.38 & 76.73 $\pm$ 1.59 & 88.49 $\pm$ 0.51 & 17.56 \\
    ACM-SGC-2 & 82.43 $\pm$ 5.44 & 86.47 $\pm$ 3.77 & 81.89 $\pm$ 4.53 & 36.04 $\pm$ 0.83 & 59.21 $\pm$ 2.22 & 40.02 $\pm$ 0.96 & 87.69 $\pm$ 1.07 & 76.59 $\pm$ 1.69 & 89.01 $\pm$ 0.6 & 17.67 \\
    Diag-NSD & \cellcolor[rgb]{ .816,  .808,  .808}\textbf{86.49 $\pm$ 7.35} & 88.63 $\pm$ 2.75 & 85.67 $\pm$ 6.95 & 37.79 $\pm$ 1.01 & 68.68 $\pm$ 1.73 & 54.78 $\pm$ 1.81 & 87.14 $\pm$ 1.06 & 77.14 $\pm$ 1.85 & 89.42 $\pm$ 0.43 & 9.00 \\
    O(d)-NSD & 84.86 $\pm$ 4.71 & \cellcolor[rgb]{ .816,  .808,  .808}\textbf{89.41 $\pm$ 4.74} & 85.95 $\pm$ 5.51 & 37.81 $\pm$ 1.15 & 68.04 $\pm$ 1.58 & 56.34 $\pm$ 1.32 & 86.90 $\pm$ 1.13 & 76.70 $\pm$ 1.57 & 89.49 $\pm$ 0.40 & 10.44 \\
    Gen-NSD & 85.68 $\pm$ 6.51 & 89.21 $\pm$ 3.84 & 82.97 $\pm$ 5.13  & 37.80 $\pm$ 1.22 & 67.93 $\pm$ 1.58 & 53.17 $\pm$ 1.31 & 87.30 $\pm$ 1.15 & 76.32 $\pm$ 1.65 & 89.33 $\pm$ 0.35 & 11.67 \\
    NLMLP  & 84.9 $\pm$ 5.7 & 87.3 $\pm$ 4.3  & 85.4 $\pm$ 3.8 & \cellcolor[rgb]{ .816,  .808,  .808}\textbf{37.9 $\pm$ 1.3} & 50.7 $\pm$ 2.2 & 33.7 $\pm$ 1.5 & 76.9 $\pm$ 1.8 & 73.4 $\pm$ 1.9 & 88.2 $\pm$ 0.5 & 16.67 \\
    NLGCN  & 57.6 $\pm$ 5.5 & 60.2 $\pm$ 5.3  & 65.5 $\pm$ 6.6 & 31.6 $\pm$ 1.0 & 70.1 $\pm$ 2.9 & 59.0 $\pm$ 1.2 & 88.1 $\pm$ 1.0 & 75.2 $\pm$ 1.4 & 89.0 $\pm$ 0.5 & 17.44 \\
    NLGAT  & 54.7 $\pm$ 7.6 & 56.9 $\pm$ 7.3 & 62.6 $\pm$ 7.1 & 29.5 $\pm$ 1.3 & 65.7 $\pm$ 1.4 & 56.8 $\pm$ 2.5 & 88.5 $\pm$ 1.8 & 76.2 $\pm$ 1.6 & 88.2 $\pm$ 0.3 & 18.56 \\
    \midrule
    ACM-GCN & 85.14 $\pm$ 6.07 & 88.43 $\pm$ 3.22 & 87.84 $\pm$ 4.4 & 36.63 $\pm$ 0.84 & 69.14 $\pm$ 1.91 & 55.19 $\pm$ 1.49 & 87.91 $\pm$ 0.95 & 77.32 $\pm$ 1.7 & 90.00 $\pm$ 0.52 & 8.11 \\
    ACMII-GCN & 85.95 $\pm$ 5.64 & 87.45 $\pm$ 3.74 & 86.76 $\pm$ 4.75 & 36.31 $\pm$ 1.2 & 68.46 $\pm$ 1.7 & 51.8 $\pm$ 1.5 & 88.01 $\pm$ 1.08 & 77.15 $\pm$ 1.45 & 89.89 $\pm$ 0.43 & 9.33 \\
    ACM-GCN+ & 85.68 $\pm$ 4.84 & 88.43 $\pm$ 2.39 & \cellcolor[rgb]{ .816,  .808,  .808}\textbf{88.38 $\pm$ 3.64} & 36.26 $\pm$ 1.34 & 74.47 $\pm$ 1.84 & 66.98 $\pm$ 1.71 & 88.05 $\pm$ 0.99 & 77.67 $\pm$ 1.19 & 89.82 $\pm$ 0.41 & 5.33 \\
    ACMII-GCN+ & 85.41 $\pm$ 5.3 & 88.04 $\pm$ 3.66 & 88.11 $\pm$ 3.24 & 36.14 $\pm$ 1.44 & 74.56 $\pm$ 2.08 & 67.07 $\pm$ 1.65 & 88.19 $\pm$ 1.17 & 77.2 $\pm$ 1.61 & 89.78 $\pm$ 0.49  & 6.78 \\
    ACM-GCN++ & 85.68 $\pm$ 5.8 & 88.24 $\pm$ 3.16 & \cellcolor[rgb]{ .816,  .808,  .808}\textbf{88.38 $\pm$ 3.43} & 37.31 $\pm$ 1.09 & 74.41 $\pm$ 1.49 & 67.06 $\pm$ 1.66 & 88.11 $\pm$ 0.96 & 77.46 $\pm$ 1.65 & 89.65 $\pm$ 0.58 & 5.33 \\
    ACMII-GCN++ & \cellcolor[rgb]{ .816,  .808,  .808}\textbf{86.49 $\pm$ 6.73} & 88.43 $\pm$ 3.66 & \cellcolor[rgb]{ .816,  .808,  .808}\textbf{88.38 $\pm$ 3.43} & 37.09 $\pm$ 1.32 & \cellcolor[rgb]{ .816,  .808,  .808}\textbf{74.76 $\pm$ 2.2} & \cellcolor[rgb]{ .816,  .808,  .808}\textbf{67.4 $\pm$ 2.21} & 88.25 $\pm$ 0.96 & 77.12 $\pm$ 1.58 & 89.71 $\pm$ 0.48 & \cellcolor[rgb]{ .816,  .808,  .808}\textbf{4.78} \\
    \bottomrule
    \bottomrule
    \end{tabular}%
    }
  \label{tab:performance_comparison_fixed_splits}%
\end{table}%

\subsection{Discussion of Random Walk and Symmetric Renormalized Filters}
\label{appendix:rw_symmetric_comparision}
\begin{table}[htbp]
  \centering
  \caption{Comparison of random walk and symmetric renormalized filters}
    \begin{tabular}{c|cc|cc}
    \toprule
    \toprule
    \multirow{2}[4]{*}{Datasets/Models} & \multicolumn{2}{c|}{RW} & \multicolumn{2}{c}{Symmetric} \\
\cmidrule{2-5}          & ACM   & ACMII & ACM   & ACMII \\
    \midrule
    Cornell & 94.75 $\pm$ 3.8 & \cellcolor[rgb]{ .647,  .647,  .647}\textbf{95.9 $\pm$ 1.83} & 94.92 $\pm$ 2.48 & 94.1 $\pm$ 2.56 \\
    Wisconsin & 95.75 $\pm$ 2.03 & \cellcolor[rgb]{ .647,  .647,  .647}\textbf{96.62 $\pm$ 2.44} & 95.63 $\pm$ 2.81 & 96.25 $\pm$ 2.5 \\
    Texas & 94.92 $\pm$ 2.88 & \cellcolor[rgb]{ .647,  .647,  .647}\textbf{95.08 $\pm$ 2.07} & 94.75 $\pm$ 2.01 & 94.59 $\pm$ 2.65 \\
    Film  & 41.62 $\pm$ 1.15 & \cellcolor[rgb]{ .647,  .647,  .647}\textbf{41.84 $\pm$ 1.15} & 41.58 $\pm$ 1.3 & 41.65 $\pm$ 0.6 \\
    Chameleon & \cellcolor[rgb]{ .647,  .647,  .647}\textbf{69.04 $\pm$ 1.74} & 68.38 $\pm$ 1.36 & 67.9 $\pm$ 2.76 & 68.03 $\pm$ 1.68 \\
    Squirrel & \cellcolor[rgb]{ .647,  .647,  .647}\textbf{58.02 $\pm$ 1.86} & 54.53 $\pm$ 2.09 & 54.18 $\pm$ 1.35 & 53.68 $\pm$ 1.74  \\
    Cora  & 88.62 $\pm$ 1.22 & \cellcolor[rgb]{ .647,  .647,  .647}\textbf{89.00 $\pm$ 0.72} & 88.65 $\pm$ 1.26 & 88.19 $\pm$ 1.38 \\
    Citeseer & 81.68 $\pm$ 0.97 & 81.79 $\pm$ 0.95 & \cellcolor[rgb]{ .647,  .647,  .647}\textbf{81.84 $\pm$ 1.15} & 81.81 $\pm$ 0.86 \\
    PubMed & 90.66 $\pm$ 0.47 & \cellcolor[rgb]{ .647,  .647,  .647}\textbf{90.74 $\pm$ 0.5} & 90.59 $\pm$ 0.81 & 90.54 $\pm$ 0.59 \\
    \bottomrule
    \bottomrule
    \end{tabular}%
  \label{tab:rw_symmetric_filter_comparison}%
\end{table}%
The definitions of the similarity matrix, (modified) aggregation similarity score and diversification distinguishability value can be extended to symmetric normalized Laplacian or other aggregation operations. Yet unfortunately, we cannot extend Theorem 1 at this moment, because we need a condition that the row sum of $\hat{A}$ is not greater than 1 in the proof. This condition is guaranteed for random walk normalized Laplacian but not for symmetric normalized Laplacian. While in practice, we evaluate our models with symmetric filters and compare them with random walk filters. From table \ref{tab:rw_symmetric_filter_comparison} we can see that, there are no big differences between these two filters.
\subsection{Ablation Study of $W_\text{mix}$}
\label{appendix:ablation_w_mix}
\begin{table}[htbp]
  \centering
  \caption{Ablation study of $W_\text{mix}$}
    \begin{tabular}{c|cc|cc}
    \toprule
    \toprule
    \multirow{2}[4]{*}{Datasets/Models} & \multicolumn{2}{c|}{With $W_\text{mix}$} & \multicolumn{2}{c}{Without $W_\text{mix}$} \\
\cmidrule{2-5}          & ACM   & ACMII & ACM   & ACMII \\
    \midrule
    Cornell & 94.75 $\pm$ 3.8 & \cellcolor[rgb]{ .647,  .647,  .647}\textbf{95.9 $\pm$ 1.83} & 93.61 $\pm$ 2.37 & 90.49 $\pm$ 2.72 \\
    Wisconsin & 95.75 $\pm$ 2.03 & 96.62 $\pm$ 2.44 & 95 $\pm$ 2.5 & \cellcolor[rgb]{ .647,  .647,  .647}\textbf{97.50 $\pm$ 1.25} \\
    Texas & 94.92 $\pm$ 2.88 & \cellcolor[rgb]{ .647,  .647,  .647}\textbf{95.08 $\pm$ 2.07} & 94.92 $\pm$ 2.79 & 94.92 $\pm$ 2.79 \\
    Film  & 41.62 $\pm$ 1.15 & \cellcolor[rgb]{ .647,  .647,  .647}\textbf{41.84 $\pm$ 1.15} & 40.79 $\pm$ 1.01 & 40.86 $\pm$ 1.48 \\
    Chameleon & \cellcolor[rgb]{ .647,  .647,  .647}\textbf{69.04 $\pm$ 1.74} & 68.38 $\pm$ 1.36 & 68.16 $\pm$ 1.79 & 66.78 $\pm$ 2.79 \\
    Squirrel & \cellcolor[rgb]{ .647,  .647,  .647}\textbf{58.02 $\pm$ 1.86} & 54.53 $\pm$ 2.09 & 55.35 $\pm$ 1.72 & 52.98 $\pm$ 1.66 \\
    Cora  & 88.62 $\pm$ 1.22 & \cellcolor[rgb]{ .647,  .647,  .647}\textbf{89.00 $\pm$ 0.72} & 88.41 $\pm$ 1.63 & 88.72 $\pm$ 1.5 \\
    Citeseer & 81.68 $\pm$ 0.97 & \cellcolor[rgb]{ .647,  .647,  .647}\textbf{81.79 $\pm$ 0.95} & 81.65 $\pm$ 1.48 & 81.72 $\pm$ 1.58 \\
    PubMed & 90.66 $\pm$ 0.47 & \cellcolor[rgb]{ .647,  .647,  .647}\textbf{90.74 $\pm$ 0.5} & 90.46 $\pm$ 0.69 & 90.39 $\pm$ 1.33 \\
    \bottomrule
    \bottomrule
    \end{tabular}%
  \label{tab:ablation_W_mix}%
\end{table}%

From table \ref{tab:ablation_W_mix} we can see that ACM(II) with $W_\text{mix}$ shows superiority in most datasets, although it is not statistically significant on some of them.

One possible explanation of the function of $W_\text{mix}$ is that it could help alleviate the dominance and bias to majority: Suppose in a dataset, most of the nodes need more information from LP channel than HP and identity channels, then $W_L, W_H, W_I$ tend to learn larger $\alpha_L$ than $\alpha_H$ and $\alpha_I$. For the minority nodes that need more information from HP or identity channels, they are hard to get large $\alpha_H$ or $\alpha_I$ values because $W_L, W_H, W_I$ are biased to the majority. And $W_\text{mix}$ can help us to learn more diverse alpha values when $W_L, W_H, W_I$ are biased.

Attention with more complicated design can be found for the node-wise adaptive channel mixing mechanism, but we do not explore this direction deeper in this paper because investigating attention function is not the main contribution of our paper.

\subsection{Learn Weights with Raw Features v.s. Combined Features}
\label{appendix:raw_combined_feature_comparison}

\begin{table}[htbp]
  \centering
  \caption{Performance comparison between raw features and combined features}
    \begin{tabular}{c|cccc}
    \toprule
    \toprule
    \multirow{2}[4]{*}{Datasets/Models} & \multicolumn{2}{c|}{With Raw Features} & \multicolumn{2}{c}{With Combined Features} \\
\cmidrule{2-5}          & ACM   & \multicolumn{1}{c|}{ACMII} & ACM   & ACMII \\
    \midrule
    Cornell & 94.75 $\pm$ 3.8 & \multicolumn{1}{c|}{\cellcolor[rgb]{ .647,  .647,  .647}\textbf{95.9 $\pm$ 1.83}} & 95.08 $\pm$ 2.64 & 93.93 $\pm$ 3.52 \\
    Wisconsin & 95.75 $\pm$ 2.03 & \multicolumn{1}{c|}{\cellcolor[rgb]{ .647,  .647,  .647}\textbf{96.62 $\pm$ 2.44}} & 96.12 $\pm$ 1.31 & 96 $\pm$ 	2 \\
    Texas & 94.92 $\pm$ 2.88 & \multicolumn{1}{c|}{\cellcolor[rgb]{ .647,  .647,  .647}\textbf{95.08 $\pm$ 2.07}} & 94.92 $\pm$ 2.48 & 94.59 $\pm$ 2.94 \\
    Film  & 41.62 $\pm$ 1.15 & \multicolumn{1}{c|}{\cellcolor[rgb]{ .647,  .647,  .647}\textbf{41.84 $\pm$ 1.15}} & 41.62 $\pm$ 1.34 & 41.44 $\pm$ 1.18 \\
    Chameleon & \cellcolor[rgb]{ .647,  .647,  .647}\textbf{69.04 $\pm$ 1.74} & \multicolumn{1}{c|}{68.38 $\pm$ 1.36} & 68.82 $\pm$ 2.18 & 68.53 $\pm$ 3.08 \\
    Squirrel & \cellcolor[rgb]{ .647,  .647,  .647}\textbf{58.02 $\pm$ 1.86} & \multicolumn{1}{c|}{54.53 $\pm$ 2.09} & 57.48 $\pm$ 1.68 & 53.28 $\pm$ 1.08 \\
    Cora  & 88.62 $\pm$ 1.22 & \multicolumn{1}{c|}{\cellcolor[rgb]{ .647,  .647,  .647}\textbf{89.00 $\pm$ 0.72}} & 88.59 $\pm$ 1.04 & 88.75 $\pm$ 0.83 \\
    Citeseer & 81.68 $\pm$ 0.97 & 81.79 $\pm$ 0.95 & \cellcolor[rgb]{ .647,  .647,  .647}\textbf{81.9 $\pm$ 1.27} & 81.76 $\pm$ 1.05 \\
    PubMed & 90.66 $\pm$ 0.47 & 90.74 $\pm$ 0.5 & \cellcolor[rgb]{ .647,  .647,  .647}\textbf{90.75 $\pm$ 0.77 } & 90.58 $\pm$ 0.64 \\
    \bottomrule
    \bottomrule
    \end{tabular}%
  \label{tab:raw_combined_feature_comparison}%
\end{table}%

Construct the combined feature ${H}^{l}_\text{Comb} = [{H}^{l}_L,{H}^{l}_H,{H}^{l}_I]$, 
Replace the first line in Step 2 by the following lines:
\begin{align*}
   & \tilde{\alpha}_L^l = \sigma\left({H}^{l}_\text{Comb} \tilde{W}^{l}_L \right),\ \tilde{\alpha}_H^l = \sigma \left({H}^{l}_\text{Comb}\tilde{W}^{l}_H \right),\ \tilde{\alpha}_I^l = \sigma \left({H}^{l}_\text{Comb} \tilde{W}^{l}_I \right),\ \tilde{W}_L^{l-1},\ \tilde{W}_H^{l-1},\ \tilde{W}_I^{l-1} \in \mathbb{R}^{3F_l \times 1}\\
   & \left[{\alpha}_L^l, {\alpha}_H^l, {\alpha}_I^l \right] = \text{Softmax}\left((\left[\tilde{\alpha}_L^l,\tilde{\alpha}_H^l,\tilde{\alpha}_I^l\right]/T) W_\text{Mix}^l \right) \in \mathbb{R}^{N\times 3},\ T \in \mathbb{R} \text{ temperature},\ W_\text{Mix}^l \in \mathbb{R}^{3\times 3}; 
\end{align*}
The performance comparison can be found in table \ref{tab:raw_combined_feature_comparison}. From the results, we do not find significant difference between the frameworks with combined features and raw features. The reason is that the necessary nonlinear information from each channel is combined in $\left[\tilde{\alpha}_L^l,\tilde{\alpha}_H^l,\tilde{\alpha}_I^l\right]$ and $W_\text{Mix}^l$ is enough to learn to mix the combined weights from different channels. The learning of redundant information in the feature extraction step for each channel will not improve the performance. Meanwhile, A disadvantage of the combined feature is that it increases the computational cost. Thus, we decide to use the raw features.
\subsection{$H_\text{node}^v$ Distributions of Different Datasets}
\begin{figure}[H]
    \centering
     {
    \subfloat[ \texttt{Cornell}]{
     \captionsetup{justification = centering}
     \includegraphics[width=0.33\textwidth]{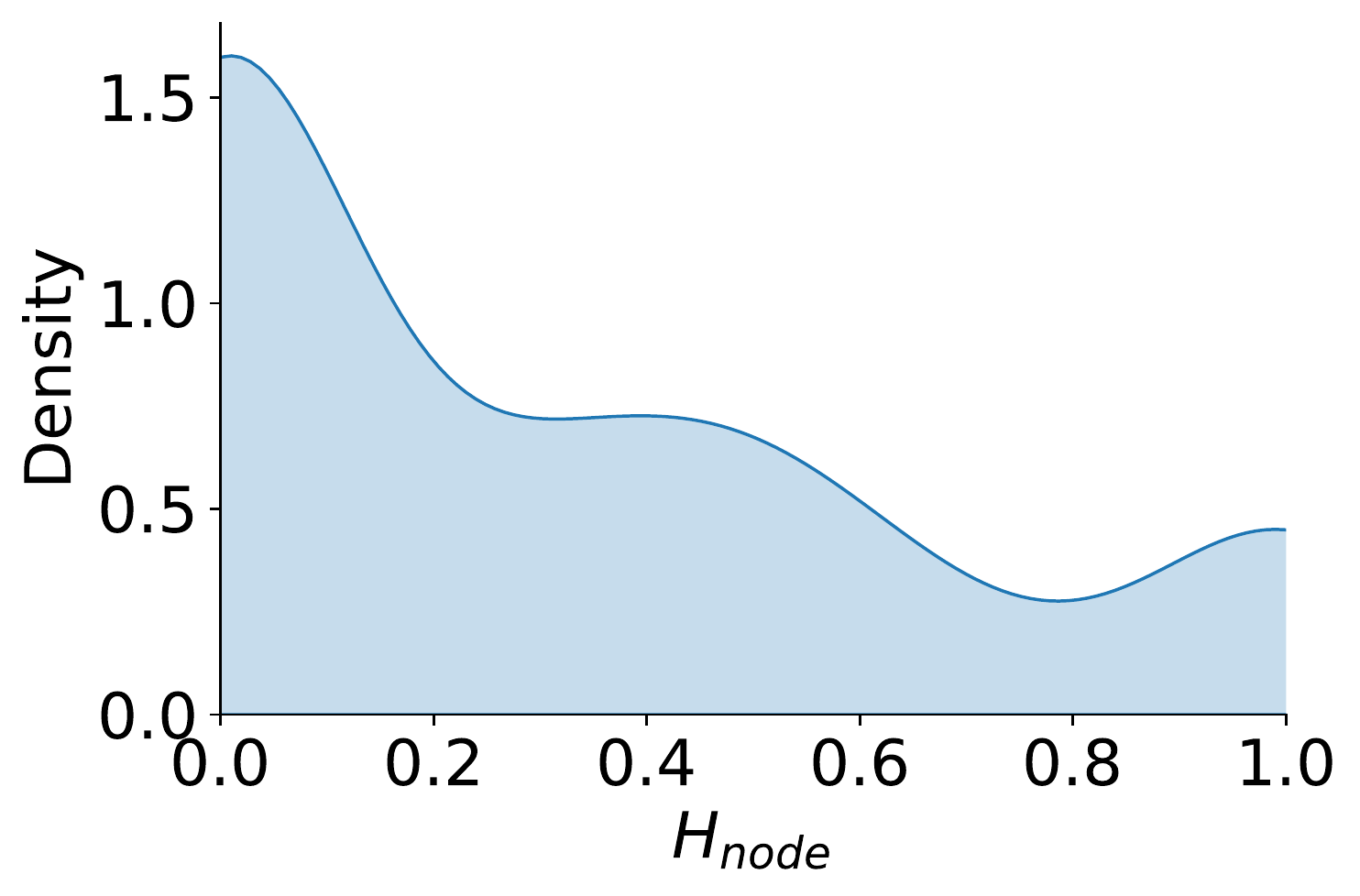}
     } 
     \subfloat[ \texttt{Wisconsin}]{
     \captionsetup{justification = centering}
     \includegraphics[width=0.33\textwidth]{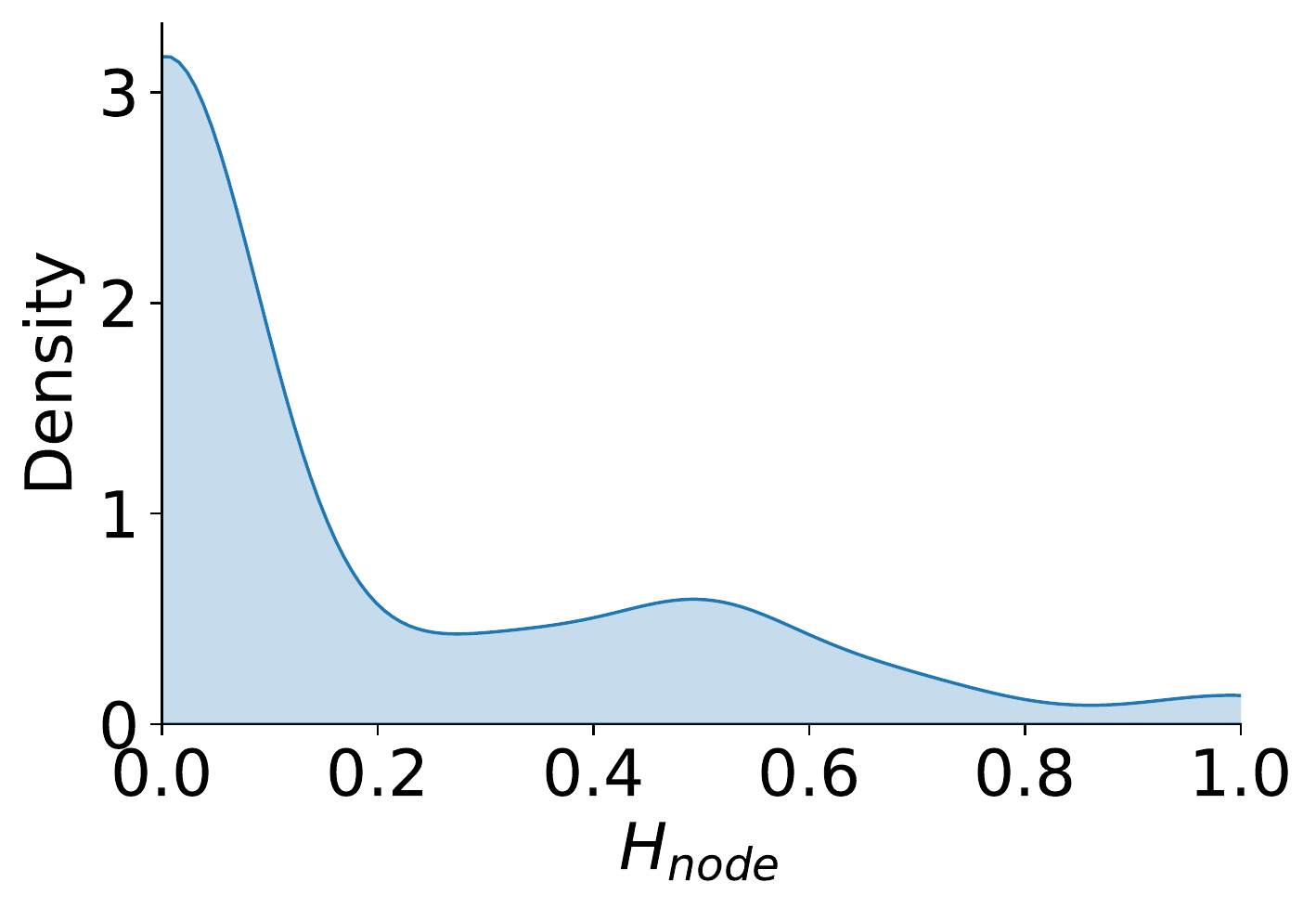}
     } 
     \subfloat[ \texttt{Texas}]{
     \captionsetup{justification = centering}
     \includegraphics[width=0.33\textwidth]{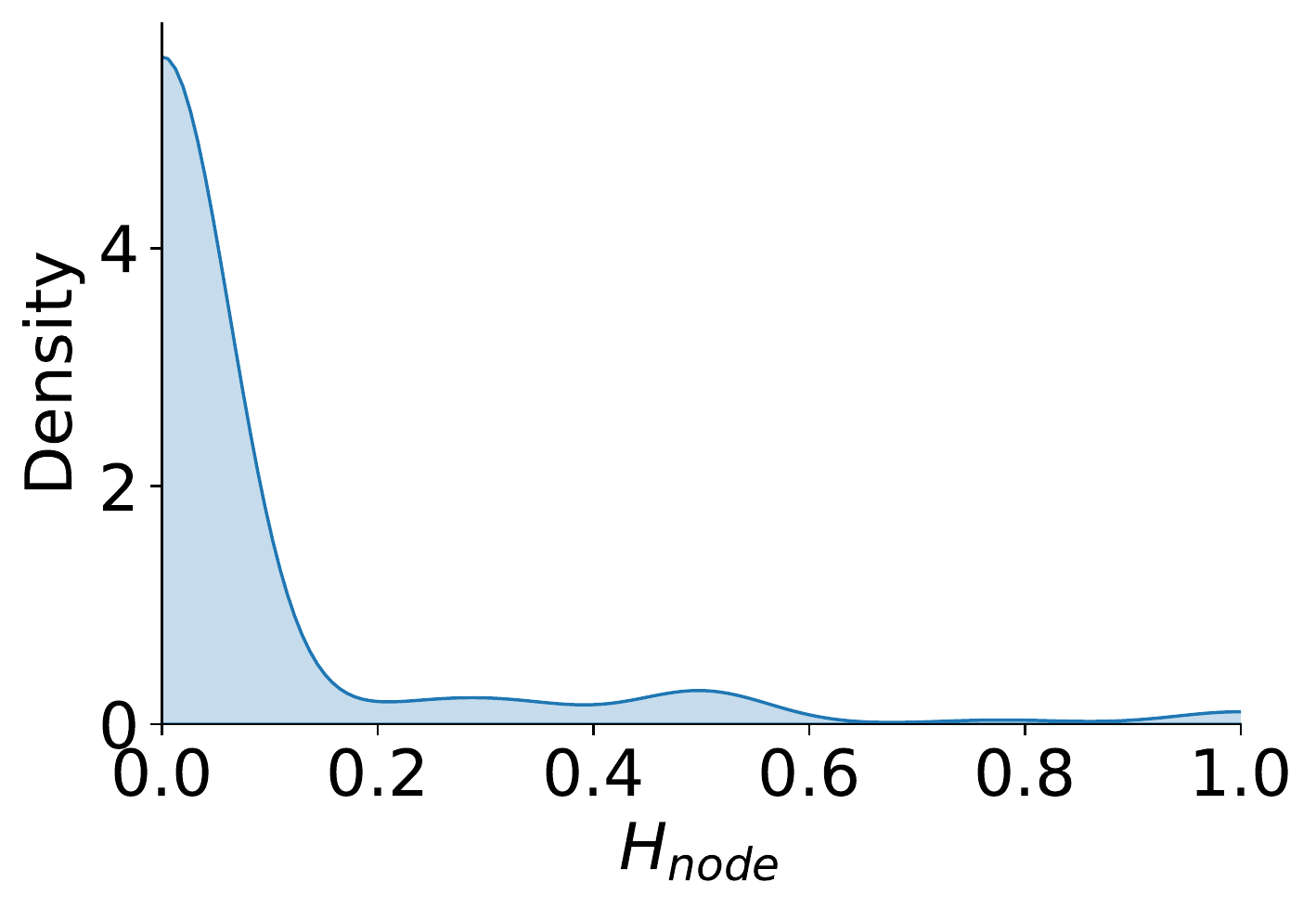}
     } \\
     \subfloat[ \texttt{Chameleon}]{
     \captionsetup{justification = centering}
     \includegraphics[width=0.33\textwidth]{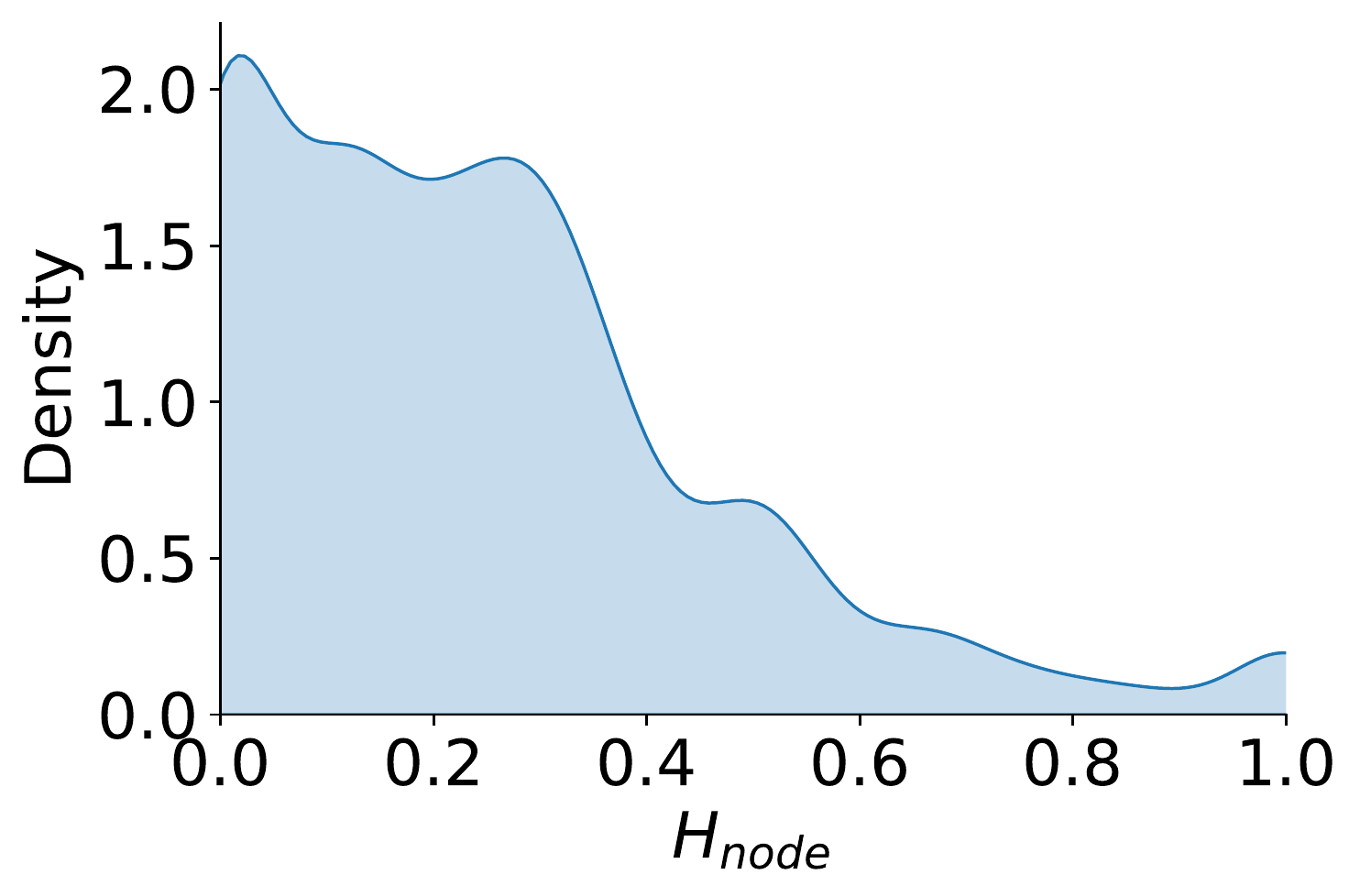}
     } 
     \subfloat[ \texttt{Squirrel}]{
     \captionsetup{justification = centering}
     \includegraphics[width=0.33\textwidth]{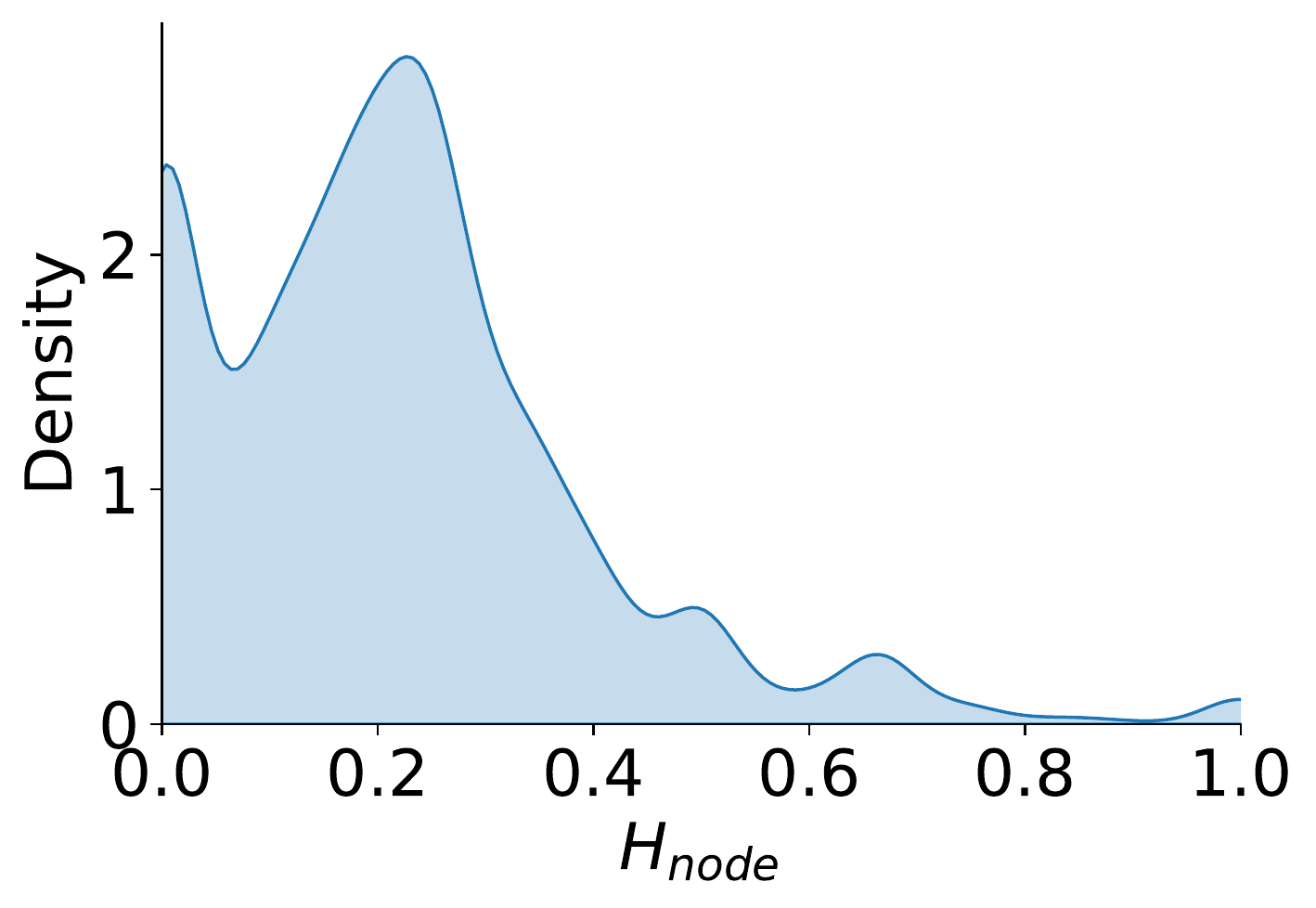}
     } 
     \subfloat[ \texttt{Film}]{
     \captionsetup{justification = centering}
     \includegraphics[width=0.33\textwidth]{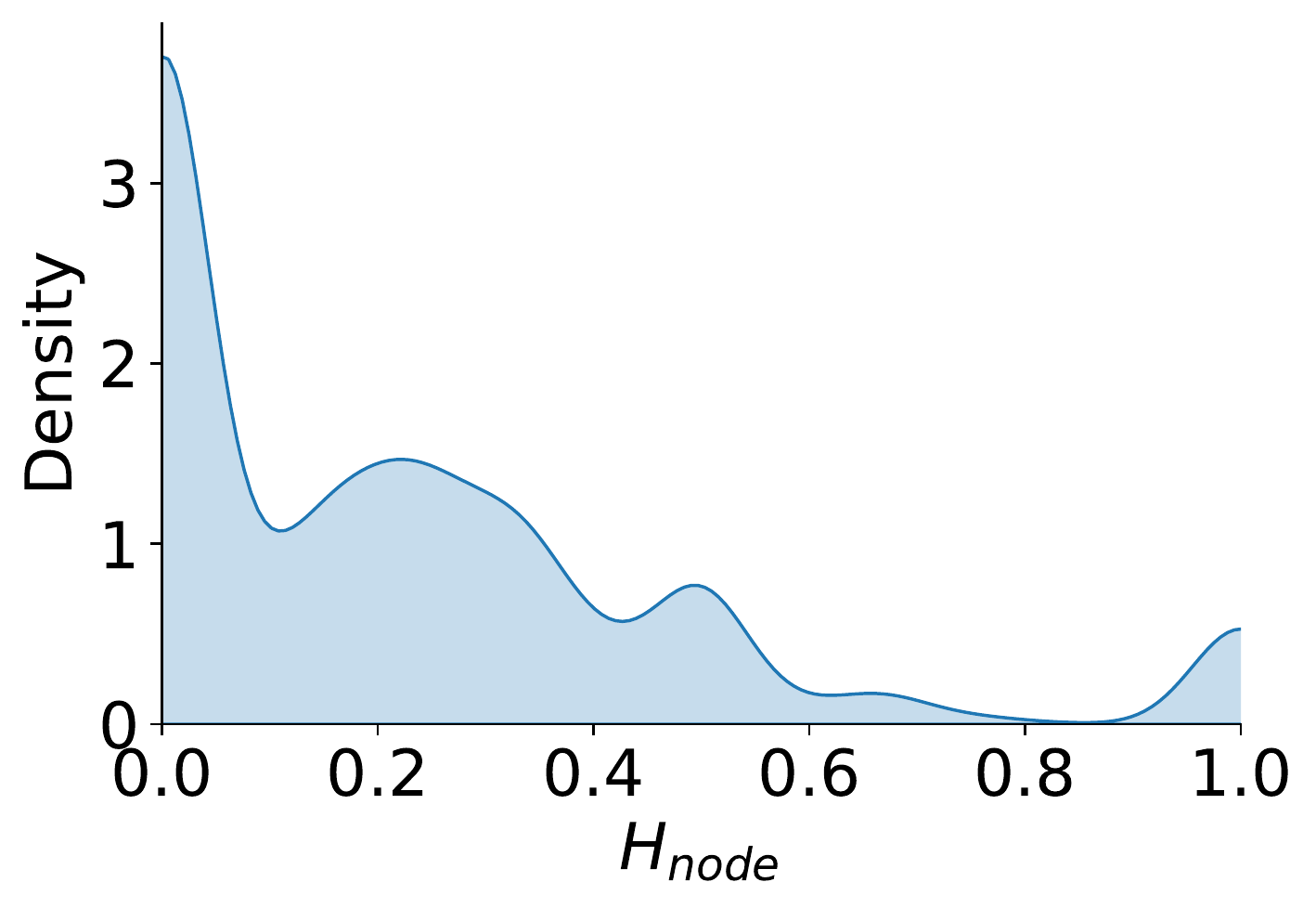}
     } \\
     \subfloat[ \texttt{Cora}]{
     \captionsetup{justification = centering}
     \includegraphics[width=0.33\textwidth]{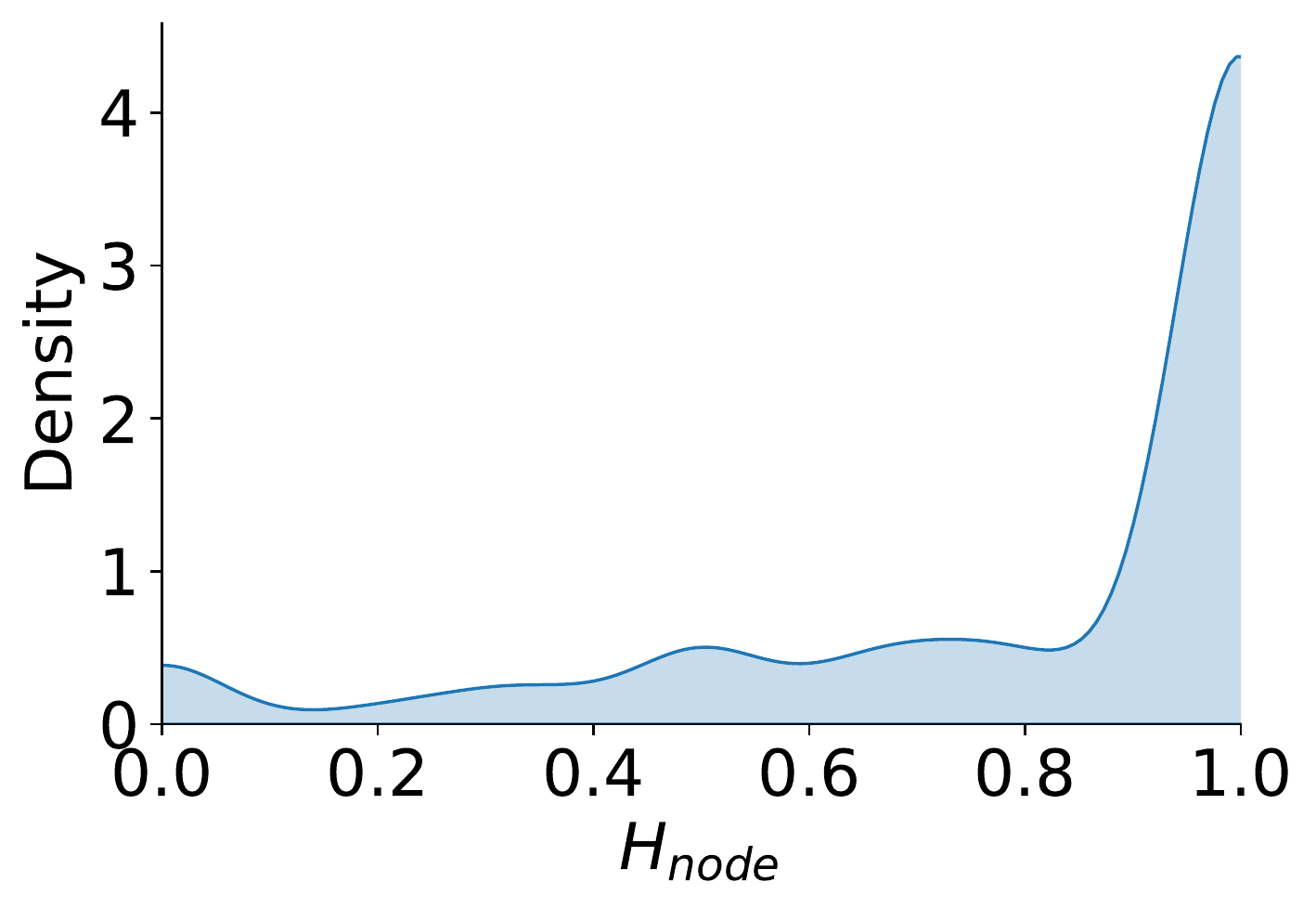}
     } 
     \subfloat[ \texttt{Citeseer}]{
     \captionsetup{justification = centering}
     \includegraphics[width=0.33\textwidth]{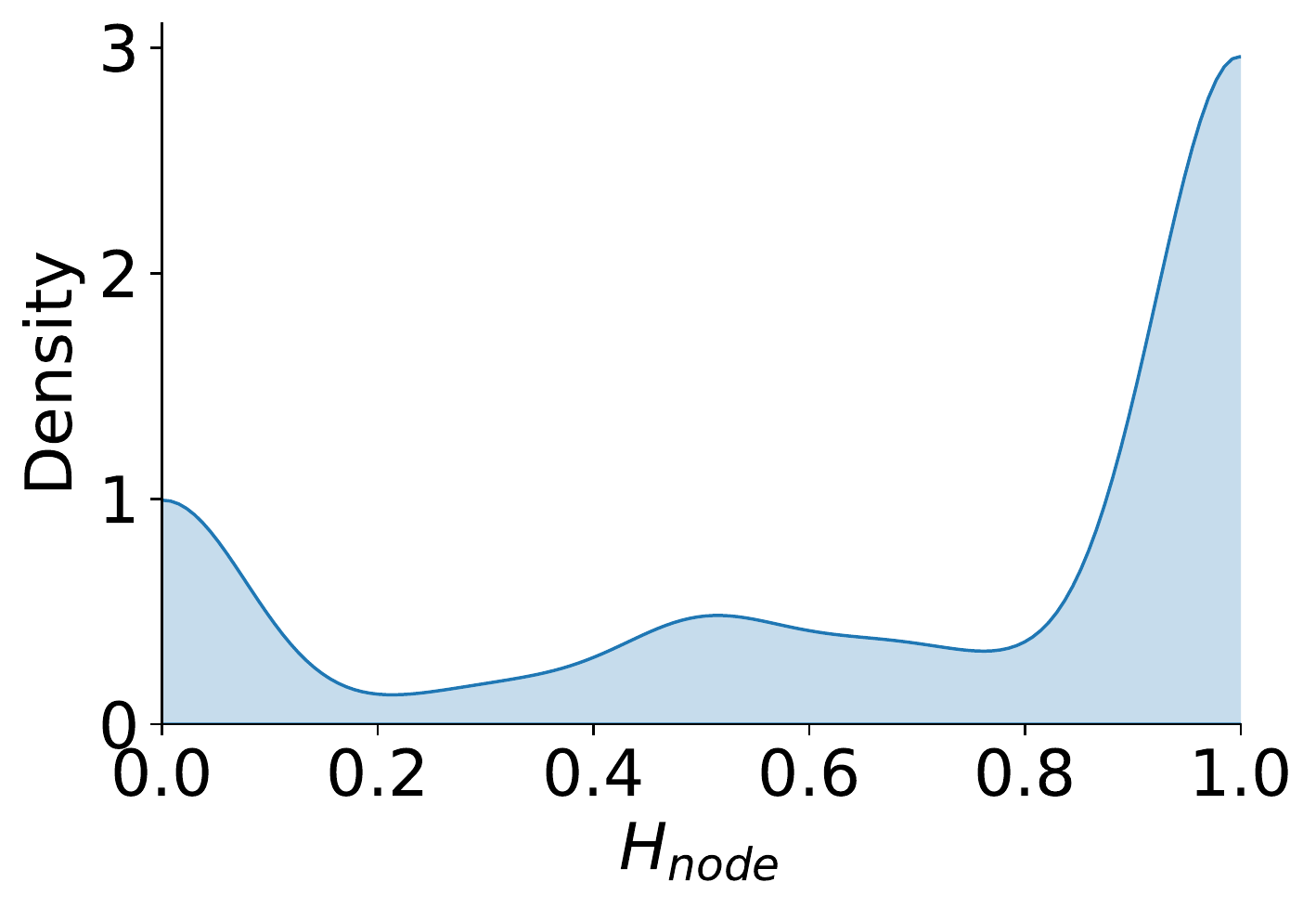}
     } 
     \subfloat[ \texttt{Pubmed}]{
     \captionsetup{justification = centering}
     \includegraphics[width=0.33\textwidth]{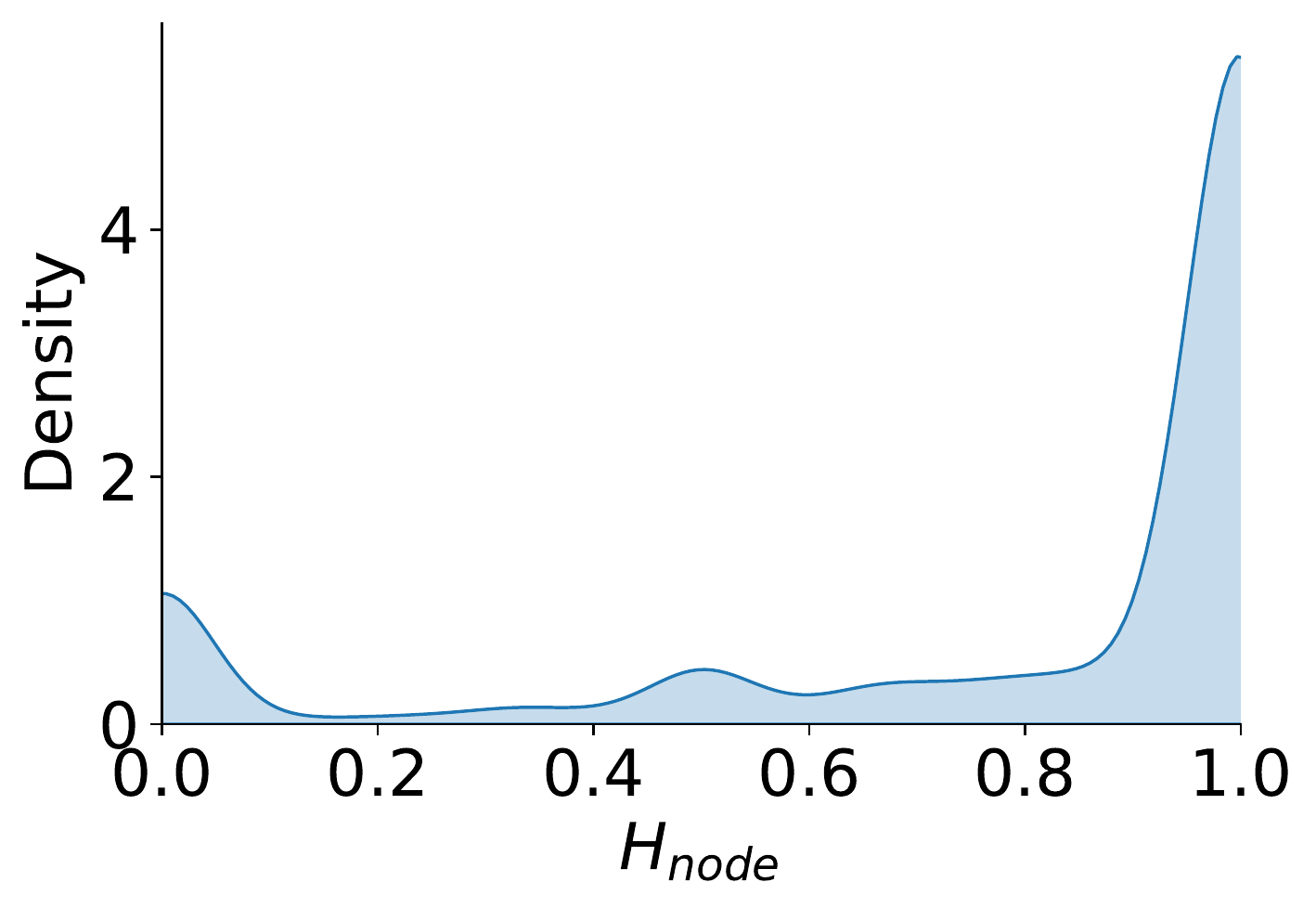}
     } 
     }
     \caption{$H_\text{node}^v$ distributions of different datasets}
     \label{fig:local_homo_distributions}
\end{figure}

See Figure \ref{fig:local_homo_distributions} for $H_\text{node}^v$ distributions. We can see that \textit{Wisconsin} and \textit{Texas} have high density in low homophily area, \textit{Cornell, Chameleon, Squirrel} and \textit{Film} have high density in low and middle homophily area, \textit{Cora, CiteSeer} and \textit{PubMed} have high density in high homophily area.

\label{appendix:alpha_values_output}

\subsection{Distributions of Learned $\alpha_L, \alpha_H, \alpha_I$ in the Hidden and Output Layers of ACN-GCN}

\begin{figure}[H]
    \centering
     {
    \subfloat[ \texttt{Cornell}]{
     \captionsetup{justification = centering}
     \includegraphics[width=0.35\textwidth]{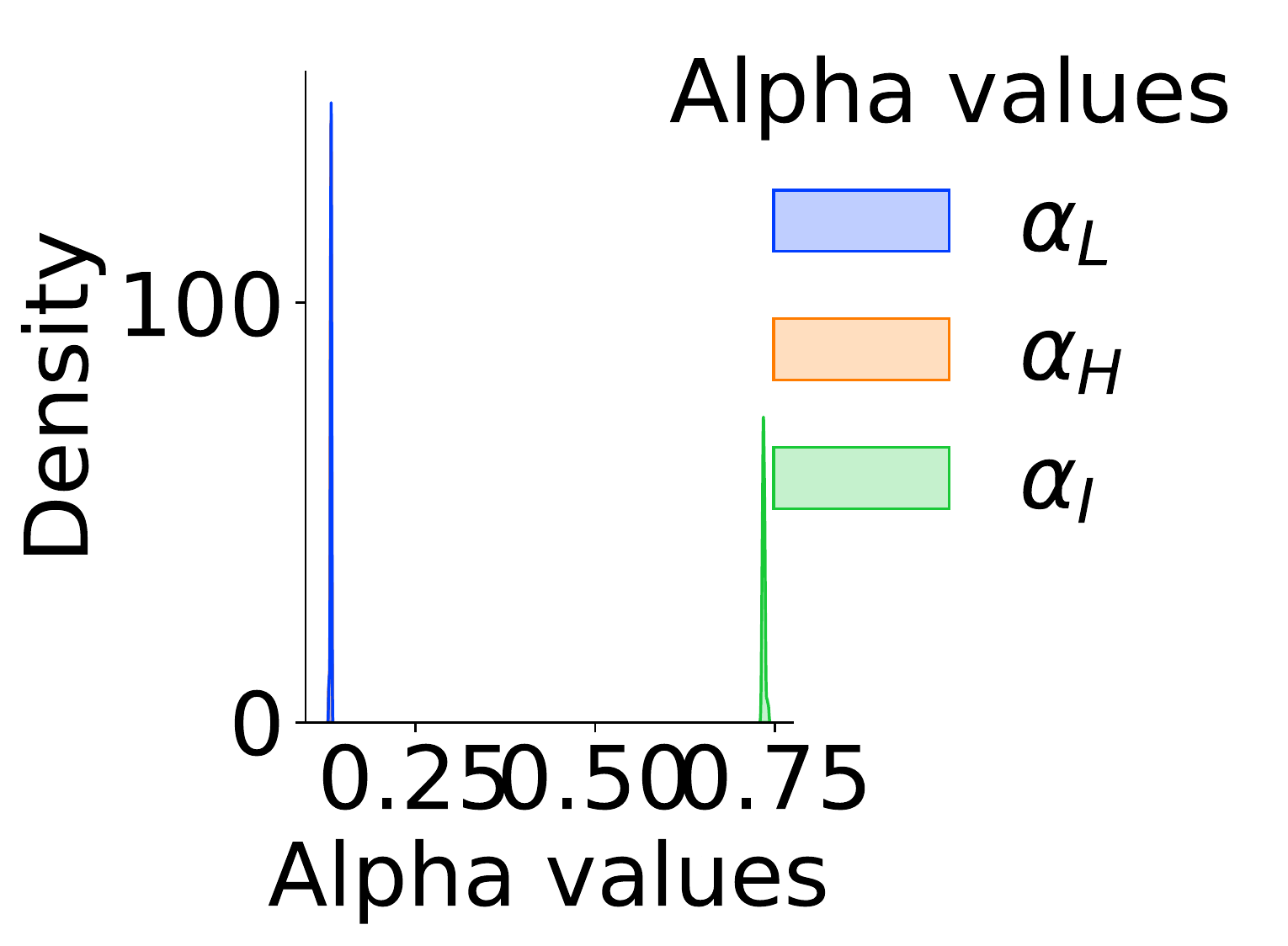}
     } 
     \subfloat[ \texttt{Wisconsin}]{
     \captionsetup{justification = centering}
     \includegraphics[width=0.35\textwidth]{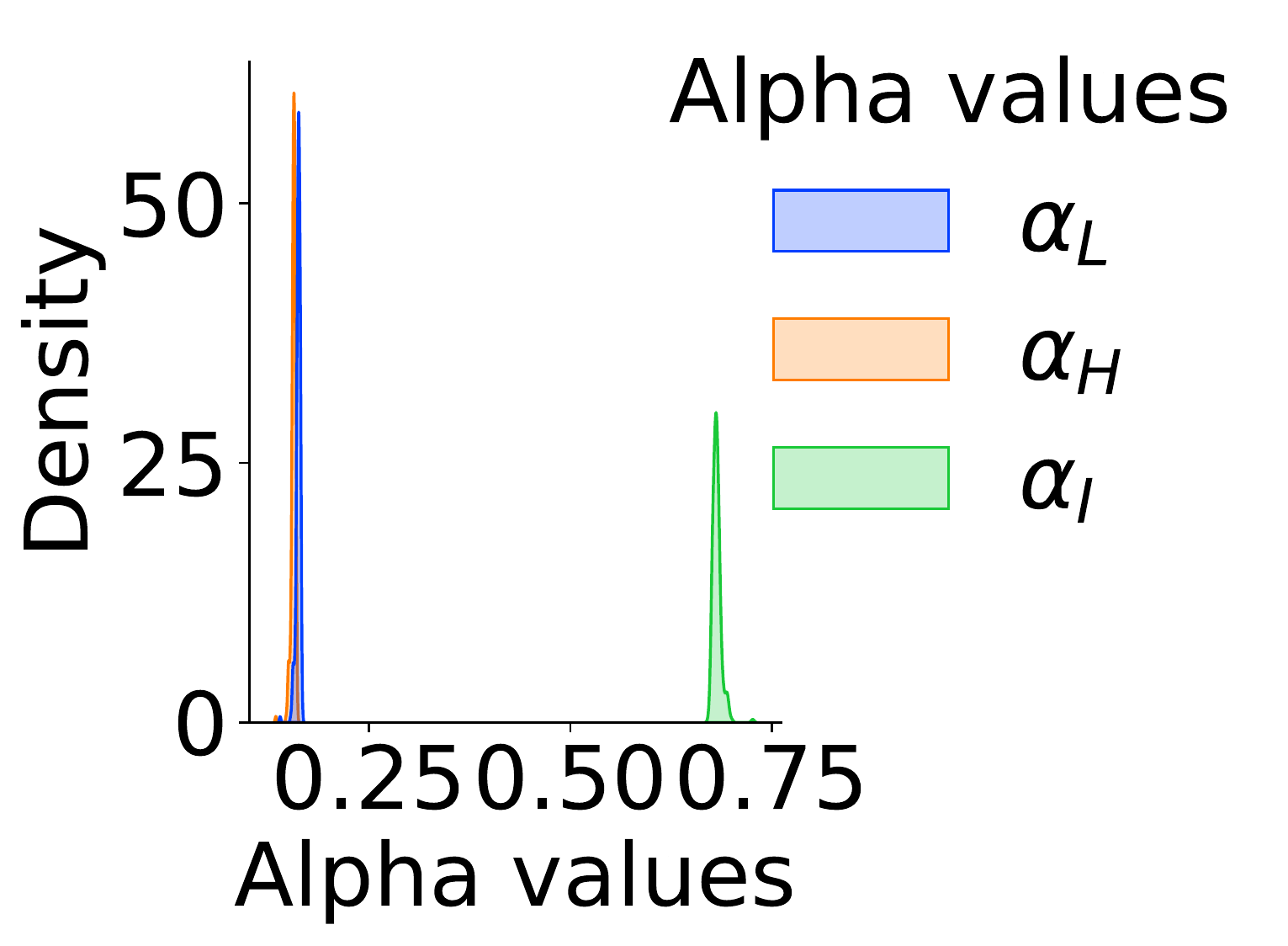}
     } 
     \subfloat[ \texttt{Texas}]{
     \captionsetup{justification = centering}
     \includegraphics[width=0.35\textwidth]{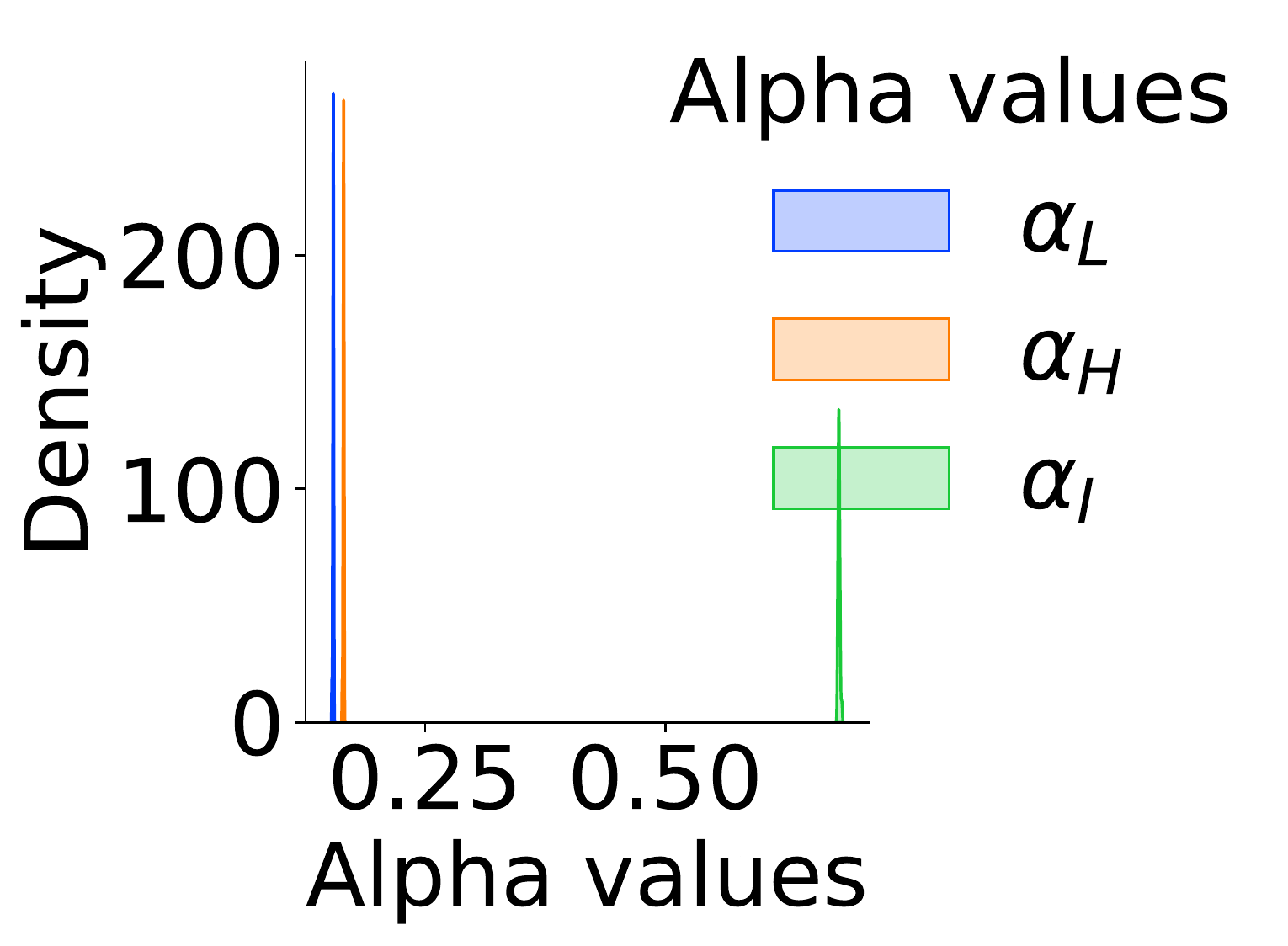}
     } \\
     \subfloat[ \texttt{Chameleon}]{
     \captionsetup{justification = centering}
     \includegraphics[width=0.35\textwidth]{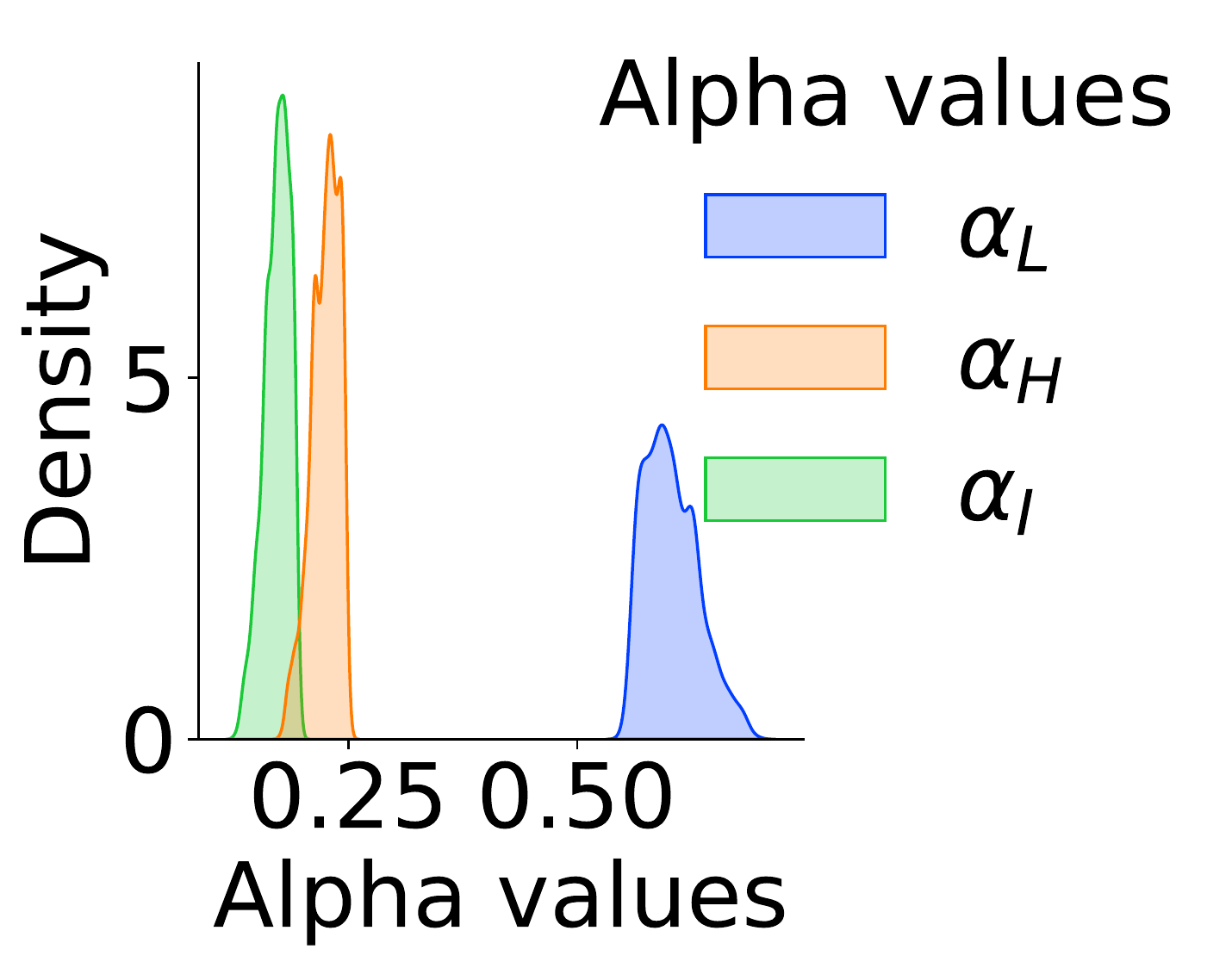}
     } 
     \subfloat[ \texttt{Squirrel}]{
     \captionsetup{justification = centering}
     \includegraphics[width=0.35\textwidth]{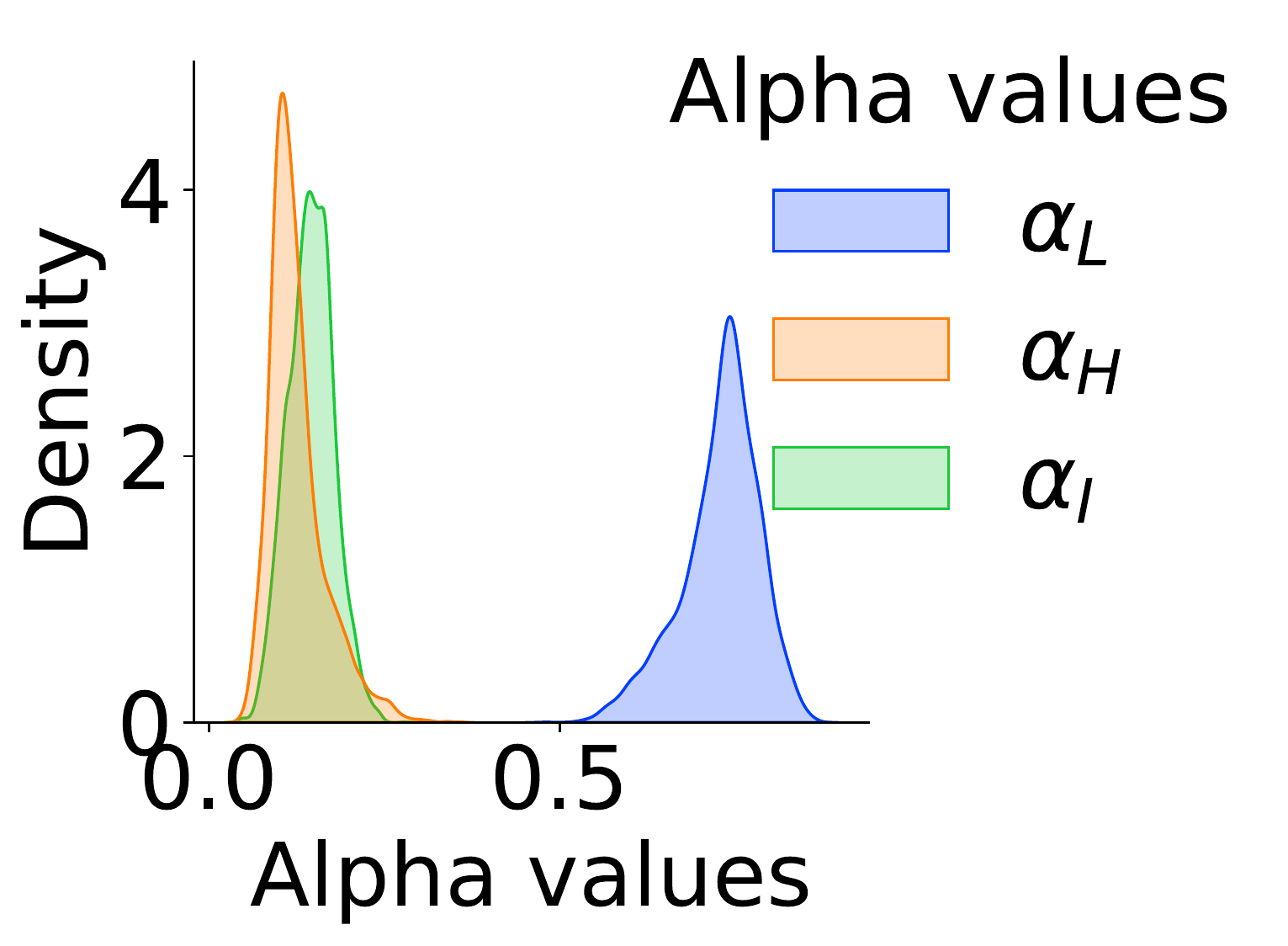}
     } 
     \subfloat[ \texttt{Film}]{
     \captionsetup{justification = centering}
     \includegraphics[width=0.35\textwidth]{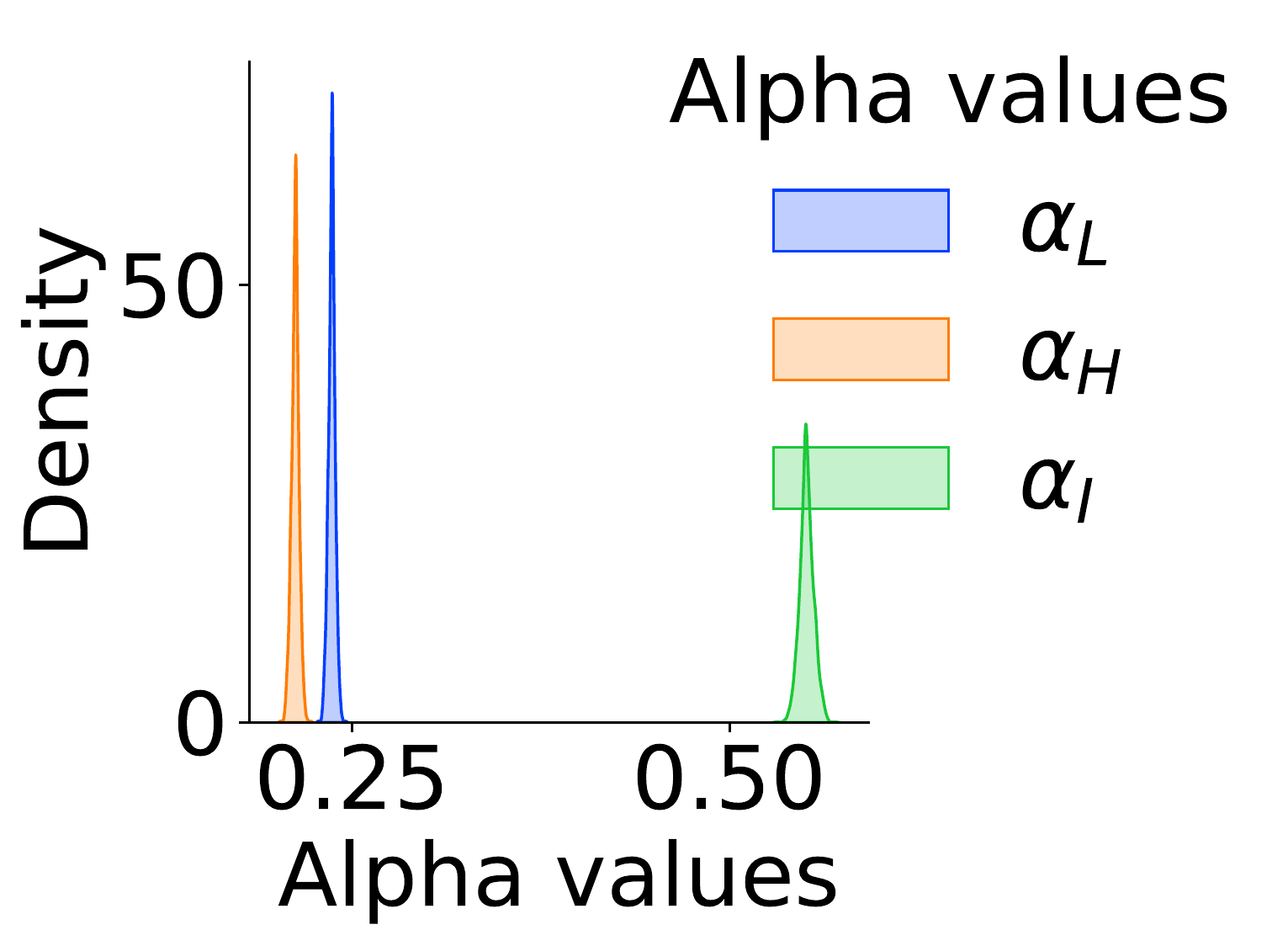}
     } \\
     \subfloat[ \texttt{Cora}]{
     \captionsetup{justification = centering}
     \includegraphics[width=0.35\textwidth]{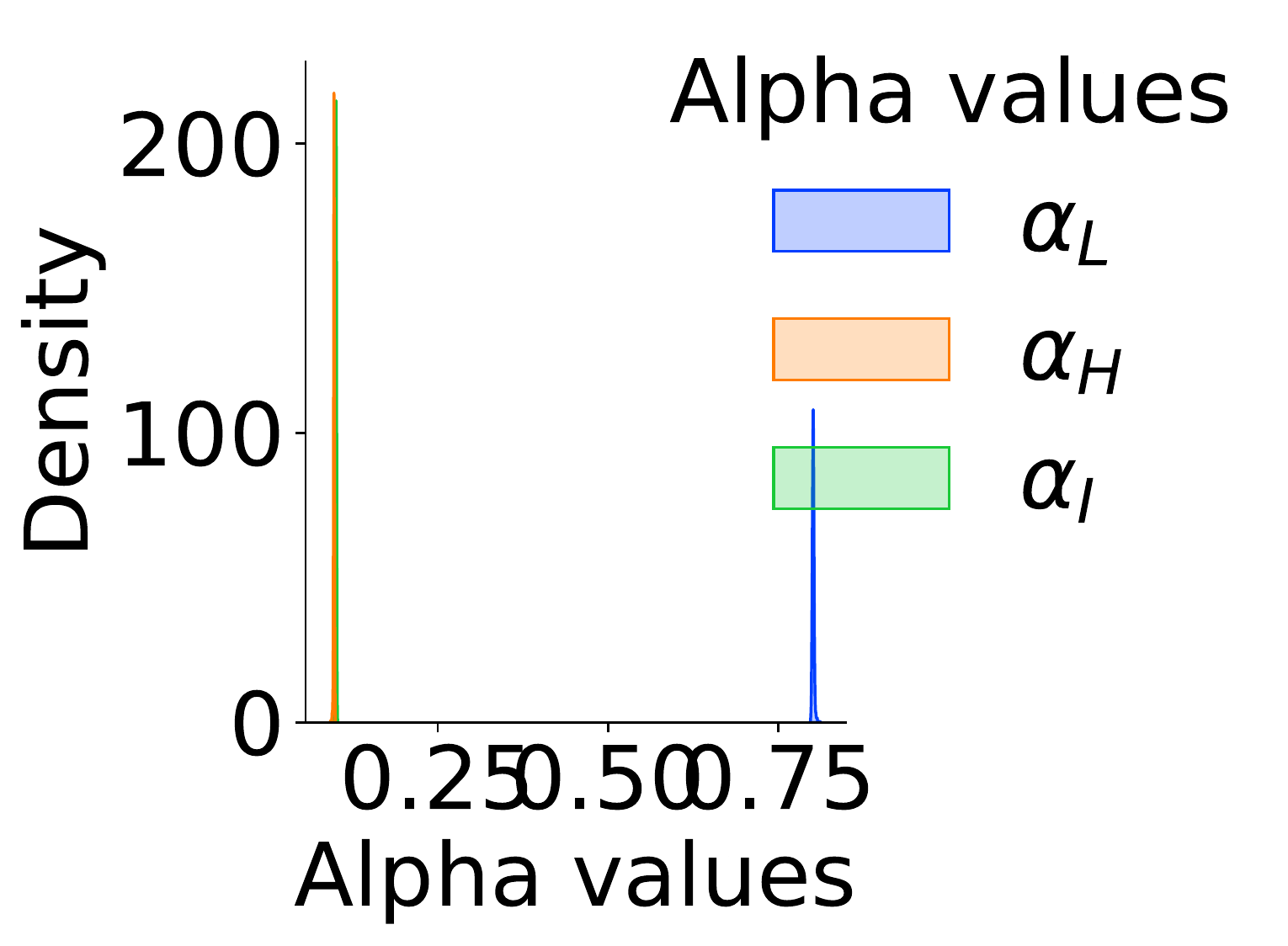}
     } 
     \subfloat[ \texttt{Citeseer}]{
     \captionsetup{justification = centering}
     \includegraphics[width=0.35\textwidth]{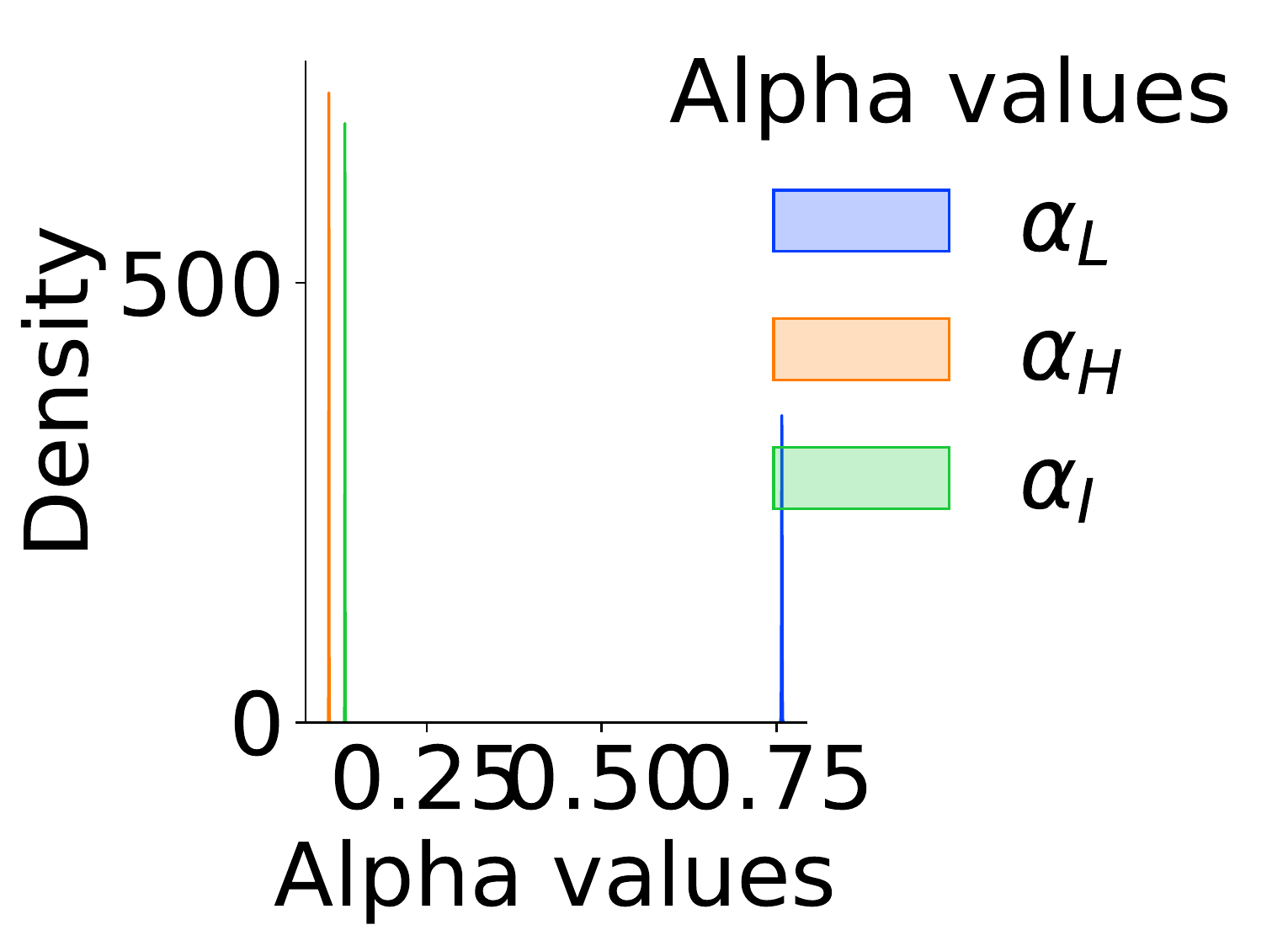}
     } 
     \subfloat[ \texttt{Pubmed}]{
     \captionsetup{justification = centering}
     \includegraphics[width=0.35\textwidth]{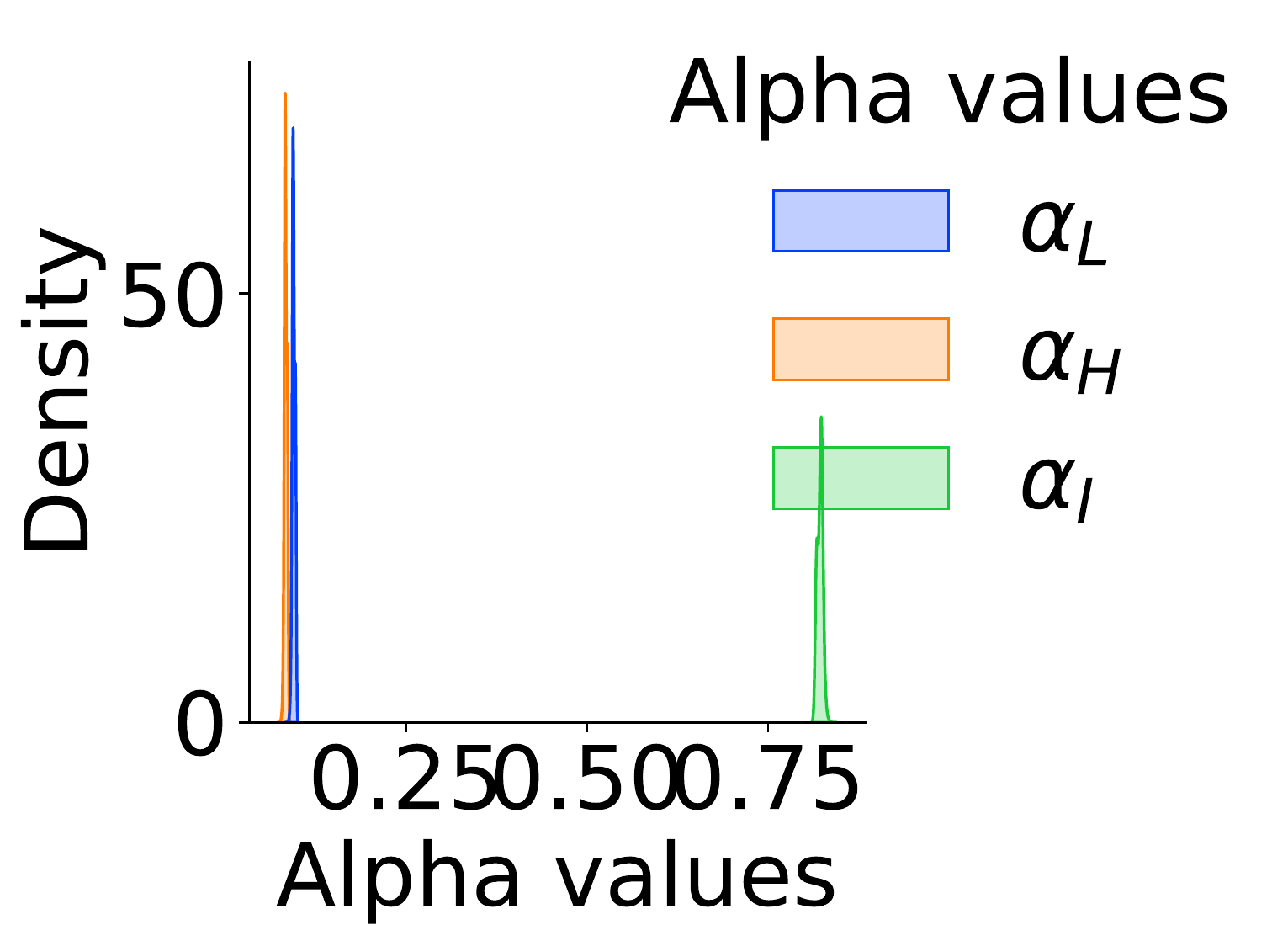}
     } 
     }
     \caption{Distributions of the learned $\alpha_L, \alpha_H, \alpha_I$  in the hidden layer of ACM-GCN}
     \label{fig:Alpha_values_hidden_layer}
\end{figure}

\begin{figure}[H]
    \centering
     {
    \subfloat[ \texttt{Cornell}]{
     \captionsetup{justification = centering}
     \includegraphics[width=0.35\textwidth]{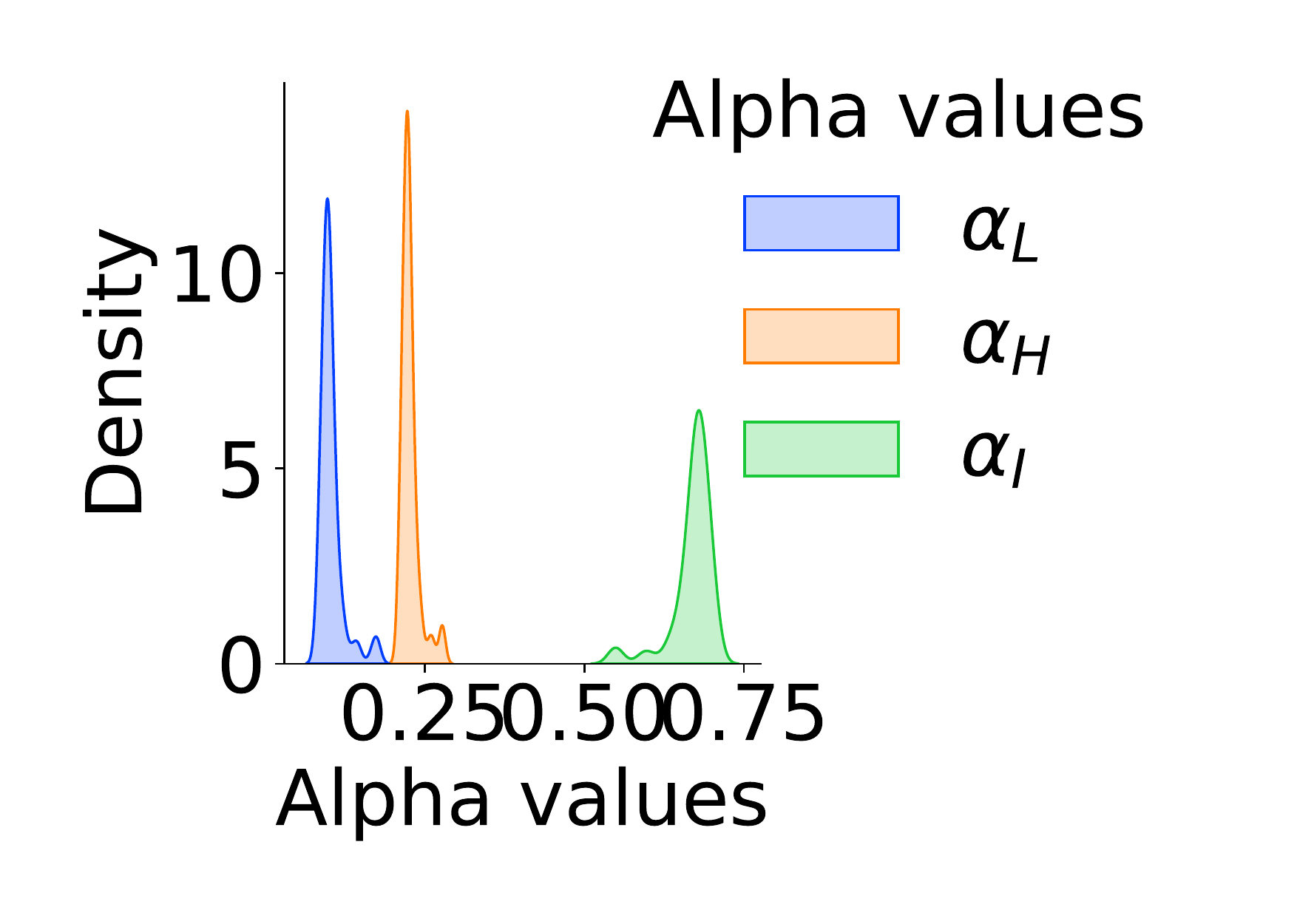}
     } 
     \subfloat[ \texttt{Wisconsin}]{
     \captionsetup{justification = centering}
     \includegraphics[width=0.35\textwidth]{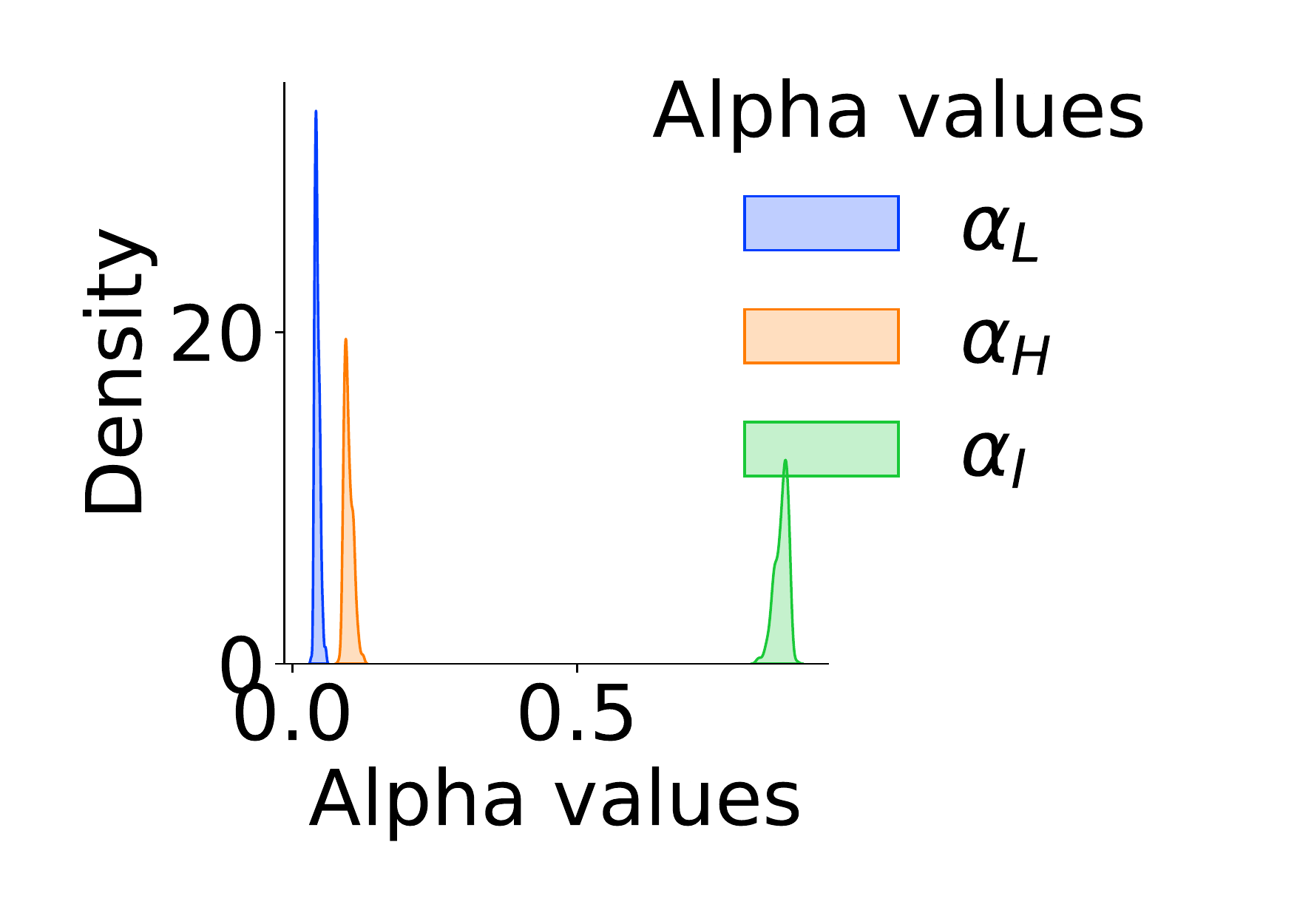}
     } 
     \subfloat[ \texttt{Texas}]{
     \captionsetup{justification = centering}
     \includegraphics[width=0.35\textwidth]{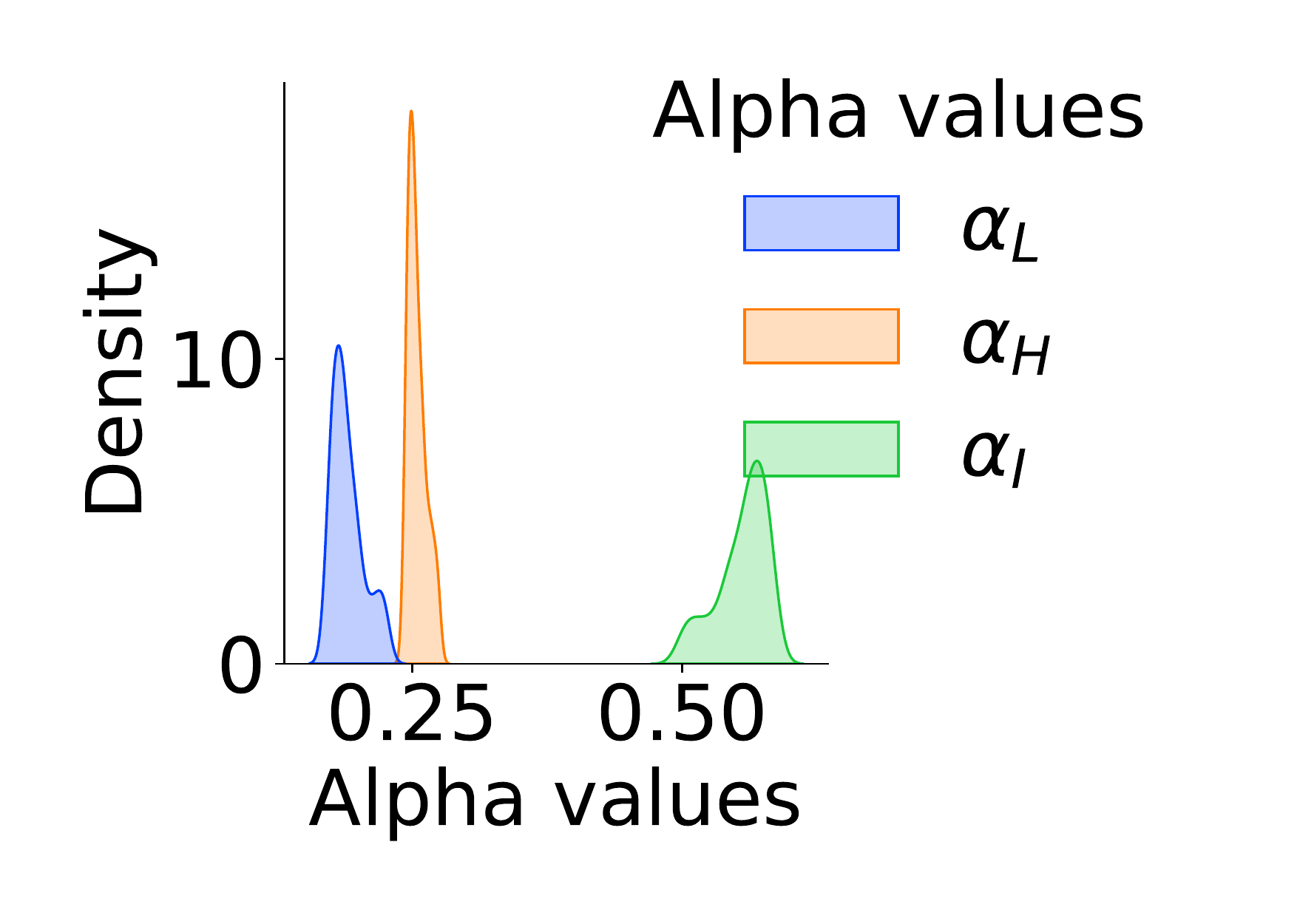}
     } \\
     \subfloat[ \texttt{Chameleon}]{
     \captionsetup{justification = centering}
     \includegraphics[width=0.35\textwidth]{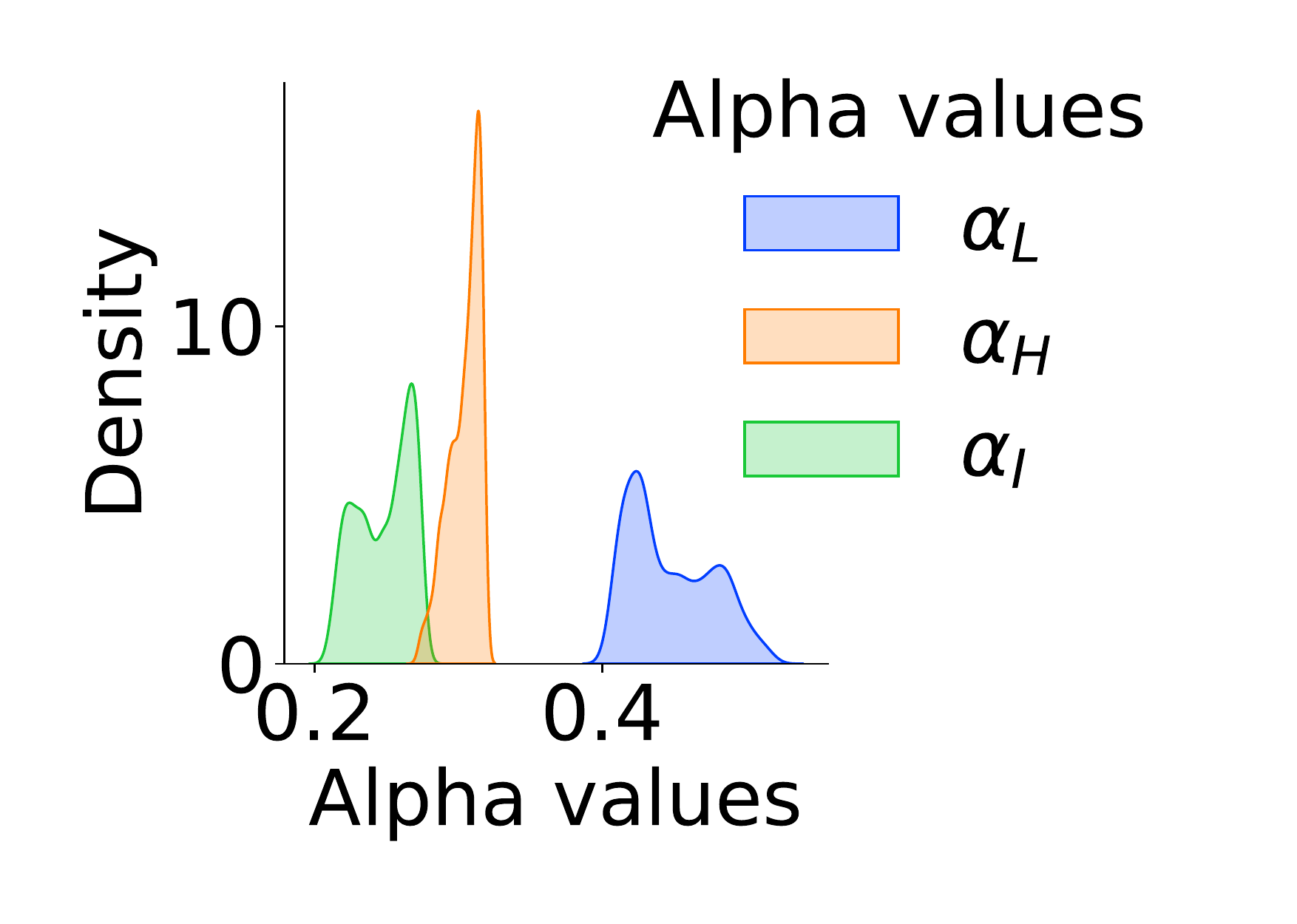}
     } 
     \subfloat[ \texttt{Squirrel}]{
     \captionsetup{justification = centering}
     \includegraphics[width=0.35\textwidth]{alpha_squirrel.pdf}
     } 
     \subfloat[ \texttt{Film}]{
     \captionsetup{justification = centering}
     \includegraphics[width=0.35\textwidth]{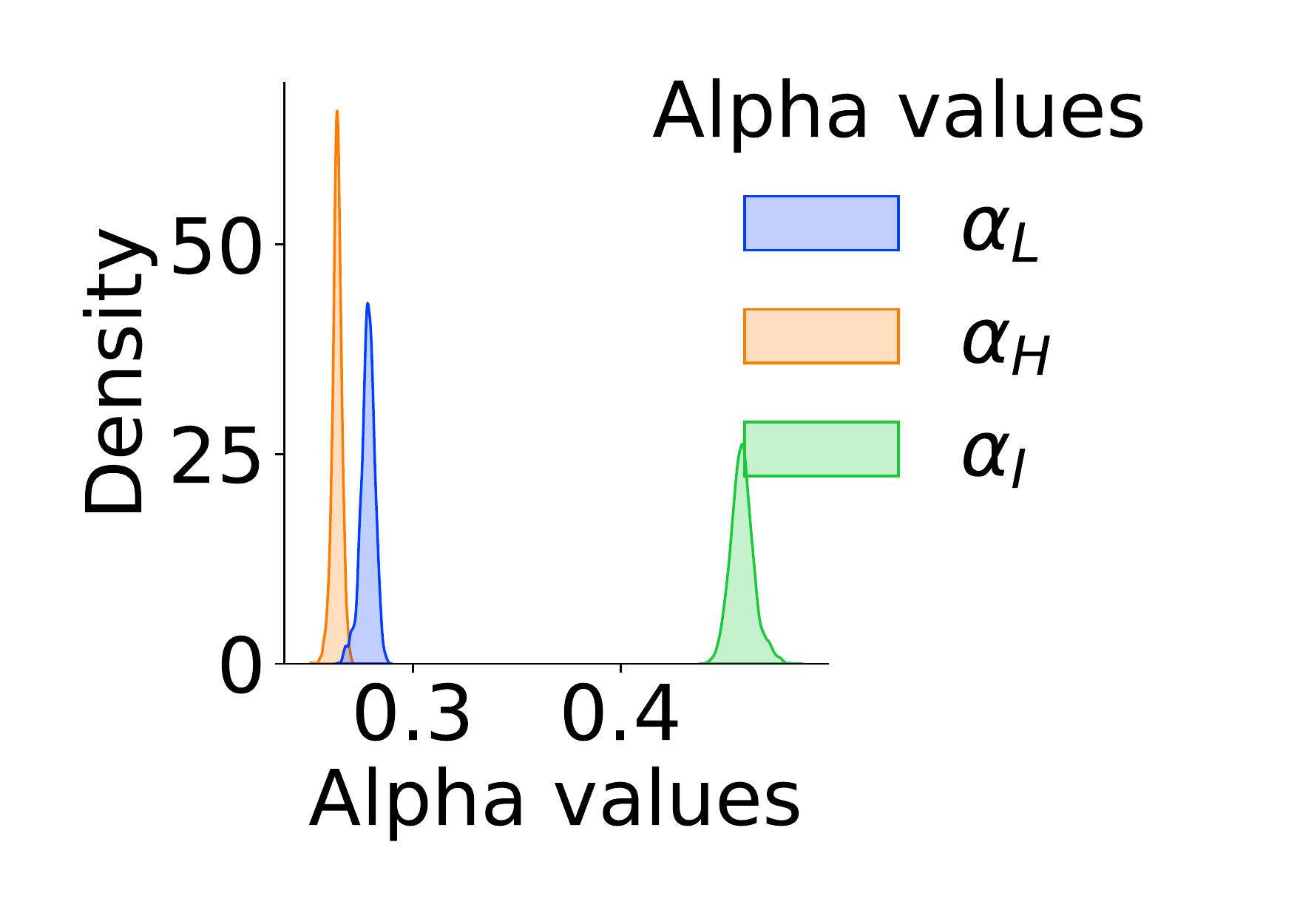}
     } \\
     \subfloat[ \texttt{Cora}]{
     \captionsetup{justification = centering}
     \includegraphics[width=0.35\textwidth]{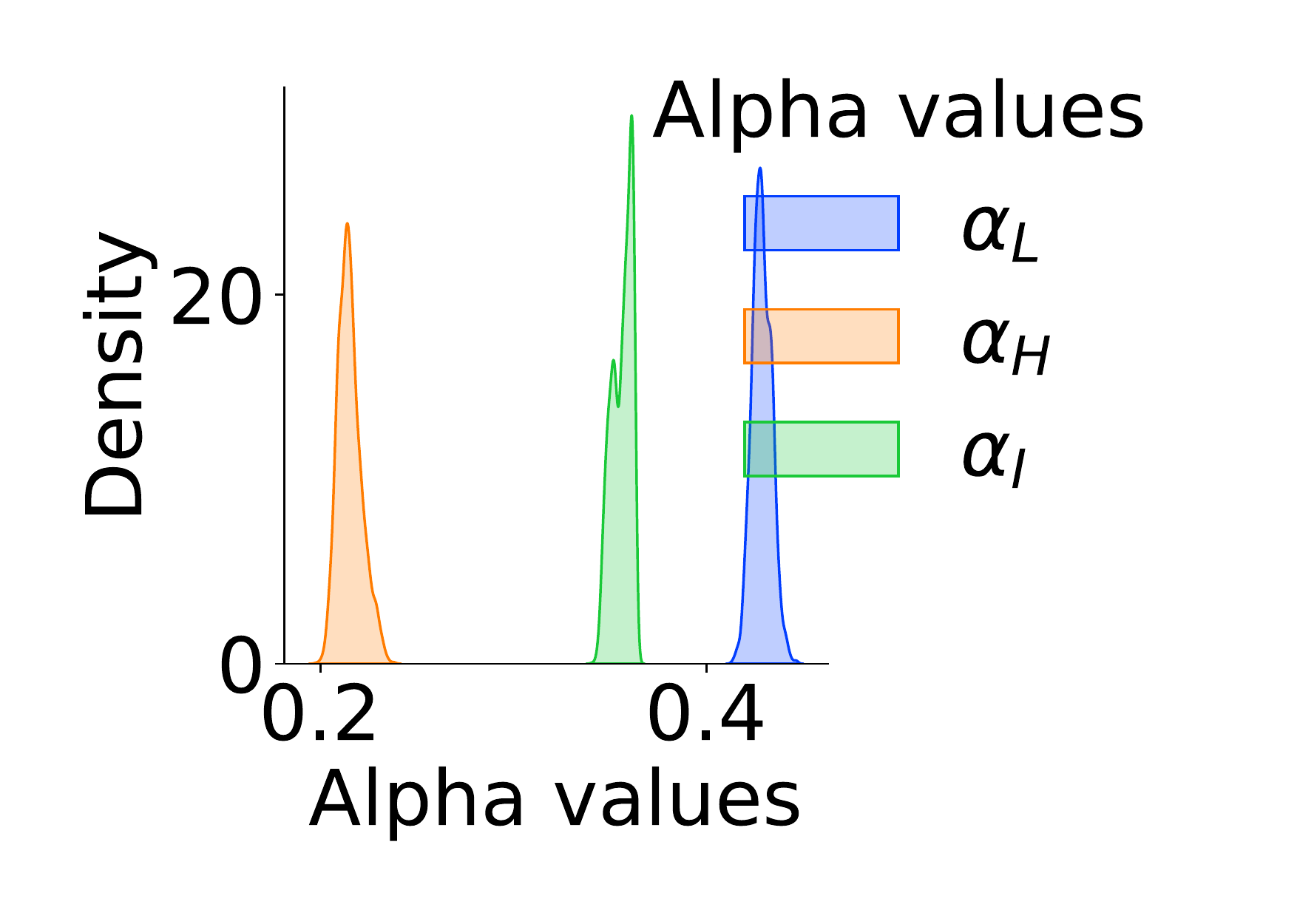}
     } 
     \subfloat[ \texttt{Citeseer}]{
     \captionsetup{justification = centering}
     \includegraphics[width=0.35\textwidth]{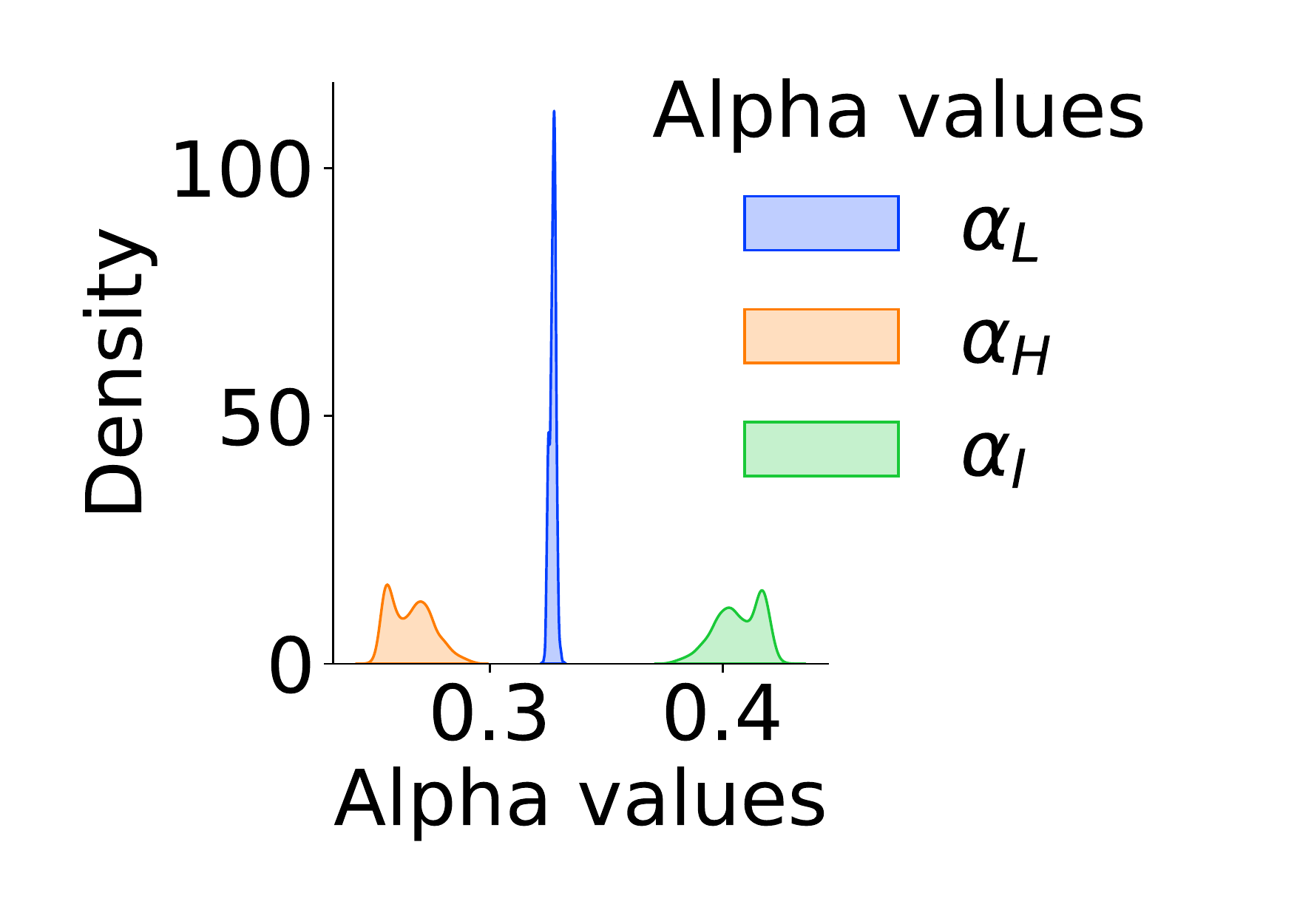}
     } 
     \subfloat[ \texttt{Pubmed}]{
     \captionsetup{justification = centering}
     \includegraphics[width=0.35\textwidth]{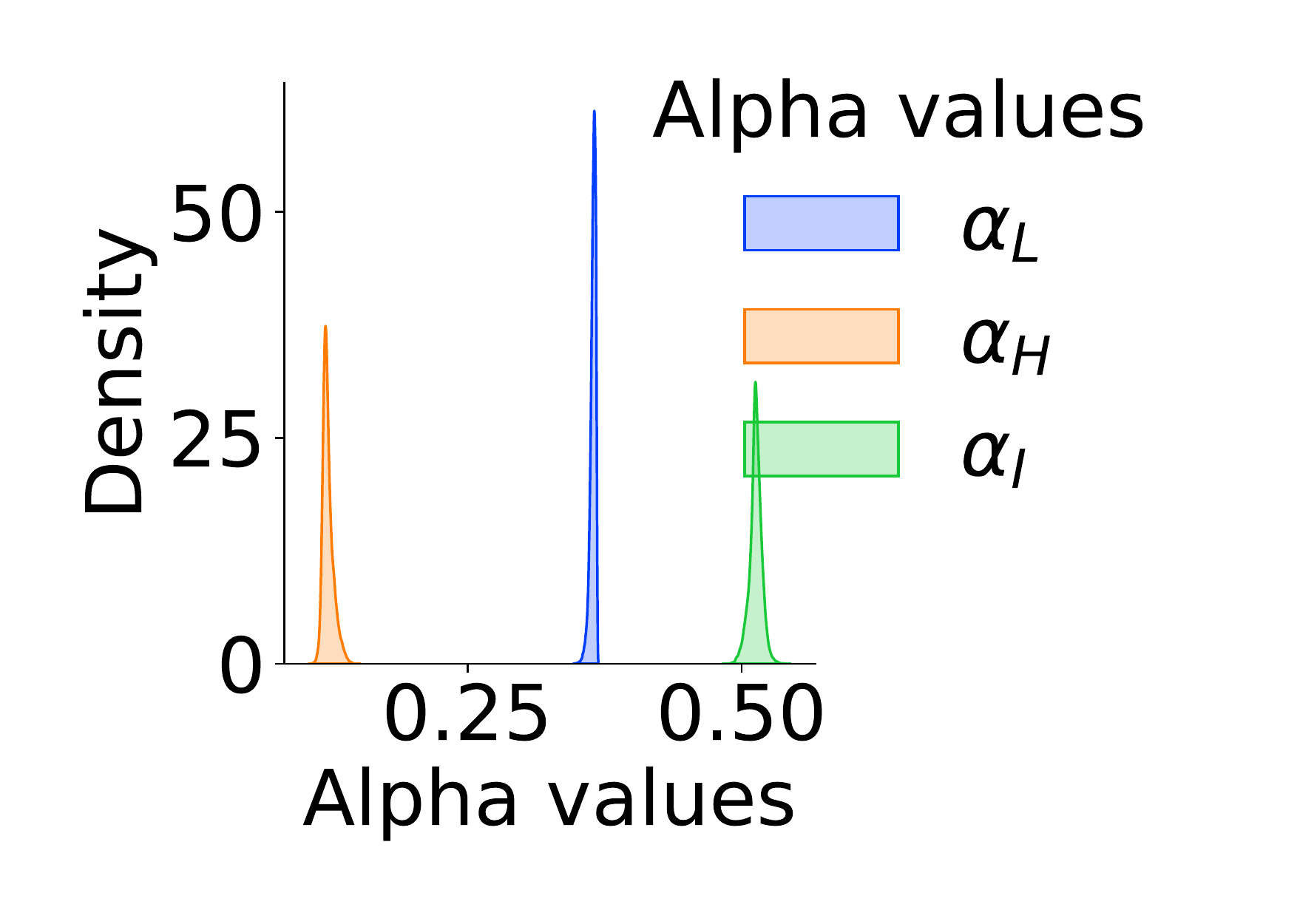}
     } 
     }
     \caption{Distributions of the learned $\alpha_L, \alpha_H, \alpha_I$  in the output layer of ACM-GCN}
     \label{fig:Alpha_values_output_layer}
\end{figure}

See Figure \ref{fig:Alpha_values_hidden_layer} for the distributions of weights in hidden layers and Figure \ref{fig:Alpha_values_output_layer} for the distributions of weights in output layers.

\label{appendix:statistical_experimental_results}

\section{Details of the Implementation}
\label{appendix:details_implementation_acm_acmII}
\subsection{Implementation of ACM-GCMII}
Unlike other baseline GNN models, GCNII and GCNII* are not able to be applied under ACMII framework and we will make an explanation as follows.

\begin{align*}
       & \text{GCNII: } \mathbf{H}^{(\ell+1)}=\sigma\left(\left(\left(1-\alpha_{\ell}\right) \hat{\mathbf{A}} \mathbf{H}^{(\ell)}+\alpha_{\ell} \mathbf{H}^{(0)}\right)\left(\left(1-\beta_{\ell}\right) \mathbf{I}_{n}+\beta_{\ell} \mathbf{W}^{(\ell)}\right)\right)\\
       & \text{GCNII*: }
       \mathbf{H}^{(\ell+1)}= \sigma\left(\left(1-\alpha_{\ell}\right) \hat{\mathbf{A}} \mathbf{H}^{(\ell)}\left(\left(1-\beta_{\ell}\right) \mathbf{I}_{n}+\beta_{\ell} \mathbf{W}_{1}^{(\ell)}\right)+\right. \left.+\alpha_{\ell} \mathbf{H}^{(0)}\left(\left(1-\beta_{\ell}\right) \mathbf{I}_{n}+\beta_{\ell} \mathbf{W}_{2}^{(\ell)}\right)\right)
\end{align*}
From the above formulas of GCNII and GCNII$^*$ we cam see that, without major modification, GCNII and GCNII* are hard to be put into ACMII framework. In ACMII framework, before apply $\hat{A}$, we first implement a nonlinear feature extractor $\sigma(H^{\ell} \mathbf{W}^{(\ell)})$. But in GCNII and GCNII*, before multiplying $W^\ell (\text{or }W_1^\ell, W_2^\ell)$ to extract features, we need to add another term including $H^{(0)}$, which are not filtered by $\hat{A}$. This makes the order of aggregator $\hat{A}$ and nonlinear extractor unexchangable and thus, incompatible with ACMII framework. So we did not implement GCNII and GCNII* in ACMII framework.

\subsection{Implementation of ACM(II)-GCN+ and ACM(II)-GCN++}
Besides the features extracted by different filters, some recent SOTA models use additional graph structure information explicitly, \ie{} $\text{MLP}_\theta(A)$ , to address heterophily problem, \eg{} LINKX \cite{lim2021large} and GloGNN \cite{li2022finding} and is found effective on some datasets, \eg{} $\textit{Chameleon, Squirrel}$. The explicit structure information can be directly incorporated into ACM and ACMII framework, and we have ACM(II)-GCN+ and ACM(II)-GCN++ as follows.
\begin{itemize}
\item ACM-GCN+ and ACMII-GCN+ have an option to include structure information channel (the 4-th channel) in each layer and their differences from ACM-GCN and ACMII-GCN are \red{highlighted in red}) as follows,
    
    \begin{equation*}
\begin{aligned}
\label{eq:acm_gnn+_spectral}
& \textbf{{Step 1. Feature Extraction for LP, HP, Identity and Structure Information Channel:}} \\
&\red{{H}^l_A = \text{ReLU}\left(A W^l_A\right), W^l_A \in \mathbb{R}^{N\times F_l}}, \text{ get ${H}^{l}_L,{H}^{l}_H,{H}^{l}_I$ with the same step as ACM-GCN and ACMII-GCN.}\\
& \textbf{Step 2. Row-wise Feature-based Weight Learning with Layer Normalization (LN)} \\
& \red{\tilde{H}^{l}_L = \text{LN} ({H}^{l}_L),\ \tilde{H}^{l}_H = \text{LN}({H}^{l}_H),\ \tilde{H}^{l}_I = \text{LN}({H}^{l}_I),\ \tilde{H}^l_A = \text{LN}({H}^l_A),} \\
&\tilde{\alpha}_L^l = \text{Sigmoid} \left(\tilde{H}^{l}_L \tilde{W}^{l}_L\right),\ \tilde{\alpha}_H^l = \text{Sigmoid} \left(\tilde{H}^{l}_H \tilde{W}^{l}_H\right), \tilde{\alpha}_I^l = \text{Sigmoid} \left(\tilde{H}^{l}_I \tilde{W}^{l}_I\right),
\tilde{\alpha}_A^l = \text{Sigmoid} \left(\tilde{H}^l_A \tilde{W}^{l}_A\right),
\\
&\tilde{W}_L^{l-1},\ \tilde{W}_H^{l-1},\ \tilde{W}_I^{l-1},\tilde{W}^{l}_A \in \mathbb{R}^{F_l \times 1}\\ 
&\textbf{Step 3. Node-wise Adaptive Channel Mixing:}&\\
& \text{Option 1: without structure information} \\
&\left[{\alpha}_L^l, {\alpha}_H^l, {\alpha}_I^l \right] = \text{Softmax}\left((\left[\tilde{\alpha}_L^l,\tilde{\alpha}_H^l,\tilde{\alpha}_I^l\right]/T) W_\text{Mix}^l \right) \in \mathbb{R}^{N\times 3}, T = 3   \text{ temperature},\ W_\text{Mix}^l \in \mathbb{R}^{3\times 3}; \\
&{H^{l}}  = \text{ReLU}\left( \text{diag}(\alpha_L^l){H}^{l}_L + \text{diag}(\alpha_H^l){H}^{l}_H + \text{diag}(\alpha_I^l){H}^{l}_I\right)\\
& \text{Option 2: with structure information}\\
&\red{\left[{\alpha}_L^l, {\alpha}_H^l, {\alpha}_I^l, {\alpha}_A^l \right] = \text{Softmax}\left((\left[\tilde{\alpha}_L^l,\tilde{\alpha}_H^l,\tilde{\alpha}_I^l,\tilde{\alpha}_A^l\right]/T) W_\text{Mix}^l \right) \in \mathbb{R}^{N\times 4}, T = 4 \text{ temperature},\ W_\text{Mix}^l \in \mathbb{R}^{4\times 4};} \\
& \red{{H^{l}}  =  \text{ReLU}\left(\text{diag}(\alpha_L^l){H}^{l}_L + \text{diag}(\alpha_H^l){H}^{l}_H + \text{diag}(\alpha_I^l){H}^{l}_I + \text{diag}(\alpha_A^l){H}^{l}_A \right) }\\
\end{aligned}
\end{equation*}

\item  ACM-GCN++ and ACMII-GCN++ have an option to include structure information channel (the 4-th channel) in each layer and residual connection and their differences from ACM-GCN+ and ACMII-GCN+ are \red{highlighted in red}) as follows,

\begin{equation*}
\begin{aligned}
\label{eq:acm_gnn++_spectral}
& \textbf{{Step 1. Feature Extraction for LP, HP, Identity and Structure Information Channel, Get ${H}_X$:}} \\
&\red{{H}_X = \text{ReLU}\left(X W_X \right) \in \mathbb{R}^{F\times F'}}, {H}^l_A = \text{ReLU}\left(A W^l_A\right), W^l_A \in \mathbb{R}^{N\times F'},\\
&\text{ get ${H}^{l}_L,{H}^{l}_H,{H}^{l}_I$ with the same step as ACM-GCN and ACMII-GCN.}\\
& \textbf{Step 2. Row-wise Feature-based Weight Learning with Layer Normalization (LN)} \\
& \tilde{H}^{l}_L = \text{LN} ({H}^{l}_L),\ \tilde{H}^{l}_H = \text{LN}({H}^{l}_H),\ \tilde{H}^{l}_I = \text{LN}({H}^{l}_I),\ \tilde{H}^l_A = \text{LN}({H}^l_A),\\
&\tilde{\alpha}_L^l = \text{Sigmoid} \left(\tilde{H}^{l}_L \tilde{W}^{l}_L\right),\ \tilde{\alpha}_H^l = \text{Sigmoid} \left(\tilde{H}^{l}_H \tilde{W}^{l}_H\right), \tilde{\alpha}_I^l = \text{Sigmoid} \left(\tilde{H}^{l}_I \tilde{W}^{l}_I\right),
\tilde{\alpha}_A^l = \text{Sigmoid} \left(\tilde{H}^l_A \tilde{W}^{l}_A\right),
\\
&\tilde{W}_L^{l-1},\ \tilde{W}_H^{l-1},\ \tilde{W}_I^{l-1},\tilde{W}^{l}_A \in \mathbb{R}^{F' \times 1}\\ 
&\textbf{Step 3. Node-wise Adaptive Channel Mixing:}&\\
& \text{Option 1: without structure information} \\
&\left[{\alpha}_L^l, {\alpha}_H^l, {\alpha}_I^l \right] = \text{Softmax}\left((\left[\tilde{\alpha}_L^l,\tilde{\alpha}_H^l,\tilde{\alpha}_I^l\right]/T) W_\text{Mix}^l \right) \in \mathbb{R}^{N\times 3}, T = 3   \text{ temperature},\ W_\text{Mix}^l \in \mathbb{R}^{3\times 3}; \\
&{H^{l}}  =  \text{ReLU}\left(\text{diag}(\alpha_L^l){H}^{l}_L + \text{diag}(\alpha_H^l){H}^{l}_H + \text{diag}(\alpha_I^l){H}^{l}_I \right) + \red{{H}_X}\\
& \text{Option 2: with structure information}\\
&\left[{\alpha}_L^l, {\alpha}_H^l, {\alpha}_I^l, {\alpha}_A^l \right] = \text{Softmax}\left((\left[\tilde{\alpha}_L^l,\tilde{\alpha}_H^l,\tilde{\alpha}_I^l,\tilde{\alpha}_A^l\right]/T) W_\text{Mix}^l \right) \in \mathbb{R}^{N\times 4}, T = 4 \text{ temperature},\ W_\text{Mix}^l \in \mathbb{R}^{4\times 4}; \\
&{H^{l}}  =  \text{ReLU}\left(\text{diag}(\alpha_L^l){H}^{l}_L + \text{diag}(\alpha_H^l){H}^{l}_H + \text{diag}(\alpha_I^l){H}^{l}_I + \text{diag}(\alpha_A^l){H}^{l}_A \right) + \red{{H}_X}\\
\end{aligned}
\end{equation*}

\end{itemize}

The results of ACM-GCN+, ACMII-GCN+, ACM-GCN++ and ACMII-GCN++ trained on random 60\%/20\%/20\% splits are reported in table \ref{tab:sota} in Appendix \ref{sec:full_result_comparison}. The results on fixed 48\%/32\%/20\% splits are reported in table \ref{tab:performance_comparison_fixed_splits} in Appendix \ref{appendix:results_on_fixed_splits_as_geomgcn}.

\paragraph{Computing Resources}
For all experiments on synthetic datasets and real-world datasets, we use NVIDIA V100 GPUs with 16/32GB GPU memory, 8-core CPU, 16G Memory. The
software implementation is based on PyTorch and PyTorch Geometric \cite{fey2019fast}.

\section{Hyperparameter Searching Range \& Optimal Hyperparameters}
\label{appendix:hyperparameter}

\subsection{Hyperparameter Searching Range for Synthetic Experiments}
\label{appendix:hyperparameter_space_synthetic_graphs}

\begin{table}[htbp]
  \centering
  \caption{Hyperparameter searching range for synthetic experiments}
    \begin{tabular}{c|cccc}
    \toprule
    \toprule
    \multicolumn{5}{c}{Hyperparameter Searching Range for Synthetic Experiments} \\
    \midrule
    Models\textbackslash{}Hyperparameters & lr    & weight\_decay & dropout & hidden \\
    \midrule
    MLP-1 & 0.05  & \{5e-5, 1e-4, 5e-4, 1e-3, 5e-3 \} & -     & - \\
    SGC-1 & 0.05  & \{5e-5, 1e-4, 5e-4, 1e-3, 5e-3\} & -     & - \\
    ACM-SGC-1 & 0.05  & \{5e-5, 1e-4, 5e-4, 1e-3, 5e-3\} &  \{ 0.1, 0.3, 0.5, 0.7, 0.9\} 
    & - \\
    \midrule
    MLP-2 & 0.05  & \{5e-5, 1e-4, 5e-4, 1e-3, 5e-3\} & \{ 0.1, 0.3, 0.5, 0.7, 0.9\} & 64 \\
    GCN   & 0.05  & \{5e-5, 1e-4, 5e-4, 1e-3, 5e-3\} & \{ 0.1, 0.3, 0.5, 0.7, 0.9\} & 64 \\
    ACM-GCN & 0.05  & \{5e-5, 1e-4, 5e-4, 1e-3, 5e-3\} & \{ 0.1, 0.3, 0.5, 0.7, 0.9\} & 64 \\
    \bottomrule
    \bottomrule
    \end{tabular}%
  \label{tab:synthetic_data_hyperparameter_searching_range}%
\end{table}%

\subsection{Hyperparameter Searching Range for Ablation Study}

\begin{table}[htbp]
  \centering
  \scalebox{0.75}{
  \caption{Hyperparameter searching range for ablation study}
  \setlength{\tabcolsep}{2pt}
    \begin{tabular}{c|cccc}
    \toprule
    \toprule
    \multicolumn{5}{c}{Hyperparameter Searching Range for Ablation Study} \\
    \midrule
    Models\textbackslash{}Hyperparameters & lr    & weight\_decay & dropout & hidden \\
    \midrule
    SGC-LP+HP & \{0.01, 0.05, 0.1\} & \{0, 5e-6, 1e-5, 5e-5, 1e-4, 5e-4, 1e-3, 5e-3, 1e-2\} & -
    & - \\
    SGC-LP+Identity & \{0.01, 0.05, 0.1\} & \{0, 5e-6, 1e-5, 5e-5, 1e-4, 5e-4, 1e-3, 5e-3, 1e-2\} & -
    & - \\
    ACM-SGC-no adaptive mixing & \{0.01, 0.05, 0.1\} & \{0, 5e-6, 1e-5, 5e-5, 1e-4, 5e-4, 1e-3, 5e-3, 1e-2\} & \{0, 0.1, 0.2, 0.3, 0.4, 0.5, 0.6, 0.7,0.8,0.9\} 
    & - \\
    GCN-LP+HP & \{0.01, 0.05, 0.1\} & \{0, 5e-6, 1e-5, 5e-5, 1e-4, 5e-4, 1e-3, 5e-3, 1e-2\} & \{0, 0.1, 0.2, 0.3, 0.4, 0.5, 0.6, 0.7,0.8,0.9\} & 64 \\
    GCN-LP+Identity & \{0.01, 0.05, 0.1\} & \{0, 5e-6, 1e-5, 5e-5, 1e-4, 5e-4, 1e-3, 5e-3, 1e-2\} & \{0, 0.1, 0.2, 0.3, 0.4, 0.5, 0.6, 0.7,0.8,0.9\} & 64 \\
    ACM-GCN-no adaptive mixing & \{0.01, 0.05, 0.1\} & \{0, 5e-6, 1e-5, 5e-5, 1e-4, 5e-4, 1e-3, 5e-3, 1e-2\} & \{0, 0.1, 0.2, 0.3, 0.4, 0.5, 0.6, 0.7,0.8,0.9\} & 64 \\
    \bottomrule
    \bottomrule
    \end{tabular}%
    }
  \label{tab:ablation_study_hyperparameters}%
\end{table}%

\subsection{Hyperparameter Searching Range for GNNs on Real-world Datasets}
\label{appendix:hyperparameter_searching_range_real_world_datasets}
See table \ref{tab:real_world_datasets_hyperparameter_searching_range} for the hyperparameter seaching range of baseline GNNs, ACM-GNNs, ACMII-GNNs and several SOTA models.

\begin{table}[htbp]
  \centering
  \tiny
  \caption{Hyperparameter searching range for training on real-world datasets}
    \begin{tabular}{p{10.875em}|cp{10.125em}p{3.625em}cp{5.065em}p{4.875em}p{3.065em}p{5.5em}p{4.25em}}
    \toprule
    \toprule
    Models\textbackslash{}Hyperparameters & {lr} & weight\_decay & dropout & {hidden} & lambda & alpha\_l & head  & layers & JK type \\
    \midrule
    H2GCN & 0.01  & {0.001} & \{0, 0.5\} & \multicolumn{1}{p{7.125em}}{\{8, 16, 32, 64\}} & -     & -  & -  & \{1, 2\} & - \\
    \midrule
    MixHop & 0.01  & {0.001} & {0.5} & \multicolumn{1}{p{7.125em}}{\{8, 16, 32\}} & -   & -   & -   & \{2, 3\} & - \\
    \midrule
    GCN+JK & \multicolumn{1}{p{4.315em}}{\{0.1, 0.01, 0.001\}} & {0.001} & {0.5} & \multicolumn{1}{p{7.125em}}{\{4, 8, 16, 32, 64\}} & -     & -     & -     & {2} & \multicolumn{1}{p{7.125em}}{\{max, cat\}} \\
    \midrule
     GAT+JK & \multicolumn{1}{p{4.315em}}{\{0.1, 0.01, 0.001\}} & {0.001} & {0.5} & \multicolumn{1}{p{7.125em}}{\{4, 8, 12, 32\}} & -     & -     & \{2,4,8\} & {2} & \multicolumn{1}{p{7.125em}}{\{max, cat\}} \\
    \midrule
    GCNII, GCNII* & 0.01  & \{0, 5e-6, 1e-5, 5e-5, 1e-4, 5e-4, 1e-3\} for Deezer-Europe and \{0, 5e-6, 1e-5, 5e-5, 1e-4, 5e-4, 1e-3, 5e-3, 1e-2\} for others & {0.5} & 64    & \{0.5, 1, 1.5\} & \{0.1,0.2,0.3,0,4,0.5\} & -     & \{4, 8, 16, 32\} for Deezer-Europe and \{4, 8, 16, 32, 64\} for others & - \\
    \midrule
    Baselines: \{SGC-1, SGC-2, GCN, Snowball-2, Snowball-3, FAGCN\}; ACM-\{SGC-1, SGC-2, GCN, GCN+, GCN++, Snowball-2, Snowball-3\}; ACMII-\{SGC-1, SGC-2, GCN, GCN+, GCN++, Snowball-2, Snowball-3\} & \multicolumn{1}{p{4.315em}}{\{0.002, 0.01, 0.05\} for Deezer-Europe and \{0.01, 0.05, 0.1\} for others} & \{0, 5e-6, 1e-5, 5e-5, 1e-4, 5e-4, 1e-3\} for Deezer-Europe and \{0, 5e-6, 1e-5, 5e-5, 1e-4, 5e-4, 1e-3, 5e-3, 1e-2\} for others & \{0, 0.1, 0.2, 0.3, 0.4, 0.5, 0.6, 0.7, 0.8, 0.9\} & 64    & -     & -     & -     & -     & - \\
    \midrule
    GraphSAGE & \multicolumn{1}{p{4.315em}}{\{0.01,0.05, 0.1\}} & \{0, 5e-6, 1e-5, 5e-5, 1e-4, 5e-4, 1e-3\} for Deezer-Europe and \{0, 5e-6, 1e-5, 5e-5, 1e-4, 5e-4, 1e-3, 5e-3, 1e-2\} for others & \{ 0, 0.1, 0.2, 0.3, 0.4, 0.5, 0.6, 0.7, 0.8, 0.9\} & \multicolumn{1}{p{7.125em}}{8 for Deezer-Europe and 64 for others} & -     & -     & -     & -     & - \\
    \midrule
    ACM-\{GCNII, GCNII*\} & 0.01  & \{0, 5e-6, 1e-5, 5e-5, 1e-4, 5e-4, 1e-3\} for Deezer-Europe and \{0, 5e-6, 1e-5, 5e-5, 1e-4, 5e-4, 1e-3, 5e-3, 1e-2\} for others & \{ 0, 0.1, 0.2, 0.3, 0.4, 0.5, 0.6, 0.7, 0.8, 0.9\} & 64    & -     & -     & -     & \{1,2,3,4\} & - \\
    \bottomrule
    \bottomrule
    \end{tabular}%
  \label{tab:real_world_datasets_hyperparameter_searching_range}%
\end{table}%

\subsection{Searched Optimal Hyperparameters for Baselines and ACM(II)-GNNs on Real-world Tasks}
\label{appendix:optimal_hyperparameter}
See the reported optimal hyperparameters on random 60\%/20\%/20\% splits for baseline GNNs in table \ref{tab:hyperparameters_baselines}, for ACM-GNNs and ACMII-GNNs in table \ref{tab:hyperparameters_acmgnns} and for ACM(II)-GCN+ and ACM(II)-GCN++ in table \ref{tab:optimal_hyperparameter_random_splits_acmgcn+_acmgcn++}.
\begin{table}[htbp]
  \centering
   \scalebox{.49}{
  \caption{Optimal hyperparameters for baseline models on random 60\%/20\%/20\% splits}
    \begin{tabular}{c|c|cccccccccccc}
    \toprule
    \toprule
    \multicolumn{14}{c}{\textbf{Hyperparameters for Baseline GNNs}} \\
    \midrule
    Datasets & Models\textbackslash{}Hyperparameters & lr    & weight\_decay & dropout & hidden & {\# layers} & {Gat heads} & {JK Type} & lambda & alpha\_l & results & std   & average epoch time/average total time \\
    \midrule
    \multirow{12}[1]{*}{\textbf{Cornell}} & SGC-1 & 0.05  & 1.00E-02 & 0     & 64    & -     & -     & -     & -     & -     & 70.98 & 8.39  & 2.53ms/0.51s \\
          & SGC-2 & 0.05  & 1.00E-03 & 0     & 64    & -     & -     & -     & -     & -     & 72.62 & 9.92  & 2.46ms/0.53s \\
          & GCN   & 0.1   & 5.00E-03 & 0.5   & 64    & 2     & -     & -     & -     & -     & 82.46 & 3.11  & 3.67ms/0.74s \\
          & Snowball-2 & 0.01  & 5.00E-03 & 0.4   & 64    & 2     & -     & -     & -     & -     & 82.62 & 2.34  & 4.24ms/0.87s \\
          & Snowball-3 & 0.01  & 5.00E-03 & 0.4   & 64    & 3     & -     & -     & -     & -     & 82.95 & 2.1   & 6.66ms/1.36s \\
          & GCNII & 0.01  & 1.00E-03 & 0.5   & 64    & 16    & -     & -     & 0.5   & 0.5   & 89.18 & 3.96  & 25.41ms/8.11s \\
          & GCNII* & 0.01  & 1.00E-03 & 0.5   & 64    & 8     & -     & -     & 0.5   & 0.5   & 90.49 & 4.45  & 15.35ms/4.05s \\
          & FAGCN & 0.01  & 1.00E-04 & 0.7   & 32    & 2     & -     & -     & -     & -     & 88.03 & 5.6   & 8.1ms/3.8858s \\
          &  Mixhop & 0.01  & 0.001 & 0.5   & 16    & 2     & - & - & -     & -     & 60.33 & 28.53 & {10.379ms/2.105s} \\
          & H2GCN & 0.01  & 0.001 & 0.5   & 64    & 1     & - & - & -     & -     & 86.23 & 4.71  & {4.381ms/1.123s} \\
          & GCN+JK & 0.1   & 0.001 & 0.5   & 64    & 2     & - & {cat} & -     & -     & 66.56 & 13.82 & {5.589ms/1.227s} \\
          &  GAT+JK & 0.1   & 0.001 & 0.5   & 32    & 2     & 8     & {max} & -     & -     & 74.43 & 10.24 & {10.725ms/2.478s} \\
    \midrule
    \multirow{14}[1]{*}{\textbf{Wisconsin}} & SGC-1 & 0.05  & 5.00E-03 & 0     & 64    & -     & -     & -     & -     & -     & 70.38 & 2.85  & 2.83ms/0.57s \\
          & SGC-2 & 0.1   & 1.00E-03 & 0     & 64    & -     & -     & -     & -     & -     & 74.75 & 2.89  & { 2.14ms/0.43s} \\
          & GCN   & 0.1   & 1.00E-03 & 0.7   & 64    & 2     & -     & -     & -     & -     & 75.5  & 2.92  & {3.74ms/0.76s} \\
          & Snowball-2 & 0.1   & 1.00E-03 & 0.5   & 64    & 2     & -     & -     & -     & -     & 74.88 & 3.42  & 3.73ms/0.76s \\
          & Snowball-3 & 0.05  & 5.00E-04 & 0.8   & 64    & 3     & -     & -     & -     & -     & 69.5  & 5.01  & {5.46ms/1.12s} \\
          & GCNII & 0.01  & 1.00E-03 & 0.5   & 64    & 8     & -     & -     & 0.5   & 0.5   & 83.25 & 2.69  &  \\
          & GCNII* & 0.01  & 1.00E-03 & 0.5   & 64    & 4     & -     & -     & 1.5   & 0.3   & 89.12 & 3.06  & 9.26ms/1.96s \\
          & FAGCN & 0.05  & 1.00E-04 & 0     & 32    & 2     & -     & -     & -     & -     & 89.75 & 6.37  & 12.9ms/4.6359s \\
          &  Mixhop & 0.01  & 0.001 & 0.5   & 16    & 2     & - & - & -     & -     & 77.25 & 7.80  & {10.281ms/2.095s} \\
          & H2GCN & 0.01  & 0.001 & 0.5   & 32    & 1     & - & - & -     & -     & 87.5  & 1.77  & {4.324ms/1.134s} \\
          & GCN+JK & 0.1   & 0.001 & 0.5   & 32    & 2     & - & {cat} & -     & -     & 62.5  & 15.75 & {5.117ms/1.049s} \\
          &  GAT+JK & 0.1   & 0.001 & 0.5   & 4     & 2     & 8     & {max} & -     & -     & 69.5  & 3.12  & {10.762ms/2.25s} \\
          & APPNP & 0.05  & 0.001 & 0.5   & 64    & 2     & - & - & -     & -     & 92    & 3.59  & {10.303ms/2.104s} \\
          & GPRGNN & 0.05  & 0.001 & 0.5   & 256   & 2     & - & - & -     & -     & 93.75 & 2.37  & {11.856ms/2.415s} \\
          \midrule
    \multirow{12}[0]{*}{\textbf{Texas}} & SGC-1 & 0.05  & 1.00E-03 & 0     & 64    & -     & -     & -     & -     & -     & 83.28 & 5.43  &  2.55ms/0.54s \\
          & SGC-2 & 0.01  & 1.00E-03 & 0     & 64    & -     & -     & -     & -     & -     & 81.31 & 3.3   & 2.61ms/2.53s \\
          & GCN   & 0.05  & 1.00E-02 & 0.9   & 64    & 2     & -     & -     & -     & -     & 83.11 & 3.2   & 3.59ms/0.73s \\
          & Snowball-2 & 0.05  & 1.00E-02 & 0.9   & 64    & 2     & -     & -     & -     & -     & 83.11 & 3.2   & 3.98ms/0.82s \\
          & Snowball-3 & 0.05  & 1.00E-02 & 0.9   & 64    & 3     & -     & -     & -     & -     & 83.11 & 3.2   & 5.56ms/1.12s \\
          & GCNII & 0.01  & 1.00E-04 & 0.5   & 64    & 4     & -     & -     & 1.5   & 0.5   & 82.46 & 4.58  &  \\
          & GCNII* & 0.01  & 1.00E-04 & 0.5   & 64    & 8     & -     & -     & 0.5   & 0.5   & 88.52 & 3.02  & 15.64ms/3.47s \\
          & FAGCN & 0.01  & 5.00E-04 & 0     & 32    & 2     & -     & -     & -     & -     & 88.85 & 4.39  & 8.8ms/6.5252s \\
          &  Mixhop & 0.01  & 0.001 & 0.5   & 32    & 2     & - & - & - & - & 76.39 & 7.66  & {11.099ms/2.329s} \\
          & H2GCN & 0.01  & 0.001 & 0.5   & 64    & 1     & - & - & - & - & 85.90 & 3.53  & {4.197ms/0.95s} \\
          & GCN+JK & 0.1   & 0.001 & 0.5   & 32    & 2     & - & {cat} & - & - & 80.66 & 1.91  & {5.28ms/1.085s} \\
          &  GAT+JK & 0.1   & 0.001 & 0.5   & 8     & 2     & 2     & {cat} & - & - & 75.41 & 7.18  & {10.937ms/2.402s} \\
          \midrule
    \multirow{12}[0]{*}{\textbf{Film}} & SGC-1 & 0.01  & 5.00E-06 & 0     & 64    & -     & -     & -     & -     & -     & 25.26 & 1.18  & 3.18ms/0.70s \\
          & SGC-2 & 0.01  & 5.00E-06 & 0     & 64    & -     & -     & -     & -     & -     & 28.81 & 1.11  &  2.13ms/0.43s \\
          & GCN   & 0.1   & 5.00E-04 & 0     & 64    & 2     & -     & -     & -     & -     & 35.51 & 0.99  & 4.86ms/0.99s \\
          & Snowball-2 & 0.1   & 5.00E-04 & 0     & 64    & 2     & -     & -     & -     & -     & 35.97 & 0.66  & 5.59ms/1.14s \\
          & Snowball-3 & 0.1   & 5.00E-04 & 0.2   & 64    & 3     & -     & -     & -     & -     & 36    & 1.36  & 7.89ms/1.60s \\
          & GCNII & 0.01  & 1.00E-04 & 0.5   & 64    & 8     & -     & -     & 1.5   & 0.3   & 40.82 & 1.79  & 15.85ms/3.22s \\
          & GCNII* & 0.01  & 1.00E-06 & 0.5   & 64    & 4     & -     & -     & 1     & 0.1   & 41.54 & 0.99  &  \\
          & FAGCN & 0.01  & 5.00E-05 & 0.6   & 32    & 2     & -     & -     & -     & -     & 31.59 & 1.37  & 45.4ms/11.107s \\
          &  Mixhop & 0.01  & 0.001 & 0.5   & 8     & 3     & 8     & {max} & - & - & 33.13 & 2.40  & {17.651ms/3.566s} \\
          & H2GCN & 0.01  & 0.001 & 0     & 64    & 1     & 8     & {max} & - & - & 38.85 & 1.17  & {8.101ms/1.695s} \\
          & GCN+JK & 0.1   & 0.001 & 0.5   & 64    & 2     & 8     & {cat} & - & - & 32.72 & 2.62  & {8.946ms/1.807s} \\
          &  GAT+JK & 0.001 & 0.001 & 0.5   & 32    & 2     & 4     & {cat} & - & - & 35.41 & 0.97  & {20.726ms/4.187s} \\
          \midrule
    \multirow{12}[0]{*}{\textbf{Chameleon}} & SGC-1 & 0.1   & 5.00E-06 & 0     & 64    & -     & -     & -     & -     & -     & 64.86 & 1.81  & 3.48ms/2.96s \\
          & SGC-2 & 0.1   & 0.00E+00 & 0     & 64    & -     & -     & -     & -     & -     & 62.67 & 2.41  &  4.43ms/1.12s \\
          & GCN   & 0.01  & 1.00E-05 & 0.9   & 64    & 2     & -     & -     & -     & -     & 64.18 & 2.62  & 4.96ms/1.18s \\
          & Snowball-2 & 1.00E-01 & 1.00E-05 & 0.9   & 64    & 2     & -     & -     & -     & -     & 64.99 & 2.39  & 4.96ms/1.00s \\
          & Snowball-3 & 0.1   & 5.00E-06 & 0.9   & 64    & 3     & -     & -     & -     & -     & 65.49 & 1.64  & 7.44ms/1.50s \\
          & GCNII & 0.01  & 5.00E-06 & 0.5   & 64    & 4     & -     & -     & 0.5   & 0.1   & 60.35 & 2.7   & 9.76ms/2.26s \\
          & GCNII* & 0.01  & 5.00E-04 & 0.5   & 64    & 4     & -     & -     & 1.5   & 0.5   & 62.8  & 2.87  & 10.40ms/2.17s \\
          & FAGCN & 0.002 & 1.00E-04 & 0     & 32    & 2     & -     & -     & -     & -     & 49.47 & 2.84  & 8.4ms/13.8696s \\
          &  Mixhop & 0.01  & 0.001 & 0.5   & 16    & 2     & 8     & {max} & - & - & 36.28 & 10.2  & {11.372ms/2.297s} \\
          & H2GCN & 0.01  & 0.001 & 0     & 32    & 1     & 8     & {max} & - & - & 52.3  & 0.48  & {4.059ms/0.82s} \\
          & GCN+JK & 0.001 & 0.001 & 0.5   & 32    & 2     & 8     & {cat} & - & - & 64.68 & 2.85  & {5.211ms/1.053s} \\
          &  GAT+JK & 0.001 & 0.001 & 0.5   & 4     & 2     & 8     & {max} & - & - & 68.14 & 1.18  & {13.772ms/2.788s} \\
          \midrule
    \multirow{12}[0]{*}{\textbf{Squirrel}} & SGC-1 & 0.05  & 0.00E+00 & 0     & 64    & -     & -     & -     & -     & -     & 47.62 & 1.27  & 4.65ms/1.44s \\
          & SGC-2 & 0.1   & 0.00E+00 & 0.9   & 64    & -     & -     & -     & -     & -     & 41.25 & 1.4   & 35.06ms/7.81s \\
          & GCN   & 0.01  & 5.00E-05 & 0.7   & 64    & 2     & -     & -     & -     & -     & 44.76 & 1.39  & 8.41ms/2.50s \\
          & Snowball-2 & 0.1   & 0.00E+00 & 0.9   & 64    & 2     & -     & -     & -     & -     & 47.88 & 1.23  & 8.96ms/1.92s \\
          & Snowball-3 & 0.1   & 0.00E+00 & 0.8   & 64    & 3     & -     & -     & -     & -     & 48.25 & 0.94  & 14.00ms/2.90s \\
          & GCNII & 0.01  & 1.00E-04 & 0.5   & 64    & 4     & -     & -     & 1.5   & 0.2   & 38.81 & 1.97  & 13.35ms/2.70s \\
          & GCNII* & 0.01  & 5.00E-04 & 0.5   & 64    & 4     & -     & -     & 1.5   & 0.3   & 38.31 & 1.3   & 13.81ms/2.78s \\
          & FAGCN & 0.05  & 1.00E-04 & 0     & 32    & 2     & -     & -     & -     & -     & 42.24 & 1.2   & 16ms/6.7961s \\
          &  Mixhop & 0.01  & 0.001 & 0.5   & 32    & 2     & - & - & - & - & 24.55 & 2.6   & {17.634ms/3.562s} \\
          & H2GCN & 0.01  & 0.001 & 0     & 16    & 1     & - & - & - & - & 30.39 & 1.22  & {9.315ms/1.882s} \\
          & GCN+JK & 0.001 & 0.001 & 0.5   & 32    & 2     & - & {max} & - & - & 53.4  & 1.9   & {14.321ms/2.905s} \\
          &  GAT+JK & 0.001 & 0.001 & 0.5   & 8     & 2     & 4     & {max} & - & - & 52.28 & 3.61  & {29.097ms/5.878s} \\
          \midrule
    \multirow{12}[1]{*}{\textbf{Cora}} & SGC-1 & 0.1   & 5.00E-06 & 0     & 64    & -     & -     & -     & -     & -     & 85.12 & 1.64  & 3.47ms/11.55s \\
          & SGC-2 & 0.1   & 1.00E-05 & 0     & 64    & -     & -     & -     & -     & -     & 85.48 & 1.48  & 2.91ms/6.85s \\
          & GCN   & 0.1   & 5.00E-04 & 0.2   & 64    & 2     & -     & -     & -     & -     & 87.78 & 0.96  & 4.24ms/0.86s \\
          & Snowball-2 & 0.1   & 5.00E-04 & 0.1   & 64    & 2     & -     & -     & -     & -     & 88.64 & 1.15  & 4.65ms/0.94s \\
          & Snowball-3 & 0.05  & 1.00E-03 & 0.6   & 64    & 3     & -     & -     & -     & -     & 89.33 & 1.3   & 6.41ms/1.32s \\
          & GCNII & 0.01  & 1.00E-04 & 0.5   & 64    & 16    & -     & -     & 0.5   & 0.2   & 88.98 & 1.33  &  \\
          & GCNII* & 0.01  & 5.00E-04 & 0.5   & 64    & 4     & -     & -     & 0.5   & 0.5   & 88.93 & 1.37  & 10.16ms/2.24s \\
          & FAGCN & 0.05  & 5.00E-04 & 0     & 32    & 2     & -     & -     & -     & -     & 88.85 & 1.36  & 8.4ms/3.3183s \\
          &  Mixhop & 0.01  & 0.001 & 0.5   & 16    & 2     & - & - & - & - & 65.65 & 11.31 & {11.177ms/2.278s} \\
          & H2GCN & 0.01  & 0.001 & 0     & 32    & 1     & - & - & - & - & 87.52 & 0.61  & {4.335ms/1.209s} \\
          & GCN+JK & 0.001 & 0.001 & 0.5   & 64    & 2     & - & {cat} & - & - & 86.90 & 1.51  & {6.656ms/1.346s} \\
          &  GAT+JK & 0.001 & 0.001 & 0.5   & 32    & 2     & 2     & {cat} & - & - & 89.52 & 0.43  & {12.91ms/2.608s} \\
    \midrule
    \multirow{12}[1]{*}{\textbf{CiteSeer}} & SGC-1 & 0.1   & 5.00E-04 & 0     & 64    & -     & -     & -     & -     & -     & 79.66 & 0.75  & 3.43ms/7.30s \\
          & SGC-2 & 0.01  & 5.00E-04 & 0.9   & 64    & -     & -     & -     & -     & -     & 80.75 & 1.15  & 5.33ms/4.40s \\
          & GCN   & 0.1   & 1.00E-03 & 0.9   & 64    & 2     & -     & -     & -     & -     & 81.39 & 1.23  & 4.18ms/0.86s \\
          & Snowball-2 & 0.1   & 1.00E-03 & 0.8   & 64    & 2     & -     & -     & -     & -     & 81.53 & 1.71  & 5.19ms/1.11s \\
          & Snowball-3 & 0.1   & 1.00E-03 & 0.9   & 64    & 3     & -     & -     & -     & -     & 80.93 & 1.32  & 7.64ms/1.69s \\
          & GCNII & 0.01  & 1.00E-03 & 0.5   & 64    & 16    & -     & -     & 0.5   & 0.2   & 81.58 & 1.3   &  \\
          & GCNII* & 0.01  & 1.00E-03 & 0.5   & 64    & 16    & -     & -     & 0.5   & 0.2   & 81.83 & 1.78  & 32.50ms/10.29s \\
          & FAGCN & 0.05  & 5.00E-04 & 0     & 32    & 2     & -     & -     & -     & -     & 82.37 & 1.46  & 9.4ms/4.7648s \\
          &  Mixhop & 0.01  & 0.001 & 0.5   & 16    & 2     & - & - & - & - & 49.52 & 13.35 & {13.793ms/2.786s} \\
          & H2GCN & 0.01  & 0.001 & 0     & 8     & 1     & - & - & - & - & 79.97 & 0.69  & {5.794ms/3.049s} \\
          & GCN+JK & 0.001 & 0.001 & 0.5   & 32    & 2     & - & {max} & - & - & 73.77 & 1.85  & {5.264ms/1.063s} \\
          &  GAT+JK & 0.001 & 0.001 & 0.5   & 8     & 2     & 4     & {max} & - & - & 74.49 & 2.76  & {12.326ms/2.49s} \\
          \midrule
    \multirow{12}[0]{*}{\textbf{PubMed}} & SGC-1 & 0.05  & 5.00E-06 & 0.3   & 64    & -     & -     & -     & -     & -     & 87.75 & 0.88  & 6.04ms/2.61s \\
          & SGC-2 & 0.05  & 5.00E-05 & 0.1   & 64    & -     & -     & -     & -     & -     & 88.79 & 0.5   & 8.62ms/3.18s \\
          & GCN   & 0.1   & 5.00E-05 & 0.6   & 64    & 2     & -     & -     & -     & -     & 88.9  & 0.32  & 5.08ms/1.03s \\
          & Snowball-2 & 0.1   & 5.00E-04 & 0     & 64    & 2     & -     & -     & -     & -     & 89.04 & 0.49  & 5.68ms/1.19s \\
          & Snowball-3 & 0.1   & 5.00E-06 & 0     & 64    & 3     & -     & -     & -     & -     & 88.8  & 0.82  & 8.54ms/1.75s \\
          & GCNII & 0.01  & 1.00E-06 & 0.5   & 64    & 4     & -     & -     & 0.5   & 0.5   & 89.8  & 0.3   & 10.98ms/3.21s \\
          & GCNII* & 0.01  & 1.00E-06 & 0.5   & 64    & 4     & -     & -     & 0.5   & 0.1   & 89.98 & 0.52  & 11.47ms/3.24s \\
          & FAGCN & 0.05  & 5.00E-04 & 0     & 32    & 2     & -     & -     & -     & -     & 89.98 & 0.54  & 14.5ms/6.411s \\
          &  Mixhop & 0.01  & 0.001 & 0.5   & 16    & 2     & - & - & - & - & 87.04 & 4.10  & {17.459ms/3.527s} \\
          & H2GCN & 0.01  & 0.001 & 0     & 64    & 1     & - & - & - & - & 87.78 & 0.28  & {8.039ms/2.28s} \\
          & GCN+JK & 0.01  & 0.001 & 0.5   & 32    & 2     & - & {cat} & - & - & 90.09 & 0.68  & {12.001ms/2.424s} \\
          &  GAT+JK & 0.1   & 0.001 & 0.5   & 8     & 2     & 4     & {max} & - & - & 89.15 & 0.87  & {20.403ms/4.125s} \\
          \midrule
    \multirow{3}[1]{*}{\textbf{Deezer-Europe}} & FAGCN & 0.01  & 0.0001 & 0     & 32    & 2     & -     & -     & -     & -     & 66.86 & 0.53  & {41.7ms/20.8362s} \\
          & GCNII & 0.01  & 5e-6,1e-5 & 0.5   & 64    & 32    & -     & -     & 0.5   & 0.5   & 66.38 & 0.45  & 126.58ms/63.16s \\
          & GCNII* & 0.01  & 1e-4,1e-3 & 0.5   & 64    & 32    & -     & -     & 0.5   & 0.5   & 66.42 & 0.56  & 134.05ms/66.89s \\
    \bottomrule
    \bottomrule
    \end{tabular}%
     \label{tab:hyperparameters_baselines}%
    }
\end{table}%

\begin{table}[htbp]
  \centering
  \scalebox{.57}{
  \caption{Optimal hyperparameters for ACM(II)-GNNs on random 60\%/20\%/20\% splits}
    \begin{tabular}{c|c|cccccccccccc}
    \toprule
    \toprule
    \multicolumn{14}{c}{\textbf{Hyperparameters for ACM-GNNs and ACMII-GNNs}} \\
    \midrule
    Datasets & Models\textbackslash{}Hyperparameters & lr    & weight\_decay & dropout & hidden & {\# layers} & {Gat heads} & {JK Type} & lambda & alpha\_l & results & std   & average epoch time/average total time \\
    \midrule
    \multirow{10}[0]{*}{\textbf{Cornell}} & ACM-SGC-1 & 0.01  & 5.00E-03 & 0.6   & 64    & -     & -     & -     & -     & -     & 93.77 & 1.91  & 5.53ms/2.31s \\
          & ACM-SGC-2 & 0.01  & 5.00E-03 & 0.6   & 64    & -     & -     & -     & -     & -     & 93.77 & 2.17  &  4.73ms/1.87s \\
          & ACM-GCN & 0.05  & 1.00E-02 & 0.2   & 64    & 2     & -     & -     & -     & -     & 94.75 & 3.8   & 8.25ms/1.69s \\
          & ACMII-GCN & 0.1   & 1.00E-02 & 0.5   & 64    & 2     & -     & -     & -     & -     & 95.25 & 2.79  & 8.43ms/1.71s \\
          & ACM-GCNII & 0.01  & 1.00E-03 & 0.5   & 64    & 1     & -     & -     & 0.5   & 0.4   & 92.62 & 3.13  & 6.81ms/1.43s \\
          & ACM-GCNII* & 0.01  & 5.00E-04 & 0.5   & 64    & 1     & -     & -     & 0.5   & 0.1   & 93.44 & 2.74  & 6.76ms/1.39s \\
          & ACM-Snowball-2 & 0.05  & 1.00E-02 & 0.2   & 64    & 2     & -     & -     & -     & -     & 95.08 & 3.11  & 9.15ms/1.86s \\
          & ACM-Snowball-3 & 0.1   & 1.00E-02 & 0.4   & 64    & 3     & -     & -     & -     & -     & 94.26 & 2.57  & 13.20ms/2.68s \\
          & ACMII-Snowball-2 & 0.05  & 1.00E-02 & 0.6   & 64    & 2     & -     & -     & -     & -     & 95.25 & 1.55  & 8.23ms/1.72s \\
          & ACMII-Snowball-3 & 0.05  & 1.00E-02 & 0.7   & 64    & 3     & -     & -     & -     & -     & 93.61 & 2.79  & 11.70ms/2.37s \\
          \midrule
    \multirow{10}[0]{*}{\textbf{Wisconsin}} & ACM-SGC-1 & 0.05  & 5.00E-03 & 0.7   & 64    & -     & -     & -     & -     & -     & {93.25} & 2.92  & 5.96ms/1.34s \\
          & ACM-SGC-2 & 0.1   & 5.00E-03 & 0.2   & 64    & -     & -     & -     & -     & -     & 94    & 2.61  & 4.60ms/0.95s \\
          & ACM-GCN & 0.1   & 5.00E-03 & 0     & 64    & 2     & -     & -     & -     & -     & 95.75 & 2.03  & {8.11ms/1.64s} \\
          & ACMII-GCN & 0.1   & 1.00E-02 & 0.2   & 64    & 2     & -     & -     & -     & -     & 96.62 & 2.44  & 8.28ms/1.68s \\
          & ACM-GCNII & 0.01  & 5.00E-03 & 0.5   & 64    & 1     & -     & -     & 1     & 0.1   & 94.63 & 2.96  & 9.31ms/2.19s \\
          & ACM-GCNII* & 0.01  & 1.00E-03 & 0.5   & 64    & 1     & -     & -     & 1.5   & 0.4   & 94.37 & 2.81  & 7.11ms/1.45s \\
          & ACM-Snowball-2 & 0.1   & 5.00E-03 & 0.1   & 64    & 2     & -     & -     & -     & -     & 96.38 & 2.59  &  8.63ms/1.74s \\
          & ACM-Snowball-3 & 0.05  & 1.00E-02 & 0.3   & 64    & 3     & -     & -     & -     & -     & 96.62 & 1.86  & 12.79ms/2.58s \\
          & ACMII-Snowball-2 & 0.1   & 1.00E-02 & 0.1   & 64    & 2     & -     & -     & -     & -     & {96.63} & 2.24  & 8.11ms/1.65s \\
          & ACMII-Snowball-3 & 0.1   & 5.00E-03 & 0.1   & 64    & 3     & -     & -     & -     & -     & 97    & 2.63  & 12.38ms/2.51s \\
          \midrule
    \multirow{10}[0]{*}{\textbf{Texas}} & ACM-SGC-1 & 0.01  & 5.00E-03 & 0.6   & 64    & -     & -     & -     & -     & -     & 93.61 & 1.55  & 5.43ms/2.18s \\
          & ACM-SGC-2 & 0.05  & 5.00E-03 & 0.4   & 64    & -     & -     & -     & -     & -     & 93.44 & 2.54  & 4.59ms/1.01s \\
          & ACM-GCN & 0.05  & 1.00E-02 & 0.6   & 64    & 2     & -     & -     & -     & -     & 94.92 & 2.88  & 8.33ms/1.70s \\
          & ACMII-GCN & 0.1   & 5.00E-03 & 0.4   & 64    & 2     & -     & -     & -     & -     & 95.08 & 2.54  & 8.49ms/1.72s \\
          & ACM-GCNII & 0.01  & 1.00E-03 & 0.5   & 64    & 1     & -     & -     & 0.5   & 0.4   & 92.46 & 1.97  & 6.47ms/1.36s \\
          & ACM-GCNII* & 0.01  & 1.00E-03 & 0.5   & 64    & 1     & -     & -     & 0.5   & 0.4   & 93.28 & 2.79  & 7.03ms/1.45s \\
          & ACM-Snowball-2 & 0.05  & 1.00E-02 & 0.1   & 64    & 2     & -     & -     & -     & -     & 95.74 & 2.22  & 8.35ms/1.71s \\
          & ACM-Snowball-3 & 0.01  & 5.00E-03 & 0.6   & 64    & 3     & -     & -     & -     & -     & 94.75 & 2.41  & 12.56ms/2.63s \\
          & ACMII-Snowball-2 & 0.1   & 1.00E-02 & 0.4   & 64    & 2     & -     & -     & -     & -     & 95.25 & 1.55  & 9.74ms/1.97s \\
          & ACMII-Snowball-3 & 0.05  & 1.00E-02 & 0.6   & 64    & 3     & -     & -     & -     & -     & 94.75 & 3.09  & 11.91ms/2.42s \\
          \midrule
    \multirow{10}[0]{*}{\textbf{Film}} & ACM-SGC-1 & 0.05  & 5.00E-05 & 0.7   & 64    & -     & -     & -     & -     & -     & 39.33 & 1.25  & 5.21ms/2.33s \\
          & ACM-SGC-2 & 0.1   & 5.00E-05 & 0.7   & 64    & -     & -     & -     & -     & -     & 40.13 & 1.21  & 12.41ms/4.87s \\
          & ACM-GCN & 0.1   & 5.00E-04 & 0.5   & 64    & 2     & -     & -     & -     & -     & 41.62 & 1.15  & 10.72ms/2.66s \\
          & ACMII-GCN & 0.1   & 5.00E-04 & 0.5   & 64    & 2     & -     & -     & -     & -     & 41.24 & 1.16  & 10.51ms/2.44s \\
          & ACM-GCNII & 0.01  & 0.00E+00 & 0.5   & 64    & 3     & -     & -     & 1.5   & 0.2   & 41.37 & 1.37  & 13.65ms/2.74s \\
          & ACM-GCNII* & 0.01  & 1.00E-05 & 0.5   & 64    & 3     & -     & -     & 1.5   & 0.1   & 41.27 & 1.24  & 14.98ms/3.01s \\
          & ACM-Snowball-2 & 0.1   & 5.00E-03 & 0     & 64    & 2     & -     & -     & -     & -     & 41.4  & 1.23  & 10.30ms/2.08s \\
          & ACM-Snowball-3 & 0.05  & 1.00E-02 & 0     & 64    & 3     & -     & -     & -     & -     & 41.27 & 0.8   & 16.43ms/3.52s \\
          & ACMII-Snowball-2 & 0.1   & 5.00E-03 & 0     & 64    & 2     & -     & -     & -     & -     & 41.1  & 0.75  & 10.74ms/2.19s \\
          & ACMII-Snowball-3 & 0.05  & 5.00E-03 & 0.2   & 64    & 3     & -     & -     & -     & -     & 40.31 & 1.6   & 16.31ms/3.29s \\
          \midrule
    \multirow{10}[0]{*}{\textbf{Chameleon}} & ACM-SGC-1 & 0.1   & 5.00E-06 & 0.9   & 64    & -     & -     & -     & -     & -     & 63.68 & 1.62  & 5.41ms/1.21s \\
          & ACM-SGC-2 & 0.1   & 5.00E-06 & 0.9   & 64    & -     & -     & -     & -     & -     & 60.48 & 1.55  & 7.86ms/1.81s \\
          & ACM-GCN & 0.01  & 5.00E-05 & 0.8   & 64    & 2     & -     & -     & -     & -     & 68.18 & 1.67  & 10.55ms/3.12s \\
          & ACMII-GCN & 0.05  & 5.00E-05 & 0.7   & 64    & 2     & -     & -     & -     & -     & 68.38 & 1.36  & 10.90ms/2.39s \\
          & ACM-GCNII & 0.01  & 5.00E-06 & 0.5   & 64    & 4     & -     & -     & 0.5   & 0.1   & 58.73 & 2.52  & 18.31ms/3.68s \\
          & ACM-GCNII* & 0.01  & 1.00E-03 & 0.5   & 64    & 1     & -     & -     & 1     & 0.1   & 61.66 & 2.29  & 6.68ms/1.40s \\
          & ACM-Snowball-2 & 0.05  & 5.00E-05 & 0.7   & 64    & 2     & -     & -     & -     & -     & 68.51 & 1.7   & 9.92ms/2.06s \\
          & ACM-Snowball-3 & 0.01  & 1.00E-04 & 0.7   & 64    & 3     & -     & -     & -     & -     & 68.4  & 2.05  & 14.49ms/3.15s \\
          & ACMII-Snowball-2 & 0.1   & 5.00E-05 & 0.6   & 64    & 2     & -     & -     & -     & -     & 67.83 & 2.63  & 9.99ms/2.10s \\
          & ACMII-Snowball-3 & 0.05  & 1.00E-04 & 0.7   & 64    & 3     & -     & -     & -     & -     & 67.53 & 2.83  & 15.03ms/3.29s \\
          \midrule
    \multirow{10}[0]{*}{\textbf{Squirrel}} & ACM-SGC-1 & 0.05  & 0.00E+00 & 0.9   & 64    & -     & -     & -     & -     & -     & 46.4  & 1.13  & 6.96ms/2.16s \\
          & ACM-SGC-2 & 0.05  & 0.00E+00 & 0.9   & 64    & -     & -     & -     & -     & -     & 40.91 & 1.39  & 35.20ms/10.66s \\
          & ACM-GCN & 0.05  & 5.00E-06 & 0.6   & 64    & 2     & -     & -     & -     & -     & 58.02 & 1.86  & 14.35ms/2.98s \\
          & ACMII-GCN & 0.05  & 0.00E+00 & 0.7   & 64    & 2     & -     & -     & -     & -     & 53.76 & 1.63  & 14.08ms/3.39s \\
          & ACM-GCNII & 0.01  & 1.00E-05 & 0.5   & 64    & 4     & -     & -     & 0.5   & 0.1   & 40.9  & 1.58  & 20.72ms/4.17s \\
          & ACM-GCNII* & 0.01  & 1.00E-03 & 0.5   & 64    & 4     & -     & -     & 0.5   & 0.3   & 38.32 & 1.5   & 21.78ms/4.38s \\
          & ACM-Snowball-2 & 0.05  & 5.00E-06 & 0.6   & 64    & 2     & -     & -     & -     & -     & 55.97 & 2.03  & 15.38ms/3.15s \\
          & ACM-Snowball-3 & 0.01  & 1.00E-04 & 0.6   & 64    & 3     & -     & -     & -     & -     & 55.73 & 2.39  & 26.15ms/5.94s \\
          & ACMII-Snowball-2 & 0.1   & 5.00E-06 & 0.6   & 64    & 2     & -     & -     & -     & -     & 53.48 & 0.6   & 15.54ms/3.19s \\
          & ACMII-Snowball-3 & 0.05  & 5.00E-05 & 0.5   & 64    & 3     & -     & -     & -     & -     & 52.31 & 1.57  & 26.24ms/5.30s \\
          \midrule
    \multirow{10}[0]{*}{\textbf{Cora}} & ACM-SGC-1 & 0.01  & 5.00E-06 & 0.9   & 64    & -     & -     & -     & -     & -     & 86.63 & 1.13  & 6.00ms/7.40s \\
          & ACM-SGC-2 & 0.1   & 5.00E-05 & 0.6   & 64    & -     & -     & -     & -     & -     & 87.64 & 0.99  & 4.85ms/1.17s \\
          & ACM-GCN & 0.1   & 5.00E-03 & 0.5   & 64    & 2     & -     & -     & -     & -     & 88.62 & 1.22  & 8.84ms/1.81s \\
          & ACMII-GCN & 0.1   & 5.00E-03 & 0.4   & 64    & 2     & -     & -     & -     & -     & 89    & 0.72  & 8.93ms/1.83s \\
          & ACM-GCNII & 0.01  & 1.00E-03 & 0.5   & 64    & 3     & -     & -     & 1     & 0.2   & 89.1  & 1.61  &  14.07ms/3.04s \\
          & ACM-GCNII* & 0.01  & 1.00E-02 & 0.5   & 64    & 4     & -     & -     & 1     & 0.2   & 89    & 1.35  & 11.36ms/2.48s \\
          & ACM-Snowball-2 & 0.05  & 1.00E-03 & 0.6   & 64    & 2     & -     & -     & -     & -     & 88.83 & 1.49  & 9.34ms/1.92s \\
          & ACM-Snowball-3 & 0.1   & 1.00E-02 & 0.3   & 64    & 3     & -     & -     & -     & -     & 89.59 & 1.58  & 13.33ms/2.75s \\
          & ACMII-Snowball-2 & 0.1   & 5.00E-03 & 0.5   & 64    & 2     & -     & -     & -     & -     & 88.95 & 1.04  & 9.29ms/1.90s \\
          & ACMII-Snowball-3 & 0.1   & 5.00E-03 & 0.5   & 64    & 3     & -     & -     & -     & -     & 89.36 & 1.26  & 14.18ms/2.89s \\
          \midrule
    \multirow{10}[0]{*}{\textbf{CiteSeer}} & ACM-SGC-1 & 0.01  & 5.00E-04 & 0.9   & 64    & -     & -     & -     & -     & -     & {80.96} & 0.93  & 5.90ms/4.31s \\
          & ACM-SGC-2 & 0.05  & 5.00E-04 & 0.9   & 64    & -     & -     & -     & -     & -     & 80.93 & 1.16  & 5.01ms/1.42s \\
          & ACM-GCN & 0.05  & 5.00E-03 & 0.7   & 64    & 2     & -     & -     & -     & -     & 81.68 & 0.97  & 11.35ms/2.57s \\
          & ACMII-GCN & 0.05  & 5.00E-05 & 0.7   & 64    & 2     & -     & -     & -     & -     & 81.58 & 1.77  & 9.55ms/1.94s \\
          & ACM-GCNII & 0.01  & 1.00E-02 & 0.5   & 64    & 3     & -     & -     & 0.5   & 0.3   & 82.28 & 1.12  & 15.61ms/3.56s \\
          & ACM-GCNII* & 0.01  & 1.00E-02 & 0.5   & 64    & 3     & -     & -     & 0.5   & 0.5   & 81.69 & 1.25  & 15.56ms/3.61s \\
          & ACM-Snowball-2 & 0.05  & 5.00E-03 & 0.7   & 64    & 2     & -     & -     & -     & -     & 81.58 & 1.23  & 11.14ms/2.50s \\
          & ACM-Snowball-3 & 0.01  & 5.00E-03 & 0.9   & 64    & 3     & -     & -     & -     & -     & 81.32 & 0.97  & 15.91ms/5.36s \\
          & ACMII-Snowball-2 & 0.05  & 5.00E-03 & 0.7   & 64    & 2     & -     & -     & -     & -     & 82.07 & 1.04  & 10.97ms/2.55s \\
          & ACMII-Snowball-3 & 0.05  & 1.00E-04 & 0.6   & 64    & 3     & -     & -     & -     & -     & 81.56 & 1.15  & 14.95ms/3.03s \\
          \midrule
    \multirow{10}[0]{*}{\textbf{PubMed}} & ACM-SGC-1 & 0.05  & 5.00E-06 & 0.3   & 64    & -     & -     & -     & -     & -     & 87.75 & 0.88  & 6.04ms/2.61s \\
          & ACM-SGC-2 & 0.05  & 5.00E-05 & 0.1   & 64    & -     & -     & -     & -     & -     & 88.79 & 0.5   & 8.62ms/3.18s \\
          & ACM-GCN & 0.1   & 5.00E-04 & 0.2   & 64    & 2     & -     & -     & -     & -     & 90.54 & 0.63  & 10.20ms/2.08s \\
          & ACMII-GCN & 0.1   & 5.00E-04 & 0.2   & 64    & 2     & -     & -     & -     & -     & 90.74 & 0.5   & 10.20ms/2.07s \\
          & ACM-GCNII & 0.01  & 1.00E-04 & 0.5   & 64    & 3     & -     & -     & 1.5   & 0.5   & 90.12 & 0.4   & 15.07ms/3.35s \\
          & ACM-GCNII* & 0.01  & 1.00E-04 & 0.5   & 64    & 3     & -     & -     & 1.5   & 0.5   & 90.18 & 0.51  & 16.62ms/3.72s \\
          & ACM-Snowball-2 & 0.1   & 1.00E-04 & 0.3   & 64    & 2     & -     & -     & -     & -     & 90.81 & 0.52  & 11.52ms/2.36s \\
          & ACM-Snowball-3 & 0.05  & 1.00E-03 & 0.2   & 64    & 3     & -     & -     & -     & -     & 91.44 & 0.59  & 18.06ms/3.69s \\
          & ACMII-Snowball-2 & 0.1   & 1.00E-04 & 0.3   & 64    & 2     & -     & -     & -     & -     & 90.56 & 0.39  & 11.74ms/2.39s \\
          & ACMII-Snowball-3 & 0.1   & 5.00E-04 & 0.2   & 64    & 3     & -     & -     & -     & -     & 91.31 & 0.6   & 18.61ms/3.88s \\
          \midrule
    \multirow{6}[1]{*}{\textbf{Deezer-Europe}} & ACM-SGC-1 & 0.05  & 0,5e-6,1e-5,5e-5 & 0.3   & 64    & -     & -     & -     & -     & -     & 66.67 & 0.56  & 146.41ms/73.06s \\
          & ACM-SGC-2 & 0.002 & {5e-5,1e-4} & 0.3   & 64    & -     & -     & -     & -     & -     & 66.53 & 0.57  & 195.21ms/97.41s \\
          & ACM-GCN & 0.002 & 5.00E-04 & 0.5   & 64    & 2     & -     & -     & -     & -     & 67.01 & 0.38  & 136.45ms/68.09s \\
          & ACMII-GCN & 0.01  & 5.00E-05 & 0.8   & 64    & 2     & -     & -     & -     & -     & 67.15 & 0.41  & 135.24ms/67.48s \\
          & ACM-GCNII & 0.01  & 0,5e-6 & 0.5   & 64    & 1     & -     & -     & 0.5   & 0.4   & 66.39 & 0.56  & 80.82ms/40.33s \\
          & ACM-GCNII* & 0.01  & 0.0001,1e-3 & 0.5   & 64    & 1     & -     & -     & 1.5   & 0.2   & 66.6  & 0.57  & 80.95ms/40.40s \\
    \bottomrule
    \bottomrule
    \end{tabular}%
    \label{tab:hyperparameters_acmgnns}%
    }
\end{table}%

\begin{table}[htbp]
  \centering
  \caption{Optimal hyperparameters for ACM(II)-GCN+ and ACM(II)-GCN++ on random 60\%/20\%/20\% splits}
  \scalebox{.75}{
    \begin{tabular}{c|c|ccccccccc}
    \toprule
    \toprule
    Datasets & Models\textbackslash{}Hyperparameters & lr    & weight\_decay & dropout & hidden & with A & results & std   & average epoch time/average total time \\
    \midrule
    \multirow{4}[1]{*}{\textbf{Cornell}} & ACM-GCN+ & 0.05  & 1.00E-02 & 0.1   & 64    & Y     & 94.92 & 2.79  & 16.66ms/3.37s \\
          & ACMII-GCN+ & 0.05  & 1.00E-02 & 0.3   & 64    & Y     & 93.93 & 1.05  & 12.55ms/2.56s \\
          & ACM-GCN++(with xX) & 0.1   & 5.00E-03 & 0.4   & 64    & N     & 93.93 & 3.03  & 12.89ms/2.62s \\
          & ACMII-GCN++(with xX) & 0.05  & 1.00E-02 & 0.6   & 64    & Y     & 92.62 & 2.57  & 18.25ms/3.69s \\
    \midrule
    \multirow{4}[1]{*}{\textbf{Wisconsin}} & ACM-GCN+ & 0.05  & 1.00E-02 & 0.3   & 64    & Y     & 96.5  & 2.08  & 16.54ms/3.35s \\
          & ACMII-GCN+ & 0.01  & 1.00E-02 & 0.1   & 64    & N     & 97.5  & 1.25  & 12.09ms/2.88s \\
          & ACM-GCN++(with xX) & 0.05  & 1.00E-02 & 0.1   & 64    & Y     & 96.75 & 1.79  & 18.12ms/3.66s \\
          & ACMII-GCN++(with xX) & 0.01  & 1.00E-02 & 0.1   & 64    & Y     & 97.13 & 1.68  & 17.32ms/3.53s \\
    \midrule
    \multirow{4}[1]{*}{\textbf{Texas}} & ACM-GCN+ & 0.05  & 1.00E-03 & 0.3   & 64    & N     & 94.92 & 2.79  & 12.05ms/2.44s \\
          & ACMII-GCN+ & 0.05  & 1.00E-02 & 0.1   & 64    & Y     & 96.56 & 2     &  22.63 ms/4.58s \\
          & ACM-GCN++(with xX) & 0.05  & 5.00E-04 & 0.2   & 64    & N     & 95.41 & 2.82  & 13.20ms/2.67s \\
          & ACMII-GCN++(with xX) & 0.05  & 5.00E-04 & 0.1   & 64    & N     & 94.75 & 2.91  & 12.82ms/2.60s \\
    \midrule
    \multirow{4}[1]{*}{\textbf{Film}} & ACM-GCN+ & 0.01  & 1.00E-03 & 0.8   & 64    & N     & 41.79 & 1.01  & 13.57ms/3.59s \\
          & ACMII-GCN+ & 0.1   & 5.00E-05 & 0.7   & 64    & N     & 41.86 & 1.48  & 13.38ms/3.59s \\
          & ACM-GCN++(with xX) & 0.002 & 5.00E-03 & 0.9   & 64    & N     & 41.5  & 1.54  & 13.76ms/2.77s \\
          & ACMII-GCN++(with xX) & 0.002 & 5.00E-03 & 0.9   & 64    & N     & 41.66 & 1.42  & 13.67ms/2.77s \\
    \midrule
    \multirow{4}[1]{*}{\textbf{Chameleon}} & ACM-GCN+ & 0.002 & 1.00E-03 & 0.4   & 64    & Y     & 76.08 & 2.13  & 18.19ms/8.60s \\
          & ACMII-GCN+ & 0.1   & 1.00E-04 & 0.7   & 64    & Y     & 75.23 & 1.72  & 17.39ms/3.57s \\
          & ACM-GCN++(with xX) & 0.1   & 5.00E-05 & 0.8   & 64    & Y     & 75.51 & 1.58  & 18.69ms/4.17s \\
          & ACMII-GCN++(with xX) & 0.01  & 1.00E-04 & 0.8   & 64    & Y     & 75.93 & 1.71  & 18.70ms/4.53s \\
    \midrule
    \multirow{4}[2]{*}{\textbf{Squirrel}} & ACM-GCN+ & 0.01  & 1.00E-04 & 0.6   & 64    & Y     & 69.26 & 1.11  & 24.71ms/4.97s \\
          & ACMII-GCN+ & 0.01  & 1.00E-04 & 0.6   & 64    & Y     & 68.56 & 1.33  & 21.21ms/4.26s \\
          & ACM-GCN++(with xX) & 0.002 & 1.00E-03 & 0.7   & 64    & Y     & 69.81 & 1.11  & 22.14ms/5.34s \\
          & ACMII-GCN++(with xX) & 0.002 & 1.00E-04 & 0.7   & 64    & Y     & 69.98 & 1.53  & 21.78ms/4.38s \\
    \midrule
    \multirow{4}[2]{*}{\textbf{Cora}} & ACM-GCN+ & 0.1   & 5.00E-03 & 0.3   & 64    & Y     & 89.75 & 1.16  & 17.29ms/3.52s \\
          & ACMII-GCN+ & 0.1   & 5.00E-03 & 0.5   & 64    & Y     & 89.33 & 0.81  & 18.08ms/3.69s \\
          & ACM-GCN++(with xX) & 0.05  & 5.00E-03 & 0.4   & 64    & Y     & 89.18 & 1.11  & 18.21ms/3.69s \\
          & ACMII-GCN++(with xX) & 0.1   & 1.00E-02 & 0.1   & 64    & Y     & 89.47 & 1.08  & 18.53ms/3.76s \\
    \midrule
    \multirow{4}[2]{*}{\textbf{CiteSeer}} & ACM-GCN+ & 0.1   & 1.00E-05 & 0.5   & 64    & N     & 81.65 & 1.48  & 12.44ms/2.50s \\
          & ACMII-GCN+ & 0.002 & 5.00E-03 & 0.8   & 64    & N     & 81.83 & 1.65  & 14.87ms/15.36s \\
          & ACM-GCN++(with xX) & 0.05  & 5.00E-03 & 0.3   & 64    & N     & 81.87 & 1.38  & 13.35ms/2.86s \\
          & ACMII-GCN++(with xX) & 0.01  & 5.00E-04 & 0.9   & 64    & N     & 81.76 & 1.25  & 14.04ms/3.88s \\
    \midrule
    \multirow{4}[2]{*}{\textbf{PubMed}} & ACM-GCN+ & 0.1   & 1.00E-04 & 0.1   & 64    & N     & 90.46 & 0.69  & 15.15ms/3.09s \\
          & ACMII-GCN+ & 0.1   & 1.00E-04 & 0.3   & 64    & N     & 90.39 & 0.33  & 17.36 ms/3.55s \\
          & ACM-GCN++(with xX) & 0.1   & 1.00E-04 & 0.1   & 64    & N     & 90.96 & 0.62  & 16.35ms/3.47s \\
          & ACMII-GCN++(with xX) & 0.1   & 1.00E-04 & 0.3   & 64    & N     & 90.63 & 0.56  & 16.18ms/3.39s \\
    \midrule
    \multicolumn{1}{c|}{\multirow{4}[2]{*}{\textbf{Deezer-Europe }}} & ACM-GCN+ & 0.002 & 1.00E-06 & 0.7   & 64    & N     & 67.4  & 0.44  & 281.97ms/140.70s \\
          & ACMII-GCN+ & 0.002 & 1.00E-04 & 0.8   & 64    & N     & 67.3  & 0.48  & 281.48ms/140.46s \\
          & ACM-GCN++(with xX) & 0.002 & 1.00E-03 & 0.7   & 64    & Y     & 67.44 & 0.31  & 332.92ms/166.13s \\
          & ACMII-GCN++(with xX) & 0.002 & 1.00E-05 & 0.8   & 64    & N     & 67.5  & 0.53  & 326.09ms/162.72s \\
    \bottomrule
    \bottomrule
    \end{tabular}%
  \label{tab:optimal_hyperparameter_random_splits_acmgcn+_acmgcn++}%
  }
\end{table}%

See the reported optimal hyperparameters on fixed 48\%/32\%/20\% splits for ACM(II)-GNNs and FAGCN in table \ref{tab:optimal_hyperparameters_fixed_splits} and for ACM(II)-GCN+ and ACM(II)-GCN++ in table \ref{tab:hyperparameters_acmgcn+_acmgcn++}.

\begin{table}[htbp]
  \centering
  \caption{Optimal hyperparameters for FAGCN and ACM(II)-GNNs on fixed 48\%/32\%/20\% splits}
   \scalebox{.675}{
    \begin{tabular}{c|c|ccccccc}
    \toprule
    Datasets & Models\textbackslash{}Hyperparameters & lr    & weight\_decay & dropout & hidden & results & std   & average epoch time/average total time \\
     \midrule
    \multirow{9}[0]{*}{\textbf{Cornell}} & ACM-SGC-1 & 0.01  & 5.00E-06 & 0     & 64    & 82.43 & 5.44  & 5.37ms/23.05s \\
          & ACM-SGC-2 & 0.01  & 5.00E-06 & 0     & 64    & 82.43 & 5.44  & 5.93ms/25.66s \\
          & ACM-GCN & 0.05  & 5.00E-04 & 0.5   & 64    & 85.14 & 6.07  & 8.04ms/1.67s \\
          & ACMII-GCN & 0.1   & 1.00E-04 & 0     & 64    & 85.95 & 5.64  & 7.83ms/2.66s \\
          & FAGCN & 0.01  & 1.00E-04 & 0.6   & 64    & 76.76 & 5.87  & 8.80ms/7.67s \\
          & ACM-Snowball-2 & 0.05  & 5.00E-03 & 0.3   & 64    & 85.41 & 5.43  & 11.50ms/2.35s \\
          & ACM-Snowball-3 & 0.05  & 5.00E-03 & 0.2   & 64    & 83.24 & 5.38  & 15.06ms/3.12s \\
          & ACMII-Snowball-2 & 0.1   & 5.00E-03 & 0.2   & 64    & 85.68 & 5.93  & 12.63ms/2.58s \\
          & ACMII-Snowball-3 & 0.05  & 5.00E-03 & 0.2   & 64    & 82.7  & 4.86  & 14.59ms/3.06s \\
          \midrule
    \multirow{9}[0]{*}{\textbf{Wisconsin}} & ACM-SGC-1 & 0.1   & 5.00E-06 & 0     & 64    & 86.47 & 3.77  & 5.07ms/14.07s \\
          & ACM-SGC-2 & 0.1   & 5.00E-06 & 0     & 64    & 86.47 & 3.77  & 5.30ms/16.05s \\
          & ACM-GCN & 0.05  & 1.00E-03 & 0.4   & 64    & 88.43 & 3.22  & {8.04ms/1.66s} \\
          & ACMII-GCN & 0.01  & 5.00E-05 & 0.1   & 64    & 87.45 & 3.74  & 8.40ms/2.19s \\
          & FAGCN & 0.01  & 5.00E-05 & 0.5   & 64    & 79.61 & 1.59  & 8.61ms/5.84s \\
          & ACM-Snowball-2 & 0.01  & 1.00E-03 & 0.4   & 64    & 87.06 & 2     & 12.51ms/2.60s \\
          & ACM-Snowball-3 & 0.01  & 1.00E-02 & 0.1   & 64    & 86.67 & 4.37  & 14.92ms/3.15s \\
          & ACMII-Snowball-2 & 0.01  & 5.00E-04 & 0.1   & 64    & 87.45 & 2.8   & 11.96ms/2.63s \\
          & ACMII-Snowball-3 & 0.01  & 5.00E-03 & 0.5   & 64    & 85.29 & 4.23  & 14.87ms/3.10s \\
           \midrule
    \multirow{9}[0]{*}{\textbf{Texas}} & ACM-SGC-1 & 0.01  & 1.00E-05 & 0     & 64    & 81.89 & 4.53  & 5.34ms/19.00s \\
          & ACM-SGC-2 & 0.05  & 1.00E-05 & 0     & 64    & 81.89 & 4.53  & 5.50ms/9.26s \\
          & ACM-GCN & 0.05  & 5.00E-04 & 0.5   & 64    & 87.84 & 4.4   & 9.62ms/1.99s \\
          & ACMII-GCN & 0.01  & 1.00E-03 & 0.1   & 64    & 86.76 & 4.75  & 9.98ms/2.22s \\
          & FAGCN & 0.01  & 1.00E-05 & 0     & 64    & 76.49 & \textcolor[rgb]{ .267,  .267,  .267}{2.87} & 10.45ms/5.70s \\
          & ACM-Snowball-2 & 0.01  & 5.00E-03 & 0.2   & 64    & 87.57 & 4.86  & 11.56ms/2.45s \\
          & ACM-Snowball-3 & 0.01  & 5.00E-03 & 0.2   & 64    & 87.84 & 3.87  & 15.17ms/3.15s \\
          & ACMII-Snowball-2 & 0.01  & 1.00E-03 & 0.2   & 64    & 86.76 & 4.43  & 11.36ms/2.30 \\
          & ACMII-Snowball-3 & 0.01  & 5.00E-03 & 0.6   & 64    & 85.41 & 6.42  & 15.84ms/3.48s \\
           \midrule
    \multirow{9}[0]{*}{\textbf{Film}} & ACM-SGC-1 & 0.05  & 5.00E-04 & 0     & 64    & 35.49 & 1.06  & 5.39ms/1.17s \\
          & ACM-SGC-2 & 0.05  & 5.00E-04 & 0.1   & 64    & 36.04 & 0.83  & 13.22ms/3.31s \\
          & ACM-GCN & 0.01  & 5.00E-03 & 0     & 64    & 36.28 & 1.09  & 8.96ms/1.82s \\
          & ACMII-GCN & 0.01  & 5.00E-03 & 0     & 64    & 36.16 & 1.11  & 9.06ms/1.83s \\
          & FAGCN & 0.01  & 5.00E-05 & 0.4   & 64    & 34.82 & 1.35  & 15.60ms/2.51s \\
          & ACM-Snowball-2 & 0.01  & 1.00E-02 & 0     & 64    & 36.89 & 1.18  & 14.77ms/3.01s \\
          & ACM-Snowball-3 & 0.01  & 1.00E-02 & 0.2   & 64    & 36.82 & 0.94  & 16.57ms/3.36s \\
          & ACMII-Snowball-2 & 0.01  & 5.00E-03 & 0.1   & 64    & 36.55 & 1.24  & 12.76ms/2.57s \\
          & ACMII-Snowball-3 & 0.05  & 5.00E-03 & 0.3   & 64    & 36.49 & 1.41  & 16.51ms/3.49s \\
           \midrule
    \multirow{9}[0]{*}{\textbf{Chameleon}} & ACM-SGC-1 & 0.1   & 5.00E-06 & 0.9   & 64    & 63.99 & 1.66  & 5.92ms/1.74s \\
          & ACM-SGC-2 & 0.1   & 0.00E+00 & 0.9   & 64    & 59.21 & 2.22  & 8.84ms/1.78s \\
          & ACM-GCN & 0.05  & 5.00E-05 & 0.7   & 64    & 66.93 & 1.85  & 8.40ms/1.71s \\
          & ACMII-GCN & 0.05  & 5.00E-06 & 0.8   & 64    & 66.91 & 2.55  & 8.90ms/2.10s \\
          & FAGCN & 0.01  & 5.00E-05 & 0     & 64    & 46.07 & 2.11  & 16.90ms/7.94s \\
          & ACM-Snowball-2 & 0.01  & 1.00E-04 & 0.7   & 64    & 67.08 & 2.04  & 12.50ms/2.69s \\
          & ACM-Snowball-3 & 0.01  & 1.00E-05 & 0.8   & 64    & 66.91 & 1.73  & 16.12ms/4.91s \\
          & ACMII-Snowball-2 & 0.01  & 5.00E-05 & 0.8   & 64    & 66.49 & 1.75  & 12.65ms/3.42s \\
          & ACMII-Snowball-3 & 0.05  & 5.00E-05 & 0.7   & 64    & 66.86 & 1.74  & 17.60ms/4.06s \\
           \midrule
    \multirow{9}[0]{*}{\textbf{Squirrel}} & ACM-SGC-1 & 0.05  & 5.00E-06 & 0.9   & 64    & 45    & 1.4   & 6.10ms/2.18s \\
          & ACM-SGC-2 & 0.05  & 0.00E+00 & 0.9   & 64    & 40.02 & 0.96  & 35.75ms/9.62s \\
          & ACM-GCN & 0.05  & 5.00E-06 & 0.7   & 64    & 54.4  & 1.88  & 10.48ms/2.68s \\
          & ACMII-GCN & 0.05  & 5.00E-06 & 0.7   & 64    & 51.85 & 1.38  & 11.69ms/2.91s \\
          & FAGCN & 0     & 5.00E-03 & 0     & 64    & 30.86 & 0.69  & 10.90ms/13.91s \\
          & ACM-Snowball-2 & 0.01  & 1.00E-04 & 0.7   & 64    & 52.5  & 1.49  & 17.89ms/5.78s \\
          & ACM-Snowball-3 & 0.01  & 5.00E-05 & 0.7   & 64    & 53.31 & 1.88  & 22.60ms/7.53s \\
          & ACMII-Snowball-2 & 0.05  & 5.00E-05 & 0.6   & 64    & 50.15 & 1.4   & 16.95ms/3.45s \\
          & ACMII-Snowball-3 & 0.01  & 5.00E-04 & 0.6   & 64    & 48.87 & 1.23  & 23.52ms/4.94s \\
           \midrule
    \multirow{9}[0]{*}{\textbf{Cora}} & ACM-SGC-1 & 0.05  & 5.00E-05 & 0.7   & 64    & 86.9  & 1.38  & 4.99ms/2.40s \\
          & ACM-SGC-2 & 0.1   & 0     & 0.8   & 64    & 87.69 & 1.07  & 5.16ms/1.16s \\
          & ACM-GCN & 0.01  & 5.00E-05 & 0.6   & 64    & 87.91 & 0.95  & 8.41ms/1.84s \\
          & ACMII-GCN & 0.01  & 1.00E-04 & 0.6   & 64    & 88.01 & 1.08  & 8.59ms/1.96s \\
          & FAGCN & 0.02  & 1.00E-04 & 0.5   & 64    & 88.05 & 1.57  & 9.30ms/10.64s \\
          & ACM-Snowball-2 & 0.01  & 1.00E-03 & 0.5   & 64    & 87.42 & 1.09  & 12.54ms/2.72s \\
          & ACM-Snowball-3 & 0.01  & 5.00E-06 & 0.9   & 64    & 87.1  & 0.93  & 15.83ms/11.33s \\
          & ACMII-Snowball-2 & 0.01  & 1.00E-03 & 0.6   & 64    & 87.57 & 0.86  & 12.06ms/2.64s \\
          & ACMII-Snowball-3 & 0.01  & 5.00E-03 & 0.5   & 64    & 87.16 & 1.01  & 16.29ms/3.62s \\
           \midrule
    \multirow{9}[0]{*}{\textbf{CiteSeer}} & ACM-SGC-1 & 0.05  & 0.00E+00 & 0.7   & 64    & 76.73 & 1.59  & 5.24ms/1.14s \\
          & ACM-SGC-2 & 0.1   & 0.00E+00 & 0.8   & 64    & 76.59 & 1.69  & 5.14ms/1.03s \\
          & ACM-GCN & 0.01  & 5.00E-06 & 0.3   & 64    & 77.32 & 1.7   & 8.89ms/1.79s \\
          & ACMII-GCN & 0.01  & 5.00E-05 & 0.5   & 64    & 77.15 & 1.45  & 8.95ms/1.80s \\
          & FAGCN & 0.02  & 5.00E-05 & 0.4   & 64    & 77.07 & 2.05  & 10.05ms/5.69s \\
          & ACM-Snowball-2 & 0.01  & 5.00E-05 & 0     & 64    & 76.41 & 1.38  & 12.87ms/2.59s \\
          & ACM-Snowball-3 & 0.01  & 5.00E-06 & 0.9   & 64    & 75.91 & 1.57  & 17.40ms/11.92s \\
          & ACMII-Snowball-2 & 0.01  & 5.00E-03 & 0.5   & 64    & 76.92 & 1.45  & 13.10ms/2.94s \\
          & ACMII-Snowball-3 & 0.1   & 5.00E-05 & 0.9   & 64    & 76.18 & 1.55  & 17.47ms/5.88s \\
           \midrule
    \multirow{9}[1]{*}{\textbf{PubMed}} & ACM-SGC-1 & 0.05  & 5.00E-06 & 0.4   & 64    & 88.49 & 0.51  & 5.77ms/3.65s \\
          & ACM-SGC-2 & 0.05  & 5.00E-06 & 0.3   & 64    & 89.01 & 0.6   & 8.50ms/5.18s \\
          & ACM-GCN & 0.01  & 5.00E-05 & 0.4   & 64    & 90    & 0.52  & 8.99ms/2.51s \\
          & ACMII-GCN & 0.01  & 1.00E-04 & 0.3   & 64    & 89.89 & 0.43  & 9.70ms/2.57s \\
          & FAGCN & 0.01  & 1.00E-04 & 0     & 64    & 88.09 & 1.38  & 10.30ms/8.75s \\
          & ACM-Snowball-2 & 0.01  & 1.00E-03 & 0.3   & 64    & 89.89 & 0.57  & 15.05ms/3.11s \\
          & ACM-Snowball-3 & 0.01  & 5.00E-03 & 0.1   & 64    & 89.81 & 0.43  & 20.51ms/4.63s \\
          & ACMII-Snowball-2 & 0.01  & 5.00E-04 & 0.4   & 64    & 89.84 & 0.48  & 15.10ms/3.2s \\
          & ACMII-Snowball-3 & 0.01  & 1.00E-03 & 0.4   & 64    & 89.73 & 0.52  & 20.46ms/4.32s \\
    \bottomrule
    \bottomrule
    \end{tabular}%
    }
  \label{tab:optimal_hyperparameters_fixed_splits}%
\end{table}%
 
\begin{table}[htbp]
  \centering
  \scalebox{.75}{
  \caption{Optimal hyperparameters for ACM(II)-GCN+ and ACM(II)-GCN++ on fixed 48\%/32\%/20\% splits}
    \begin{tabular}{c|c|cccccccc}
    \toprule
    \toprule
    Datasets & Models\textbackslash{}Hyperparameters & lr    & weight\_decay & dropout & hidden & with A & results & std   & average epoch time/average total time \\
    \midrule
    \multirow{4}[0]{*}{{Cornell}} & ACM-GCN+ & 0.05  & 1.00E-03 & 0.1   & 64    & N     & 85.68 & 4.84  & 10.86ms/2.28s \\
          & ACMII-GCN+ & 0.05  & 5.00E-03 & 0     & 64    & Y     & 85.41 & 5.3   & 14.42ms/2.97s \\
          & ACM-GCN++ & 0.01  & 5.00E-04 & 0.1   & 64    & N     & 85.68 & 5.8   & 14.15ms/3.17s \\
          & ACMII-GCN++ & 0.01  & 5.00E-03 & 0.3   & 64    & N     & 86.49 & 6.73  & 14.11ms/3.19s \\
          \midrule
    \multirow{4}[0]{*}{{Wisconsin}} & ACM-GCN+ & 0.01  & 1.00E-03 & 0.1   & 64    & Y     & 88.43 & 2.39  & 14.50ms/3.18s \\
          & ACMII-GCN+ & 0.01  & 5.00E-03 & 0.3   & 64    & Y     & 88.04 & 3.66  & 17.71ms/3.75s \\
          & ACM-GCN++ & 0.05  & 5.00E-03 & 0.1   & 64    & Y     & 88.24 & 3.16  & 20.61ms/4.29s \\
          & ACMII-GCN++ & 0.01  & 5.00E-03 & 0.2   & 64    & Y     & 88.43 & 3.66  & 18.28ms/3.75s \\
          \midrule
    \multirow{4}[0]{*}{{Texas}} & ACM-GCN+ & 0.01  & 5.00E-04 & 0.2   & 64    & Y     & 88.38 & 3.64  & 22.63ms/4.63s \\
          & ACMII-GCN+ & 0.05  & 1.00E-02 & 0.4   & 64    & Y     & 88.11 & 3.24  & 16.92ms/3.44s \\
          & ACM-GCN++ & 0.01  & 5.00E-03 & 0.3   & 64    & Y     & 88.38 & 3.43  & 20.69ms/4.25s \\
          & ACMII-GCN++ & 0.01  & 5.00E-03 & 0.6   & 64    & Y     & 88.38 & 3.43  & 18.58ms/3.84s \\
          \midrule
    \multirow{4}[0]{*}{{Film}} & ACM-GCN+ & 0.05  & 5.00E-03 & 0     & 64    & N     & 36.13 & 1.19  & 18.33ms/3.68s \\
          & ACMII-GCN+ & 0.05  & 5.00E-03 & 0     & 64    & N     & 35.95 & 1.33  &  19.07ms/3.83s \\
          & ACM-GCN++ & 0.01  & 5.00E-03 & 0     & 64    & N     & 37.31 & 1.09  & 18.57ms/3.73s \\
          & ACMII-GCN++ & 0.01  & 5.00E-03 & 0     & 64    & N     & 36.68 & 1.35  & 15.79ms/3.17s \\
          \midrule
    \multirow{4}[0]{*}{{Chameleon}} & ACM-GCN+ & 0.05  & 1.00E-04 & 0.7   & 64    & Y     & 74.23 & 2.25  & 25.31ms/5.14s \\
          & ACMII-GCN+ & 0.05  & 1.00E-04 & 0.7   & 64    & Y     & 74.3  & 2.03  & 25.04ms/5.04s \\
          & ACM-GCN++ & 0.002 & 5.00E-04 & 0.8   & 64    & Y     & 74.3  & 2.23  & 19.44ms/8.58s \\
          & ACMII-GCN++ & 0.01  & 1.00E-04 & 0.8   & 64    & Y     & 74.45 & 1.34  & 21.24ms/4.92s \\
          \midrule
    \multirow{4}[0]{*}{{Squirrel}} & ACM-GCN+ & 0.002 & 1.00E-04 & 0.6   & 64    & Y     & 66.06 & 2.16  & 36.96ms/7.82s \\
          & ACMII-GCN+ & 0.01  & 5.00E-04 & 0.8   & 64    & Y     & 65.95 & 1.74  & 35.56ms/9.18s \\
          & ACM-GCN++ & 0.01  & 1.00E-04 & 0.8   & 64    & Y     & 66.45 & 1.83  & 26.34ms/6.20s \\
          & ACMII-GCN++ & 0.002 & 5.00E-04 & 0.8   & 64    & Y     & 66.75 & 1.82  &  24.55ms/10.49s \\
          \midrule
    \multirow{4}[0]{*}{{Cora}} & ACM-GCN+ & 0.002 & 0.00E+00 & 0.6   & 64    & N     & 88.05 & 0.99  & 15.21ms/5.00s \\
          & ACMII-GCN+ & 0.002 & 5.00E-05 & 0.7   & 64    & Y     & 88.19 & 1.17  & 13.74ms/5.67s \\
          & ACM-GCN++ & 0.002 & 5.00E-06 & 0.7   & 64    & N     & 88.11 & 0.96  & 14.59ms/5.05s \\
          & ACMII-GCN++ & 0.002 & 5.00E-05 & 0.7   & 64    & N     & 88.25 & 0.96  & 15.75ms/5.87s \\
          \midrule
    \multirow{4}[0]{*}{{CiteSeer}} & ACM-GCN+ & 0.01  & 5.00E-05 & 0.3   & 64    & N     & 77.67 & 1.19  & 17.36ms/3.49s \\
          & ACMII-GCN+ & 0.01  & 5.00E-03 & 0.2   & 64    & Y     & 77.2  & 1.61  & 22.99ms/4.74s \\
          & ACM-GCN++ & 0.002 & 5.00E-06 & 0.6   & 64    & N     & 77.46 & 1.65  & 14.51ms/3.88s \\
          & ACMII-GCN++ & 0.01  & 5.00E-05 & 0.6   & 64    & N     & 77.12 & 1.58  & 18.69ms/3.76s \\
          \midrule
    \multirow{4}[1]{*}{{PubMed}} & ACM-GCN+ & 0.05  & 5.00E-05 & 0.3   & 64    & N     & 89.82 & 0.41  & 24.63ms/4.95s \\
          & ACMII-GCN+ & 0.01  & 1.00E-04 & 0.3   & 64    & N     & 89.78 & 0.49  & 25.10ms/5.61s \\
          & ACM-GCN++ & 0.01  & 5.00E-05 & 0.3   & 64    & N     & 89.65 & 0.58  & 18.36ms/3.76s \\
          & ACMII-GCN++ & 0.002 & 5.00E-06 & 0.4   & 64    & N     & 89.71 & 0.48  & 16.98ms/9.44s \\
    \bottomrule
    \bottomrule
    \end{tabular}%
  \label{tab:hyperparameters_acmgcn+_acmgcn++}%
  }
\end{table}%

\clearpage

\section{Experimental Setup and Further Discussion on Synthetic Graphs}
\label{appendix:synthetic_setup_discussion}

\subsection{Detailed Description of Data Generation Process}
\begin{itemize}
    \item For each node $v$, we first randomly generate its degree $d_v$.
    \item Given $d_v$, for any $h$, we sample $hd_v$ intra-class edges and $(1-h)d_v$ inter-class edges.
\end{itemize}

More specifically in our synthetic experiments, for a given $h$,
\begin{itemize}
    \item we generate node degree $d_v$ for nodes in each class from multinomial distribution with $\texttt{numpy.random.multinomial(800/h, numpy.ones(400)/400, size=1)[0]}$. 
    \item For a sampled $d_v$, we generate intra-class edges from (does not include self-loop) $\texttt{numpy.random.multinomial(h$d_v$, numpy.ones(399)/399, size=1)[0]}$  and inter-class edges from $\texttt{numpy.random.multinomial((1-h) $d_v$, numpy.ones(1600)/1600,}$  \texttt{size=1)[0]}.
\end{itemize}

For each generated graph, we calculate their $H_\text{node},H_\text{class}, H_\text{agg}^M$. Then, we reorder the value of the metrics in ascend order for x-axis and plot the corresponding test accuracy.

Here is a simplified example of how we draw Figure \ref{fig:comparison_homophily_metrics}. Suppose we generate 3 graphs with $H_\text{edge}=0.1,0.5,0.9$, the test accuracy of GCN on these 3 synthetic graphs are $0.8,0.5,0.9$. For the generated graphs, we calculate their $H_\text{agg}^M$, and suppose we get $H_\text{agg}^M=0.7,0.4,0.8$. Then we will draw the performance of GCN under $H_\text{agg}^M$ with ascend x-axis order $[0.4,0.7,0.8]$ and the corresponding reordered y-axis is $[0.5,0.8,0.9]$. Other figures are drawn with the same process.
\subsection{Model Comparison on Synthetic Graphs}
\label{appendix:model_comparison_synthetic_datasets}

\begin{figure}[H]
    \centering
     {
    \subfloat[ \texttt{syn-Cora}]{
     \captionsetup{justification = centering}
     \includegraphics[width=0.32\textwidth]{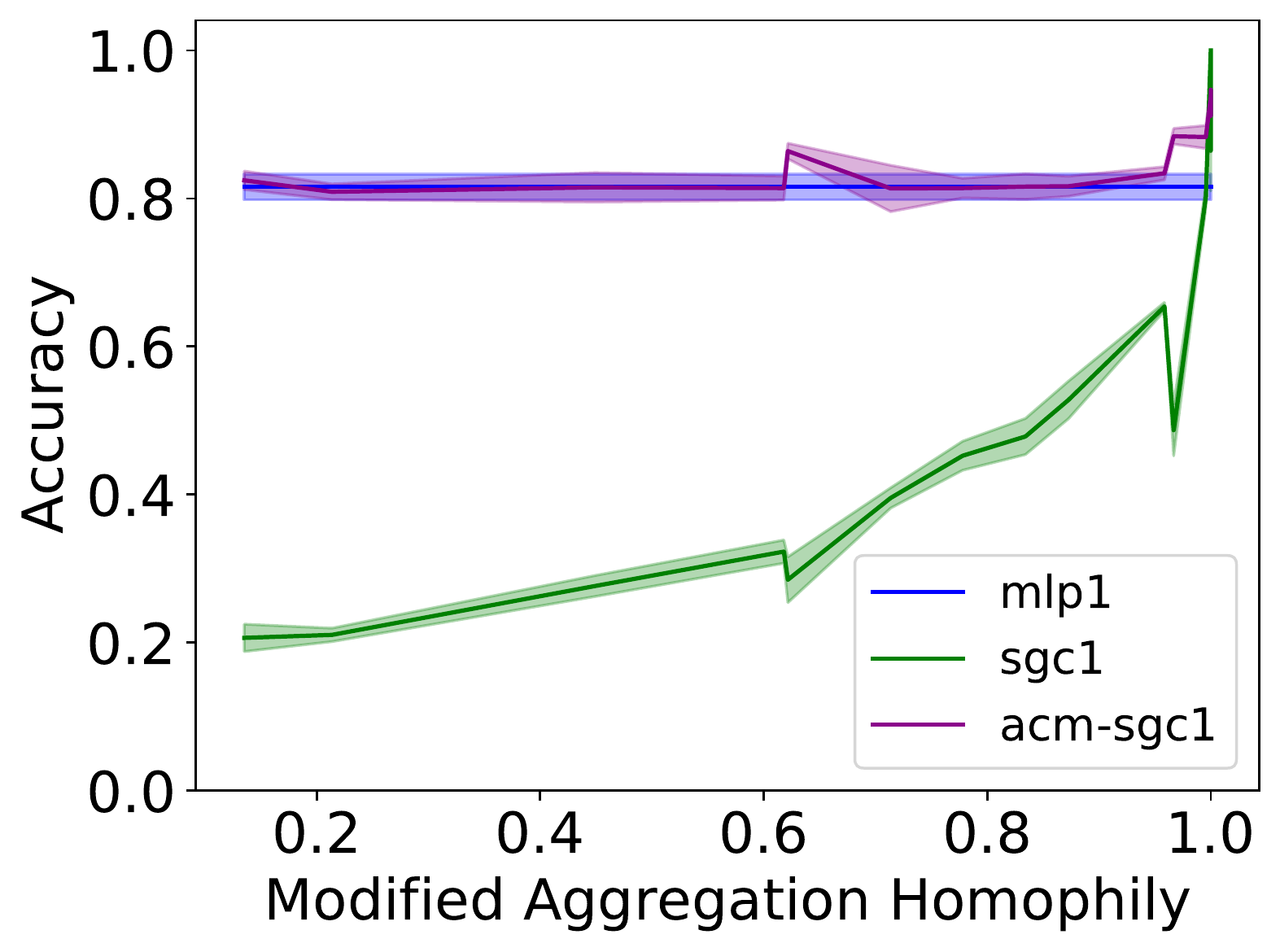}
     } 
     \subfloat[ \texttt{syn-CiteSeer}]{
     \captionsetup{justification = centering}
     \includegraphics[width=0.32\textwidth]{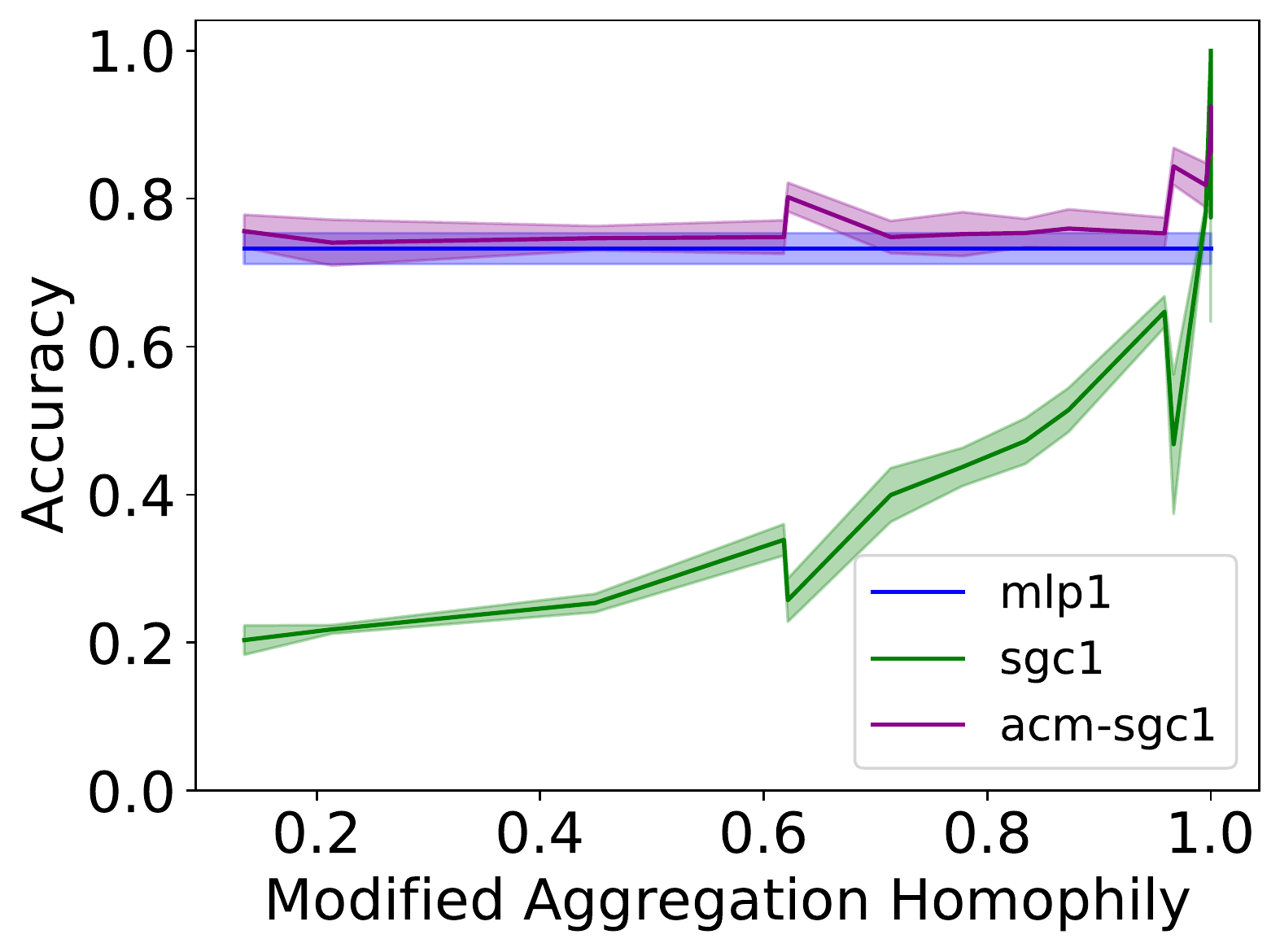}
     } 
     \subfloat[ \texttt{syn-PubMed}]{
     \captionsetup{justification = centering}
     \includegraphics[width=0.32\textwidth]{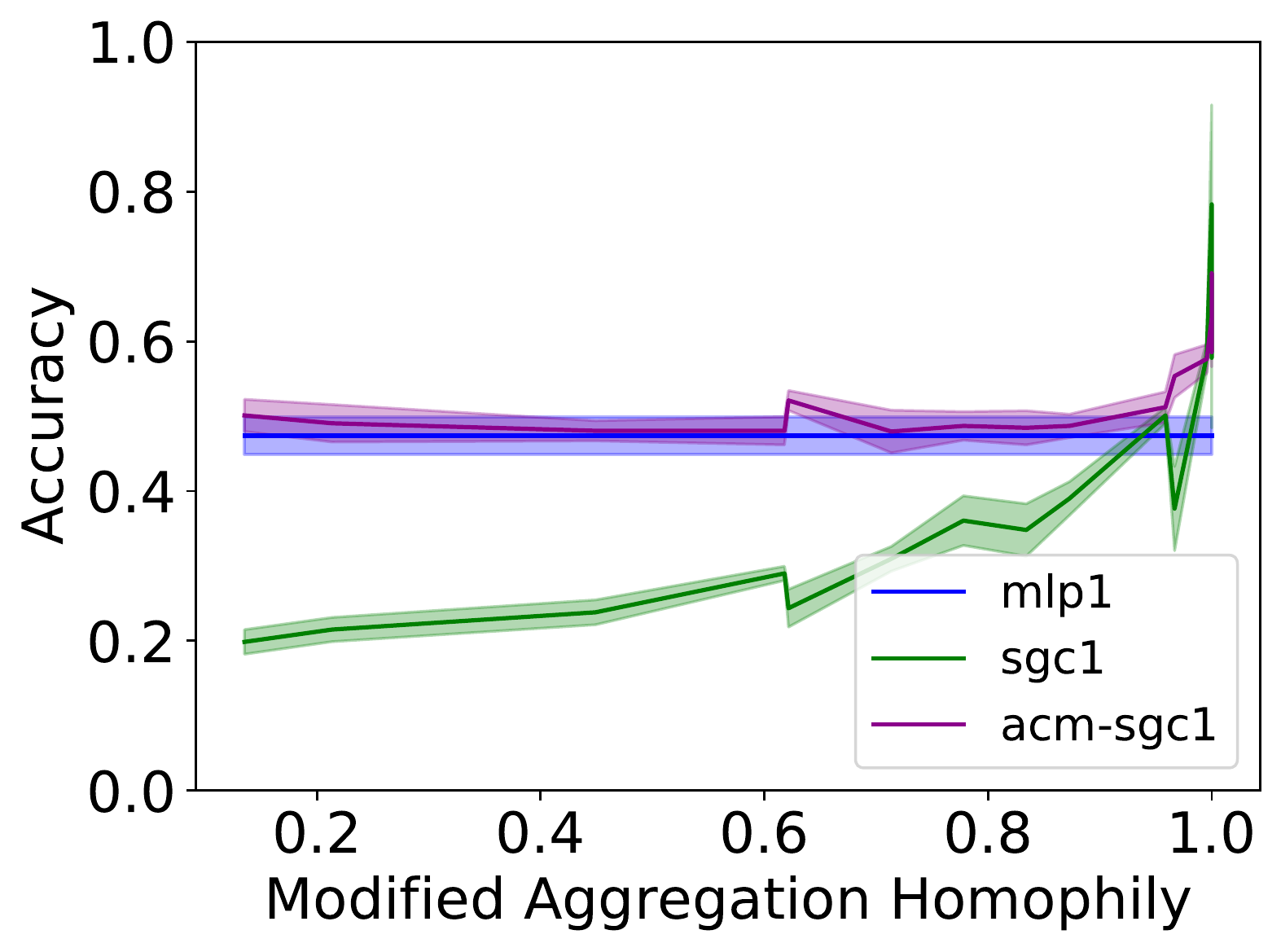}
     } \\
     \subfloat[ \texttt{syn-Chameleon}]{
     \captionsetup{justification = centering}
     \includegraphics[width=0.32\textwidth]{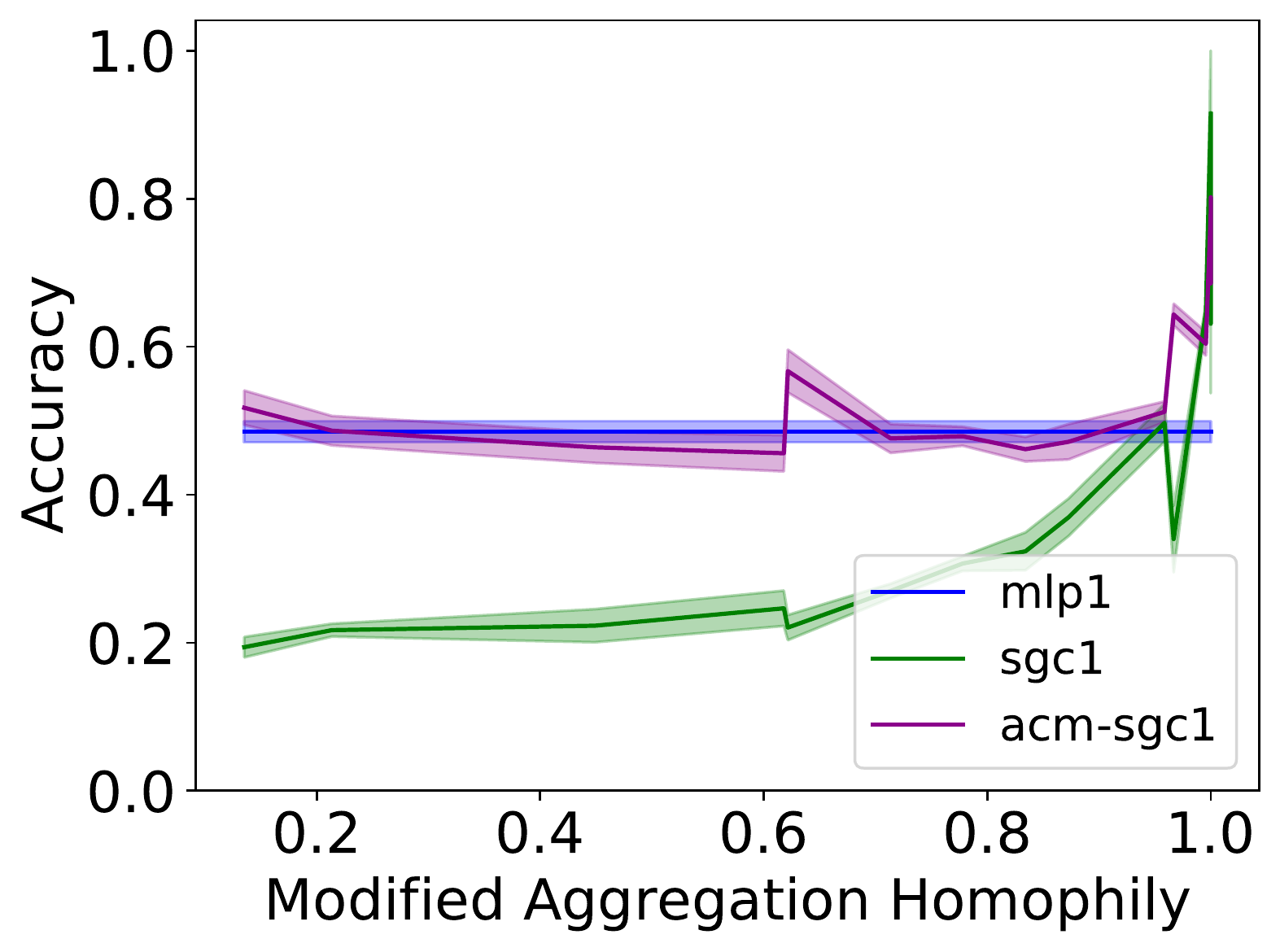}
     } 
     \subfloat[ \texttt{syn-Squirrel}]{
     \captionsetup{justification = centering}
     \includegraphics[width=0.32\textwidth]{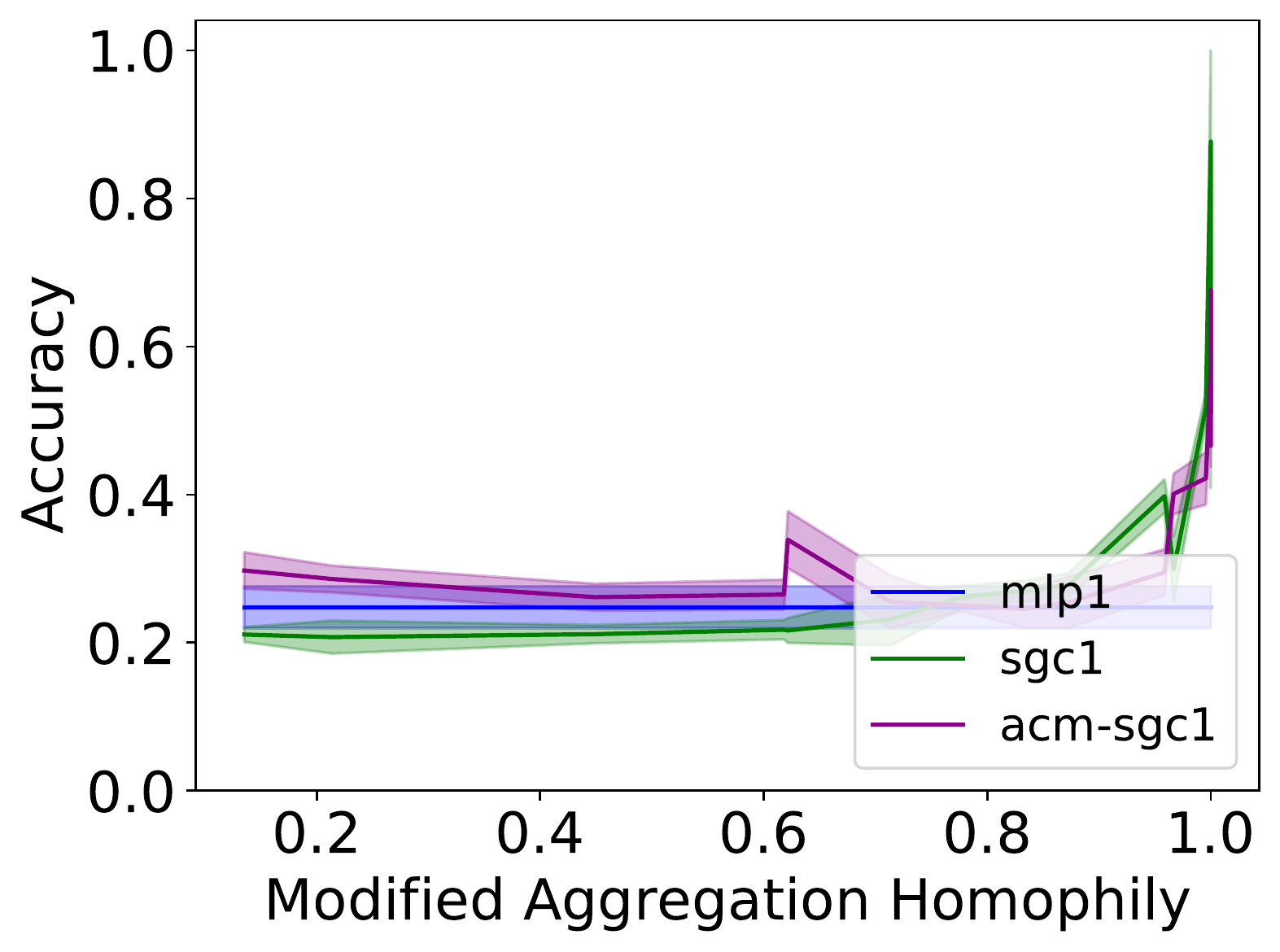}
     } 
     \subfloat[ \texttt{syn-Film}]{
     \captionsetup{justification = centering}
     \includegraphics[width=0.32\textwidth]{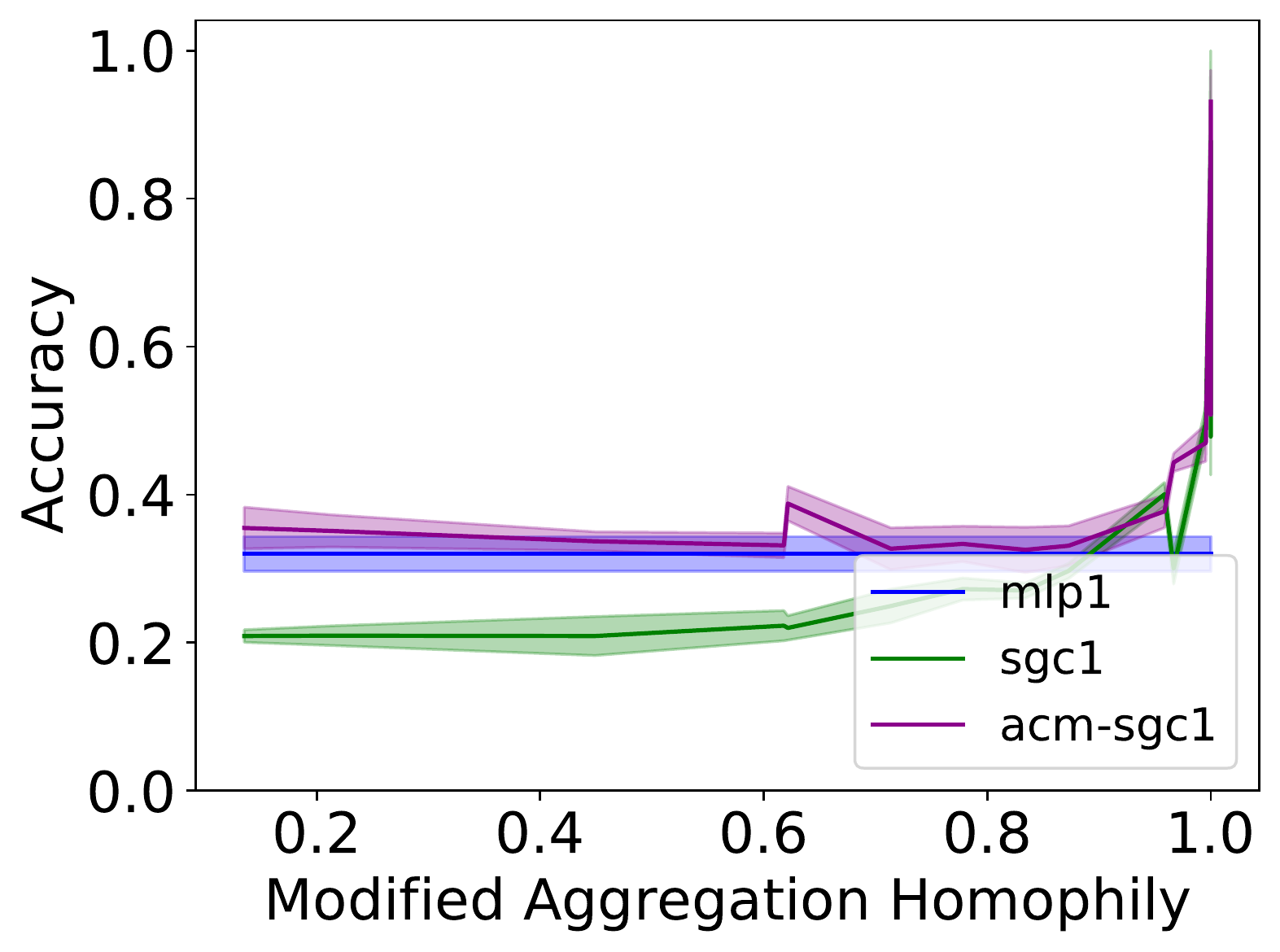}
     } 
     }
     \caption{Comparison of test accuracy (mean $\pm$ std) of MLP-1, SGC-1 and ACM-SGC-1 on synthetic datasets}
     \label{fig:sgc_acmsgc_synthetic_comparison}
\end{figure}

\begin{figure}[H]
    \centering
     { 
     \subfloat[\texttt{syn-Cora}]{
     \captionsetup{justification = centering}
     \includegraphics[width=0.32\textwidth]{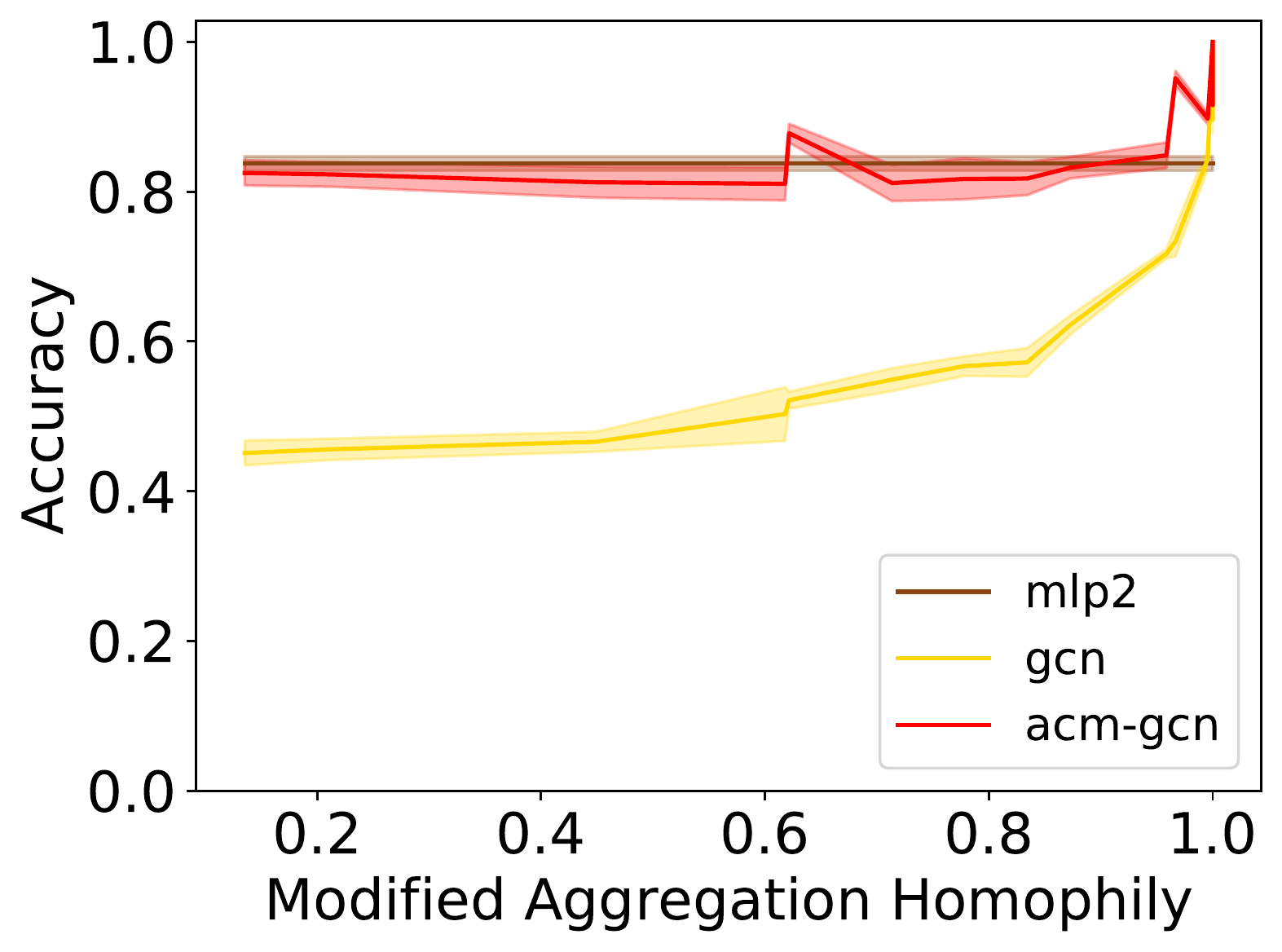}
     }
     \subfloat[\texttt{syn-CiteSeer}]{
     \captionsetup{justification = centering}
     \includegraphics[width=0.32\textwidth]{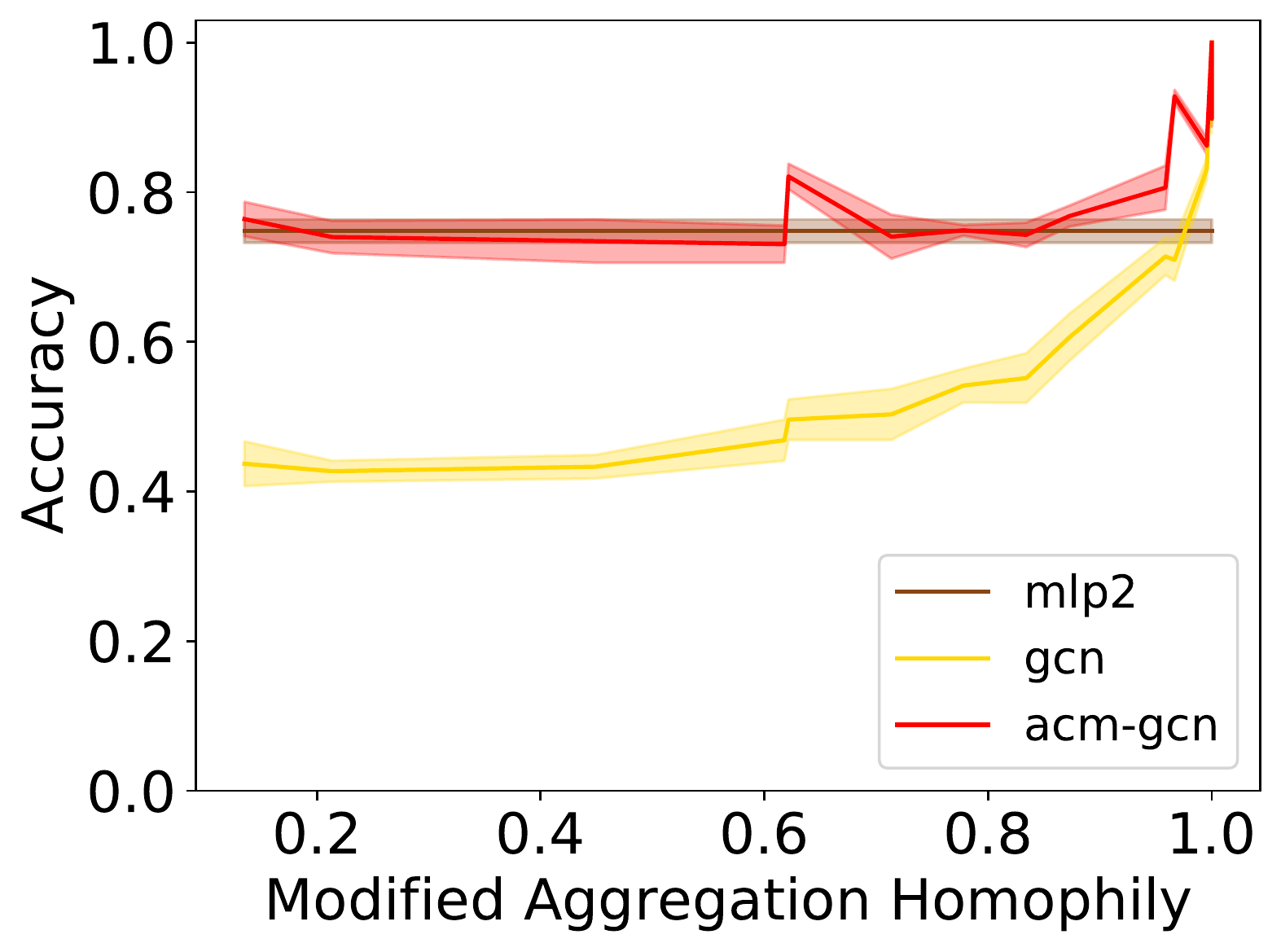}
     }
     \subfloat[\texttt{syn-PubMed}]{
     \captionsetup{justification = centering}
     \includegraphics[width=0.32\textwidth]{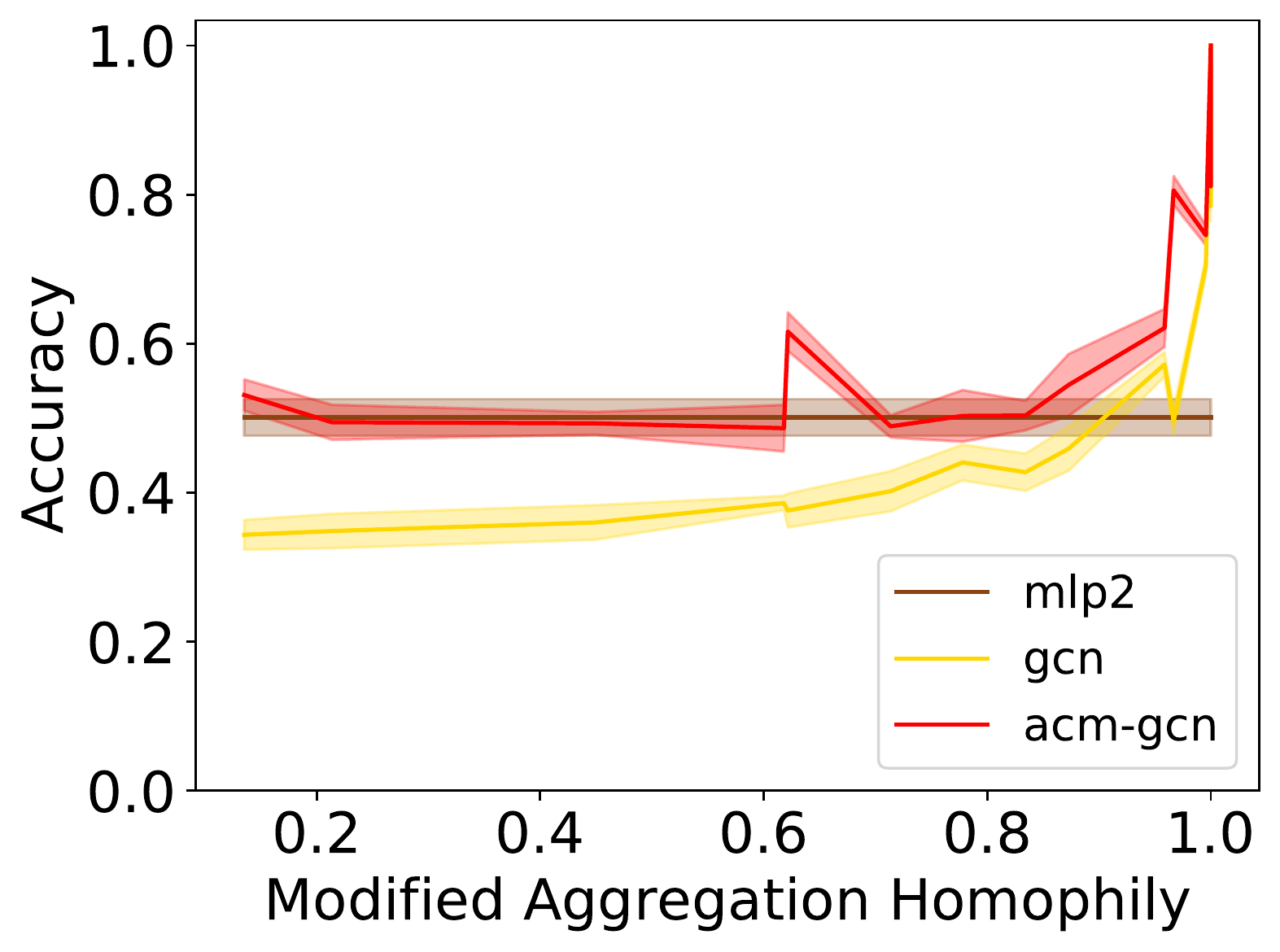}
     } \\
     \subfloat[\texttt{syn-Chameleon}]{
     \captionsetup{justification = centering}
     \includegraphics[width=0.32\textwidth]{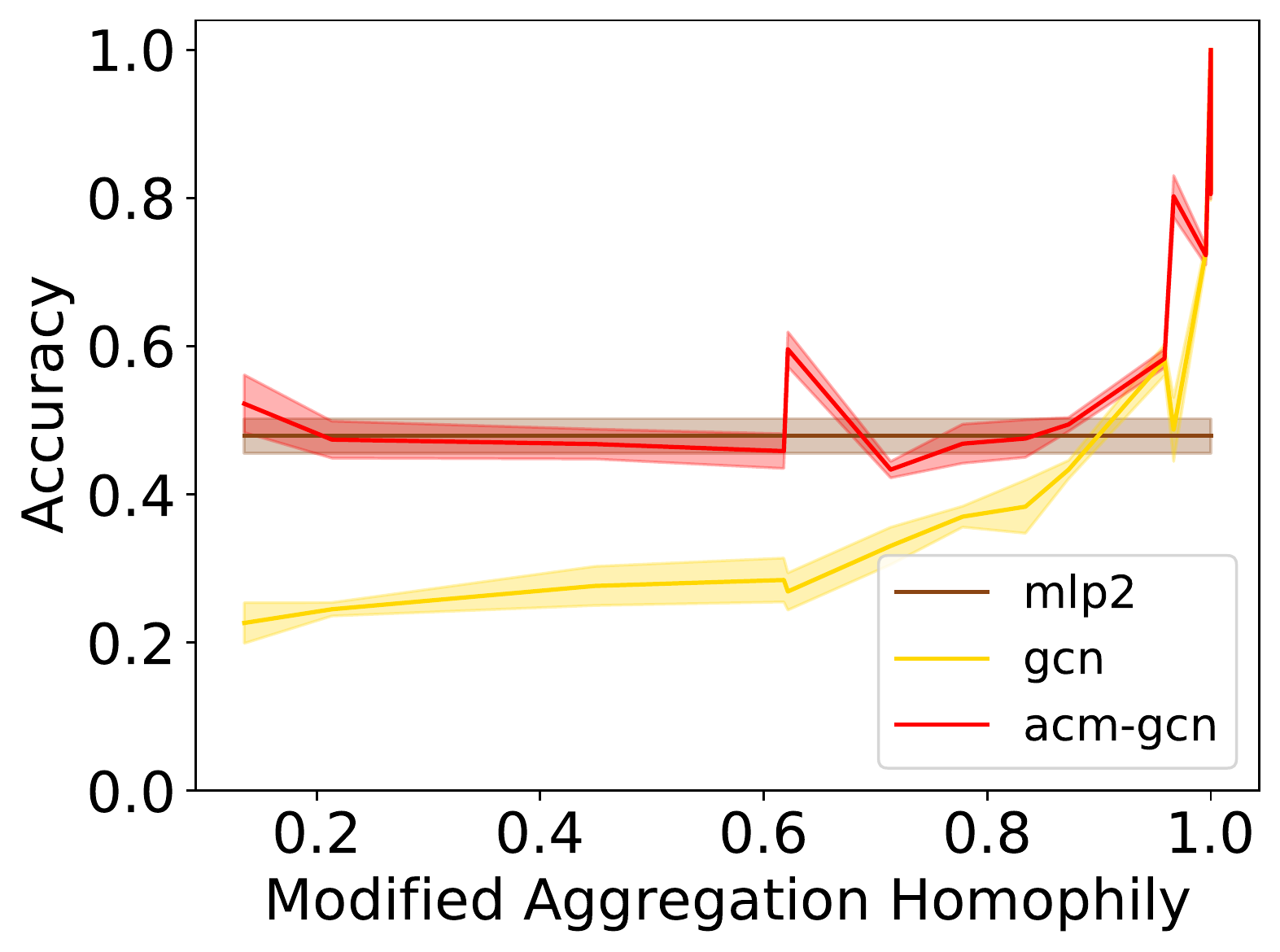}
     } 
     \subfloat[\texttt{syn-Squirrel}]{
     \captionsetup{justification = centering}
     \includegraphics[width=0.32\textwidth]{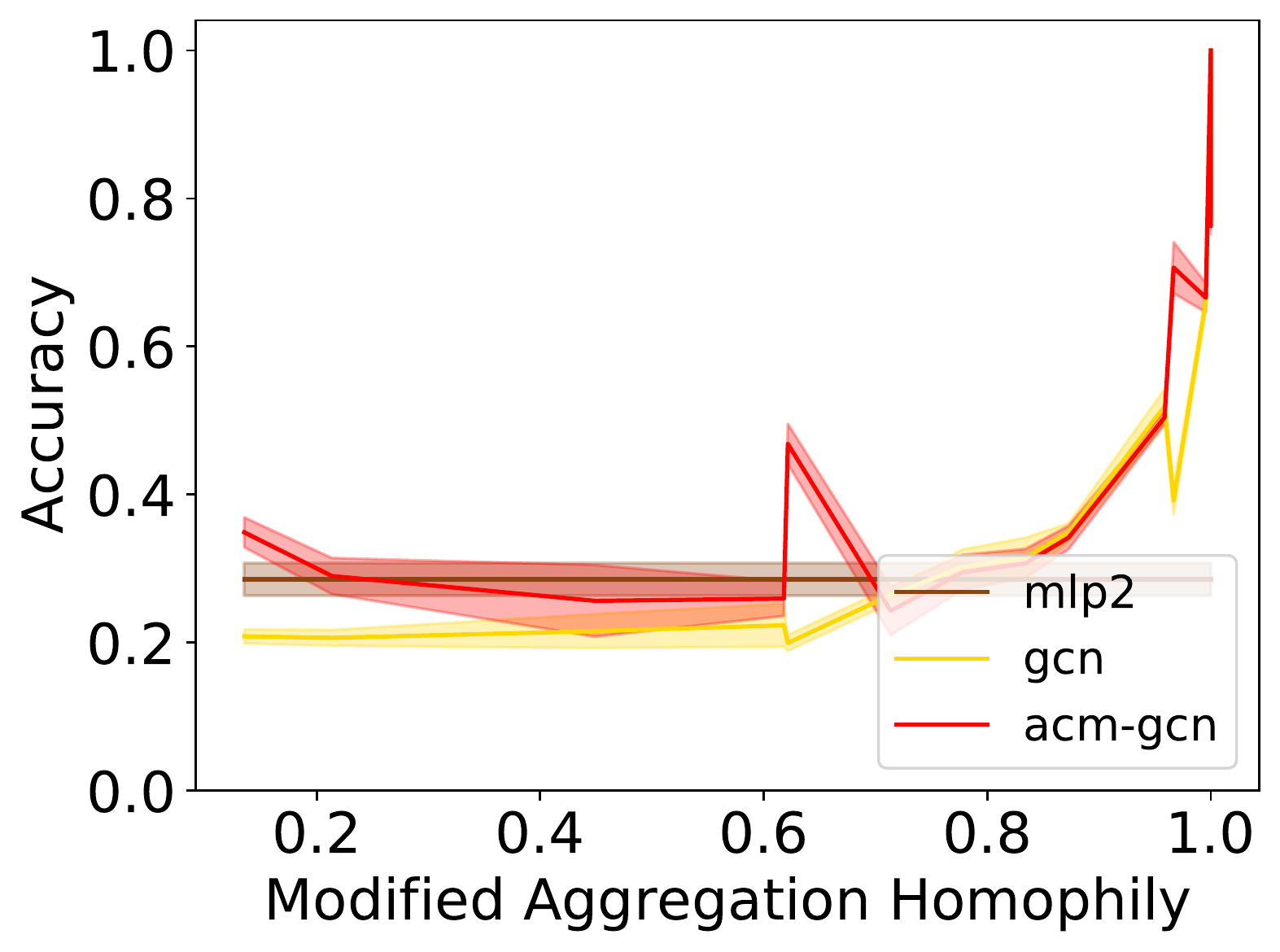}
     }
     \subfloat[\texttt{syn-Film}]{
     \captionsetup{justification = centering}
     \includegraphics[width=0.32\textwidth]{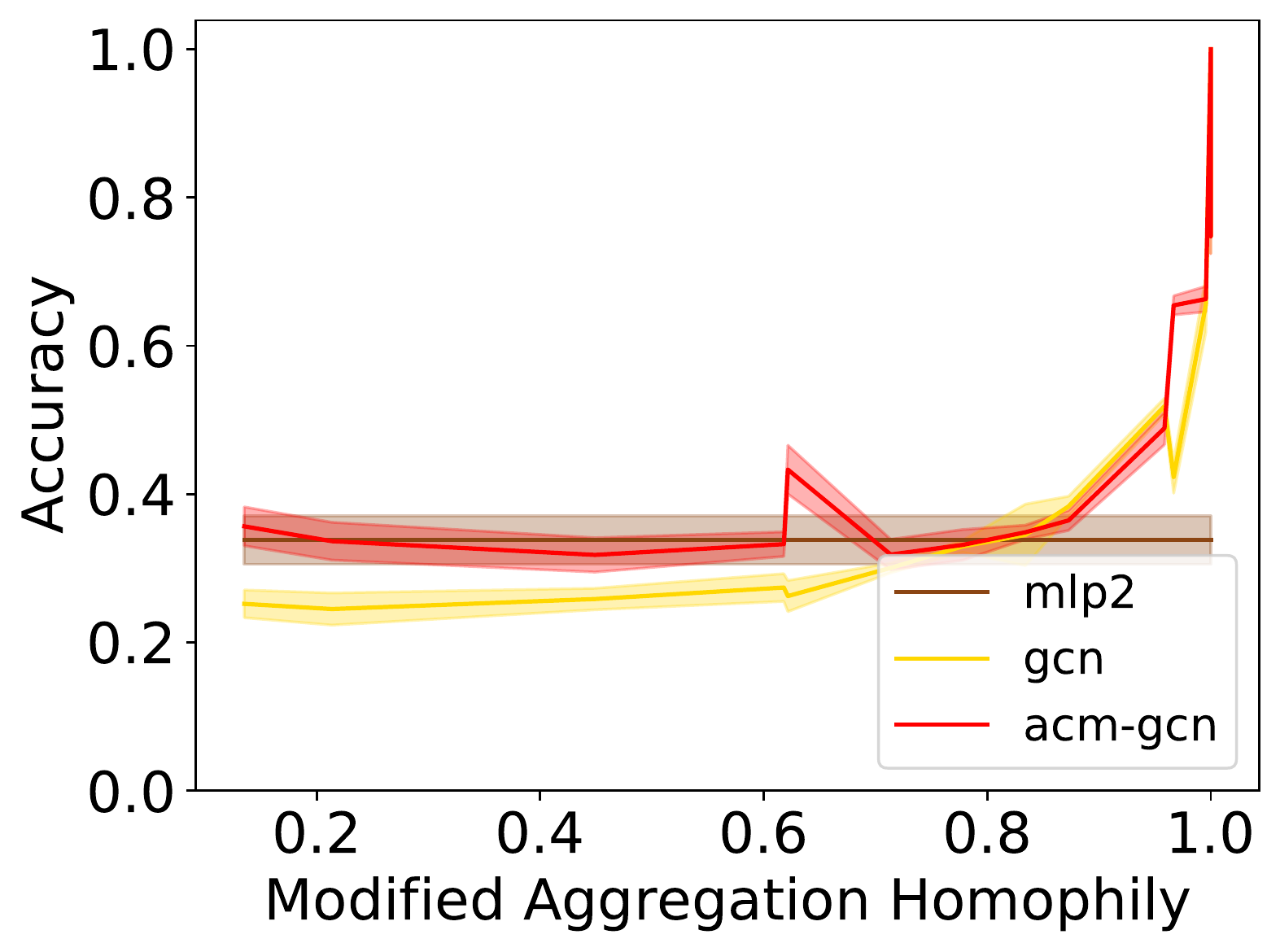}
     } 
     }
     \caption{Comparison of test accuracy (mean $\pm$ std) of MLP-2, GCN and ACM-GCN on synthetic datasets}
     \label{fig:gcn_acmgcn_synthetic_comparison}
\end{figure}

In order to separate the effects of nonlinearity and graph structure, we compare SGC with 1 hop (sgc-1) with MLP-1 (linear model). For GCN which includes nonlinearity, we use MLP-2 as its corresponding graph-agnostic baseline model. We train the above GNN models, graph-agnostic baseline models and ACM-GNN models on all synthetic datasets and plot the mean test accuracy with standard deviation on each dataset. From Figure \ref{fig:sgc_acmsgc_synthetic_comparison} and Figure \ref{fig:gcn_acmgcn_synthetic_comparison}, we can see that on each $H_\text{agg}^M(\mathcal{G})$ level, ACM-GNNs will not underperform baseline GNNs and the graph-agnostic models. But when $H_\text{agg}^M(\mathcal{G})$ is small, baseline GNNs will be outperformed by graph-agnostic models by a large margin. This demonstrate that the ACM framework can help GNNs to perform well on harmful graphs while keep competitive on less harmful graphs.

\subsection{Further Discussion of Aggregation Homophily on Regular Graphs}

\begin{figure}[h]
     {
     \captionsetup{justification = centering}
     \includegraphics[height=0.20\textheight]{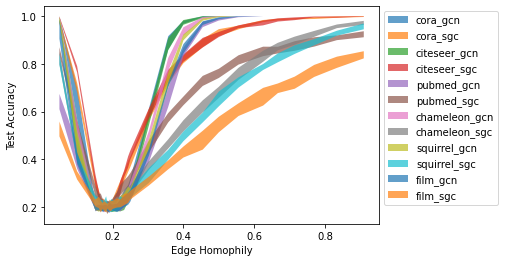}
     }
     \caption{Synthetic experiments for edge homophily on regular graphs.}
     \label{fig:synthetic_regular_graphs}
\end{figure}

We notice that in Figure \ref{fig:comparison_homophily_metrics}(a), the performance of SGC-1 and GCN both have a turning point, \ie{} when $H_{\text{edge}}(\mathcal{G})$ is smaller than a certain value, the performance will get better instead of getting worse. With some extra restriction on node degree in data generation process, we find that this interesting phenomenon can be theoretically explained by the following proposition 1 based on our proposed similarity matrix which can verify the usefulness of $H_{\text{agg}}^M(\mathcal{G})$. We first generate regular graphs ,\ie{} each node has the same degree, as follows,

\paragraph{Generate Synthetic Regular Graphs} We first generate 180 graphs in total with 18 edge homophily levels varied from 0.05 to 0.9, each corresponding to $10$ graphs. For every generated graph, we have 5 classes with 400 nodes in each class. For each node, we randomly generate 10 intra-class edges and [$\frac{10}{H_\text{edge}(\mathcal{G})} -10$] inter-class edges. The features of nodes in each class are sampled from node features in the corresponding class of the base dataset. Nodes are randomly split into 60\%/20\%/20\% for train/validation/test. We train 1-hop SGC (\textit{sgc-1}) \cite{wu2019simplifying} and GCN \cite{kipf2016classification} on synthetic data  
(see Appendix \ref{appendix:hyperparameter_space_synthetic_graphs} for hyperparameter searching range). For each value of $H_\text{edge}(\mathcal{G})$, we take the average test accuracy and standard deviation of runs over 10 generated graphs.  We plot the performance curves in Figure \ref{fig:synthetic_regular_graphs}.

From Figure \ref{fig:synthetic_regular_graphs} we can see that the turning point is a bit less than 0.2. We derive the following proposition for $d$-regular graph to explain and predict it.

\begin{proposition} 1
(See Appendix \ref{appendix:proof_theorem1} for proof). Suppose there are $C$ classes in the graph $\cal G$ and $\cal G$ is a $d$-regular graph (each node has $d$ neighbors). Given $d$, edges for each node are \iid generated, such that each edge of any node has probability $h$ to connect with nodes in the same class and probability $1-h$ to connect with nodes in different classes. Let the aggregation operator $\hat{A} = \hat{A}_\text{rw}$. Then, for nodes $v$, $u_1$ and $u_2$, where $Z_{u_1,:}=Z_{v,:}$ and $Z_{u_2,:} \neq Z_{v,:}$, we have
\begin{equation}
\label{eq:theorem1_expectation_differences}
    g(h)\equiv 
    \mathbb{E}\left(S(\hat{A},Z)_{v,u_1}  \right) 
    - \mathbb{E}\left(S(\hat{A},Z)_{v,u_2}  \right)
   = \left(\frac{(C-1)(hd+1)-(1-h)d}{(C-1)(d+1)} \right)^2
\end{equation}
and the minimum of $g(h)$ is reached at 
$$
h=\frac{d+1-C}{Cd} = \frac{d_{\text{intra}}/h + 1 -C}{C (d_{\text{intra}}/h)}  \Rightarrow h=\frac{d_{\text{intra}}}{C d_{\text{intra}} +C-1}
$$
where $d_{\text{intra}}=dh$, which is the expected number of neighbors of a node that have the same label as the node.
\end{proposition}

The value of $g(h)$ in \eqref{eq:theorem1_expectation_differences} is the expected differences of the similarity values between nodes in the same class as $v$ and nodes in other classes. $g(h)$ is strongly related to the definition of aggregation homophily and its minimum potentially implies the turning point of performance curves. In the synthetic experiments, we have $d_{\text{intra}} = 10, C=5$ and the minimum of $g(h)$ is reached at $h = 5/27 \approx 0.1852$, which corresponds to the lowest point in the performance curve in Figure \ref{fig:synthetic_regular_graphs}. In other words, the $H_\text{edge}(\mathcal{G})$ where SGC-1 and GCN perform worst is where $g(h)$ gets the smallest value, instead of the point with the smallest edge homophily value, \ie{} $H_\text{edge}(\mathcal{G})=0$. This reveals the advantage of $H_{\text{agg}}(\mathcal{G})$ over $H_\text{edge}(\mathcal{G})$ by taking use of the similarity matrix.

\section{Details of Gradient Calculation in \eqref{eq:gradient_descent_update}}
\label{appendix:details_of_nll_loss_explanation}
\subsection{Derivation in Matrix Form} This derivation is similar to \cite{luan2020training}.

In output layer, we have
\begin{align*}
Y & = \text{softmax} (\hat{A} X W ) \equiv  \text{softmax} (Y') = \left(\exp(Y') \bm{1}_C \bm{1}_C^T \right)^{-1}  \odot \exp(Y') > 0 \\   
\mathcal{L} & = -\trace(Z^T \log Y)    
\end{align*}
where $\bm{1}_C \in \mathcal{R}^{C\times 1}$, $(\cdot)^{-1}$ is point-wise inverse function and each element of $Y$ is positive. Then
\begin{align*}
d \mathcal{L} = -\trace\left(Z^T ((Y)^{-1} \odot d Y) \right) = -\trace\left(Z^T \left( \left(\text{softmax} (Y') \right)^{-1} \odot d\ \text{softmax} ( Y') \right) \right) 
\end{align*}
Note that
\begin{equation}
\begin{aligned}
d\ \text{softmax} (Y') 
= & - \left(\exp(Y') \bm{1}_C \bm{1}_C^T \right)^{-2} \odot [(\exp(Y') \odot d Y') \bm{1}_C \bm{1}_C^T] \odot \exp(Y')  \\ \nonumber
& + \left(\exp(Y') \bm{1}_C \bm{1}_C^T \right)^{-1}  \odot (\exp(Y') \odot d Y')\\ \nonumber
 = & - \text{softmax} (Y') \odot \left(\exp(Y') \bm{1}_C \bm{1}_C^T \right)^{-1} \odot [(\exp(Y') \odot d Y') \bm{1}_C \bm{1}_C^T]   \\
& + \text{softmax} (Y') \odot d Y'\\ \nonumber
 =  & \ \text{softmax} (Y') \odot \left(  - \left(\exp(Y') \bm{1}_C \bm{1}_C^T \right)^{-1} \odot \left[(\exp(Y') \odot d Y') \bm{1}_C \bm{1}_C^T \right] + d Y' \right) \nonumber
\end{aligned}
\end{equation}
Then,
\begin{equation}
\begin{aligned}
d \mathcal{L} =\ & -\trace\Bigg(Z^T \bigg((\text{softmax} (Y'))^{-1} \odot \bigg[ \text{softmax} (Y') \odot \bigg(  - \left(\exp(Y') \bm{1}_C \bm{1}_C^T \right)^{-1} \\
& \odot \left[(\exp(Y') \odot d Y') \bm{1}_C \bm{1}_C^T \right] + d Y' \bigg) \bigg] \bigg) \Bigg)\\ \nonumber
=\ & -\trace\left(Z^T \left(  - \left(\exp(Y') \bm{1}_C \bm{1}_C^T \right)^{-1} \odot \left[(\exp(Y') \odot d Y') \bm{1}_C \bm{1}_C^T \right] + d Y' \right) \right) \\ \nonumber
 =\ & \trace\left( \left( \left(Z \odot \left(\exp(Y') \bm{1}_C \bm{1}_C^T \right)^{-1} \right) \bm{1}_C \bm{1}_C^T \right)^T  \left[\exp(Y') \odot d Y'  \right] - Z^T d Y' \right)\\ \nonumber
=\ &  \trace\left( \left(  \exp(Y') \odot \left( \left(Z \odot \left(\exp(Y') \bm{1}_C \bm{1}_C^T \right)^{-1} \right) \bm{1}_C \bm{1}_C^T \right) \right)^T  d Y' - Z^T d Y' \right)\\ \nonumber
=\ &  \trace\left( \left(  \exp(Y') \odot \left(\exp(Y') \bm{1}_C \bm{1}_C^T \right)^{-1} \right)^T   d Y' - Z^T d Y' \right)\\ \nonumber
=\ & \trace\left( (\text{softmax}(Y') - Z)^T dY' \right)  \nonumber
\end{aligned}
\end{equation}
where the  4-th equation holds due to $\left(Z \odot \left(\exp(Y') \bm{1}_C \bm{1}_C^T \right)^{-1} \right) \bm{1}_C \bm{1}_C^T = \left(\exp(Y') \bm{1}_C \bm{1}_C^T \right)^{-1}$. Thus, we have
\begin{equation*}
    \frac{d \mathcal{L} }{d Y'} = \text{softmax}(Y') - Z = Y-Z
\end{equation*}
For $Y'$ and $W$, we have
\begin{align*}
   d Y' & = \hat{A} X d W \text{  and  } d\mathcal{L}= \text{trace}\left(\frac{d \mathcal{L} }{d Y'}^T d Y' \right) = \text{trace}\left(\frac{d \mathcal{L} }{d Y'}^T \hat{A}  X \ d W \right) = \text{trace}\left(\frac{d \mathcal{L} }{d W}^T \ d W \right) 
\end{align*}
To get $\frac{d \mathcal{L} }{d W}$ we have, 
\begin{equation}\label{eq5}
\begin{aligned}
   \frac{d \mathcal{L}}{d W} = X^T \hat{A}^T \frac{d \mathcal{L} }{ d Y'} = X^T \hat{A}^T (Y-Z)
\end{aligned}
\end{equation}

\subsection{Component-wise Derivation}
Denote $\tilde{X}=XW$. 
We rewrite $\cal L$ as follows:
\begin{align*}
    \label{eq:nll_loss_explanation_details}
    \mathcal{L} & = -\trace\left(Z^T \log \left((\exp({Y'}) \bm{1}_C \bm{1}_C^T )^{-1}  \odot \exp({Y'})\right) \right) \\
    & = -\trace\left(Z^T  \left(-\log(\exp({Y'}) \bm{1}_C \bm{1}_C^T ) + {Y'} \right) \right) \\
    & = -\trace\left(Z^T {Y'} \right) + \trace\left(Z^T  \log\left(\exp({Y'}) \bm{1}_C \bm{1}_C^T \right)  \right)\\
    & = -\trace\left(Z^T \hat{A} X W  \right) + \trace\left(Z^T  \log\left(\exp({Y'}) \bm{1}_C \bm{1}_C^T \right)  \right)\\
    & = -\trace\left(Z^T \hat{A} X W  \right) + \trace\left(\bm{1}_C^T  \log\left(\exp({Y'}) \bm{1}_C \right)  \right)\\
\end{align*}
Expand $\mathcal{L}$ component-wisely, we have
\begin{align*}
   \mathcal{L}  & = - \sum\limits_{i=1}^N \sum\limits_{j\in \mathcal{N}_i}  \hat{A}_{i,j} Z_{i,:}\tilde{X}_{j:}^T + \sum\limits_{i=1}^N \log \left( \sum\limits_{c=1}^C \exp(\sum\limits_{j\in \mathcal{N}_i} \hat{A}_{i,j} \tilde{X}_{j,c}) \right)\\
    & = - \sum\limits_{i=1}^N \log \left( \exp\left( \sum\limits_{c=1}^C \sum\limits_{j\in \mathcal{N}_i}  \hat{A}_{i,j} Z_{i,c}\tilde{X}_{j,c} \right) \right) + \sum\limits_{i=1}^N \log \left( \sum\limits_{c=1}^C \exp\left(\sum\limits_{j\in \mathcal{N}_i} \hat{A}_{i,j} \tilde{X}_{j,c} \right) \right)\\
    &= - \sum\limits_{i=1}^N \log \frac{\exp \left(\sum\limits_{c=1}^C \sum\limits_{j\in \mathcal{N}_i} \hat{A}_{i,j} Z_{i,c}\tilde{X}_{j,c}\right)}{\left(\sum\limits_{c=1}^C \exp(\sum\limits_{j\in \mathcal{N}_i} \hat{A}_{i,j} \tilde{X}_{j,c}) \right)}
\end{align*}
Note that $\sum\limits_{c=1}^C Z_{j,c} = 1$ for any $j$. 
Consider the derivation of $\mathcal{L}$ over $\tilde{X}_{j',c'}$:
\begin{equation}
\begin{split}
    \allowdisplaybreaks
    &\frac{d \mathcal{L}}{d \tilde{X}_{j',c'}} 
     =  - \sum\limits_{i=1}^N \frac{\sum\limits_{c=1}^C \exp(\sum\limits_{j\in \mathcal{N}_i} \hat{A}_{i,j} \tilde{X}_{j,c}) }{\exp \left(\sum\limits_{c=1}^C \sum\limits_{j\in \mathcal{N}_i} \hat{A}_{i,j} Z_{i,c}\tilde{X}_{j,c}\right) } \\ 
& \times \left( \frac{ \left( \hat{A}_{i,j'} Z_{i,c'}  \right) \exp \left(\sum\limits_{c=1}^C \sum\limits_{j\in \mathcal{N}_i} \hat{A}_{i,j} Z_{i,c}\tilde{X}_{j,c}\right) \left(\sum\limits_{c=1}^C \exp(\sum\limits_{j\in \mathcal{N}_i} \hat{A}_{i,j} \tilde{X}_{j,c}) \right) }{ \left(\sum\limits_{c=1}^C \exp(\sum\limits_{j\in \mathcal{N}_i} \hat{A}_{i,j} \tilde{X}_{j,c}) \right)^2} \right.\\
    &\left. - \frac{\left( \hat{A}_{i,j'} \right) \exp \left(\sum\limits_{c=1}^C \sum\limits_{j\in \mathcal{N}_i} \hat{A}_{i,j} Z_{i,c}\tilde{X}_{j,c}\right) \left( \exp(\sum\limits_{j\in \mathcal{N}_i} \hat{A}_{i,j} \tilde{X}_{j,c'}) \right) }{ \left(\sum\limits_{c=1}^C \exp(\sum\limits_{j\in \mathcal{N}_i} \hat{A}_{i,j} \tilde{X}_{j,c}) \right)^2} \right) \\
     =&  - \sum\limits_{i=1}^N \left( \frac{ \left( \hat{A}_{i,j'} Z_{i,c'}  \right) \left(\sum\limits_{c=1}^C \exp(\sum\limits_{j\in \mathcal{N}_i} \hat{A}_{i,j} \tilde{X}_{j,c}) \right) - \left( \hat{A}_{i,j'} \right) \left( \exp(\sum\limits_{j\in \mathcal{N}_i} \hat{A}_{i,j} \tilde{X}_{j,c'}) \right)}{ \left(\sum\limits_{c=1}^C \exp(\sum\limits_{j\in \mathcal{N}_i} \hat{A}_{i,j} \tilde{X}_{j,c}) \right)} \right)\\
     = & - \sum\limits_{i=1}^N \left( \hat{A}_{i,j'}  \frac{\left(\sum\limits_{c=1, c\neq c'}^C (Z_{i,c'}) \exp(\sum\limits_{j\in \mathcal{N}_i} \hat{A}_{i,j} \tilde{X}_{j,c}) \right) +  \left( Z_{i,c'}-1 \right) \left( \exp(\sum\limits_{j\in \mathcal{N}_i} \hat{A}_{i,j} \tilde{X}_{j,c'}) \right)}{ \left(\sum\limits_{c=1}^C \exp(\sum\limits_{j\in \mathcal{N}_i} \hat{A}_{i,j} \tilde{X}_{j,c}) \right)} \right)\\
     =& - \sum\limits_{i=1}^N \hat{A}_{i,j'} \left( Z_{i,c'} \hat{P}(Y_i \neq c') + (Z_{i,c'} - 1) \hat{P} (Y_i = c') \right)\\
     = & - \sum\limits_{i=1}^N \hat{A}_{i,j'} \left( Z_{i,c'} - \hat{P} (Y_i = c') \right)
    \end{split}
\end{equation}
Writing the above in matrix form, we have
\begin{equation}
    \frac{d \mathcal{L}}{d \tilde{X} } = \hat{A}(Z- Y), \  \frac{d \mathcal{L}}{d \tilde{W} } = X^T \hat{A}^T (Z-Y), \ \Delta Y' \propto \hat{A}XX^T\hat{A}^T(Z-Y)
\end{equation}

\section{Proof of Proposition 1}
\label{appendix:proof_theorem1}
\begin{proof}

According to the given assumptions, for node $v$, we have $\hat{A}_{v,k}=\frac{1}{d+1}$, the expected number of intra-class edges is $dh$ (here the self-loop edge introduced by $\hat{A}$ is not counted based on the definition of edge homophily and data generation process) and inter-class edges is $(1-h)d$. Suppose there are $C \geq 2$ classes. Consider matrix $\hat{A}Z$,

Then, we have $\mathbb{E}\left[(\hat{A}Z)_{v,c}\right] =  \mathbb{E}\left[\sum\limits_{k \in \mathcal{V}} \hat{A}_{v,k} \textbf{1}_{\{Z_{k,:}= e_c^T\}}\right] = \sum\limits_{k \in \mathcal{V}} \frac{\mathbb{E}\left[\textbf{1}_{\{Z_{k,:}= e_c^T\}}\right] }{d+1}$, where $\bm{1}$ is the indicator function.

When $v$ is in class $c$, we have $\sum\limits_{k \in \mathcal{V}} \frac{\mathbb{E}\left[ \textbf{1}_{\{Z_{k,:}= e_c^T\}}\right]}{d+1} = \frac{hd+1}{d+1}$ ($hd+1=hd$ intra-class edges $+$ 1 self-loop introduced by $\hat{A}$).

When $v$ is not in class $c$, we have $\sum\limits_{k \in \mathcal{V}} \frac{\mathbb{E}\left[ \textbf{1}_{\{Z_{k,:}= e_c^T\}}\right]}{d+1} = \frac{(1-h)d}{(C-1)(d+1)}$ ($(1-h)d$ inter-class edges uniformly distributed in the other $C-1$ classes).

For nodes $v,u$, we have $(\hat{A}Z)_{v,:}, (\hat{A}Z)_{u,:} \in \mathbb{R}^C$ and since elements in $\hat{A}_{v,k}$ and $\hat{A}_{u,k'}$ are independently generated for all $k,k' \in \mathcal{V}$, we have
\begin{align*}
    \mathbb{E}\left[(\hat{A}Z)_{v,c} (\hat{A}Z)_{u,c}\right] & =  \mathbb{E}\left[(\sum\limits_{k \in \mathcal{V}} \hat{A}_{v,k} \textbf{1}_{\{Z_{k,:}= e_c^T\}}) (\sum\limits_{k' \in \mathcal{V}} \hat{A}_{u,k'} \textbf{1}_{\{Z_{k',:}= e_c^T\}}) \right] \\
    & = \mathbb{E}\left[(\sum\limits_{k \in \mathcal{V}} \hat{A}_{v,k} \textbf{1}_{\{Z_{k,:}= e_c^T\}}) \right] \mathbb{E} \left[ (\sum\limits_{k' \in \mathcal{V}} \hat{A}_{u,k'} \textbf{1}_{\{Z_{k',:}= e_c^T\}}) \right]
\end{align*}
Thus,

\begin{align*}
  \mathbb{E}\left[ S(\hat{A},Z)_{v,u} \right] 
  & = \mathbb{E}\left[<(\hat{A}Z)_{v,:}, (\hat{A}Z)_{u,:}> \right] = \sum\limits_c \mathbb{E}\left[(\sum\limits_{k \in \mathcal{V}} \hat{A}_{v,k} \textbf{1}_{\{Z_{k,:}= e_c^T\}}) \right] \mathbb{E} \left[ (\sum\limits_{k' \in \mathcal{V}} \hat{A}_{u,k'} \textbf{1}_{\{Z_{k',:}= e_c^T\}}) \right]\\
  & = \left\{
             \begin{array}{ll}
              \left(\frac{hd+1}{d+1}\right)^2 + \frac{\left( (1-h)d \right)^2}{(C-1)(d+1)^2}  , & \text{$u,v$ are in the same class} \\
               \frac{2(hd+1)(1-h)d}{(C-1)(d+1)^2}+ \frac{(C-2)(1-h)^2 d^2}{(C-1)^2 (d+1)^2}  , & \text{$u,v$ are in different classes} 
             \end{array}\right.
\end{align*}
For nodes $u_1$, $u_2$, and $v$,
where $Z_{u_1,:}=Z_{v,:}$ and $Z_{u_2,:}\neq Z_{v,:}$, 
\begin{align}
\label{eq:expectation_of_similarity_element}
    g(h) & \equiv \mathbb{E}\left[ S(\hat{A},Z)_{v,u_1} \right] - \mathbb{E}\left[ S(\hat{A},Z)_{v,u_2} \right] \\ \nonumber
    & = \frac{  (C-1)^2(hd+1)^2 + (C-1)\left[(1-h)d\right]^2 - (C-1)\left(2(hd+1)(1-h)d \right) - (C-2)\left[(1-h)d\right]^2}{(C-1)^2(d+1)^2} \\ \nonumber
 & 
 = \left(\frac{(C-1)(hd+1)-(1-h)d}{(C-1)(d+1)} \right)^2
\end{align}
Setting $g(h)=0$, we obtain the optimal $h$:
\beq \label{eq:optimalh}
h=\frac{d+1-C}{Cd} 
\eeq
For the data generation process in the synthetic experiments, 
we fix $d_\text{intra}$, then $d=d_\text{intra}/h$, which is a function of $h$.
We change $d$ in \eqref{eq:optimalh} to $d_\text{intra}/h$, leading to
\beq \label{eq:newh}
h = \frac{d_\text{intra}/h+1-C}{Cd_\text{intra}/h}
\eeq
It is easy to observe that $h$ satisfying \eqref{eq:newh} still makes $g(h)=0$, when $d$ in $g(h)$ 
is replaced by $d_\text{intra}/h$.
From \eqref{eq:newh} we obtain the optimal $h$ in terms of $d_\text{intra}$:
$$
h=\frac{d_{\text{intra}}}{C d_{\text{intra}} +C-1}
$$

\end{proof}
\subsection{An Extension of Proposition 1}
\label{appendix:extension_of_proposition1}
Base on the definition of aggregation similarity, we have
\begin{align*}
    S_\text{agg}\left(S(\hat{A},Z)\right) &= \frac{\left| \left\{v   \,\big| \,
    \mathrm{Mean}_u\big( \{S(\hat{A},Z)_{v,u} | Z_{u,:}=Z_{v,:} \}\big) 
    \geq \mathrm{Mean}_u\big(\{S(\hat{A},Z)_{v,u} | Z_{u,:} \neq Z_{v,:} \} \big) \right\} \right|}{\left| \mathcal{V} \right|}\\
    & = \frac{ \sum\limits_{v\in \mathcal{V}} \bm{1}_{ \left\{
    \mathrm{Mean}_u \big( \{S(\hat{A},Z)_{v,u} | Z_{u,:}=Z_{v,:} \}\big) 
    \geq \mathrm{Mean}_u\big(\{S(\hat{A},Z)_{v,u} | Z_{u,:} \neq Z_{v,:} \} \big) \right\}}}{\left| \mathcal{V} \right|}
\end{align*}
Then,
\begin{align*}
    \mathbb{E}\left(S_\text{agg}\left(S(\hat{A},Z)\right)\right) & = \mathbb{E} \left(\frac{ \sum\limits_{v\in \mathcal{V}} \bm{1}_{ \left\{
    \mathrm{Mean}_u \big( \{S(\hat{A},Z)_{v,u} | Z_{u,:}=Z_{v,:} \}\big) 
    \geq \mathrm{Mean}_u\big(\{S(\hat{A},Z)_{v,u} | Z_{u,:} \neq Z_{v,:} \} \big) \right\}}}{\left| \mathcal{V} \right|} \right)\\
    & =  \frac{ \sum\limits_{v\in \mathcal{V}} \mathbb{P}\left(  
    \mathrm{Mean}_u \big( \{S(\hat{A},Z)_{v,u} | Z_{u,:}=Z_{v,:} \}\big) 
    \geq \mathrm{Mean}_u\big(\{S(\hat{A},Z)_{v,u} | Z_{u,:} \neq Z_{v,:} \} \big)   \right)}{\left| \mathcal{V} \right|}  \\
    & = \mathbb{P}\left(\mathrm{Mean}_u \big( \{S(\hat{A},Z)_{v,u} | Z_{u,:}=Z_{v,:} \}\big) - \mathrm{Mean}_u\big(\{S(\hat{A},Z)_{v,u} | Z_{u,:} \neq Z_{v,:} \} \big) \geq 0 \right) \\
\end{align*}
Consider the random variable
\begin{equation*}
    RV = \mathrm{Mean}_u \big( \{S(\hat{A},Z)_{v,u} | Z_{u,:}=Z_{v,:} \}\big) - \mathrm{Mean}_u\big(\{S(\hat{A},Z)_{v,u} | Z_{u,:} \neq Z_{v,:} \} \big)
\end{equation*}
Since $RV$ is symmetrically distributed and under the conditions in proposition 1, its expectation is $\mathbb{E}[RV] = g(h)$ as showed in \eqref{eq:expectation_of_similarity_element}. Since the minimum of $g(h)$ is $0$ and $RV$ is symmetrically distributed, we have $\mathbb{P}(RV \geq 0) \geq 0.5$ and this can explain why $H_{\text{agg}}(\mathcal{G})$ is always greater than 0.5 in many real-world tasks.
\section{Proof of Theorem 1}
\label{appendix:proof_theorem2}
\begin{proof}
Define $ W_v^{c}=(\hat{A}Z)_{v,c}$. 
Then,
\begin{equation*}
    W_v^{c}=
    \sum\limits_{k\in \mathcal{V}} \hat{A}_{v,k} \bm{1}_{\{Z_{k,:} = e_c^T\}} \in [0,1], \ \ \sum\limits_{c=1}^C W_v^c = 1
\end{equation*}
Note that
\begin{equation} \label{eq:S}
S(I-\hat{A},Z) = (I-\hat{A})ZZ^T(I-\hat{A})^T = ZZ^T + \hat{A}ZZ^T\hat{A}^T - \hat{A}ZZ^T - ZZ^T\hat{A}^T 
\end{equation}
For any node $v$, let the class $v$ belongs to be denoted by $c_v$.
For two nodes $v,u$, if $Z_{v,:} \neq Z_{u,:}$, we have
\begin{align*}
    &(ZZ^T)_{v,u} = 0\\
    & (\hat{A}ZZ^T\hat{A}^T)_{v,u} = \sum\limits_{c=1}^C W_v^c W_u^c \\
    & (\hat{A}ZZ^T)_{v,u} = W_v^{c_u} \\
    & (ZZ^T\hat{A}^T)_{v,u} = (\hat{A}ZZ^T)_{u,v} =  W_u^{c_v}
\end{align*}
Then, from \eqref{eq:S} it follows that
\begin{align*}
    (S(I-\hat{A},Z))_{v,u} = \sum\limits_{c=1}^C W_v^c W_u^c  - W_v^{c_u} - W_u^{c_v}
\end{align*}
When $C=2$,  
\begin{align*}
    S(I-\hat{A},Z)_{v,u} = W_v^{c_u}( W_u^{c_u}-1) + W_u^{c_v} (W_v^{c_v}-1) \leq 0
\end{align*}
If $Z_{v,:} = Z_{u,:}$, \ie{} $c_v=c_u$, we have
\begin{align*}
    &(ZZ^T)_{v,u} = 1\\
    & (\hat{A}ZZ^T\hat{A}^T)_{v,u} = \sum\limits_{c=1}^C W_v^c W_u^c \\
    & (\hat{A}ZZ^T)_{v,u} = W_v^{c_v} \\
    & (ZZ^T\hat{A}^T)_{v,u} = (\hat{A}ZZ^T)_{u,v} =  W_u^{c_u} = W_u^{c_v}
\end{align*}
Then, from \eqref{eq:S} it follows that
\begin{align*}
    S(I-\hat{A},Z)_{v,u} &= 1+\sum\limits_{c=1}^C W_v^c W_u^c  - W_v^{c_v} - W_u^{c_v}\\
    & = \sum\limits_{c=1,c\neq c_v}^C W_v^c W_u^c + 1+ W_v^{c_v}W_u^{c_v} - W_v^{c_v} - W_u^{c_v}\\
    & = \sum\limits_{c=1,c\neq c_v}^C W_v^c W_u^c + (1- W_v^{c_v})(1-W_u^{c_v}) \geq 0
\end{align*}
Thus, if $C=2$, for any $v\in\mathcal{V}$, if $Z_{u,:} \neq Z_{v,:}$, we have $S(I-\hat{A},Z)_{v,u} \leq 0$; if $Z_{u,:} = Z_{v,:}$, we have $S(I-\hat{A},Z)_{v,u} \geq 0$. Apparently, the two conditions in \eqref{eq:diversification_distinguishability} are satisfied.
Thus $v$ is  diversification distinguishable and $\mathrm{DD}_{\hat{A},X}(\mathcal{G})=1$.
The theorem is proved.
\end{proof}

\section{Discussion of the Limitations of Diversification Operation}
\label{appendix:limitation_diversification}
\begin{figure}[htbp]
\centering
{
\captionsetup{justification = centering}
\includegraphics[width=1\textwidth]{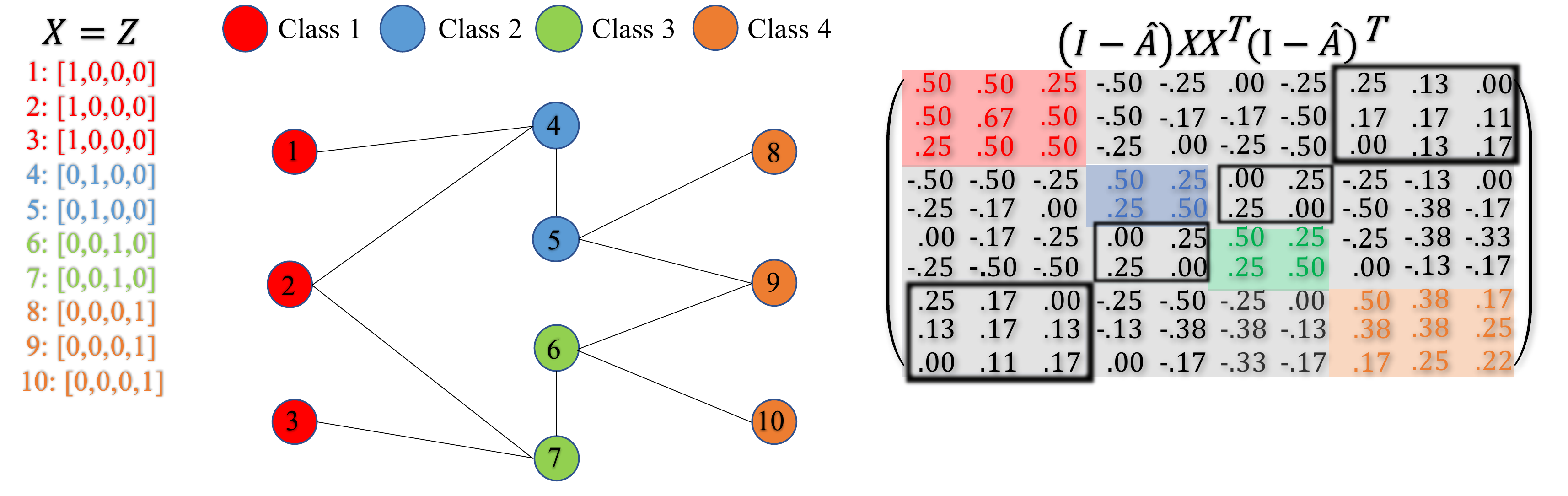}}
{%
  \caption{Example of the case (the area in black box) that HP filter does not work well for harmful heterophily}%
  \label{fig:unsuccessful_example_hp_filter}
}
\end{figure}
From the black box area of $S(I-\hat{A},X)$ in the example in Figure \ref{fig:unsuccessful_example_hp_filter} we can see that nodes in class 1 and 4 assign non-negative weights to each other although there is no edge between them; nodes in class 2 and 3 assign non-negative weights to each other as well. This is because the surrounding differences of class 1 are similar as class 4, so are class 2 and 3. In real-world applications, when nodes in several small clusters connect to a large cluster, the surrounding differences of the nodes in the small clusters will become similar. In such case, HP filter are not able to distinguish the nodes from different small clusters. 

\section{The Similarity, Homophily and $\mathrm{DD}_{\hat{A},X}(\mathcal{G})$ Metrics and Their Estimations}
\label{appendix:estimation_similarity_homophily_diversification_metrics}
Firstly, we would like to clarify that, for each curve in the synthetic experiments, the node features are fixed and we only change the homophily values. But in real-world tasks, different datasets have different features and aggregated features. Thus, to get more instructive information for different datasets and compare them, we need to consider more metrics, e.g. feature-label consistency and aggregated-feature-label consistency. With the similarity score of the features $S_\text{agg}\left(S(I,X)\right)$ and aggregated features $S_\text{agg}\left(S(\hat{A},X)\right)$ listed in Table \ref{tab:estimation_dataset_stats}, our methods open up a new perspective on analyzing and comparing the performance of graph-agnostic models and graph-aware models in real-world tasks. Here are 2 examples.

\paragraph{Example 1:} People observe that GCN (graph-aware model) underperforms MLP-2 (graph-agnostic model) on $\textit{Cornell, Wisconsin, Texas, Film}$ and people commonly believe that the bad graph structure (low $H_\text{edge}, H_\text{node}, H_\text{class}$ values) is the reason for performance degradation. But based on the high aggregation homophily values, the graph structure inconsistency is not the main cause of the performance degradation. And from Table \ref{tab:estimation_dataset_stats} we can see that the $S_\text{agg}\left(S(\hat{A},X)\right)$ for those 4 datasets are lower than their corresponding $S_\text{agg}\left(S(I,X)\right)$, which implies that it is the aggregated-feature-label inconsistency that causes the performance degradation, i.e. the aggregation step actually decrease the quality of node features rather than making them more distinguishable.

For the rest 5 datasets $\textit{Chameleon, Squirrel, Cora, Citeseer, PubMed}$, we all have $S_\text{agg}\left(S(\hat{A},X)\right)$ larger than $S_\text{agg}\left(S(I,X)\right)$ except $\textit{PubMed}$, which means the aggregated features have higher quality than raw features. We can see that the proposed metrics are much more instructive than the existing ones.

\paragraph{Example 2:} According to $H_\text{edge}, H_\text{node}, H_\text{class}$, the value for $\textit{Chameleon}$, and $\textit{Squirrel}$ are extremely low indicating graph structure are bad for GNNs. But on contrary, GCN outperforms MLP-2 on those 2 datasets. Traditional homophily metrics fail to explain such phenomenon but our method can give an explanation from different angles: For Chameleon, its modified aggregation homophily is not low and its $S_\text{agg}\left(S(\hat{A},X)\right)$ is higher than its $S_\text{agg}\left(S(I,X)\right)$, which means its graph-label consistency together with aggregated-feature-label consistency help the graph-aware model obtain the performance gain; for Squirrel, its modified aggregation homophily is low but its $S_\text{agg}\left(S(\hat{A},X)\right)$ is higher than its $S_\text{agg}\left(S(I,X)\right)$, which means although its graph-label consistency is bad, the aggregated-feature-label consistency is the key factor to help the graph-aware model perform better.

We also need to point out that (modified) aggregation similarity score, $S_\text{agg}\left(S(\hat{A},X)\right)$ and $S_\text{agg}\left(S(I,X)\right)$ are not deciding values because they do not consider the nonlinear structure in the features. In practice, a low score does not tell us the GNN models will definitely perform bad.

\begin{table}[htbp]
  \centering
  \tiny
  \setlength{\tabcolsep}{2pt}
  \caption{Additional metrics and their estimations with only training labels (mean $\pm$ std)}
    \begin{tabular}{c|ccccccccc}
    \toprule
    \toprule
          & Cornell & Wisconsin & Texas & Film  & Chameleon & Squirrel & Cora  & CiteSeer & PubMed \\
    \midrule
    $H_\text{agg}(\mathcal{G})$ & 0.9016 & 0.8884 & 0.847 & 0.8411 & 0.805 & 0.6783 & 0.9952 & 0.9913 & 0.9716 \\
    $S_\text{agg}\left(S(\hat{A},X)\right)$ & 0.8251 & 0.7769 & 0.6557 & 0.5118 & 0.8292 & 0.7216 & 0.9439 & 0.9393 & 0.8623 \\
     $S_\text{agg}\left(S(I,X)\right)$ & 0.9672 & 0.8287 & 0.9672 & 0.5405 & 0.7931 & 0.701 & 0.9103 & 0.9315 & 0.8823 \\
    $DD_{\hat{A},X}(\mathcal{G})$ & 0.3497 & 0.6096 & 0.459 & 0.3279 & 0.3109 & 0.2711 & 0.2681 & 0.4124 & 0.1889 \\
    \midrule
    $\hat{H}_\text{agg}(\mathcal{G})$ & 0.9046 $\pm$ 0.0282 & 0.9147 $\pm$ 0.0260 & 0.8596 $\pm$ 0.0299 & 0.8451 $\pm$ 0.0041 & 0.8041 $\pm$ 0.0078 & 0.6788 $\pm$ 0.0077 & 0.9959 $\pm$ 0.0011 & 0.9907 $\pm$ 0.0015 & 0.9724 $\pm$ 0.0015 \\
    $\hat{S}_\text{agg}\left(S(\hat{A},X)\right)$ & 0.8266 $\pm$ 0.0526 & 0.8280 $\pm$ 0.0351 & 0.6835 $\pm$ 0.0498 & 0.5345 $\pm$ 0.0421 & 0.8433 $\pm$ 0.0070 & 0.7352 $\pm$ 0.0132 & 0.9487 $\pm$ 0.0023 & 0.9451 $\pm$ 0.0038 & 0.8626 $\pm$ 0.0021 \\
    $\hat{S}_\text{agg}\left(S(I,X)\right)$ & 0.9752 $\pm$ 0.0174 & 0.8680 $\pm$ 0.0270 & 0.9661 $\pm$ 0.0336 & 0.5438 $\pm$ 0.0184 & 0.8257 $\pm$ 0.0050 & 0.7472 $\pm$ 0.0089 & 0.9204 $\pm$ 0.0044 & 0.9441 $\pm$ 0.0036 & 0.8835 $\pm$ 0.0019 \\
    $\hat{DD}_{\hat{A},X}(\mathcal{G})$ & 0.3936 $\pm$ 0.0663 & 0.6073 $\pm$ 0.0436 & 0.4817 $\pm$ 0.0762 & 0.3300 $\pm$ 0.0136 & 0.3329 $\pm$ 0.0151 & 0.3021 $\pm$ 0.0101 & 0.3198 $\pm$ 0.0225 & 0.4424 $\pm$ 0.0136 & 0.1919 $\pm$ 0.0046 \\
    \bottomrule
    \bottomrule
    \end{tabular}%
  \label{tab:estimation_dataset_stats}%
\end{table}%

Furthermore, in most real-world applications, not all labels are available to calculate the dataset statistics. Thus, we randomly split the data into 60\%/20\%/20\% for training/validation/test, and only use the training labels for the estimation of the statistics. We repeat each estimation for 10 times and report the mean with standard deviation. The results are shown in table \ref{tab:estimation_dataset_stats}.
\paragraph{Analysis}
From the reported results we can see that the estimations are accurate and the errors are within the acceptable range, which means the proposed metrics and similarity scores can be accurately estimated with a subset of labels and this is important for real-world applications.

\section{A Detailed Explanation of the Differences Between ACM(II)-GNNs and GPRGNN, FAGCN}
\label{appendix:difference_with_sota_methods}

Differences with GPRGNN \cite{chien2021adaptive}:
\begin{itemize}
    \item \emph{GPRGNN does not feed distinct $\textbf{node-wise feature transformation}$ to different "multi-scale channels"}
    
    We first rewrite GPRGNN as 
$$\mathbf{Z} = \sum\limits_{k=0}^{K} \gamma_{k} \mathbf{H}^{(k)} = \sum\limits_{k=0}^{K} \gamma_{k} I \mathbf{H}^{(k)} = \sum\limits_{k=0}^{K} diag(\gamma_{k}, \gamma_{k},\dots,\gamma_{k}) \mathbf{H}^{(k)}, \text{ where } \mathbf{H}^{(k)} = \hat{A}_{\text{sym}}\mathbf{H}^{(k-1)}, \mathbf{H}^{(0)}_{i:} = f_\theta(X_{i:}).$$ 
From the above equation we can see that $\mathbf{Z} = \sum\limits_{k=0}^{K} \gamma_{k} \hat{A}_{\text{sym}}^k f_\theta(X_{i:})$, \ie{} the $\textbf{node-wise feature transformation}$ in GPRGNN is only learned by the same $\theta$ for all the "multi-scale channels". But in the ACM framework, different channels extract distinct information with different parameters separately.

    \item \emph{GPRGNN does not have node-wise mixing mechanism.}
    
There is no node-wise mixing in GPRGNN. The mixing mechanism in GPRGNN is $\mathbf{Z} = \sum\limits_{k=0}^{K} diag(\gamma_{k}, \gamma_{k},\dots,\gamma_{k}) \mathbf{H}^{(k)}$, i.e. for each "multi-scale channel $k$", all nodes share the same mixing parameter $\gamma_{k}$. But in the ACM framework, the node-wise channel mixing can be written as $\mathbf{Z} = \sum\limits_{k=0}^{K} diag(\gamma_{k}^1,\gamma_{k}^2,\dots,\gamma_{k}^N) \mathbf{H}^{(k)}$ where $K$ is the number of channels, $N$ is the number of nodes and $\gamma_{k}^i, i=1,\dots,N$ are the mixing weights that are learned by node $i$ to mix channel $k$. ACM and ACMII allow GNNs to learn more diverse mixing parameters in diagonal than GPRGNN and thus, have stronger expressive power than GPRGNN.
\end{itemize}

Differences with FAGCN \cite{bo2021beyond}:
\begin{itemize}
    \item  \emph{The targets of node-wise operations in ACM (channel mixing) and FAGCN (negative message passing) are different.}
    
   Instead of using a fixed low-pass filter $\hat{A}$, FAGCN tries to learn a more powerful aggregator $\hat{A}'$ based on $\hat{A}$ by allowing negative message passing. The node-wise operation in FAGCN is similar to GAT [3] which is trying to modify the $\textbf{node-wise filtering (message passing) process}$, i.e. for each node $i$, it assigns different weights $\alpha_{ij} \in [-1,1]$ to different neighborhood nodes (equation 7 in FAGCN paper). The goal of this node-wise operation in FAGCN is $\textbf{to learn a new filter during the filtering process node-wisely}$. But in ACM, the node-wise operation is to mix the $\textbf{filtered information}$ from each channel which is processed by different fixed filters. The targets of two the node-wise operations are actually different things.
    \item \emph{FAGCN does not learn distinct information from different "channels". FAGCN only uses simple addition to mix information instead of node-wise channel mixing mechanism}
    
   The learned filter $\hat{A}'$ can be decomposed as $\hat{A}'=\hat{A}_1' + (-\hat{A}_2')$, where $\hat{A}_1'$ and $-\hat{A}_2'$ represent  positive and negative edge (propagation) information, respectively. But FAGCN does not feed distinct information to $\hat{A}_1'$ and $-\hat{A}_2'$. Moreover, the aggregated $\hat{A}_1' X$ and "diversified" information $(-\hat{A}_2') X$ are simply added together instead of using any node-wise mixing mechanism. In ACM, we learn distinct information separately in each channel with different parameters and add them adaptively and node-wisely instead of just adding them together. In section 6.1, the ablation study has empirically shown that node-wise adaptive channel mixing is better than simple addition.
\end{itemize}

\end{document}